\documentclass[
  10pt,
  longbibliography,
  twocolumn,
  twoside,
]{article}

\usepackage[numbers,round,comma,sort&compress]{natbib}
\usepackage{natbib}
\setcitestyle{square}
\usepackage[
  left=0.65in,
  right=0.65in,
  top=0.65in,
  bottom=0.65in,
]{geometry}

\usepackage{sectsty}
\sectionfont{\raggedright\large\bfseries}
\subsectionfont{\raggedright\large}

\usepackage[font=small,labelfont=bf]{caption}

\usepackage{fancyhdr}
\usepackage{environ}

\usepackage{rotating}

\usepackage{floatrow}

\usepackage[table]{xcolor}

\usepackage{multirow}

\usepackage{url}
\PassOptionsToPackage{hyphens}{url}\usepackage{hyperref}
\makeatletter
\g@addto@macro{\UrlBreaks}{\UrlOrds}
\makeatother

\usepackage{hyperref}
\hypersetup{
    colorlinks=true,
    linkcolor=grey,
    filecolor=grey,
    urlcolor=grey,
    citecolor=todoblue,
    breaklinks=true,
}

\usepackage[normalem]{ulem}

\usepackage{soul}

\makeatletter
\let\ftype@table\ftype@figure
\makeatother

\usepackage{array}

\newcommand{\PreserveBackslash}[1]{\let\temp=\\#1\let\\=\temp}

\usepackage{colortbl}

\makeatletter

\def\CT@@do@color{%
  \global\let\CT@do@color\relax
  \@tempdima\wd\z@
  \advance\@tempdima\@tempdimb
  \advance\@tempdima\@tempdimc
  \advance\@tempdimb\tabcolsep
  \advance\@tempdimc\tabcolsep
  \advance\@tempdima2\tabcolsep
  \kern-\@tempdimb
  \leaders\vrule
  \hskip\@tempdima\@plus  1fill
  \kern-\@tempdimc
  \hskip-\wd\z@ \@plus -1fill }
\makeatother

\usepackage{xcolor}

\definecolor{olivegreen}{rgb}{0.33333,.41961,0.18431}
\definecolor{forestgreen}{rgb}{0.13333,.5451,0.13333}

\definecolor{lightgrey}{rgb}{0.7,0.7,0.7}
\definecolor{verylightgrey}{rgb}{0.90,0.90,0.90}
\definecolor{veryverylightgrey}{rgb}{0.95,0.95,0.95}
\definecolor{grey}{rgb}{0.5,0.5,0.5}
\definecolor{darkgrey}{rgb}{0.3,0.3,0.3}
\definecolor{verydarkgrey}{rgb}{0.15,0.15,0.15}

\definecolor{headerblue}{HTML}{33367E}
\definecolor{unitednationsblue}{HTML}{4D88FF}

\definecolor{charcoal}{HTML}{36454F}
\definecolor{cinerous}{HTML}{98817B}
\definecolor{feldgrau}{HTML}{4D5D53}
\definecolor{glaucous}{HTML}{6082B6}
\definecolor{arsenic}{HTML}{3B444B}
\definecolor{xanadu}{HTML}{738678}

\definecolor{firebrick}{HTML}{B22222}
\definecolor{orangered}{HTML}{FF4500}
\definecolor{tomato}{HTML}{FF6347}

\definecolor{purpletaupe}{HTML}{3B444B}
\definecolor{headerorange}{RGB}{255,116,0}
\definecolor{headergray}{RGB}{230,230,230}

\definecolor{headerpop}{RGB}{230,230,230}

\definecolor{magmalight}{RGB}{252,251,195}
\definecolor{magmalightalt}{RGB}{250,240,184}
\definecolor{magmamedium}{RGB}{245,200,146}
\definecolor{magmadark}{RGB}{224,106,98}

\definecolor{icelight}{RGB}{223,242,244}
\definecolor{icelightalt}{RGB}{189,222,226}
\definecolor{icemedium}{RGB}{132,184,204}
\definecolor{icedark}{RGB}{103,153,191}

\definecolor{traitrowcolor}{RGB}{223,242,244}
\definecolor{traitrowcoloralt}{RGB}{189,222,226}

\definecolor{characterrowcolor}{RGB}{252,251,195}
\definecolor{characterrowcoloralt}{RGB}{250,240,184}

\definecolor{archetyperowcolor}{RGB}{255,213,212} %% pale pink
\definecolor{archetyperowcoloralt}{RGB}{255,182,179} %% melon

\definecolor{datasetrowcolor}{RGB}{232,244,234}
\definecolor{datasetrowcoloralt}{RGB}{210,231,214}

\usepackage{etoolbox}
\AtBeginEnvironment{quotation}{%
  \setlength{\parindent}{0pt}%
  \setlength{\parskip}{0.5\baselineskip}%
}

\usepackage{placeins}

\NewEnviron{excerpt}{
    \begin{quote}
      \medskip
      \BODY
      \medskip
    \end{quote}
}

\NewEnviron{editnote}{
    \begin{quote}
      \color{rose}
      $\blacksquare$
      \BODY
      \medskip
    \end{quote}
}

\newcommand{\command}[1]{
  \lstinline[language={[LaTeX]TeX},basicstyle=\ttfamily]{#1}
}

\newcommand{\editbox}[2]{
}

\newcommand{\editboxwithlatex}[2]{
}

\usepackage{fancyvrb}

\usepackage{tcolorbox}
\tcbuselibrary{skins,breakable,listings,breakable}

\usepackage{tikz}
\usetikzlibrary{shapes}

\tikzstyle{mybox} = [draw=lightblue!70, fill=lightblue!7, very thick,
    rectangle, rounded corners, inner sep=10pt, inner ysep=20pt]

\tikzstyle{editortitle} =[draw=archetyperowcoloralt, fill=archetyperowcoloralt, text=black]

\newcommand\Loadedframemethod{default}
\usepackage[framemethod=\Loadedframemethod]{mdframed}

\mdfsetup{skipabove=\topskip,skipbelow=\topskip}

\tikzstyle{loglinetitle} =[draw=icedark, fill=icemedium!50, text=black]

\newenvironment{loglinebox}[1][]{

  \ifstrempty{#1}%
  {\mdfsetup{%
    frametitle={%
       \tikz[baseline=(current bounding box.east),outer sep=0pt]
        \node[loglinetitle, anchor=east,rectangle]
        {\strut~~#1:~~\strut};}}
  }%
  {\mdfsetup{%
     frametitle={%
       \tikz[baseline=(current bounding box.east),outer sep=0pt]
        \node[loglinetitle,anchor=east,rectangle]
        {\strut~~#1:~~\strut};}}%
   }%
   \mdfsetup{innertopmargin=5pt,linecolor=icedark,%
             linewidth=0.5pt,topline=true,
             frametitleaboveskip=\dimexpr-\ht\strutbox\relax,}
   \begin{mdframed}[backgroundcolor=icelight,nobreak=true]\relax%
     \raggedright
}{\end{mdframed}}

\tikzstyle{abstracttitle} =[draw=magmadark!75, fill=magmamedium!75, text=black]

\newenvironment{abstractbox}[1][]{

  \ifstrempty{#1}%
  {\mdfsetup{%
    frametitle={%
       \tikz[baseline=(current bounding box.east),outer sep=0pt]
        \node[abstracttitle, anchor=east,rectangle]
        {\strut~~#1:~~\strut};}}
  }%
  {\mdfsetup{%
     frametitle={%
       \tikz[baseline=(current bounding box.east),outer sep=0pt]
        \node[abstracttitle,anchor=east,rectangle]
        {\strut~~#1:~~\strut};}}%
   }%
   \mdfsetup{innertopmargin=5pt,linecolor=magmadark,%
             linewidth=0.5pt,topline=true,
             frametitleaboveskip=\dimexpr-\ht\strutbox\relax,}
   \begin{mdframed}[backgroundcolor=magmalight,nobreak=true]\relax%
     \raggedright
}{\end{mdframed}}

\tikzstyle{infotitle} =[draw=darkgrey, fill=lightgrey!50, text=black]

\newenvironment{infobox}[1][]{

  \ifstrempty{#1}%
  {\mdfsetup{%
    frametitle={%
       \tikz[baseline=(current bounding box.east),outer sep=0pt]
        \node[infotitle, anchor=east,rectangle]
        {\strut~~#1:~~\strut};}}
  }%
  {\mdfsetup{%
     frametitle={%
       \tikz[baseline=(current bounding box.east),outer sep=0pt]
        \node[infotitle,anchor=east,rectangle]
        {\strut~~#1:~~\strut};}}%
   }%
   \mdfsetup{innertopmargin=5pt,linecolor=grey,%
             linewidth=0.5pt,topline=true,
             frametitleaboveskip=\dimexpr-\ht\strutbox\relax,}
   \begin{mdframed}[backgroundcolor=lightgrey!25,nobreak=true]\relax%
     \raggedright
}{\end{mdframed}}

\tikzstyle{changelogtitle} =[draw=darkgrey, fill=lightgrey!50, text=black]

\newenvironment{changelogbox}[1][]{

  \ifstrempty{#1}%
  {\mdfsetup{%
    frametitle={%
       \tikz[baseline=(current bounding box.east),outer sep=0pt]
        \node[changelogtitle, anchor=east,rectangle]
        {\strut~~#1:~~\strut};}}
  }%
  {\mdfsetup{%
     frametitle={%
       \tikz[baseline=(current bounding box.east),outer sep=0pt]
        \node[changelogtitle,anchor=east,rectangle]
        {\strut~~#1:~~\strut};}}%
   }%
   \mdfsetup{innertopmargin=5pt,linecolor=grey,%
             linewidth=0.5pt,topline=true,
             frametitleaboveskip=\dimexpr-\ht\strutbox\relax,}
   \begin{mdframed}[backgroundcolor=lightgrey!25,nobreak=true]\relax%
     \raggedright
     \begin{enumerate}[
         noitemsep,
         leftmargin=10pt,
       ]
}{
     \end{enumerate}
   \end{mdframed}
}

\newcommand{\revisioncolor}{forestgreen}
  
\newif\ifhighlightrevisions
\highlightrevisionsfalse  %% set false to remove all margin bars

\ifhighlightrevisions
\newtcolorbox{revisionbar}{
  enhanced jigsaw,
  boxrule=0pt,
  colframe=white,
  colback=white,
  left=0pt,
  right=0pt,
  top=2pt,
  bottom=2pt,
  breakable,
  before upper={%
    \setlength{\parskip}{1\baselineskip plus .1\baselineskip minus .1\baselineskip}%
    \setlength{\parindent}{0pt}%
  },
  before skip=6pt,
  after skip=6pt,
  overlay={%
    \draw[\revisioncolor, line width=2pt]
      ([xshift=-5pt]frame.north west) --
      ([xshift=-5pt]frame.south west);
  }
}
\else
\newenvironment{revisionbar}{}{}  % disables markup
\fi

\newcommand{\revisionnote}[1]{}

\usepackage{listings}

\usepackage{stringstrings}
\usepackage{readarray}

\def\firstchar#1#2|{#1}
\edef\tbs{\detokenize{\X}}
\edef\tbs{\expandafter\firstchar\tbs|}
\edef\tlb{\detokenize{{}}}
\edef\tlb{\expandafter\firstchar\tlb|}
\edef\tus{\detokenize{_}}
\newcounter{index}
\newcommand\detokenizeplus[1]{%
  \def\temparg{\detokenize{#1}}%
  \getargsC{\temparg}%
  \setcounter{index}{0}%
  \def\prevmacro{F}%
  \whiledo{\value{index} < \narg}{%
    \stepcounter{index}%
    \isnextbyte[q]{\tbs}{\csname arg\roman{index}\endcsname}%
    \if T\theresult%
      \if T\prevmacro\unskip\else\fi%
      \def\prevmacro{T}%
    \else%
      \def\prevmacro{F}%
   \fi%
    \isnextbyte[q]{\tlb}{\csname arg\roman{index}\endcsname}%
    \if T\theresult\unskip\else\fi%
    \isnextbyte[q]{\tus}{\csname arg\roman{index}\endcsname}%
    \if T\theresult\unskip\else\fi%
    \csname arg\roman{index}\endcsname~%
  }%
}

\usepackage[bbgreekl]{mathbbol}
\usepackage{bbm}

\usepackage{colortbl}

\definecolor{goodblue}{RGB}{0, 91, 187}
\hypersetup{
  colorlinks=true,
  allcolors=goodblue,
  urlcolor=goodblue,
  citecolor=goodblue,
  pdfborder={0 0 0},
  breaklinks=true,
}

\newenvironment{textblock}{\renewcommand{\item}{}}{}

\newcommand{\newphrase}{\newline}

\usepackage{changepage}

\newcommand{\padzero}[1]{\ifnum #1 < 10 0\fi #1}

\usepackage{siunitx}

\newcommand\zeropad[2]{%
  \ifnum#2<0\relax%
    {\ensuremath-}\zeropadA{#1}{\the\numexpr#2*-1\relax}%
  \else%
    \zeropadA{#1}{#2}%
  \fi%
}
\def\zeropadA#1#2{%
  \ifnum1#2<1#1
    \zeropadA{#1}{0#2}%
  \else%
    #2%
  \fi%
}

\usepackage{enumitem}

\usepackage{graphicx}

\usepackage{verbatim}

\usepackage{amssymb}
\usepackage{amsmath}

\usepackage{ifthen}

\usepackage{overpic}

\usepackage{longtable}

\usepackage{mathtools}
\usepackage{gensymb}

\usepackage{textgreek}

\newcommand{\greekousia}{\textomikron$\overset{\mbox{,}}{\mbox{\textupsilon}}$\textsigma$\acute{\mbox{\textiota}}$\textalpha}

\newboolean{twocolswitch}

\makeatletter
\let\ftype@table\ftype@figure
\makeatother

\usepackage[export]{adjustbox}

\makeatletter
\newcommand*{\centerfloat}{%
  \parindent \z@
  \leftskip \z@ \@plus 1fil \@minus \textwidth
  \rightskip\leftskip
  \parfillskip \z@skip}
\makeatother

\newcommand{\sindex}[1]{}
\newcommand{\nindex}[1]{}

\newcommand{\etal}{\textit{et al.}}
\newcommand{\www}[1]{\url{#1}}

\newcommand{\Req}[1]{Eq.~(\ref{#1})}

\usepackage{lettrine}

\newcommand{\onehalf}{\frac{1}{2}}

\newcommand{\colvec}[1]{
  \left[
    \begin{array}{c}
      #1
    \end{array}
  \right]
}

\newcommand{\mymatrix}[2]{
  \left[
    \begin{array}{#1}
      #2
    \end{array}
  \right]
}

\newcommand{\frequency}{f}
\newcommand{\probsymbol}{p}

\newcommand{\bigrank}{R}

\newcommand{\systemsymbol}{\Omega}
\newcommand{\elementsymbol}{\tau}

\newcommand{\lexicallens}{\mathcal{L}}

\newcommand{\wordvar}[1]{\{\unskip\textnormal{\unskip\textbf{\unskip#1}}\}}

\newcommand{\semdiffsign}{\Leftrightarrow}
\newcommand{\semdiff}[2]{\{#1\,$\semdiffsign$\,#2\}}

\newcommand{\Ntypes}{N_{\textnormal{types}}}
\newcommand{\Nsd}{N_{\textnormal{differentials}}}

\newcommand{\meaningsymbol}{M}
\newcommand{\avgmeaning}{\meaningsymbol_{\textnormal{avg}}}

\newcommand{\Mvalence}{\textnormal{\textbf{Va}}}
\newcommand{\Marousal}{\textnormal{\textbf{Ar}}}
\newcommand{\Mdominance}{\textnormal{\textbf{Dm}}}

\newcommand{\Mgoodness}{\textnormal{\textbf{Gd}}}

\newcommand{\Maggression}{\textnormal{\textbf{Ag}}}

\newcommand{\Mpower}{\textnormal{\textbf{Pw}}}

\newcommand{\Mdanger}{\textnormal{\textbf{Dg}}}

\newcommand{\Mstructure}{\textnormal{\textbf{St}}}

\setboolean{twocolswitch}{true}

\raggedright

\newcommand{\suppmaterial}{Supplementary Materials}

\newcommand{\onlineappendicesplain}{Online Appendices}
\newcommand{\onlineappendiceslocation}{\href{https://compstorylab.org/ousiometry/}{compstorylab.org/ousiometry/}}%% \newcommand{\suppmaterial}{Supplementary Material}
\usepackage{authblk}

\begin{document}

\title{
  \protect\revisionnote{Revised title (129 characters):}
  \protect Ousiometrics:\\
The essence of meaning aligns with a power-danger-structure framework\\
instead of valence-arousal-dominance

  \bigskip

  \large
  \protect\textit{Science Advances},
\textbf{12}(9): eadr4039,
2026
\\
\href{https://doi.org/10.1126/sciadv.adr4039}{https://doi.org/10.1126/sciadv.adr4039}

%% \href{https://www.science.org/doi/10.1126/sciadv.adr4039}{10.1126/sciadv.adr4039}

%%  \protect\revisionnote{Short title (41 characters):
%%  \protect\input{true-dimensions-of-meaning.title-50characters.tex}}
%%  \protect\revisionnote{Teaser (122 characters):
%%  \protect\input{true-dimensions-of-meaning.teaser.tex}}
}

%% https://tex.stackexchange.com/questions/214404/add-affiliations-to-the-authors-name-in-the-article-class
%% https://tex.stackexchange.com/questions/259375/customize-author-and-affiliation-using-authblk

\renewcommand*{\Authsep}{, }
\renewcommand*{\Authand}{, }
\renewcommand*{\Authands}{, }
\renewcommand*{\Affilfont}{\normalsize\normalfont}
\renewcommand*{\Authfont}{\bfseries}
\setlength{\affilsep}{2em}

\author[1,2,3,4,5*]{Peter~Sheridan~Dodds\thanks{Corresponding author: peter.dodds@uvm.edu}}
\author[6,7]{Thayer~Alshaabi}
\author[8]{Mikaela~Irene~Fudolig}
\author[1,2,9]{Julia~Witte~Zimmerman}
\author[2,3,5,9]{Juniper~Lovato}
\author[2,3]{Shawn~Beaulieu}
\author[10]{Joshua~R.~Minot}
\author[1,2]{Michael~V.~Arnold}
\author[10]{Andrew~J.~Reagan}
\author[1,2,11]{Christopher~M.~Danforth}

\affil[1]{
  Computational Story Lab,
  Vermont Advanced Computing Center,
  University of Vermont,
  Burlington,
  VT 05405,
  United States
}

\affil[2]{
  Vermont Complex Systems Institute,
  MassMutual Center of Excellence for Complex Systems and Data Science,
  University of Vermont,
  Burlington,
  VT 05405,
  United States
  }

\affil[3]{
  Department of Computer Science,
  University of Vermont,
  Burlington,
  VT 05405,
  United States
}

\affil[4]{
  Santa Fe Institute,
  1399 Hyde Park Rd,
  Santa Fe,
  NM 87501,
  United States
}

\affil[5]{
  Complexity Science Hub,
  Metternichgasse 8,
  1030 Vienna,
  Austria
}

\affil[6]{
  Howard Hughes Medical Institute,
  Janelia Research Campus,
  Ashburn,
  VA 20147,
  United States
} 

\affil[7]{
  Advanced Bioimaging Center,
  University of California Berkeley,
  Berkeley,
  CA 94720,
  United States
}

\affil[8]{
  School of Computer and Mathematical Sciences,
  University of Adelaide,
  Adelaide,
  SA 5005,
  Australia
}

\affil[9]{
  Computational Ethics Lab,
  University of Vermont,
  Burlington,
  VT 05405,
  United States
}

\affil[10]{
  MassMutual Data Science,
  Amherst,
  MA 01002,
  United States
}

\affil[11]{
  Department of Mathematics \& Statistics,
  University of Vermont,
  Burlington,
  VT 05405,
  United States
  }

\date{\today}

\maketitle

%%%%%%%%%%%%%%%%%%%%%%%%%%%%%%%%%%%%%
%% logline and excerpt graphic
%%%%%%%%%%%%%%%%%%%%%%%%%%%%%%%%%%%%%

%% turn off for Science Advances

%% \begin{comment}

\mbox{}

\vspace*{-40pt}
\hspace*{20pt}
\begin{minipage}{450pt}

  \begin{tabular}{>{\raggedright\arraybackslash}p{180pt}p{30pt}p{270pt}}

    %% excerpt graphic
    \parbox{180pt}{    
      \includegraphics[width=180pt,valign=m,frame]{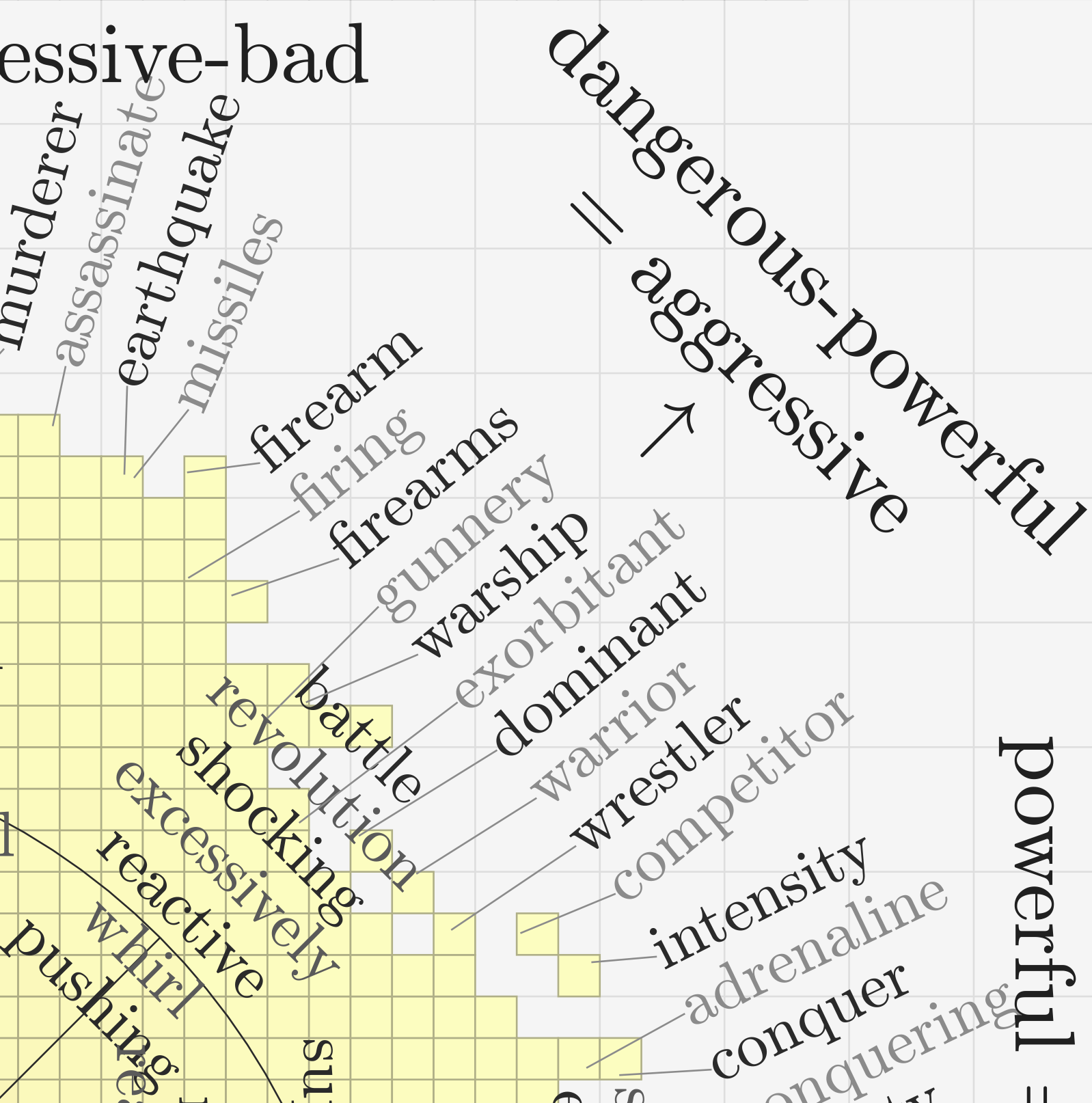}
    }

    &
    
    &
    
    \parbox{270pt}{    
      \begin{loglinebox}[Logline]
        \raggedright
        We show that the essential meaning \newphrase
conveyed by individual words \newphrase
is best represented by a compass-like plane \newphrase
described by interrelated differentials of
bad-good,
weak-powerful,
gentle-aggressive,
and
safe-dangerous,
joined with a third dimension
of structured-unstructured
(GPADS).

\bigskip

We uncover a linguistic `safety bias' \newphrase
by examining how words are used \newphrase
in large-scale, diverse corpora.

\bigskip

We find the power-danger-structure framework
to be naturally aligned with token usage in real corpora
as well as with seemingly disparate frameworks.

%% including
%% the circumplex model of affect
%% and
%% the archetypes of fictional characters.

\bigskip

We construct and demonstrate the use of an `ousiometer' for
measuring time series of essential meaning.

%% We show that the essential meaning \newphrase
%% conveyed by individual words \newphrase
%% maps to a compass-like plane \newphrase
%% with major axes of \newphrase
%% powerful-weak and dangerous-safe.
%% 
%% \bigskip
%% 
%% We uncover a linguistic `safety bias' \newphrase
%% by examining how words are used \newphrase
%% in large-scale, diverse corpora.

%% We show that the essential meaning conveyed by individual around 20,000 English words
%% maps to a compass-like plane with major axes of
%% powerful-weak and dangerous-safe.
%% The power-danger (PD) framework is modified by a minor third dimension of structure (PDS).
%% We uncover a linguistic `safety bias' by examining
%% how words are used in large-scale, diverse corpora.
%% To enable exploration and description,
%% we define ousiometrics to be the study of essential meaning,
%% develop ousiograms along with
%% synousionyms and antousionyms,
%% and construct and test a prototype `ousiometer' for
%% measuring essential meaning in texts.
%% We also find that a power-danger framework variously revises
%% and accords with models of personality and psychological states,
%% and detail a range of future areas of research.

%% We re-analyze a 20,000 word lexicon scored according to
%% longstanding conceptions of essential meaning.

        \smallskip
      \end{loglinebox}
    }
    
  \end{tabular}
  
\end{minipage}

\clearpage

\newgeometry{
  left=2in,
  right=2in,
  top=0.65in,
  bottom=0.65in,
  }

\onecolumn

\renewcommand{\baselinestretch}{1.2}
\selectfont

%% \end{comment}

%%%%%%%%%%%%%%%%%%%
%% Abstract:
%%%%%%%%%%%%%%%%%%%

%% for arXiv

\begin{abstractbox}[Abstract]
  \raggedright

  \bigskip
  
  \begin{textblock}
\item
  From work emerging through the middle of the 20th century,
  the essence of meaning has become widely accepted
  as being described by the three orthogonal dimensions of
  valence, arousal, and dominance (VAD).
\item 
  These essential dimensions have become the cornerstone
  of sentiment analysis across many fields.
\end{textblock}

\begin{textblock}
\item 
  By re-examining first types and then tokens for the English language,
  and through the use of automatically annotated histograms---`ousiograms'---we
  find here that:
\end{textblock}
\begin{enumerate}
\item 
  The essence of meaning conveyed by words
  is instead best described by a
  goodness-power-aggression-danger-structure circumplex framework (GPADS);
\item 
  That large-scale English language corpora
  reveal a systematic bias toward safe, low-danger words;
\item[]
  and
\item 
  That the power-danger-structure (PDS)
  framework is the minimal framework
  that represents essential meaning.
\end{enumerate}

\begin{textblock}
\item
  We find remarkable congruences between the GPADS framework
  and other spaces including
  mental states and fictional archetypes,
  and we construct and demonstrate a prototype ousiometer.
\end{textblock}

  \bigskip
  
\end{abstractbox}

%% \input{true-dimensions-of-meaning.abs-long.tex}

%% for Science Advances
%% 
%% \hspace*{110pt}
%% \begin{minipage}{300pt}
%% 
%%   \vspace{40pt}
%%   
%%   \begin{center}
%%     \large\textbf{\protect\revisionnote{Revised} Abstract \protect\revisionnote{(92 words)}}
%%   \end{center}
%% 
%%   \protect\revisionnote{NOTE: The abstract is revised overall and much shorter. Important changes are highlighted:}
%%   
%%   \noindent\input{true-dimensions-of-meaning.abs.tex}
%% 
%% \end{minipage}

%%%%%%%%%%%%%%%%%%%
%% Keywords:
%%%%%%%%%%%%%%%%%%%

%% turn off for Science Advances
\begin{infobox}[Keywords]
  \raggedright
  meaning;
essential~meaning;
ousiometrics;
ousiometry;
ousiograms;
language;
words;
semantic~differentials;
Osgood;
best-worst~scaling;
essential~meaning~frameworks;
VAD;
GAS;
PDS;
GPAD;
GPADS;
valence;
arousal;
dominance;
goodness;
aggression;
power;
danger;
structure;
emotion;
mental~states;
lexical~instruments;
hedonometrics;
computational~humanities;
stories;
archetypes;
archetypometrics;
telegnomics

 \smallskip
\end{infobox}

\renewcommand{\baselinestretch}{1}
\selectfont

\twocolumn

\restoregeometry

%% turn off for Science Advances
\clearpage

\tableofcontents

\clearpage

%% \maybe{Note that half plain for sem diff matters}

%% TODO: make this true:
%% 
%% to make all figures:
%% 
%% cd ~/work/meaning/2021-01ousiometrics/figures/
%% make_all_the_figures
%%
%%
%% DAPGS => GPADS (G-A-S properly weaved with P-D-S)

%%%%%%%%%%%%%%%%%%%%%%%%%%%%%%%%%%%%%%%%%%%
%%

\section{Introduction}
\label{sec:meaning.introduction}

As encoded by human language,
meaning spans a high dimensional semantic space
that is continually expanding and evolving,
bearing
complex hierarchical and networked structures~\cite{lakoff1980a,miller1995a,mikolov2013b}.
In attempting to understand any quantified complex system,
a most basic step is to apply a method of dimensional reduction.
If we distill meaning to its essence, boiling off all higher dimensions of meaning---the focus of our work here---do we
find fundamental dimensions of meaning space that are interpretable
and, moreover,
reliably experienced, conceptualized, and conveyed by people~\cite{osgood1952a,solomon1954a,osgood1957a,osgood1969a,osgood1975a,russell1980a,russell2003a}?

\begin{revisionbar}
  The short answer is yes---Osgood et al.'s methods~\cite{osgood1957a}
  (see below) have proved to
be successful and the Osgood framework has come to be widely used
across many fields.
But as we will show, Osgood's framework does not fully hold up to a
modern computational analysis using large-scale datasets and new methodologies.
\end{revisionbar}

%% More broadly,
%% how might essential meaning vary
%% for entities across all levels of cognition and complexity,
%% from individual organisms,
%% groups at all scales,
%% and
%% to artificial, algorithmic systems?

Here, we define `ousiometrics' to be the quantitative study of
the essential meaningful components of an entity, however represented and perceived.
Used in philosophical and theological settings, the word `ousia' comes from Ancient Greek
\greekousia\
and is the etymological root of
the word `essence' whose more modern usage is our intended reference.

Introducing the terminology of ousiometrics in particular helps
us distinguish from general fields which study meaning
such as semantics or semiotics.
And for framing purposes, being able to thematically
name specialized tools and instruments will be helpful
in the presentation of our work.
In particular, we will develop and explore a series of `ousiograms'
which are annotated representations of two dimensional slices of meaning space.

For our purposes here,
and in the tradition of Osgood et al.~\cite{solomon1954a,osgood1957a}
and the many who have followed~\cite{mehrabian1974a,mehrabian1974b,russell1980a,russell2003a,bradley1999a,dodds2011e,warriner2013a,warriner2014a,warriner2017a,mohammad2018a,mohammad2018b},
our measurement of essential meaning
will rest on---and is constrained by---the map presented by language.

%% foundational work, EPA and VAD

\subsection{How the measurement of essential meaning has been operationalized}
\label{sec:meaning.intro-history}

To help explain the purpose of our paper---why our
approach to ousiometrics is warranted---we
outline the relevant history of measuring essential meaning,
and describe the longstanding problematic aspects of
experiment design and measurement instruments.

The quantitative measurement of the essence of meaning
was primarily developed by researchers in the middle of the 1900s,
particularly by Osgood and colleagues,
with their foundational work published in the 1950s~\cite{solomon1954a,osgood1957a}.
The majority of ensuing research~\cite{osgood1969a,osgood1975a,bradley1999a}
has been built around 
human evaluation of words or phrases using the instrument of
the semantic differential~\cite{karwoski1938a,solomon1954a,osgood1957a}.
In a typical study, surveyed participants are
asked to rate individual words on Likert scales with
endpoints described by `bipolar adjectival pairs' (BAPs)
such as
\semdiff{soft}{hard},
\semdiff{rough}{smooth},
and
\semdiff{cold}{hot}.
Throughout, we will indicate semantic differentials as per the
examples of the preceding sentence,
bracketing
and
connecting bipolar adjectival pairs with the symbol $\semdiffsign$.
The measurement of essential meaning is thus
operationalized via surveys which ask people
to respond to isolated words and phrases,
given some semantic differential and some scale.

\begin{revisionbar}
Modern support for semantic differentials
comes from findings that large language models (LLMs)
encode semantic axes consistent with human
judgment.
Grand et al.~\cite{grand2022a} demonstrate that ``semantic
projection''---mapping embeddings onto differential axes like
size, age, and wealth---recovers
feature-level conceptual knowledge.
\end{revisionbar}

Each semantic differential is considered a dimension (an axis) in
a potentially high dimensional space,
and researchers then apply some variant of factor analysis
to the average scores,
such as principal component analysis (PCA)
or singular value decomposition (SVD),
or more sophisticated
methods~\cite{thurstone1931a,joreskog1969a,fabrigar1999a,visinescu2014a,jennrich2002a}.
In general, factor analysis determines a set of
dimensions that are linear combinations of the study's
semantic differentials, which must then be interpreted.
As we detail below, we are able to use basic SVD
and it is in the interpretation step
that we develop a new systematic methodology.

Based on a range of studies,
Osgood \etal~\cite{osgood1957a} identified
three orthogonal dimensions for the essence of meaning.
In order of variance explained for the studies at the time, the three dimensions were dubbed:\\
1.~Evaluation (e.g., \semdiff{positive}{negative}),\\
2.~Potency (e.g., \semdiff{dominant}{submissive}),
and \\
3.~Activity (e.g., \semdiff{active}{passive}).

Though the `EPA' framework has been
challenged in various ways~\cite{russell1980a,barrett1999a,yik2011a},
as have semantic differentials themselves~\cite{russell1980a,mohammad2018b},
researchers were increasingly drawn to
take the EPA framework as a ground truth
when carrying out new studies~\cite{bradley1999a,warriner2013a,mohammad2018b}.

In the focused context of studying emotion,
a theoretical concept of a three dimensional representation of emotion
goes back to (at least)
Wundt in the late 1800s~\cite{wundt1922a,reisenzein2000a}.
For emotion, the EPA dimensions were re-ordered and recast
by Mehrabian and Russell as:
1. Pleasure or Valence,
2. Arousal,
and
3. Dominance
(PAD or VAD)~\cite{mehrabian1974a,mehrabian1974b}.
To make clear that this was the authors' intention,
from the summary of Ref.~\cite{mehrabian1974b}:
\begin{quote}
  ``Semantic differential studies, in particular, have shown that human
  judgments of diverse samples of stimuli can be characterized in terms
  of three dimensions: evaluation, activity, and potency. We have termed
  the corresponding emotional responses pleasure, arousal, and
  dominance.''
\end{quote}
%% just linear independence:
%% And to exemplify the presumption of orthogonality of the three core dimensions,
%% from p.~292 of Ref.~\cite{mehrabian1974b} we have:
%% \begin{quotation}
%%  ``Thus, each dimension is, in principle, functionally independent of
%%   the other two; none of the three dimensions could be subsumed by the
%%   others.''
%% \end{quotation}
Subsequent work has tended to use the term valence instead of pleasure,
and we will follow the VAD nomenclature.

The VAD framework has seen extensive application across diverse disciplines.
In psychology~\cite{warriner2013a,bradley1994measuring}, 
neuroscience~\cite{gannouni2020a,lang1997iaps}, 
and natural language processing (NLP)~\cite{el-haj2023a,mohammad2018b},
VAD provides dimensional models to capture affective meaning.
Other areas include
sentiment analysis~\cite{mantyla2016a,mohammad2025a},
emotion detection~\cite{yang2024a,park2021a},
marketing~\cite{mehrabian1974a,lang1993looking}, 
human-computer interaction~\cite{verma2017a}, 
and education~\cite{li2022emotion,guendil2017vad}.

Now, while VAD was intended to be a scoped version of EPA,
the two frameworks have been conflated.
Generally, VAD has become the framework presented
in studies, even when essential meaning, rather than emotion, has been
the focus~\cite{bradley1999a,warriner2013a,mohammad2018b}.
Elsewhere, the original connection between VAD and EPA has been overlooked
or considered broken,
leading to re-analyses about whether or not
the match between EPA and VAD holds at all~\cite{bakker2014a}.

Nevertheless, to be consistent with the direction taken by the literature,
we will refer to VAD rather than the more general EPA going forward.

\subsection{The major problems with measuring essential meaning}
\label{sec:meaning.intro-problems}

We describe a set of problems that we contend have thwarted
the full development of ousiometry over time.

%% \smallskip
%% \noindent
%% \textbf{1. Scale:}\\

\subsubsection{Scale}
\label{subsec:meaning.intro-problems-scale}

Given that the EPA framework was developed before and during the 1950s,
the foundational studies were limited in size,
both in lexicon analyzed and the number of participants surveyed.
For example, as part of the research that led to the EPA framework,
Osgood \etal~\cite{osgood1957a} report on a study of 20 concept nouns
evaluated on 50 semantic differentials by 100 undergraduates.
Published in 1980, Russell's circumplex model of affect (which we
examine later in Sec.~\ref{subsec:meaning.circumplex})
was based on the scoring
of 28 words and phrases~\cite{russell1980a}.
The Affective Norms for English Words (ANEW) study of the late 1990s
moved the lexicon size up to 1,034, but with VAD as the accepted
fundamental framework, and still using surveys of undergraduates.
In work carried out around 2010 involving two of the present authors,
an order of magnitude
jump to over 10,000 English words was conducted online
through Amazon's Mechanical Turk with 50 evaluations per word
along the single semantic differential
of valence interpreted as happiness
(discussed further below)~\cite{dodds2011e}.
This data set, labMT (language assessment by Mechanical Turk),
was later expanded to 10 languages, each with over 10,000 words scored online
by participants around the world~\cite{dodds2015a}.
Crucially, and in contrast to the ANEW word lists,
the labMT words analyzed were chosen according to frequency of usage
(again, discussed further below).
In 2013, Warriner \etal~\cite{warriner2013a}
published scores for close to 14,000 English words with VAD scores.
Finally, in 2018,
Mohammad produced what will be the basis of
our analysis here, the NRC VAD lexicon:
Over 20,000 English words and phrases with VAD scores~\cite{mohammad2018b}.

So, it is only in the last 15 years that studied lexicons have begun to represent
the scale of human vocabularies.
We are consequently now well placed to perform the necessary
work of re-examining the findings of the field's foundational research.

%% \smallskip
%% \noindent
%% \textbf{2. The focus on types alone and not tokens:}\\

\subsubsection{The focus on types alone and not tokens}
\label{subsec:meaning.intro-problems-types-tokens}

We use the standard type-token language for describing entities~\cite{herdan1960a}:
Type refers to an entity's class (or species) while
token refers to an entity itself as an instance of that class.
Beyond language, the type-token distinction appears across all
complex systems with heavy-tailed distributions of component frequencies.
Perhaps in settings not involving words and texts,
the problems with studying only types would be more apparent.
For example, in determining some overall measure of a forest,
we would not want to assign equal weight to the most common
and the most rare species.
Here, we will study both lexicons (types) and large-scale texts (tokens),
gaining separate results from both.

Almost all essential meaning studies have been at the level of types, each word or concept
given equal weighting.
However, we must consider the weight of types in a text
according to the frequency of their
corresponding tokens~\cite{herdan1960a}.
Only then can we make defensible observations about a whole space of communication.
The ANEW study~\cite{bradley1999a}, for example, is based on 1,034 expert chosen words
which proved to be a poor fit for natural language~\cite{reagan2017a}.
By contrast, with careful consideration of word usage, we were able
to show that the Polyanna Principle~\cite{boucher1969a} 
manifests a linguistic positivity bias
across 24 corpora spanning 10 languages~\cite{dodds2015a}.

%% \smallskip
%% \noindent
%% \textbf{3. The use of Likert scales for semantic differentials:}\\

\subsubsection{The use of Likert scales for semantic differentials}
\label{subsec:meaning.intro-problems-likert}

The use of a Likert scale for evaluations
of semantic differentials has long been standard practice.
Relatively recently, best-worst scaling  has
been suggested to be a more robust instrument than the Likert scale,
as well as a far more efficient one~\cite{louviere2015a}.
To our great benefit,
Mohammad's survey of over 20,000 words and phrases
preferentially uses best-worst scaling,
finding appreciable improvement in
split-half reliabilities over studies using Likert scales.

%% \smallskip
%% \noindent
%% \textbf{4. Limitations of factor analysis for a large number of categorical dimensions:}\\

\subsubsection{Limitations of factor analysis for a large number of categorical dimensions}
\label{subsec:meaning.intro-problems-factor-analysis}

While tables of factor analysis weightings can be exhaustively informative for small-scale studies,
we will not be able to make much sense of point clouds of tens of thousands of
unlabeled words in two or three dimensions.
Here, we will show how a kind of automatically annotated
histogram---an ousiogram---coupled with ranked word lists
provides an instrument that will help us explore, describe, and
support our assessments of the dimensions of essential meaning.

%% \smallskip
%% \noindent
%% \textbf{5. The misalignment between expert-chosen, end-point descriptors
%%  and dimensions of essential meaning:}\\

\subsubsection{The misalignment between expert-chosen, end-point descriptors
  and dimensions of essential meaning}
\label{subsec:meaning.intro-problems-experts}

We come to a critical problem with any essential meaning study
that starts from a presumption of the EPA/VAD framework.
We go back to basics and outline the
four-step experimental process that has been used to
extract essential dimensions of meaning in the first place:
\begin{enumerate}[label=\roman*.]
\item 
  Participants are asked to rate a set of $\Ntypes$ types (e.g.,
  words, images) using a set of $\Nsd$ semantic differentials defined
  by bipolar adjectival pairs.
  Some examples from Osgood \etal's
  50 semantic differentials
  for the study mentioned above
  include
  \semdiff{large}{small},
  \semdiff{clean}{dirty},
  \semdiff{brave}{cowardly},
  \semdiff{bass}{treble},
  and
  \semdiff{near}{far}
  (p.~43 in Ref.~\cite{osgood1957a}).
\item 
  Some variant of factor analysis (e.g., PCA, SVD) is then employed to
  obtain an ordered set of dimensions that are linear combinations of
  the semantic differential dimensions.
\item 
  Researchers interpret the main dimensions and ascribe them with
  both descriptive names (e.g., `evaluation') and, crucially, sets of
  `end-point descriptors'
  (e.g., happiness, pleasure, contentedness for high valence
  and unhappiness, annoyance, negativeness for low valence).
  These new semantic differentials are not then
  described by simple bipolar adjectival pairs
  but rather clusters of words and phrases at each end.
\item 
  Researchers reduce the meaning space to 2 or 3 of the most prominent
  dimensions (e.g., by variance explained through singular values).
\end{enumerate}
With ousiometric dimensions so determined (e.g., EPA), researchers then
move on to new studies using only a modified version of step I:
\begin{enumerate}[label=\roman*.]
\item 
  Participants are asked to rate a set of $\Ntypes$ types along
  2 or 3 expert-chosen dimensions that are defined by
  expert-identified sets of end-point descriptors.
\end{enumerate}
As such, 
there is then no assurance that the expert-identified
end-point descriptors
will be construed by participants in a way that
imposes the expert-defined dimensions.

Indeed, we observe that across many studies,
raters have been
presented with end-point descriptors that
render the three VAD dimensions
with problematic imprecision~\cite{osgood1957a,bradley1999a,russell1980a,bakker2014a,mohammad2018b}.
For example, for the ANEW study, valence was described
to participants as a \semdiff{happy}{unhappy} scale
as follows (emphasis added):
\begin{quote}
``At one extreme of [this \semdiff{happy}{unhappy}] scale, you are
  \textbf{happy, pleased, satisfied, contented, hopeful}.
  \ldots\
  The other end of the scale is when you feel completely
  \textbf{unhappy, annoyed, unsatisfied, melancholic, despaired, or bored}.''
\end{quote}
The meaning captured by both ends is broad,
the numbers of descriptors differ,
and the word `bored' clearly overlaps
with the dimension of arousal.

For the NRC VAD lexicon, raters were guided by end-point descriptors
(`paradigm terms')
which were taken from Refs.~\cite{osgood1957a}, \cite{russell1980a},
and \cite{bradley1999a}.
We list all descriptors for the six end-points used in Ref.~\cite{mohammad2018b}
in Tab.~\ref{tab:meaning.VADlimits}.
As for the ANEW study, we see the end-points for each dimension
combine to create coarse semantic limits.
For example, for low arousal,
there is clear semantic separation between
`sluggishness' and `calmness',
as there is for
`weak' and `guided' for low dominance.

Our remedy is simple: Always carry out steps 1--4 above even
when attempting to impose a minimal ousiometric framework.
Factor analysis will then accommodate a reasonable lack of exactness 
in how dimensions are prescribed.
And if we find
that the VAD framework is in fact perfectly prescribable,
we will have done the work needed to make this clear.

\setlength{\tabcolsep}{6pt}
\begin{table*}[t!]
  \renewcommand*{\arraystretch}{1.4}
  \rowcolors{2}{gray!5}{gray!10}
  \begin{tabular}{rl} %% >{\raggedright\arraybackslash}p{0.7\textwidth}}
    \rowcolor{gray!25}
    \textbf{VAD end-points} & \textbf{Paradigm words and phrases presented to raters}
    \\
    %%    \hline
    highest valence &
    happiness, pleasure, positiveness, satisfaction, contentedness, hopefulness \\
    lowest valence &
    unhappiness, annoyance, negativeness, dissatisfaction, melancholy, despair \\
    highest arousal &
    arousal, activeness, stimulation, frenzy, jitteriness, alertness \\ 
    lowest arousal &
    unarousal, passiveness, relaxation, calmness, sluggishness, dullness, sleepiness \\
    highest dominance &
    dominant, in control of the situation, powerful, influential, important, autonomous \\
    lowest dominance &
    submissive, controlled by outside factors, weak, influenced, cared-for, guided \\
    %%    \hline
  \end{tabular}
  \caption{
    End-point descriptors used in Ref.~\cite{mohammad2018b}
    for the survey leading to the NRC VAD lexicon.
    As for many studies presuming an orthogonal VAD framework,
    the end-points are semantically broad and imprecise.
  }
  \label{tab:meaning.VADlimits}
\end{table*}

%% \smallskip
%% \noindent
%% \textbf{6. Presuming that the VAD framework does capture essential meaning and that the three dimensions are orthogonal:}\\

\subsubsection{Presuming that the VAD framework does capture essential meaning and that the three dimensions are orthogonal}
\label{subsec:meaning.intro-presumption-of-VAD}

As we have observed, Osgood \etal's~\cite{osgood1957a} EPA/VAD framework
has become generally accepted as valid.
However, 
modern, large-scale VAD
evaluations of words and phrases have increasingly pointed toward the VAD framework
being non-orthogonal.
Leaving aside problematic sampling of words,
the ANEW study~\cite{bradley1999a}
found evidence that arousal was mildly positively correlated with the magnitude of valence.
The near 14,000 lemma VAD study of Warriner \etal~\cite{warriner2013a} 
found correlations between the three VAD dimensions,
the strongest being between valence and dominance with
$r_{\Mvalence,\Mdominance} \simeq 0.72$ (Pearson's correlation),
which prompted
the authors to call into question 
the orthogonality of the VAD framework.

Most recently, using best-worst scaling for the NRC VAD lexicon,
Mohammad~\cite{mohammad2018b} 
found a somewhat weaker correlation 
of $r_{\Mvalence,\Mdominance}\simeq 0.49$,
and then asserted that 
valence and dominance were only ``slightly correlated'',
a view with which we do not agree.
In reference to the valence-dominance correlation
in the Warriner \etal\ study~\cite{warriner2013a}, Mohammad stated:
\begin{quote}
  ``Given that the split-half reliability score for their dominance
  annotations is only 0.77, the high V–D correlations raises the
  suspicion whether annotators sufficiently understood the difference
  between dominance and valence.''
\end{quote}
So the suggestion here is that the problem is not that the VAD framework
is not orthogonal, but that participants failed to grasp
the definitions of dimensions.

Our position, per problem 5 above,
is that imposing VAD dimensions experimentally
through end-point descriptors
is a difficult task and that factor analysis
is always required.
And in challenging the VAD framework,
we will show that these observed correlations are real and understandable,
and ultimately lead to a revised framework
we will identify to be power-danger-structure (PDS).

\begin{revisionbar}
  We note that we do not use the expanded NRC VAD lexicon
  described in Ref.~\cite{mohammad2025a}
  as the lexicon merges words scored with best-worst scaling
  and traditional Likert scale.
\end{revisionbar}

\subsection{Roadmap for the paper}
\label{sec:meaning.intro-road map}

We first describe the data sets we analyze and explore
in Sec.~\ref{sec:meaning.data}.
We make the key distinction between text corpora
that are type-based (i.e., lexicons) or
token-based (written or recorded expression)~\cite{herdan1960a}.

\begin{revisionbar}
Through a series of integrated figures,
we then demonstrate our four main findings:
1. The framework of valence-arousal-dominance (VAD)
is far from being an orthogonal system~\cite{warriner2013a},
and this failure is due to the difficulties
of constructing semantic differentials for essential
dimensions of meaning
(Secs.~\ref{subsec:meaning.ousiograms}
and~\ref{subsec:meaning.DVousiogram});
2. A goodness-aggression-structure (GAS) framework and a
power-danger-structure (PDS) framework
both provide two alternative, interpretable, and interconnected orthogonal systems
that agree with earlier circumplex formulations,
and which we will combine as a
goodness-power-aggression-danger-structure (GPADS) framework
(Secs.~\ref{subsec:meaning.extra-analysis},
~\ref{subsec:meaning.VADfail},
\ref{subsec:meaning.GAS},
\ref{subsec:meaning.PDS},
and
\ref{subsec:meaning.GPADS});
and
3. Only the power-danger-structure framework aligns with
the essential meaning patterns of real corpora when 
we properly account for frequency of usage by considering tokens;
and
4. Diverse, large-scale text corpora present
a systematic, low-danger  `safety bias'
(Sec.~\ref{subsec:meaning.safetybias}).

With the GAS, PDS, and GPADS frameworks established, we identify
remarkable congruences in two other areas which are far removed
from how people respond to isolated words:
1. The influential circumplex model of affect~\cite{russell1980a}
(Sec.~\ref{subsec:meaning.circumplex});
%% 2. The Five Factor model of personality
%% (Sec.~\ref{subsec:meaning.personality});
and
2. Archetypometrics---the data driven determination
of fictional character archetypes~\cite{dodds2025archetypometrics,dodds2025archetypometrics-dataset-zenodo}
(Sec.~\ref{subsec:meaning.fictional-characters}).
\end{revisionbar}

We share ousiometric word scores for all frameworks,
additional figures,
and scripts
on Gitlab and/or in the \suppmaterial.

We summarize our results and offer thoughts on future work
in Sec.~\ref{sec:meaning.concludingremarks}.

\section{Description of data sets}
\label{sec:meaning.data}

We build our findings in two stages using two distinct kinds of word lists:
1. Types: A lexicon for the English language (each word is of equal importance),
and
2. Tokens: Frequency-rank distributions~\cite{zipf1949a} for large-scale corpora
(words are ranked by frequency of usage with most used word having rank 1;
words with tied frequency are given an average of corresponding ranks if they were not tied).
In general, observations made solely by examining a lexicon (the level of types)
will be given a stringent test
when confronted by real-world word usage (the level of tokens).

The type stage:
As indicated in the introduction, we use the
NRC VAD lexicon comprising around 20,000 words and terms~\cite{mohammad2018b}.
The lexicon was compiled from a variety of sources and largely contains
lower-case, latin-character words along with some 2-, 3-, and 4-grams
(an $n$-gram is a phrase made up of $n$-terms).
Proper nouns and function words have generally been excluded.
Some words are evidently hashtag constructions from
social media (with the hashtag removed).
The lexicon is a union of existing lexicons,
some of which were expert-compiled (e.g., the ANEW study~\cite{bradley1999a})
and others based on frequency of usage.
While the presence of expert-compiled lexicons is not ideal,
we will see that the coverage of real corpora is sufficient
for the purposes of our work here.

For each term in the NRC VAD lexicon,
scores
within the VAD framework~\cite{mehrabian1974a,mehrabian1974b}
were assessed by
survey using best-worst scaling~\cite{louviere2015a}.
Terms were presented in groups of four
and participants were asked to rank the highest and lowest
according to one of the three VAD dimensions
(see Ref.~\cite{mohammad2018b} for full details).
Each term's score is in the interval [0,1].
To accommodate singular value decomposition,
we remove the mean from each dimension,
which by the nature of best-worst scaling is $\onehalf$.
We thus shift the VAD 
scores from [0,1] to [-$\onehalf$,+$\onehalf$].

The token stage:
With findings from studying the NRC VAD lexicon,
we then analyze seven corpora---where frequency of word usage now matters---which
vary broadly in kind, formality, scale, and historical time frame.
\begin{enumerate}
\item
  English Fiction (1900--2019) from the Google Books project,
  with each book contributing words equally,
  and then each year's size-rank distribution weighted equally~\cite{michel2011a,pechenick2015a};
\item 
  Jane Austen's six novels with all books merged,
  sourced from the Gutenberg Project,
  \url{http://www.gutenberg.org}.
\item
  The majority of Arthur Conan Doyle's Sherlock Holmes stories
  with all stories merged,
  (4 novels and 44 short stories
  taken from
  \url{https://sherlock-holm.es/},
  missing the 12 short stories contained in ``The Case-Book of Sherlock Holmes'');
\item 
  The New York Times (1987--2007) Frequency-rank distributions
  merged without reweighting across all years~\cite{nytimescorpus2008a};
\item
  Wikipedia (English language, 2019/03 snapshot)~\cite{semenov2019a};
\item
  RadioTalk (transcriptions of talk radio broadcasts in the US,
  2018/10--2019/03)~\cite{beeferman2019a};
\item
  Twitter (approximately 10\% of all tweets
  identified as English in 2020---including retweets---with each day's
  frequency-rank distribution contributing equally)~\cite{alshaabi2021c}.
\end{enumerate}

%% We acknowledge that we
%% have represented the dominance and danger dimensions by the same variable $\Mdanger$,
%% and potency and power by $\Mpower$.
%% We opt for this notational collision in preference to more cumbersome expressions
%% that would largely only be of service in the present paper.
%% To maintain clarity, we repeatedly express
%% the contexts of VAD and PDS in text and figures,
%% and we will use full names for dimensions where needed.

\section{Analysis, Results, and Discussion}
\label{sec:meaning.results}

\begin{figure*}[tp!]

  %% build with figousiometer9010_lexicon012
   
  \includegraphics[width=\textwidth]{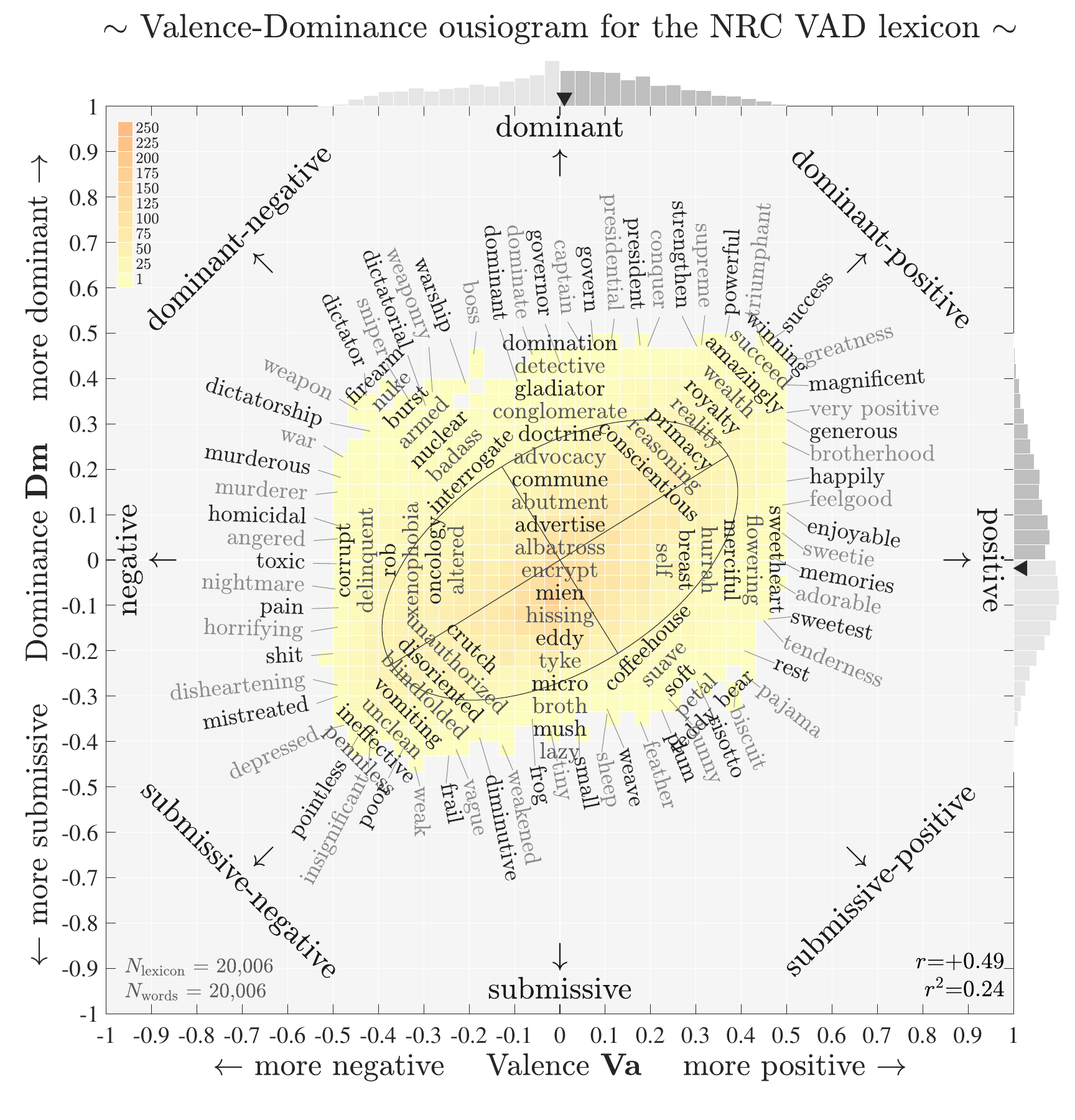}
  \caption{
    \raggedright
    Valence-dominance `ousiogram' for the NRC VAD lexicon
    of around 20,000 words
    scored within the
    valence-arousal-dominance (VAD) framework~\protect\cite{mehrabian1974a,mehrabian1974b}
    using best-worst scaling~\cite{louviere2015a,mohammad2018b}.
    Ousiograms are annotated two-dimensional histograms
    of two essential dimensions describing any collection
    of labeled entities.
    Here, we arrange words according to their valence-dominance scores,
    collapsing the third dimension of arousal.
    We use a bin width of 1/30, and we have shifted all
    $\Mvalence$, $\Marousal$, and $\Mdominance$
    scores from [0,1] to [-$\onehalf$,+$\onehalf$].
    To enable comparisons, we use limits of [-1,1] throughout the paper.
    We plot marginal distributions of
    $\Mvalence$
    and
    $\Mdominance$
    along the top and right sides, with darker gray
    indicating positive values, and 
    solid dark triangles locating the
    medians
    of
    $\Mvalence$
    and
    $\Mdominance$.
    The ellipse 
    represents the axes determined by singular value decomposition (SVD)
    acting on the $\Mvalence$-$\Mdominance$ plane,
    and shows a strong departure from
    the $\Mvalence$ and $\Mdominance$ axes.
    We label words around the edge of the
    $\Mvalence$-$\Mdominance$
    distribution aligned with normals to the distribution's convex hull,
    and add example words
    at internal locations along the main axes and the two diagonals.
    Upon inspection, the words shown are reasonably located according
    to their essential
    values of $\Mvalence$ and $\Mdominance$.
    Notes:
    See
    Figs.~\ref{fig:meaning.ousiometer9010_VAD012_1_supp1}
    and~\ref{fig:meaning.ousiometer9010_VAD012_3_supp1}
    in the 
    \protect\suppmaterial\
    for large-scale ousiograms of
    $\Mvalence$-$\Marousal$
    and
    $\Marousal$-$\Mdominance$.
    Labeled words are not restricted in their value of
    the third dimension, arousal $\Marousal$, which may vary unevenly.
    Alternating shades of gray are for readability.
    For these larger ousiograms, we automatically label the four cardinal and inter-cardinal directions
    with their endpoint adjective (e.g, `dominant-positive' in the northeast corner).
  }
  \label{fig:meaning.ousiometer9010_VAD012_2_noname}
\end{figure*}

%%%%%%%%%%%%%%%%%%%%%%%%%%%%%%%%%%%%%%%%%%%%%%%%%%%%%%%%%%%%%%
%% first figure: D-V ousiogram
%%%%%%%%%%%%%%%%%%%%%%%%%%%%%%%%%%%%%%%%%%%%%%%%%%%%%%%%%%%%%%

\subsection{Ousiograms}
\label{subsec:meaning.ousiograms}

Complex systems are often manifested from
a set of distinct, named entities---types---whose frequencies of
occurrence as interacting tokens roughly obey a heavy-tailed distribution,
and whose characteristics
reside in some high dimensional space~\cite{zipf1949a,simon1955a,williams2015a,newman2005b,dodds2023a}.
Language is a canonical example with
words as types and meanings as one of their characteristics.
One approach to better understanding such high dimensional complex systems,
is thorough dimensional reduction
where we maintain the set of all types
but seek to distill the characteristics of these types
down to an essential few.

%% That is, we work to extract the essential characteristics
%% of the types which constitute a complex system.

To inform and help validate our analysis,
we will use `ousiograms'.
We define an ousiogram as a systematically
and informatively annotated two-dimensional histogram
for two essential quantities of a complex system's component entities.
The entities represented in ousiograms
may be either types or tokens~\cite{herdan1960a},
with types contributing equally
while  a token's contribution would be proportional to
the frequency of that token's appearance within
a given system.

In Fig.~\ref{fig:meaning.ousiometer9010_VAD012_2_noname},
we present an ousiogram for
valence $\Mvalence$
and
dominance $\Mdominance$
for the NRC VAD lexicon~\cite{mohammad2018b}.
We use valence
and
dominance
as an example to demonstrate the non-orthogonality
of the VAD framework with best-worst scaling.
In
Figs.~\ref{fig:meaning.ousiometer9010_VAD012_1_supp1}--\ref{fig:meaning.ousiometer9010_PDS012_3_supp1}
in the 
\suppmaterial,
we provide the
corresponding
$\Mvalence$-$\Marousal$
and
$\Marousal$-$\Mdominance$
large-scale ousiograms.
For our main analysis,
we present smaller versions of these ousiograms in
Figs.~\ref{fig:meaning.ousiometer9310_lexicon001}A--C,
which we discuss below in 
Sec.~\ref{subsec:meaning.VADfail}.

We first briefly describe the ousiogram 
in Fig.~\ref{fig:meaning.ousiometer9010_VAD012_2_noname}
(see the Figure's caption for more detail),
and then contend
with the non-orthogonality of the VAD framework.

As a guide,
we label the cardinal directions
for valence $\Mvalence$
and
dominance $\Mdominance$
by the standard (if problematic) bipolar adjectival pairs
anchoring the semantic differentials
\semdiff{negative}{positive}
and
\semdiff{submissive}{dominant}.
The intercardinal directions are then
combinations of these adjectives
(e.g., submissive-positive).
We label all other ousiograms in the same fashion
with appropriate bipolar adjectival pairs.

Given that we have shifted the VAD scores to lie in [-$\onehalf$,+$\onehalf$],
the two dimensional histogram of Fig.~\ref{fig:meaning.ousiometer9010_VAD012_2_noname}
shows that the NRC VAD lexicon 
accesses much of the available
$\Mvalence$-$\Mdominance$ plane.
The marginal distributions at the top and right show that
both valence and dominance are well dispersed,
with dominance exhibiting some minor asymmetry.
The dark triangles indicate the medians for each marginal.

We show words using two kinds of annotations:
At the extremes of the histogram's boundary
and 
internally along the cardinal and intercardinal axes.
(See Ref.~\cite{warriner2013a} for scatter plots with perimeter annotations.)
For words on the boundary, we automatically
construct and segment a convex hull for the histogram,
determine normals to each segment, and annotate the closest word.
Internally, we find words closest to points along the eight outgoing lines.
We leave the third dimension aside (here, arousal $\Marousal$).
Both the bin width for the underlying histogram and the spacing of
annotations are tunable, and we avoid annotating a word more than once.

Ousiograms will have two main benefits for us.
First, they give us a way to check that words line up with prescribed axes.
Second, and crucially for our later work here, when we move to a potential new framework,
ousiograms will help us to interpret the underlying axes.

In the first sense of acting as a check,
the ousiogram of Fig.~\ref{fig:meaning.ousiometer9010_VAD012_2_noname}
shows that word ratings performed
by the survey participants in Ref~\cite{mohammad2018b}
are reasonably sensible.
Travelling around the histogram's boundary,
we see how the essential meaning of
words incrementally changes.
Starting in the `dominant-positive' direction (upper right),
we see `triumphant', 'success', and `greatness'.
As we move clockwise going down the right side of the boundary,
the annotated words become softer while remaining pleasant: `generous' to `memories' to `pajama'.
Moving left along the bottom boundary, positive gives way to negative, and
we reach the extreme of negative-submissive:
`feather' to `weakened' to `pointless'.
Moving up the left side, we see a string of negative words which grow
in strength, partly because of the scope of dominance:
`depressed',
`nightmare',
`murderous',
`dictator'.
Returning across the top of the ousiogram,
we move through
martial, leadership, and power terms
that gradually lessen in violence:
`weaponry',
`dominate',
`president',
`powerful',
and
back to
`success'.

Internally, each of the eight directions leading out from the center
also reflect changes in the strength of essential meaning.
For the full negative-submissive to dominant-positive axis, for example,
we track from 
`penniless',
`vomiting',
`disoriented',
and
`crutch,'
up through to
`conscientious',
`qualifying',
`amazingly',
and
`success'.

The words `encrypt' and `albatross' are neutral in
the $\Mvalence$-$\Mdominance$ plane, and are worth reflecting on.
These are certainly meaningful words.
And as for all words, these examples could take on
a strong meaning in the right context.
An albatross for sailors
is a dire omen whereas an albatross in golf is a rare, extraordinary success.
But raters are asked to compare the essential meaning of
words based on the perceived
meaning in isolation, which is to say, in the context of
the rater's knowledge of the word.

\subsection{The Valence-Arousal-Dominance (VAD) framework is not orthogonal}
\label{subsec:meaning.DVousiogram}

We turn now to the issue of orthogonality,
a longstanding point of contention for the EPA and VAD frameworks~\cite{osgood1957a,mehrabian1974a,mehrabian1974b,russell1980a,warriner2013a,mohammad2018b}.
For the NRC VAD lexicon, 
we find that the VAD dimensions as interpreted by raters
are not close to being orthogonal.
We observe that standard correlation coefficients
for the three pairs of VAD variables are
\begin{gather}
  r_{\Mvalence,\Marousal}\simeq -0.27,
  \
  r_{\Marousal,\Mdominance}\simeq 0.30,
  \
  \mbox{and}
  \
  r_{\Mvalence,\Mdominance}\simeq 0.49,
  \label{eq:meaning.correlations}
\end{gather}
where the corresponding $p$-values are computed to be essentially 0.
If the VAD framework were orthogonal,
these three correlation coefficients should
be statistically indistinguishable from 0.

We note that the linkages between the VAD dimensions are not simple,
with valence and arousal being anticorrelated with
the other two pairs being positively correlated.

\begin{figure*}[tp!]

  %% build with figousiometer9310_lexicon012
  
  \includegraphics[width=\textwidth]{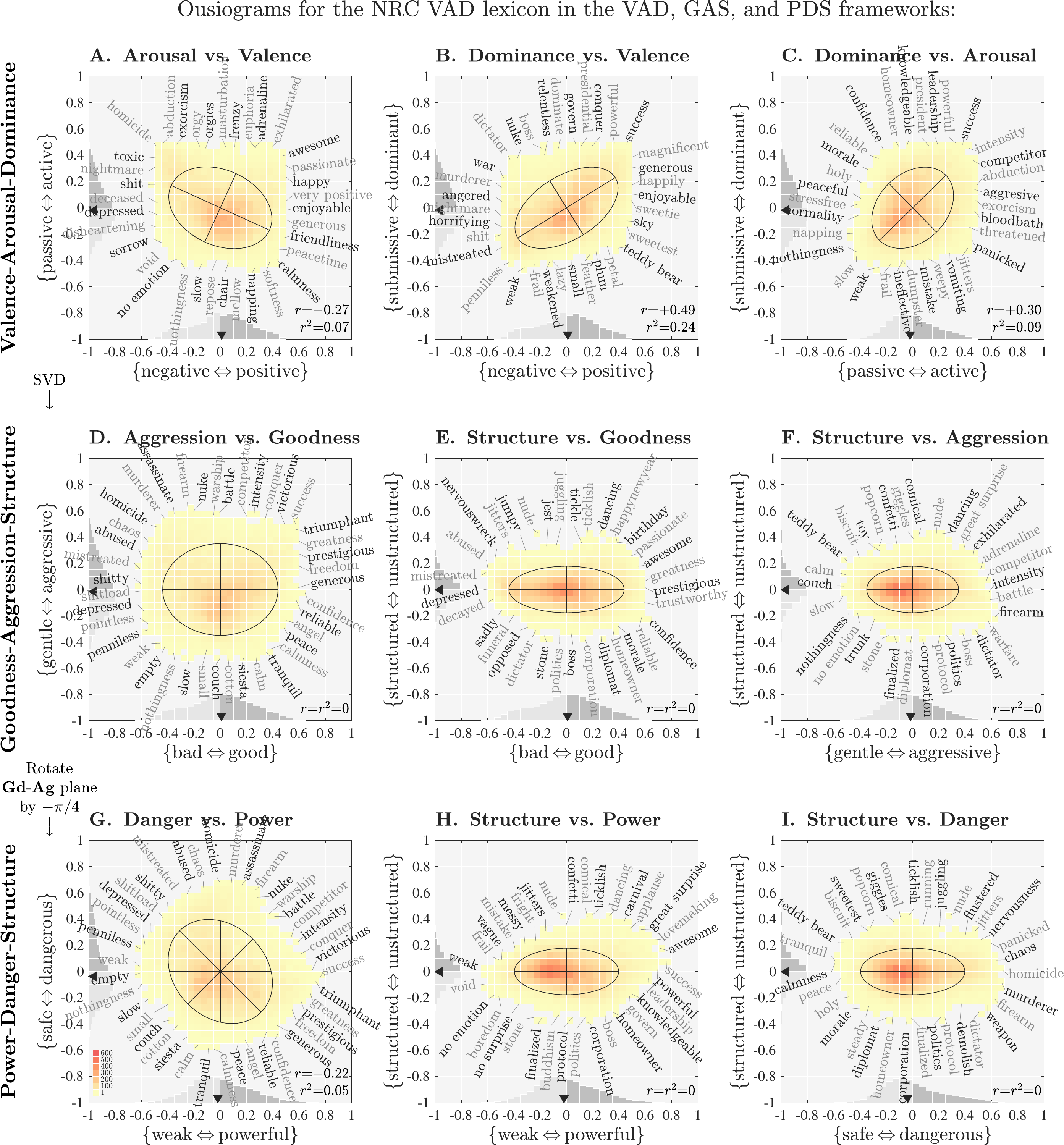}

  \caption{
    \raggedright
    Ousiograms showing the analytic sequence moving
    from the 
    valence-arousal-dominance
    (VAD) framework (top row)
    to the
    goodness-aggression-structure
    (GAS)
    and 
    power-danger-structure
    (PDS)
    frameworks (second and third rows).
    \textbf{Row 1, panels A, B, and C:}
    Ousiograms for the three pairs of variables
    $\Mvalence$, $\Marousal$, and $\Mdominance$
    for $\sim$ 20,000 words in the VAD NRC lexicon~\cite{mohammad2018b}
    (panel B corresponds to Fig.~\ref{fig:meaning.ousiometer9010_VAD012_2_noname}).
    We determine the ellipses
    by using singular value decomposition (SVD) in each plane, ignoring the third dimension.
    The ill fit of the
    VAD
    framework is apparent for the misalignments of ellipse axes.
    Word annotations along the edges of the nine pairwise distributions,
    coupled with ranked words lists by component size
    (Figs.~\ref{fig:meaning.ousiometrics_opposite_cubes001_001}
    to~\ref{fig:meaning.ousiometrics_opposite_cubes001_005}),
    enable interpretation of the new frameworks of GAS and PDS.
    \textbf{Row 2, panels D, E, and F:}
    We perform SVD on the full matrix
    formed by the
    $\Mvalence$, $\Marousal$, and $\Mdominance$ scores,
    and identify
    goodness
    $\Mgoodness$, 
    aggression
    $\Maggression$, 
    and    
    structure
    $\Mstructure$,
    with the first two dimensions accounting for over 90\% of explained variance.
    \textbf{Row 3, panels G, H, and I:}
    Rotating the goodness-aggression plane by $-\pi/4$,
    we uncover a framework with
    \semdiff{weak}{powerful} and
    \semdiff{safe}{dangerous}.
    The GA and PD dimensions interlink to form an
    interpretable circumplex model GPADS.
    See Fig.~\ref{fig:meaning.ousiometer9010_PDS012_1}
    for a larger, more detailed
    power-danger ousiogram.
    As any lexicon reflects only the possible but not the used language (types versus tokens),
    whether or not
    the VAD, GAS, or PDS frameworks are sensible
    must be tested by considering real corpora.
    See Sec.~\ref{subsec:meaning.VADfail}
    and
    Eqs.~\ref{eq:meaning.VAD-to-GAS}
    and~\ref{eq:meaning.transformVAD-to-PDS}
    for interpretation
    of the VAD, GAS, and PDS relationships.
    }
  \label{fig:meaning.ousiometer9310_lexicon001}
\end{figure*}

For a visual guide, and one that we will use throughout the paper,
the ellipse in Fig.~\ref{fig:meaning.ousiometer9010_VAD012_2_noname}
represents the coordinate system uncovered by singular value decomposition (SVD)~\cite{strang2009a}
in the
$\Mvalence$-$\Mdominance$ plane
(we ignore $\Marousal$ for this example calculation).
The ellipse is clearly off axis.
For the equivalent ellipses for the
$\Mvalence$-$\Marousal$
and
$\Marousal$-$\Mdominance$ planes,
see the ousiograms in
Figs.~\ref{fig:meaning.ousiometer9310_lexicon001}A and C
as well as in \suppmaterial.

Now, given that we do not see orthogonality for the VAD framework for
the largest lexicon ever studied coupled with a markedly improved rating system,
we are compelled to investigate why VAD (equivalently EPA)
fails as an orthogonal framework and what alternate framework
might be revealed in doing so.

The root cause of confusion lies in the difficulty
of ascribing stable and meaningful end-point descriptors
for VAD (and EPA) variables.
As was true for Osgood \etal's work that led to the EPA framework~\cite{osgood1957a},
from the start in developing the VAD framework~\cite{mehrabian1974a,mehrabian1974b},
Mehrabian and Russell were concerned with both orthogonality and finding
suitable end-point descriptors.
As explored in Ref.~\cite{bakker2014a},
researchers have continued to use a varying array of end-point descriptors for EPA and VAD,
including the same researchers
over time~\cite{russell1980a,bradley1999a,mohammad2018b}.

Problematically, and as we noted in the introduction,
some end-point descriptors have the effect of correlating different dimensions.
For example, in the ANEW study of Ref.~\cite{bradley1999a}, the negative valence
end-point was presented to participants
as a state of feeling
``completely unhappy, annoyed, unsatisfied, melancholic, despaired, or bored.''
The last descriptor `bored' evidently would be elicited at the low end-point
of the arousal dimension which itself was framed as
``completely relaxed, calm, sluggish, dull, sleepy, or unaroused.''

For the NRC VAD lexicon we study here~\cite{mohammad2018b},
the end-points were described by 6 or 7 words or phrases,
unavoidably broadening them away from being sharply defined
(Tab.~\ref{tab:meaning.VADlimits}).
For example, the words `happiness' and `hopefulness' are used for high valence,
`unhappiness' and `despair' for low valence,
`activeness' and `frenzy' for high arousal,
and
`relaxation' and `sluggishness' for low arousal
(see Tab.~\ref{tab:meaning.VADlimits} for all descriptors).
There is a gap in meaning between all of these pairs of words,
and how participants might perform at rating or ranking words
is not a priori clear.

A further complication is that the names of the dimensions themselves do not track
well within the VAD framework.
While not strictly necessary that they do so, if the name of dimension is a
word with a common meaning, then raters may be guided
away from an intended direction in meaning space.
For example, the word `arousal' is itself high on arousal
($\Marousal$=0.44)
but also registers on the valence and dominance dimensions
$\Mvalence$=0.29,
$\Mdominance$=0.23.
And while the word `dominance' scores strongly in dominance
and neutrally for valence,
it does pick up in the arousal dimension with
($\Mvalence$,$\Marousal$,$\Mdominance$) = (0.04,0.28,0.34).
By contrast, `valence' is sufficiently rare---it is not part of the NRC VAD lexicon---that
it does not color how it is defined for the measurement of emotion.
We are of course not suggesting that there is a simple solution to such ousiometric
nomenclature issues---we are after all using words to define words
as well as kinds of meanings of words.
%% We discuss related mismatches between common and ousiometric meaning
%% later in Sec.~\ref{sec:meaning.synousionyms}
%% in the context of what we will call synousionyms and antousionyms.

While we have critiqued how end-point descriptors have been used,
we are not saying such an approach is invalid.
Rather, we point out that:
1. The EPA dimensions were originally
outputs of relatively small studies involving numerous semantic differentials,
and
2. The attempt to then
make these dimensions controlled inputs to new studies is an entirely different exercise.

In sum, the NRC VAD lexicon, the output of Ref.~\cite{mohammad2018b}'s study,
does not align with the VAD framework, even
though the VAD framework was the intended input.

To move forward, we observe that for any  essence-of-meaning study,
if participants are guided by some 
well constructed set of end-point descriptors,
then we can always
compare and re-consider how well these descriptors perform.
Moreover, we must allow that a distinct framework may emerge over time
as far larger and more sophisticated studies are carried out.
We are effectively maintaining the approach of the founding experiments,
allowing the outcomes to remain informative and be potentially corrective.

%%%%%%%%%%%%%%%%%%%%%%%%%%%%%%%%%%%%%%%%%%%%%%%%%%%%%%%%%%%%%%
%% second figure: Analysis of lexicon
%% VAD -> GAS -> PDS
%%%%%%%%%%%%%%%%%%%%%%%%%%%%%%%%%%%%%%%%%%%%%%%%%%%%%%%%%%%%%%

\begin{revisionbar}
  
\subsection{Additional analytic processes for the level of types}
\label{subsec:meaning.extra-analysis}

For the following sections,
we complement our analysis of the results of SVD
with four pieces in the \suppmaterial:
\begin{itemize}
\item
  Sec.~\ref{sec:meaning.appendix-ousiogram-three-frameworks}:
  Large-scale ousiograms in
  Figs.~\ref{fig:meaning.ousiometer9010_VAD012_1_supp1}--\ref{fig:meaning.ousiometer9010_PDS012_3_supp1}.
\item
  Sec.~\ref{sec:meaning.synousionyms}:
  Explorations of synousionyms and antousionyms, words that are similar or opposite in terms
  of essential meaning.
\item
  Sec.~\ref{sec:meaning.MRIs}:
  An `MRI' of meaning space rendered as a flipbook
  in Figs.~\ref{fig.meaning:figousiometer_3d_MRI_slices100_01_noname}--\ref{fig.meaning:figousiometer_3d_MRI_slices100_19_noname}.
\item
  Sec.~\ref{sec:meaning.appendix-cube-tables}:
  Lists of the top 20 words 
  ranked by component size along the 13 axes of a 3$\times$3 cube that is
  aligned with the SVD dimensions.
  See
  Figs.~\ref{fig:meaning.ousiometrics_opposite_cubes001_001}--\ref{fig:meaning.ousiometrics_opposite_cubes001_013}.
  Each axis is a semantic differential, making for 26 end point cubes
  (the neutral center cube brings the total to 27=3$\times$3).
  Words are ordered by component size with the restriction that
  their vector representations lie within a cone of half angle
  $
  \frac{1}{2}
  \frac{180}{\pi}
  \cos^{-1}(2/\sqrt{6})
  \simeq
  17.6\degree
  $
  around a given axis.
  We present the full cube below in Sec.~\ref{subsec:meaning.GPADS}
  and Fig.~\ref{fig:meaning.ousiometric_cube3d002}.
\end{itemize}

\end{revisionbar}

\subsection{Assessing the failure of the Valence-Arousal-Dominance (VAD) framework}
\label{subsec:meaning.VADfail}

In the present and following two sections,
we show how the NRC VAD lexicon
affords two possible alternate and mutually consistent frameworks:
Goodness-Aggression-Structure (GAS)
and Power-Danger-Structure (PDS).
The steps of our analysis are represented by the rows of
Fig.~\ref{fig:meaning.ousiometer9310_lexicon001},
which we explain as follows.

We first note that for the NRC VAD lexicon,
the overall contributions to variance explained by the three VAD dimensions of meaning are
approximately
44.4\%,
28.0\%,
and
27.6\%.
Valence is clearly the leading dimension
with arousal and dominance balanced.

To determine the uncorrelated orthogonal dimensions
for the NRC VAD lexicon,
we perform singular value decomposition (SVD) on
the 3 by 20,006 matrix
$\mathbb{A}$
of average VAD scores
($\mathbb{A}
=
\mathbb{U}
\mathbb{\Sigma}
\mathbb{V}^{\textnormal{T}}$).
We find singular values
$\sigma_{1} \simeq 34.1$,
$\sigma_{2} \simeq 27.2$,
and
$\sigma_{3} \simeq 13.8$,
which correspond to explained variances of
55.6\%,
35.3\%,
and
9.1\%.
The first two dimensions now explain 90.9\% as opposed
to 72.4\% explained by $\Mvalence$ and $\Marousal$.

The point cloud of VAD scores is thus a non-axis-aligned
ellipsoid, strongly flattened in one dimension.
In Fig.~\ref{fig:meaning.ousiometer9310_lexicon001},
the first row of ousiograms show projected histograms of the
ellipsoid in VAD space for each pair of dimensions
(Fig.~\ref{fig:meaning.ousiometer9310_lexicon001}B
corresponds to Fig.~\ref{fig:meaning.ousiometer9010_VAD012_2_noname}).
The SVD ellipses in all three projections demonstrate
the correlations in \Req{eq:meaning.correlations} above.

As for all ousiograms, the word annotations help us understand
how raters have responded to the end-point descriptors.
Here, these annotations may be diagnostic (VAD) or illuminating (GAS and PDS, below).
For VAD, we have already considered
$\Mvalence$-$\Mdominance$ ousiogram's annotation
(Fig.~\ref{fig:meaning.ousiometer9310_lexicon001}B),
finding them to be sensible,
and
we see that annotations for the other dimension pairs
are similarly interpretable within the VAD framework
(Figs.~\ref{fig:meaning.ousiometer9310_lexicon001}A and C).

\begin{revisionbar}
\subsection{The Goodness-Aggression-Structure (GAS) framework}
\label{subsec:meaning.GAS}
\end{revisionbar}

Moving to the middle row
of panels
(Figs.~\ref{fig:meaning.ousiometer9310_lexicon001}D--F),
we show ousiograms for word scores represented by the orthogonal basis determined by SVD
acting on the VAD word scores.
By construction, all three SVD ellipses are now aligned with the underlying axes.

Upon considering the annotated words, along
with the ranked word lists in
Figs.~\ref{fig:meaning.ousiometrics_opposite_cubes001_002},
\ref{fig:meaning.ousiometrics_opposite_cubes001_004},
and
\ref{fig:meaning.ousiometrics_opposite_cubes001_005},
we interpret these three new
essence-of-meaning dimensions
to be
goodness $\Mgoodness$,
aggression $\Maggression$,
and
structure $\Mstructure$.
(For annotations internal to each histogram,
see the larger ousiograms in \suppmaterial.)

To arrive at the goodness dimension, we look to words on the left and right side
of the ousiogram in Fig.~\ref{fig:meaning.ousiometer9310_lexicon001}D.
On the left,
we see
`shitty',
`penniless',
`mistreated',
and
`abused';
and on the right,
`reliable',
`confidence',
`freedom',
and
`triumphant'.

Words at the bottom and top of the same ousiogram
in Fig.~\ref{fig:meaning.ousiometer9310_lexicon001}D
are connected in essential meaning by their
signifying of low and high aggression:
`slow',
`couch',
`siesta',
and
`calm',
versus
`assassinate',
`battle',
`competitor',
and
`conquer'.

Finally, we distill the vertical dimension in the ousiograms of
Figs.~\ref{fig:meaning.ousiometer9310_lexicon001}E
and~\ref{fig:meaning.ousiometer9310_lexicon001}F
as structure.
We choose the alignment of the third dimension
to be
\semdiff{structured}{unstructured},
moving upwards.
At the bottom of these ousiograms,
we have words connoting organization, rigidity, and systematic form:
`stone',
`protocol',
`corporation',
`dictator',
and
`diplomat'.
At the top, we see terms that convey
lack of structure:
`jest',
`confetti',
`dancing',
`popcorn',
and
`great surprise'.
To support the choice of orientation for the structure axis,
we make a thermodynamic analogy where
rigid organization is akin to a zero temperature frozen state,
and a growing lack of structure corresponds to increasing temperature.
We also see that
\semdiff{serious}{playful}
and
\semdiff{predictable}{unpredictable}
differentials
are subsets of the more general
\semdiff{structured}{unstructured}
differential.
Broadly speaking,
we view the third dimension of essential meaning \semdiff{structured}{unstructured}
as encoding evolvability.

For purposes of clarity of argument,
we have sought to choose valid but distinct names for the three dimensions in GAS
to distinguish them from VAD (or EPA).
We acknowledge that valence, evaluation, and goodness are conceptually similar
as are activity, arousal, and aggression.
And as we discuss below, in the realm of emotion,
valence is often taken to be analogous to a
\semdiff{happiness}{sadness}
dimension~\cite{bradley1999a,dodds2009b}.

The linear transformation between VAD and GAS obtained from SVD
is:
%% \begin{align}
%%   &
%%   \colvec{
%%     \goodnesssymbol \\
%%     \aggressionsymbol \\
%%     \structuresymbol
%%   }
%%   \simeq
%%   \nonumber
%%   \\
%%   &
%%   \qquad
%%   \mymatrix{ccc}{
%%     +0.86 & -0.15 & +0.48 \\
%%     -0.16 & +0.83 & +0.54 \\
%%     +0.48 & +0.55 & -0.69 
%%   }
%%   \colvec{
%%     \valencesymbol \\
%%     \arousalsymbol \\
%%     \dominancesymbol
%%   }.
%%   \label{eq:meaning.VAD-to-GAS}
%% \end{align}
\begin{equation}
  \hspace*{-20pt}
  \colvec{
    \Mgoodness \\
    \Maggression \\
    \Mstructure
  }
  \simeq
  \mymatrix{ccc}{
    +0.86 & -0.15 & +0.48 \\
    -0.16 & +0.83 & +0.54 \\
    +0.48 & +0.55 & -0.69 
  }
  \colvec{
    \Mvalence \\
    \Marousal \\
    \Mdominance
  }.
  \label{eq:meaning.VAD-to-GAS}
\end{equation}

In moving to the GAS framework,
we have goodness most connected with valence ($+$0.86)
and dominance ($+$0.48), with a minor negative linkage to arousal ($-$0.15).
Aggression is most connected with arousal ($+$0.83) and also, like
goodness, with dominance ($+$0.54), but is somewhat at odds with valence ($-$0.16).
And what we have identified as an increasing lack of structure
corresponds roughly equally to increases in valence and arousal ($+$0.48 and $+$0.55)
while increasing dominance points in the direction of more structure ($-$0.69).

We can now see that what separates the GAS framework
from the VAD framework (or EPA framework)
is that 
the dominance (or potency) dimension
lies within the goodness-aggression plane.
That is, the three conceptual dimensions of VAD
are in fact collapsed into the two dimensions of goodness and aggression,
with a new third and less important dimension of structure being revealed.

When dominance is near zero,
\Req{eq:meaning.VAD-to-GAS} shows that
goodness and aggression approximate valence and arousal.
However, the correlations between valence and dominance as well
as arousal and dominance mean that dominance increasing in magnitude
will move goodness-aggression away from valence-arousal.

Returning to the ousiogram in Fig.~\ref{fig:meaning.ousiometer9310_lexicon001}D,
we see that the four intercardinal axes carry distinguishable essential meanings, 
interpolating between the
\semdiff{good}{bad}
and
\semdiff{high-aggression}{low-aggression}
axes.

The diagonal axis 
running from
`weak' and `empty' to
`success' and `triumphant'
is a
\semdiff{weak}{powerful}
axis,
while the orthogonal diagonal axis traveling from
`calmness' and `peace' to `murderer' and `homicide'
is, we argue, a
\semdiff{safe}{dangerous}
axis.

%% $\weaknesssymbol$ $\propto$ - $\goodnesssymbol$ - $\aggressionsymbol$,
%% and
%% $\safetysymbol$ $\propto$ + $\goodnesssymbol$ - $\aggressionsymbol$.

\begin{figure*}[tp!]

  %% create with:
  %% figousiometer9010_lexicon002.m
  
  \centering
  \includegraphics[width=\textwidth]{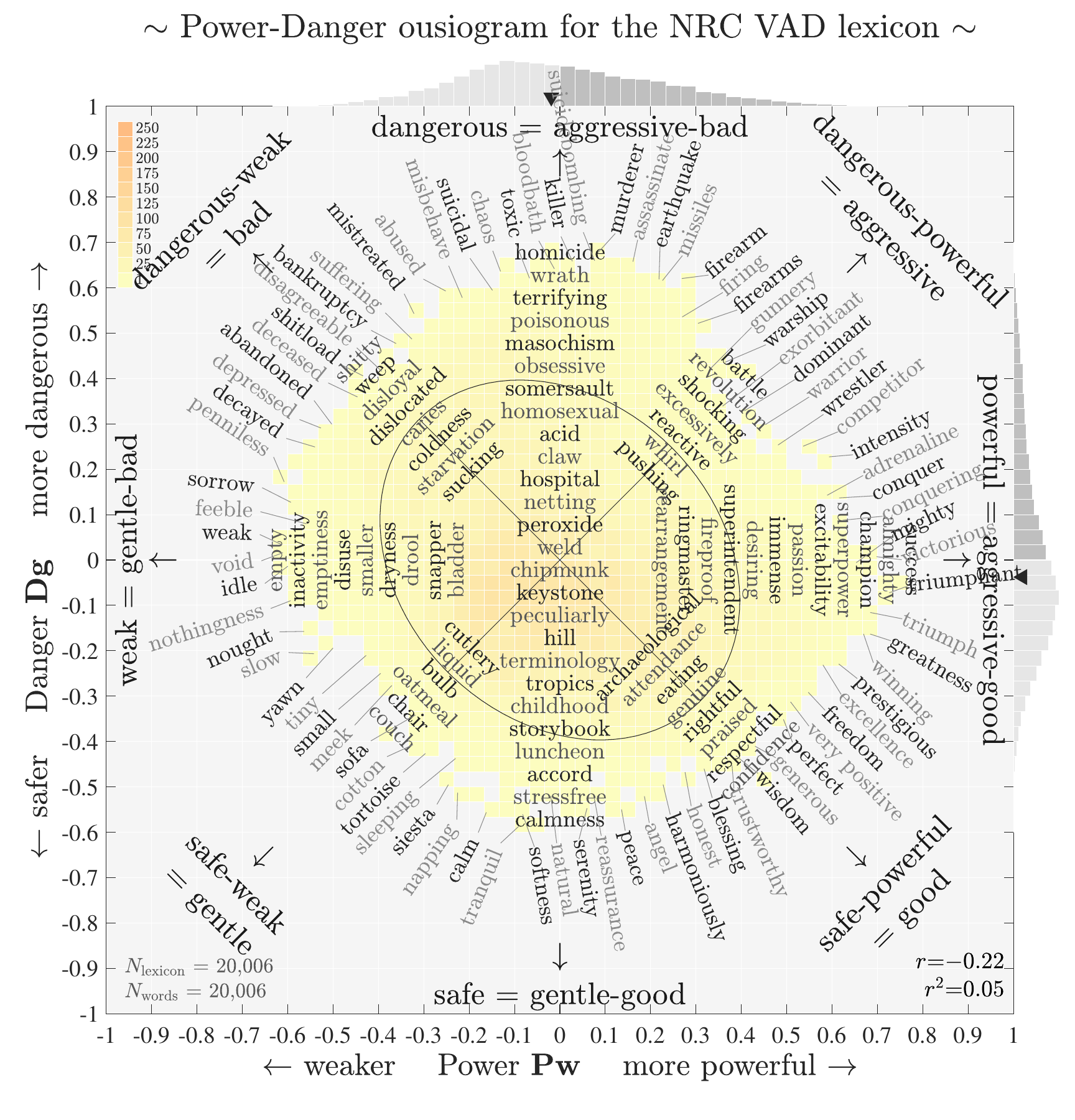}
  \caption{
    \raggedright
    Power-danger ousiogram for the NRC VAD lexicon~\cite{mohammad2018b}
    for the PDS framework,
    an expanded version of
    Fig.~\ref{fig:meaning.ousiometer9310_lexicon001}G
    with internal annotations.
    The diagonal endpoints match the axis endpoints for the GAS framework:
    safe-powerful $\sim$ good,
    dangerous-weak $\sim$ bad,
    powerful-dangerous $\sim$ aggressive,
    and
    safe-weak $\sim$ gentle.
    PDS and GAS interpolate between each other in
    the primary plane
    which can be viewed as a kind of circumplex model, GPADS~\cite{russell1980a},
    which can be viewed as a cube model (see Fig.~\ref{fig:meaning.ousiometric_cube3d002}).
    Both power and danger reach further
    into positive values than negative with
    $-0.612 \le \Mpower \le  0.758$
    and
    $-0.591 \le \Mdanger \le 0.681$.
    The modes and the medians
    indicate a slight safe-weak
    tendency for the meanings of words in the NRC VAD lexicon
    (medians: -0.019 and -0.038, dark triangles),
    which is cautioned as an observation preliminary to later measurements
    where, in accounting for frequency of usage,
    we find a bias towards safety in real corpora
    (see Figs.~\ref{fig:meaning.ousiometer9310_googlebooks001}
    and
    \ref{fig:meaning.ousiometer9400_PDS_tableau004}).
    In the \suppmaterial, we provide larger ousiograms with internal labels
    for all corpora and all three frameworks
    (for the NRC VAD lexicon examples, we use the same color map across
    all figures).\\
    ~\\
    ~\\
  }
  \label{fig:meaning.ousiometer9010_PDS012_1}
\end{figure*}

\subsection{The Power-Danger-Structure (PDS) framework}
\label{subsec:meaning.PDS}

For reasons we explain below,
we are drawn to consider
\semdiff{weak}{powerful}
and
\semdiff{safe}{dangerous}
as an alternate
essence-of-meaning axes,
which we achieve in the third row of ousiograms
in Fig.~\ref{fig:meaning.ousiometer9310_lexicon001}
by a simple clockwise rotation of the
$\Mgoodness$-$\Maggression$ plane by $-\pi/4$.
We call this rotation of the GAS framework
the Power-Danger-Structure (PDS) framework,
and we will then also consider a combined framework: GPADS.

At this stage, we do not view either GAS or PDS to be correct
but rather complementary, interrelated frameworks.
Even though we of course only need two basis vectors
to describe two dimensions, viewing word scores
in the combined GPADS framework is valuable.

Expressed as a simple linear transformation,
we have
\begin{equation}
  \colvec{
    \Mpower \\
    \Mdanger \\
    \Mstructure
  }
  =
  \frac{1}{\sqrt{2}}
  \mymatrix{ccc}{
    1 & 1 & 0 \\
    -1 & 1 & 0 \\
    0 & 0 & 1 \\
  }
  \colvec{
    \Mgoodness \\
    \Maggression\\
    \Mstructure
  }.
  \label{eq:meaning.transformGA-to-PD}
\end{equation}
We supply a more detailed power-danger ousiogram
in Fig.~\ref{fig:meaning.ousiometer9010_PDS012_1}.
When later considering large-scale corpora, we will see
that the PDS framework
rather than GAS conforms to real word usage (tokens instead of types).
But we first must explore its characteristics for
the unamplified NRC VAD lexicon.

In the PDS framework, the variance explained is now evenly divided
between power and danger (45.5\% each)
while structure's contribution remains the same.
Further, the words with the largest magnitude are
now aligned with the positive axes of power and danger.
For the largest overall magnitude word `success',
the PDS coordinates 
are
$(\Mpower,\Mdanger,\Mstructure)$=$(0.76,-0.05,0.10)$.
For `murderer',
$(\Mpower,\Mdanger,\Mstructure)$=$(0.08,0.68,-0.09)$.
Of the top 20 words by vector magnitude,
12 are strongly aligned with the power direction
and 8 with the danger direction
(see Figs.~\ref{fig:meaning.ousiometrics_opposite_cubes001_001}
and
\ref{fig:meaning.ousiometrics_opposite_cubes001_003}).
However, we suffer one drawback as we have
reintroduced a non-zero correlation, $r$=$-0.22$,
as indicated by the rotated ellipse in
Fig.~\ref{fig:meaning.ousiometer9310_lexicon001}G
and Fig.~\ref{fig:meaning.ousiometer9010_PDS012_1}.

The rotated and internal annotations
in the power-danger ousiogram
in Fig.~\ref{fig:meaning.ousiometer9010_PDS012_1},
are now in line with our interpretation
of the two axes being
\semdiff{weak}{powerful}
and
\semdiff{safe}{dangerous}.
The horizontal axis, for example, runs from
`void',
`nothingness',
and
`empty'
to
`powerful',
`success',
and
`almighty'.
We find high danger in `earthquake',
`suicidebombing',
and
`toxic',
and
safety in
`serenity',
`softness',
and
`tranquil'.

As for the valence-dominance ousiogram
in Fig.~\ref{fig:meaning.ousiometer9010_VAD012_2_noname},
traveling around the boundary of the power-danger ousiogram
loops us through an ousiometrically sensible sequence of terms.
Moving upwards and around from `triumphant', words take
on increasingly violent connotations,
while moving down, success begins to ebb while peaceful aspects build.

Crucially, and as we have described,
the GAS and PDS frameworks form GPADS, a kind of circumplex model.
We list the four axes in the primary plane
in order from danger at the top of the PD plane,
moving clockwise by $\pi/4$ through aggression, power, and goodness.

Each of the four directions in GA and PD are mutually intelligible
by their adjacent directions in the alternate framework.
For example,
powerful is aggressive-good,
dangerous is aggressive-bad,
good is safe-powerful (`wisdom' and `generous'),
bad is dangerous-weak (`deceased' and `bankruptcy'),
and
gentle is weak-safe.

Combining SVD and the $-\pi/4$ rotation,
we have the linear transformation connecting the VAD and PDS frameworks:
\begin{equation}
  \hspace*{0mm}
  \colvec{
    \Mpower \\
    \Mdanger \\
    \Mstructure
  }
  \simeq
  \mymatrix{ccc}{
    +0.50 & +0.48 & +0.72 \\
    -0.72 & +0.69 & +0.04 \\
    +0.48 & +0.55 & -0.69 
  }
  \colvec{
    \Mvalence \\
    \Marousal \\
    \Mdominance
  }.
  \label{eq:meaning.transformVAD-to-PDS}
\end{equation}
We see that power is roughly a direct sum of
valence, arousal, and dominance (+0.50, +0.48, and +0.72).
Danger is a near equally weighted linear combination of
negative valence and positive arousal ($-$0.72 and +0.69),
and has little connection to dominance (+0.04).
Structure's connection to VAD remains the same
as in \Req{eq:meaning.VAD-to-GAS} since we have rotated a plane
orthogonal to its axis.

We have thus determined that the VAD framework
was effectively interpreted as strongly correlated
by participants in the NRC VAD lexicon study of
Ref.~\cite{mohammad2018b}.
We also now have two complementary frameworks
in GAS and PDS that are potential candidates
for a fundamental minimal representation of essential meaning.

We emphasize that at this stage,
we do not know if the VAD, GAS, or PDS frameworks---or indeed none of them---may be suitable
when confronted with real word usage when we consider tokens instead of types.
What we do have is a circumplex-like model in GPADS
which contains two clear orthogonal frameworks.

\begin{figure*}[t!]

  \includegraphics[width=\columnwidth]{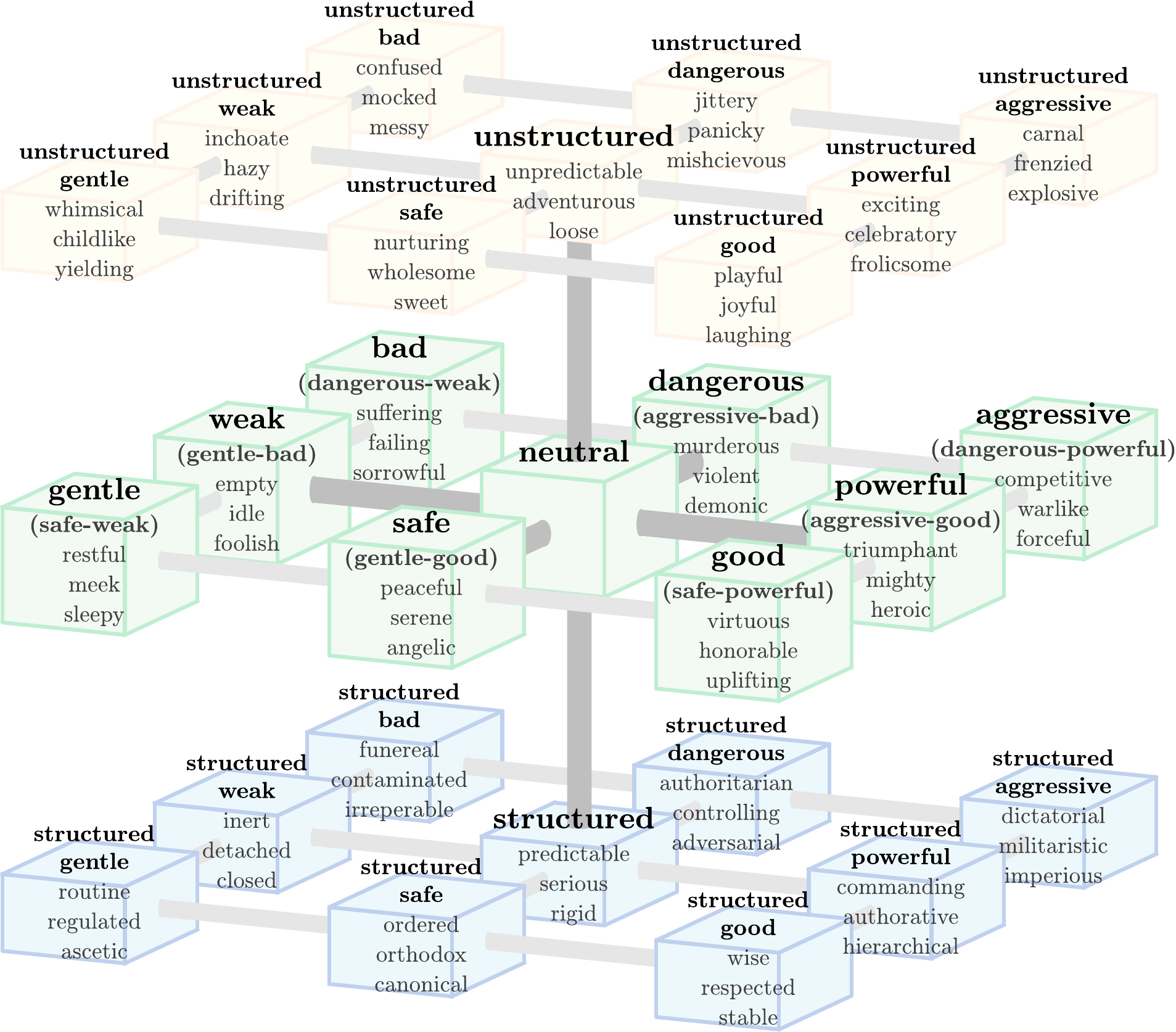}
    \caption{
    \raggedright
      An exploded cube representing the GPADS framework.
    In Figs.~\ref{fig:meaning.ousiometrics_opposite_cubes001_001}--\ref{fig:meaning.ousiometrics_opposite_cubes001_013}
    in the \suppmaterial, we provide complementary lists of 20 words for each end of the 13 cube axes.
    Words are ranked by component size conditioned on sufficient alignment.
    Each of these paired lists indicates a differential with a small version of the exploded cube depicted.
    The example words on each block are a mixture of words on these lists
    and adjectives which capture a category.
    For the cube, blue represents structure (cold, low temperature, rigid) and yellow represents lack of structure (warm, high temperature, loose),
    while block thickness qualitatively indicates variance explained.
  }
  \label{fig:meaning.ousiometric_cube3d002}
\end{figure*}

\begin{revisionbar}
  
\subsection{A cube model of meaning for the Goodness-Power-Aggression-Danger-Structure (GPADS) framework}
\label{subsec:meaning.GPADS}

In Fig.~\ref{fig:meaning.ousiometric_cube3d002},
we combine all of our framework identifications in
a unified GPADS cube model of essential meaning.

The middle primary plane shows the four
interconnected axes of GAS and PDS: GPAD.
In anticipation of our later findings,
we have aligned the cube within the PDS framework,
though it could be readily rotated to align with GAS.

The cube model is
complementary to the map-like ousiograms in
Fig.~\ref{fig:meaning.ousiometer9010_PDS012_1}
and
Figs.~\ref{fig:meaning.ousiometer9010_VAD012_1_supp1}--\ref{fig:meaning.ousiometer9010_PDS012_3_supp1}
and  is also informed by and connected
with lists of ranked words
in Figs.~\ref{fig:meaning.ousiometrics_opposite_cubes001_001}--\ref{fig:meaning.ousiometrics_opposite_cubes001_013}.

%% The sequence of lists in the \suppmaterial\ also functions 
%% Figs.~\ref{fig:meaning.ousiometer9010_VAD012_1_supp1}--\ref{fig:meaning.ousiometer9010_PDS012_3_supp1}
%% as a flipbook.

For the word lists, we show the top 20 words by component size for each
of the 13 cube axes,
9 of which connect the structured and unstructured levels.
Three axes connect face cubes,
six connect edge cubes,
and
four connect corner cubes.
To avoid overlap we restrict words to cones around axes
of half angle
$
\frac{1}{2}
\frac{180}{\pi}
\cos^{-1}(2/\sqrt{6})
\simeq
17.6\degree
$.
For example,
Fig.~\ref{fig:meaning.ousiometrics_opposite_cubes001_011}
shows the top 20 words for unstructured-gentle vs.\ structured-aggressive,
which is the axis connecting the corner cubes in the front top left and back lower right.

For each cube (except the neutral one), we show three adjectives
that capture the words associated with that cube and their
location within the GPADS framework.
The adjectives are in some cases words directly taken from the relevant list
but more usually are general descriptors.

\end{revisionbar}

\begin{figure*}[tp!]
  \centerfloat
  \includegraphics[width=\textwidth]{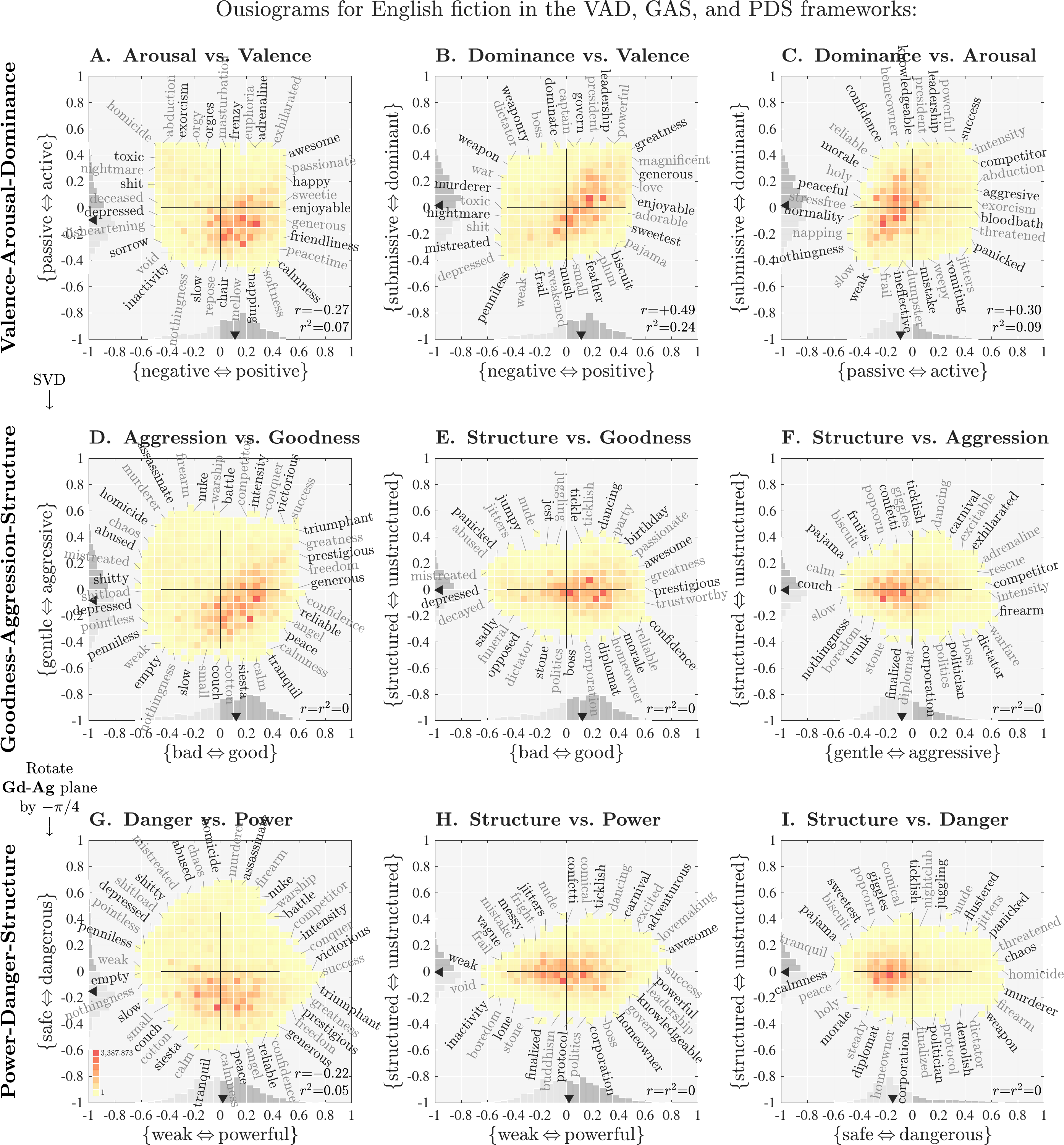}
  \caption{
    \raggedright
    Ousiograms for
    English fiction (1900--2019)
    arranged in the same analytic sequence format as Fig.~\ref{fig:meaning.ousiometer9310_lexicon001}.
    We now allow
    each word's contribution to be its overall frequency of usage within a given corpus.
    We form a single frequency-rank distribution~\cite{zipf1949a}
    for the entire corpus
    by equally weighting each year's frequency-rank distribution~\cite{michel2011a,pechenick2015a}.
    The sequence indicates that:
    1. Overall, the Google Books English fiction corpus is best aligned with the PDS framework,
    and that
    2. Expressed language exhibits a `safety bias', a generalization
    of the Pollyanna principle~\cite{boucher1969a,kloumann2012b,dodds2015a}.
    \textbf{Row 1, panels A, B, and C:}
    In the VAD framework,
    the histograms are clearly misaligned with the main axes.
    \textbf{Row 2, panels D, E, and F:}
    The histograms are again poorly aligned with the main axes of
    $\Mgoodness$, 
    $\Maggression$, 
    and
    $\Mstructure$.
    The marginal distributions for Goodness and Aggression in panel D
    show an apparent `goodness bias' and a `low-aggression bias'.
    The goodness bias is an instantiation
    of the
    Pollyanna principle for language~\cite{boucher1969a,dodds2015a}.
    \textbf{Row 3, panels G, H, and I:}
    Rotation to the power-danger framework
    shows that words used in English fiction conform to a safety bias
    with the preponderance of words falling on the safe side of the power-danger plane (panel G).
    Both the goodness and aggression biases in panel D are revealed to be
    one dimensional projections of an underlying safety bias.
    Words are distributed broadly in the power-structure plane (panel H)
    and are on the safe side of the danger-structure plane (panel I).
  }
  \label{fig:meaning.ousiometer9310_googlebooks001}
\end{figure*}

\begin{figure*}[t!]
  \includegraphics[width=\textwidth]{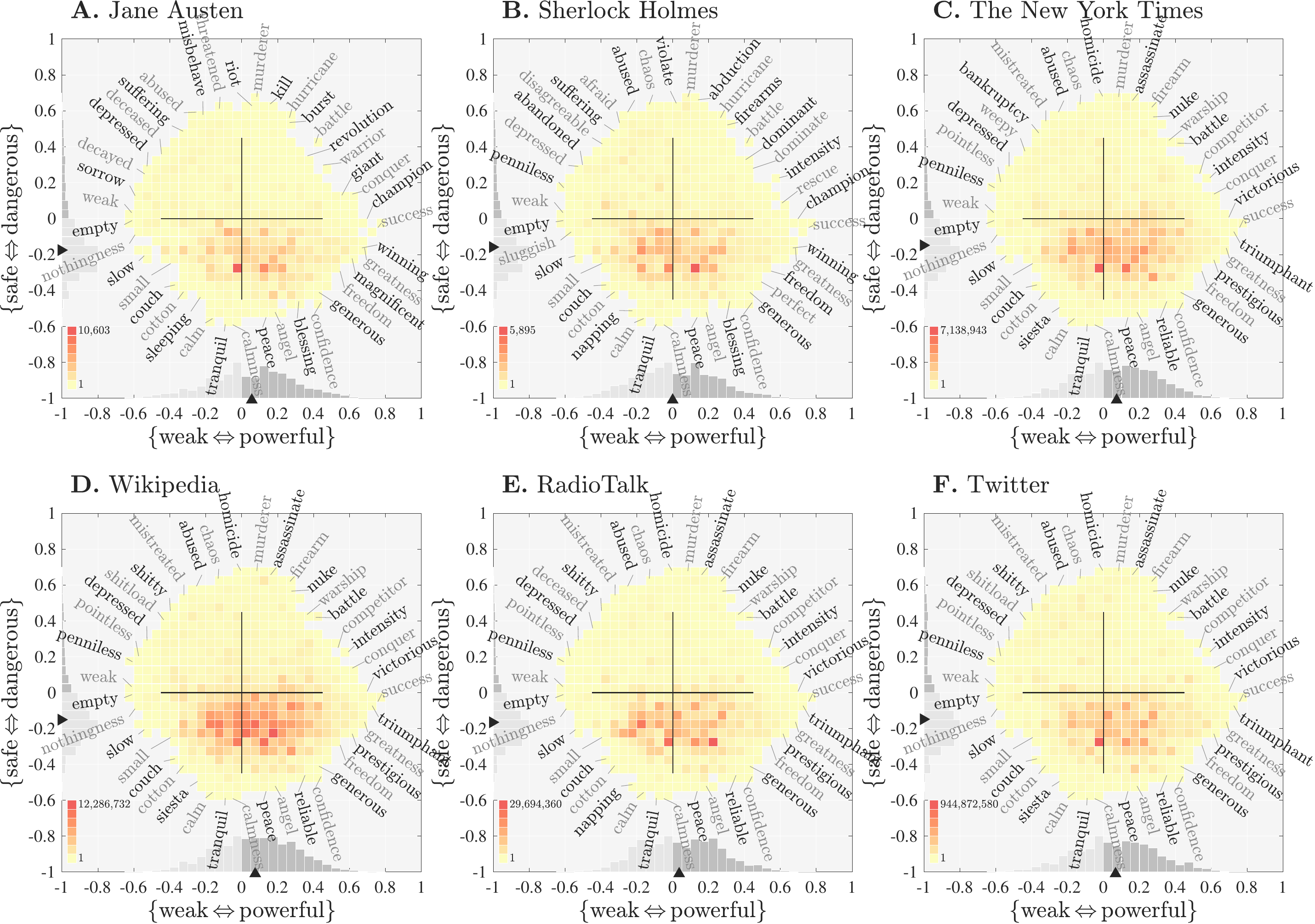}
  \caption{
    \raggedright
    Ousiograms for power-danger space for six corpora of varying type and scale:
    \textbf{A.}
    Jane Austen's novels;
    \textbf{B.}
    Arthur Conan Doyle's Sherlock Holmes novels and short stories;
    \textbf{C.}
    The New York Times (1987--2007)~\cite{nytimescorpus2008a};
    \textbf{D.}
    Wikipedia (March, 2019)~\cite{semenov2019a};
    \textbf{E.}
    Talk radio transcripts
    (2018/10--2019/03)~\cite{beeferman2019a};
    and
    \textbf{F.}
    Twitter (approximately 10\% of all English tweets in 2020,
    with each day weighted equally)~\cite{alshaabi2021b,alshaabi2021c}.
    Words of the six corpora all strongly canvass power-danger space
    with a marked bias towards safe.
    Jane Austen's novels, the New York Times, and Wikipedia are all
    author-side corpora
    in that their frequency-rank distributions do not incorporate popularity of
    books, sections, or entries.
    By contrast, Twitter incorporates a reader-side measure of popularity
    through
    amplification by retweets.
    Each ousiogram's color map is linearly normalized to the highest count bin,
    and the maximum bin count is indicated at the top of each color bar.
    The highest count bin in panels A, C, and F
    is due to the word `be'
    ($\Mpower$=-0.001, $\Mdanger$=-0.300).
    See Sec.~\ref{sec:meaning.data} for description of data sets.
    For the
    six corpora here,
    we provide the full VAD-GAS-PDS analytic sequence
    of Figs.~\ref{fig:meaning.ousiometer9310_lexicon001}
    and~\ref{fig:meaning.ousiometer9310_googlebooks001}
    in
    Figs.~\ref{fig:meaning.ousiometer9310_jane_austen001},
    \ref{fig:meaning.ousiometer9310_sherlock_holmes001},
    \ref{fig:meaning.ousiometer9310_nyt001},
    \ref{fig:meaning.ousiometer9310_wikipedia001},
    \ref{fig:meaning.ousiometer9310_radiotalk001},
    and
    \ref{fig:meaning.ousiometer9310_twitter001}.
    }
    \label{fig:meaning.ousiometer9400_PDS_tableau004}
\end{figure*}

\subsection{The linguistic `safety bias' of disparate large-scale, corpora}
\label{subsec:meaning.safetybias}

Having established the GAS and PDS frameworks as alternatives to VAD,
we turn to real, large-scale corpora.
By intent, we have so far only considered
the essential meaning of words and terms in the NRC VAD lexicon---the level of types.

We now aim to incorporate frequency of usage of words---tokens---for a collection of
well-defined corpora.
We can only do so sensibly within each structured corpus---we cannot meaningfully combine,
for example, the New York Times and Twitter.

For an initial example corpus, we investigate
the ousiometric content of $1$-grams used in English fiction from 1900--2019
per the Google Books project~\cite{michel2011a}.
We note that we have earlier argued and demonstrated 
that the Google Books project generates problematic corpora
in that 1.\ Each book is in principle counted once (popularity is not measured)
and that 2.\ For all English books combined, the corpus is clouded
by a growing preponderance of scientific literature~\cite{pechenick2015a}.
To use the framing of types and tokens for the former point,
the books are themselves types, containing $n$-grams as tokens, but we do
not have the books as tokens by knowing, for example, numbers of copies sold.
Nevertheless, for our purposes here, the relatively-science-free 2019 English fiction corpus
provides a raw large-scale body of text to examine.

%% even while we fully acknowledge that it is does not bear any signatures of social amplification.

We generate ousiograms in the same fashion as before, but
we now weight words by their frequency of usage.
The NRC VAD lexicon acts as a lexical lens on the
frequency-rank distribution---we only take word counts for those words
we have VAD/GAS/PDS scores for.
In Fig.~\ref{fig:meaning.ousiometer9310_googlebooks001},
we reprise the analytic sequence
of Fig.~\ref{fig:meaning.ousiometer9310_lexicon001}
for words used in English fiction.

While the histograms were relatively uniform for the NRC VAD lexicon, 
we now see uneven distributions.
For the VAD row, we see the distributions are not aligned with the underlying
axes of the VAD framework (Figs.~\ref{fig:meaning.ousiometer9310_googlebooks001}A--C).
The main ousiogram for goodness-aggression (Fig.~\ref{fig:meaning.ousiometer9310_googlebooks001}D)
still does not align well, showing an off-axis bias towards goodness and low aggression,
the former being a linguistic signature of the
Pollyanna principle~\cite{boucher1969a,kloumann2012b,dodds2015a}.
We discuss both biases further below.
The goodness-structure and aggression-structure ousiograms
(Figs.~\ref{fig:meaning.ousiometer9310_googlebooks001}E and F)
show biases towards goodness and low aggression that appear more aligned.

\begin{figure*}[t!]

  \centering

  \includegraphics[width=\textwidth]{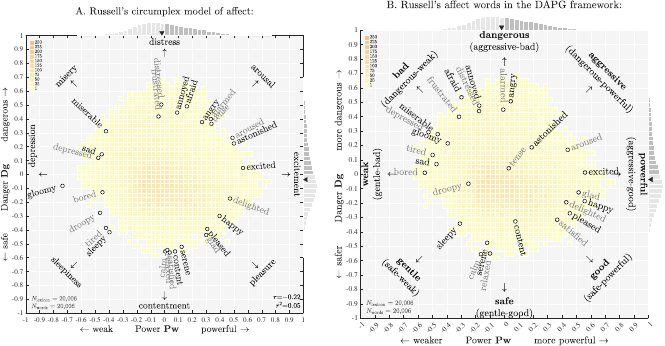}

  \caption{    
    \raggedright
    Comparison of Russell's circumplex model of affect~\cite{russell1980a}
    with the GPAD framework.
    The two frameworks show impressive agreement given
    how differently words were scored
    in Refs.~\cite{russell1980a} and~\cite{mohammad2018b}.
    %%     In the main text, we argue
%%     that the affect concepts of the circumplex model
%%     are better interpreted as states-of-being concepts,
%%     and that
%%     depression, arousal, and excitement
%%     may be revised as
%%     boredom, aggression, and powerful
%%     (see Sec.~\ref{subsec:meaning.circumplex}).
    \textbf{A.}
    Reconstruction of the circumplex model scores for 27 affect words
    for the first survey presented in Fig.~2 in Ref.~\cite{russell1980a}.
    We obtained data for 27 of 28 terms by visual inspection of
    Fig.~2 in~\cite{russell1980a},
    omitting the 2-gram `at ease'.
    To enable comparison,
    we rotate Russell's scores by $-\pi/4$,
    and also underlie both plots with the power-danger histogram
    per Fig.~\ref{fig:meaning.ousiometer9010_PDS012_1}.
    The comparison is nevertheless with the GPAD framework,
    and the PD alignment is for consistency.
    Leaving angles unchanged, we uniformly rescale the magnitude of
    scores for the 27 affect words in the circumplex model
    to give an approximate fit to the
    power-danger scale---only
    angles and relative magnitudes may be sensibly compared.
    \textbf{B.}
    Locations in the power-danger plane for the 27 affect words of Ref.~\cite{russell1980a},
    all of
    which are also found in the NRC VAD lexicon.
    We indicate the full GPAD framework with directions.
  }
  \label{fig:meaning.figousiometer9010_russell1980a}
\end{figure*}

It is in the PDS framework
(Figs.~\ref{fig:meaning.ousiometer9310_googlebooks001}G--I),
that we see 
robust agreement between ousiograms and the underlying axes.
In the main power-danger ousiogram (Fig.~\ref{fig:meaning.ousiometer9310_googlebooks001}G),
the histogram shows a definitive bias towards safe, low-danger words.
As shown by the marginal on the left axis, the danger distribution is skewed strongly
towards safer words, and the median danger score is well below 0.
By contrast, power presents a symmetric marginal distribution with a median slightly above 0.
The power-structure ousiogram shows a general spread
(Fig.~\ref{fig:meaning.ousiometer9310_googlebooks001}H)
while the
danger-structure ousiogram again shows a clear safety bias
(Fig.~\ref{fig:meaning.ousiometer9310_googlebooks001}I).

In Fig.~\ref{fig:meaning.ousiometer9400_PDS_tableau004},
we expand our analysis to show power-danger ousiograms for six more corpora:
The novels of Jane Austen,
a subset of Arthur Conan Doyle's Sherlock Holmes stories,
the New York Times,
Wikipedia,
transcriptions of talk radio in the US,
and
Twitter (see Sec.~\ref{sec:meaning.data} for details).
These corpora vary widely in size and kind:
written versus spoken, news, literature, formal and informal,
bearing social amplification or not (e.g., the inclusion of retweets from Twitter
encodes one form of echoing,
but the other corpora carry no such equivalent signature of popularity).
For each corpus, we provide the full analytic sequence in the manner of
Figs.~\ref{fig:meaning.ousiometer9310_lexicon001}
and~\ref{fig:meaning.ousiometer9310_googlebooks001}
in
Figs.~\ref{fig:meaning.ousiometer9310_jane_austen001}--\ref{fig:meaning.ousiometer9310_twitter001}.

The power-danger ousiograms for these six distinct corpora
in Fig.~\ref{fig:meaning.ousiometer9400_PDS_tableau004}
all present the same safety bias for words
as we saw for English fiction in Fig.~\ref{fig:meaning.ousiometer9310_googlebooks001}G.
While varying in detail as they must,
the six histograms in Fig.~\ref{fig:meaning.ousiometer9400_PDS_tableau004}
all show a weight toward words below the horizontal
\semdiff{weak}{powerful}
axis,
and the danger marginals on the left axes of all ousiograms
are skewed toward safety.
There is no such bias for the power dimension,
though median power is at or above zero in all cases.

We emphasize again that our initial determination of the PDS framework
was performed only at the level of types, using the NRC VAD lexicon.
In these subsequent tests with real corpora, we have found
that our hypothesized ousiometric PDS framework has been borne out
to be fundamental.

\section{Congruences}
\label{sec:meaning.congruences}

\subsection{Russell's circumplex model of emotion}
\label{subsec:meaning.circumplex}

\begin{revisionbar}

%% \todo{Align with bad-good and gentle-aggressive}

We consider Russell's highly cited circumplex model of affect~\cite{russell1980a},
in light of the GPADS framework.
We find general accordance
with one region of disagreement
being in the aggressive direction (the dangerous-powerful quadrant).
%% We make the observations that negative emotions cannot be adequately
%% represented in a two dimensional framework.
%% and
%% 2.
%% The circumplex model of affect is a map of general states of being,
%% not just emotional states.

Affective states are representations of emotional states,
and may be external (e.g., facial expressions) or internal (conscious awareness).
In linking to essential meaning,
in 1952, Schlosberg~\cite{schlosberg1952a} was one of the first to suggest
that emotion---as conveyed by facial expressions---could be
well represented by two dimensions,
suggesting
\semdiff{pleasantness}{unpleasantness}
and
\semdiff{attention}{rejection}.
Two years later, Schlosberg then posited a third dimension of level of activation
while also asserting that
``the field [of emotion] is chaotic''~\cite{schlosberg1954a}.
Certain emotions would seem to readily connect with locations in
the power-danger framework.
Fear is a particular response to danger,
contentment is a possible state in a safe environment,
and so on.
We examine such connections carefully below.

We consider Russell's original, unrevised model because of its historical
and continued importance
to the field~\cite{russell1997a,barrett1998a,barrett1999a,yik2011a}
as well as the challenge delivered
by such a distinct kind of study.
Indeed, the approximate agreement
between the studies of Russell and Mohammad
is remarkable given
the differences between them:
Era
(late 1970s versus late 2010s),
subjects
(undergraduate students at the University of British Columbia
versus online crowdsourcing),
assessment type (various in Ref.~\cite{russell1980a} versus best-worst scaling),
scale (28 versus $\sim$ 20,000 terms),
and
framing
(the specific of affect versus the general of essential meaning).

Building on earlier work~\cite{schlosberg1952a,fillenbaum1971a},
Russell asserted that eight fundamental affect concepts
could be arranged as compass points on a circle (see Fig.~1 in Ref.~\cite{russell1980a}).
As we indicate in Fig.~\ref{fig:meaning.figousiometer9010_russell1980a}A,
starting from pointing upwards and 
stepping around clockwise,
these concepts are
distress ($\sim$ danger),
arousal ($\sim$ aggression),
excitement ($\sim$ power),
pleasure ($\sim$ goodness),
contentment ($\sim$ safety),
sleepiness ($\sim$ gentleness),
depression ($\sim$ weakness),
and
misery ($\sim$ badness).
In line with the VAD framework,
Russell took the underlying horizontal and vertical
dimensions to be
\semdiff{pleasure}{displeasure}
and
\semdiff{arousal}{sleepiness},
aligning in the GPAD frameworks
\semdiff{goodness}{badness}
and
\semdiff{aggression}{gentleness}
The alignment with the GAS framework notwithstanding,
we have facilitated comparison with the GPAD framework,
by rotating Russell's framework by $-\pi/4$.

Russell then carried out a series of varying types of surveys
on perceptions of 28 affect terms
(e.g.,
`afraid',
`glad',
`serene',
`bored').
In Fig.~\ref{fig:meaning.figousiometer9010_russell1980a}A,
we show the locations of 27 words 
according to the results presented in Fig.~2 of Ref.~\cite{russell1980a}
(we exclude the 2-gram `at ease').
In Fig.~\ref{fig:meaning.figousiometer9010_russell1980a}B,
we show the same words located by their power-danger scores.

In general, we see that words in the circumplex and GPAD frameworks
are reasonably well aligned.
A number of words show strong congruence across the two studies,
including
`sleepy',
`excited',
`aroused',
and
`miserable'.
Angles of affect words are generally similar with a maximum
discrepancy of around $\pi/4$.
For example, `tired' is in the direction of
gentle  in the circumplex model and weak in GPAD framework
(`sleepy' aligns with gentle in both, and the added hue of danger
for `tired' in the GPAD framework is sensible).
Apart from `tense' and to a lesser extent `astonished' and `droopy',
affect words register strong power-danger magnitudes,
and are consequently located around an approximate circle.

The word that most stands out as differing between the two studies is `tense'.
On top of the major distinctions between the studies listed above,
without the context of working with a small set of emotion-themed words,
participants in the NRC VAD study would be more likely to interpret
words and phrases by their most general, dominant meaning.
While many of the affect words have clear meanings that are emotional
(e.g., `miserable'),
the word `tense' might not be as strongly construed as `stressed'.
Over four decades, we might also expect meanings of some words to shift somewhat.
And in any case, the four surveys in Russell only show rough agreement
with each other
(see Figs.~2--5 in Ref.~\cite{russell1980a}).

\end{revisionbar}

\begin{figure*}[tp!]

  \includegraphics[width=\textwidth]{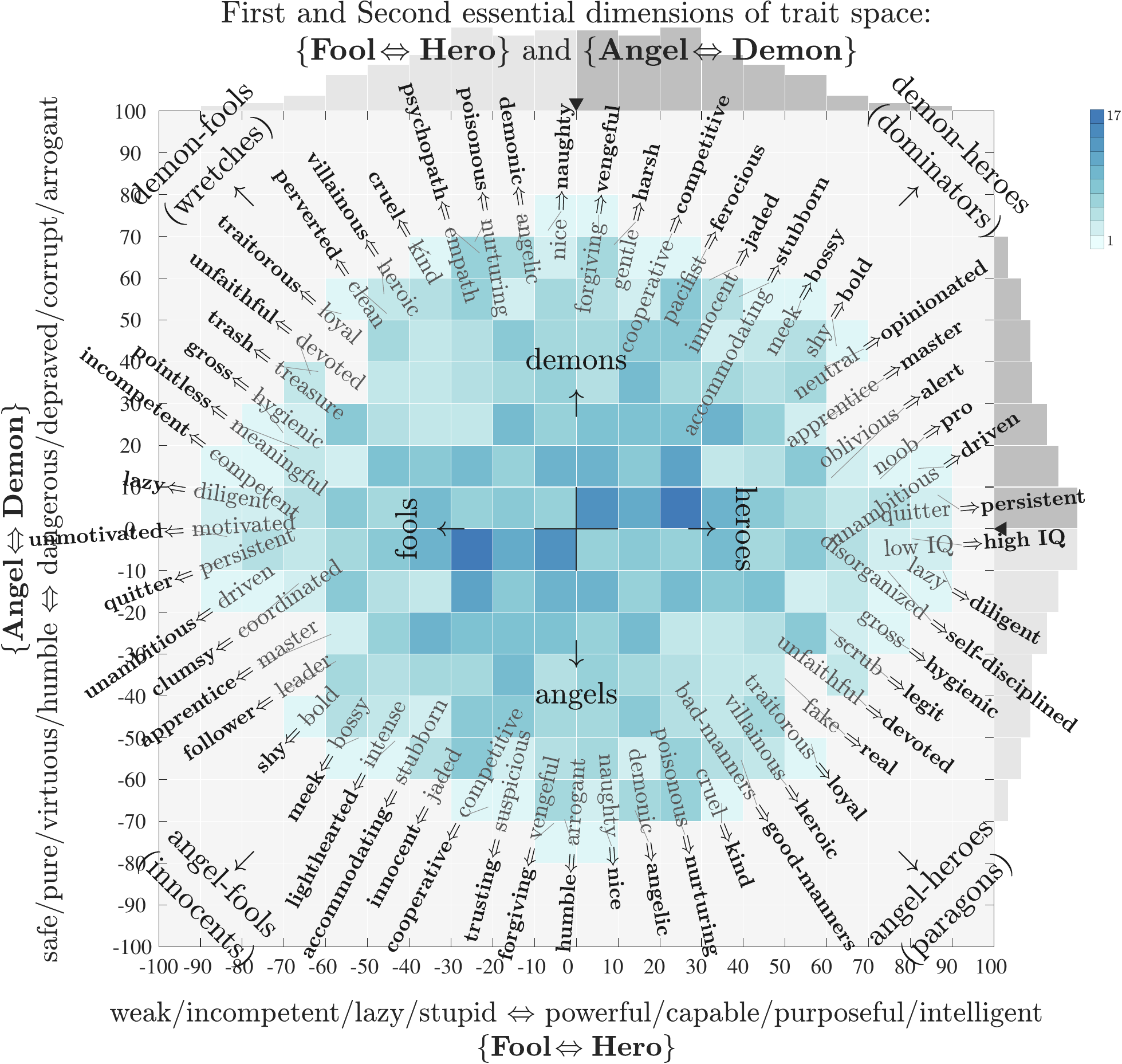}

  \caption{
    \raggedright
    Ousiogram for the first two singular dimensions of essential trait space
    based on 2000 fictional characters evaluated across 464 traits each.
    We observe a clear congruence between the GPADS framework
    and the primary archetypes of fictional characters.
    Compare with Tab.~\ref{tab:meaning.lists-for-archetype-traits}
    which lists
    the top ranked traits for the first three dimensions.
    The first three dimensions of essential trait space align with PDS
    as
    \semdiff{Fool}{Hero}
    for
    \semdiff{weak}{powerful},
    \semdiff{Angel}{Demon}
    for
    \semdiff{safe}{dangerous},
    and
    \semdiff{Traditionalist}{Adventurer}
    for
    \semdiff{structured}{unstructured}.
    Because the orientation of each trait's semantic differentials was arbitrary,
    we duplicated all semantic differentials with their bipolar adjectival pairs swapped.
    Hence, the traits and their reverse appear on opposite sides of the ousiogram.
    See Ref.~\cite{dodds2025archetypometrics} for a full exploration
    of archetype space with analysis that shows the first
    six singular dimensions rise to the level of identifiable archetypes.
  }
  \label{fig:meaning.ousiogram-for-archetype-traits}
\end{figure*}

\begin{table*}

  \setlength{\tabcolsep}{0pt}
  
  \begin{tabular}{ccccc}

    \rowcolors{2}{traitrowcolor}{traitrowcoloralt}
    \begin{tabular}{rl}
      \rowcolor{headerpop}
      & Essential trait dimension 1
      \\
      \rowcolor{headerpop}
      & \semdiff{Fool}{Hero}
      \\
      \rowcolor{headerpop}
      & $\sim$ \semdiff{weak}{powerful}
      \\
      1. & \semdiff{lazy}{diligent}
      \\
      2. & \semdiff{quitter}{persistent}
      \\
      3. & \semdiff{unmotivated}{motivated}
      \\
      4. & \semdiff{unambitious}{driven}
      \\
      5. & \semdiff{incompetent}{competent}
      \\
      6. & \semdiff{low IQ}{high IQ}
      \\
      7. & \semdiff{absentminded}{focused}
      \\
      8. & \semdiff{helpless}{resourceful}
      \\
      9. & \semdiff{unobservant}{perceptive}
      \\
      10. & \semdiff{slacker}{workaholic}
      \\
      11. & \semdiff{disorganized}{self-disciplined}
      \\
      12. & \semdiff{noob}{pro}
      \\
      13. & \semdiff{slugabed}{go-getter}
      \\
      14. & \semdiff{underachiever}{overachiever}
      \\
      15. & \semdiff{gross}{hygienic}
    \end{tabular}
    &
    ~~~
    &
    \rowcolors{2}{traitrowcolor}{traitrowcoloralt}
    \begin{tabular}{rl}
      \rowcolor{headerpop}
      & Essential trait dimension 2
      \\
      \rowcolor{headerpop}
      & \semdiff{Angel}{Demon}
      \\
      \rowcolor{headerpop}
      & $\sim$ \semdiff{safe}{dangerous}
      \\
      1. & \semdiff{nice}{naughty}
      \\
      2. & \semdiff{forgiving}{vengeful}
      \\
      3. & \semdiff{humble}{arrogant}
      \\
      4. & \semdiff{nurturing}{poisonous}
      \\
      5. & \semdiff{gentle}{harsh}
      \\
      6. & \semdiff{angelic}{demonic}
      \\
      7. & \semdiff{warm}{quarrelsome}
      \\
      8. & \semdiff{cooperative}{competitive}
      \\
      9. & \semdiff{empath}{psychopath}
      \\
      10. & \semdiff{kind}{cruel}
      \\
      11. & \semdiff{wholesome}{salacious}
      \\
      12. & \semdiff{altruistic}{selfish}
      \\
      13. & \semdiff{sweet}{bitter}
      \\
      14. & \semdiff{respectful}{rude}
      \\
      15. & \semdiff{pure}{debased}
    \end{tabular}
    &
    ~~~
    &
    \rowcolors{2}{traitrowcolor}{traitrowcoloralt}
    \begin{tabular}{rl}
      \rowcolor{headerpop}
      & Essential trait dimension 3
      \\
      \rowcolor{headerpop}
      & \semdiff{Traditionalist}{Adventurer}
      \\
      \rowcolor{headerpop}
      & $\sim$ \semdiff{structured}{unstructured}
      \\
      1. & \semdiff{scheduled}{spontaneous}
      \\
      2. & \semdiff{stick-in-the-mud}{adventurous}
      \\
      3. & \semdiff{uncreative}{open to new experiences}
      \\
      4. & \semdiff{serious}{bold}
      \\
      5. & \semdiff{monotone}{expressive}
      \\
      6. & \semdiff{lifeless}{spirited}
      \\
      7. & \semdiff{corporate}{freelance}
      \\
      8. & \semdiff{geriatric}{vibrant}
      \\
      9. & \semdiff{serious}{playful}
      \\
      10. & \semdiff{stoic}{expressive}
      \\
      11. & \semdiff{shy}{playful}
      \\
      12. & \semdiff{humorless}{funny}
      \\
      13. & \semdiff{deliberate}{spontaneous}
      \\
      14. & \semdiff{orderly}{chaotic}
      \\
      15. & \semdiff{withdrawn}{outgoing}
    \end{tabular}
  \end{tabular}
  
  \caption{
    The leading trait composition 
    for the three primary dimensions of archetype space~\cite{dodds2025archetypometrics}.
    Traits are ranked by component size in each dimension.
    Compare with the ousiogram in Fig.~\ref{fig:meaning.ousiogram-for-archetype-traits}
    for the first two dimensions.
  }

  \label{tab:meaning.lists-for-archetype-traits}
  
\end{table*}

\begin{revisionbar}

%% \subsection{The Five Factor Model of personality}
%% \label{subsec:meaning.personality}
%% 
%% \begin{textblock}
%% \item
%%   The power-danger-structure framework
%%   could be considered in the context of personality.
%% \item
%%   The well established framework
%%   of the ``Big Five'' maintains
%%   the core personality traits
%%   of
%%   openness,
%%   conscientiousness,
%%   extroversion,
%%   agreeableness,
%%   and
%%   neuroticism
%%   (OCEAN)~\cite{john1999a}.
%% \item
%%   These traits align with the PDS framework:
%%   conscientious as powerful,
%%   agreeableness as safe,
%%   and
%%   openness as unstructured (playful).
%% \item
%%   The opening three letter ordering would then be CAO---contrary to OCEAN.
%% \item
%%   Conscientiousness and agreeableness would have equal weight
%%   while openness would be a third lesser dimension.
%% \end{textblock}

\subsection{Fictional characters}
\label{subsec:meaning.fictional-characters}

In separate work---archetypometrics---we have carried out
extensive analysis of a dataset of 2000 characters
from 341 popular stories from literature, movies, and television
which have over 70 million ratings across
464 semantic differential traits~\cite{dodds2025archetypometrics,dodds2025archetypometrics-dataset-zenodo}.
As per our methodology here,
we performed SVD on the 464$\times$2000 matrix
and then generated a suite of ousiograms and ranked lists
for both essential trait and character space.
Here, we focus on a few core elements of essential trait space.

In Fig.~\ref{fig:meaning.ousiogram-for-archetype-traits},
we show an ousiogram for the first two singular dimensions
of essential trait space,
and
in Tab.~\ref{tab:meaning.lists-for-archetype-traits},
we list the top 15 traits for the first three dimensions.

Remarkably, the PDS framework is aligned with the first
three dimensions of archetype space.
We emphasize that we again kept the factor analysis simple,
and that we did not perform any rotations or further manipulations.

We have named the archetype pairs of the first three dimensions
\semdiff{Fool}{Hero},
\semdiff{Angel}{Demon},
and
\semdiff{Traditionalist}{Adventurer}.
For even more agreement,
we find that in the GPADS framework,
the words `foolish', `heroic', `angelic', and `demonic'
are all located in strong alignment with their corresponding
archetypes in essential trait space.
We have included these words as descriptors
in the meaning cube in Fig.~\ref{fig:meaning.ousiometric_cube3d002}.

These archetype names are of course general names which may register variably.
We further define them
by distilling the dominant traits into four pairs of semantic differentials each.
For \semdiff{Fool}{Hero},
we have
\semdiff{weak}{powerful},
\semdiff{incompetent}{capable},
\semdiff{lazy}{purposeful},
and
\semdiff{stupid}{intelligent};
for
\semdiff{Angel}{Demon},
we have
\semdiff{safe}{dangerous},
\semdiff{pure}{depraved},
\semdiff{virtuous}{corrupt},
and
\semdiff{humble}{arrogant};
and for 
\semdiff{Traditionalist}{Adventurer},
we have
\semdiff{serious}{playful},
\semdiff{predictable}{unpredictable},
\semdiff{humorless}{funny},
and
\semdiff{uncreative}{creative}.

One differential that might first appear at odds with our framework
is \semdiff{villainous}{heroic}, which is aligned with \semdiff{bad}{good}.
However, the differential correctly interpolates between
the directions of \wordvar{demon} (villain) and \wordvar{hero}
which are not opposites but rather at right angles to each other.
That is,
\semdiff{villainous}{heroic}
$\sim$
\semdiff{weak}{powerful}
-
\semdiff{safe}{dangerous}
=
\semdiff{bad}{good}.

Finally, for a circumplex model of archetypes that is aligned with the GPAD framework,
and as indicated in Fig.~\ref{fig:meaning.ousiogram-for-archetype-traits},
we suggest (moving around the circle starting at the top):
Demons (dangerous),
Dominators (aggressive),
Heroes (powerful),
Paragons (good),
Angels (safe),
Innocents (or Lambs) (gentle),
Fools (weak),
and
Wretches (bad).

\end{revisionbar}

\begin{revisionbar}
  
\section{Ousiometer}
\label{sec:meaning.ousiometer}

\end{revisionbar}

%% run figtelegnomic_timeseries_ousiometrics_DAPGS_flipbook100.m
%% in ~/work/stories/2025-06ousiometric-time-series-books/figures

\begin{figure*}[t!]
  \includegraphics[width=\textwidth]{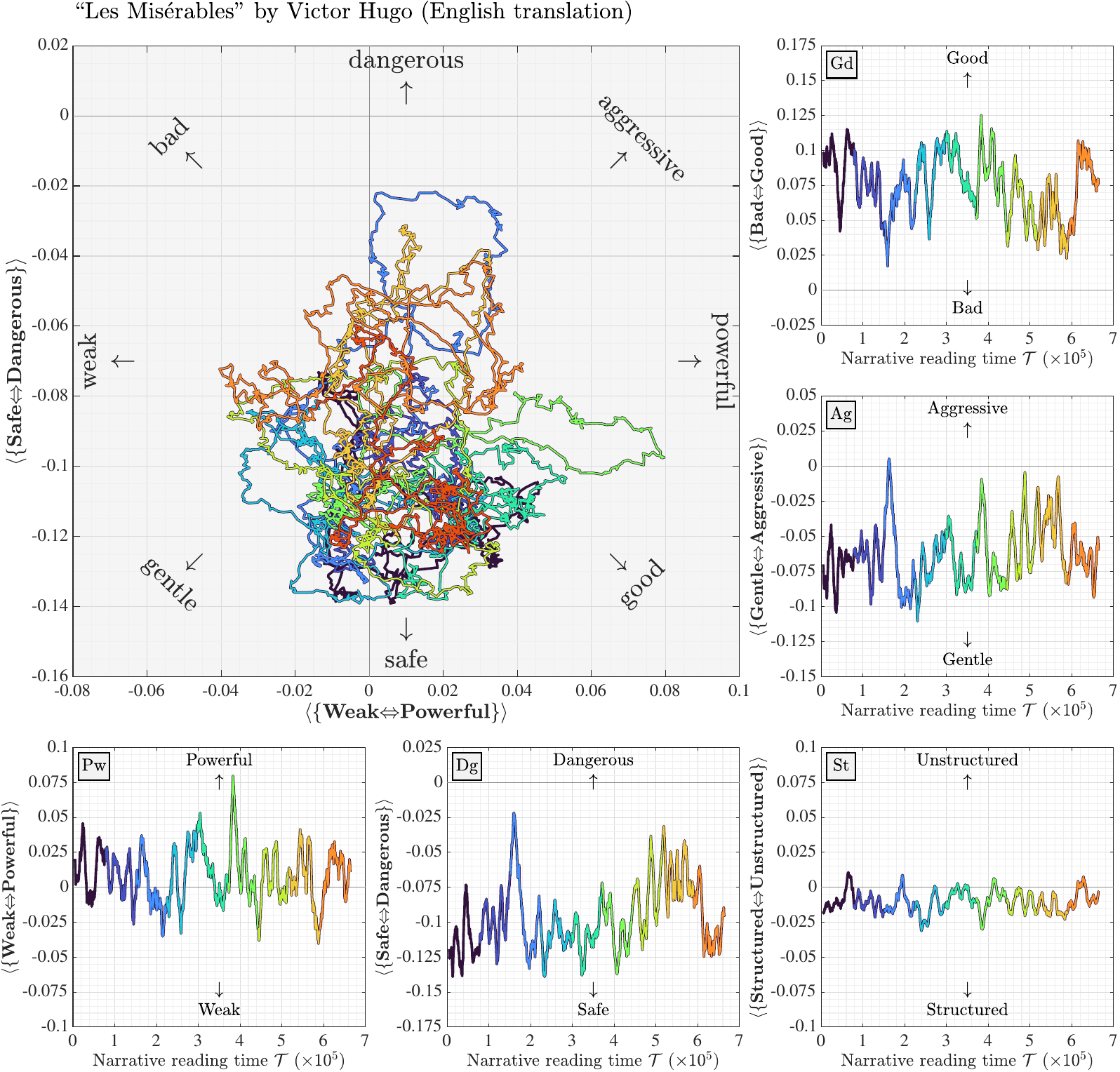}
  \caption{
    \raggedright
    \textbf{
      The ousiometer as a telegnomic lexical instrument for literature:
      Example ousiometric time series for Victor Hugo's Les Mis\'{e}rables.
    }
    Main plot: Ousiometric trajectory of the novel in the PD plane
    with the GPAD framework indicated.
    The five individual time series comprise the GPADS framework.
    We break the timeline into 10 equal epochs and align the colors on
    all time series, including that of the trajectory in the PD plane.
    Narrative reading time $\mathcal{T}$ is in terms of 1-grams
    which includes numbers, punctuation, and other
    non-word elements.
    All vertical ranges for the time series are the same (0.20),
    and are shifted as needed.
    As expected per variance explained,
    the first four time series of the GPADS framework show
    similar variation while Structure is muted.
    Smoothing is at 10,000 1-grams with steps of 100 1-grams.
    See the accompanying flipbook in Figs.~\ref{fig.meaning:figtelegnomic_timeseries_flipbook9017_001_10000_100_ousiometrics_GPADS100_LesMis_noname}--\ref{fig.meaning:figtelegnomic_timeseries_flipbook9017_025_10000_100_ousiometrics_GPADS100_LesMis_noname} in the \suppmaterial.
    }
  \label{fig:meaning.ousiometer-les-miserables}
\end{figure*}

We construct an elementary `ousiometer',
a lexical instrument for measuring
the average essential meaning
of large-scale texts.
We take a similar approach 
to that of our hedonometer~\cite{edgeworth1881a,dodds2009b,dodds2011e,dodds2015a,reagan2016c,gallagher2021a}.
We view the ousiometer and hedonometer as
example `telegnomic' lexical instruments
capable of remotely sensing meaning, knowledge, and stories.

We use $\meaningsymbol$ to represent
one of the essential meaning dimensions within
a specified ousiometric framework.
For a simple ousiometer, we compute the average meaning score
$\avgmeaning(\systemsymbol)$
for a text $\systemsymbol$ in the following way.
We consider only the 1-grams of the NRC VAD lexicon,
leaving aside $n$-grams for $n \ge 2$ for possible future improvements.
For any given text $\systemsymbol$, we apply a
`lexical lens' $\lexicallens$, a simple operator
that filters the text's 1-grams,
returning the subset 1-gram lexicon that intersects with the
NRC VAD 1-gram lexicon.
We denote the lensed text as
$\lexicallens(\systemsymbol)$.
We write
the resultant lensed lexicon as
$\bigrank_{\lexicallens(\systemsymbol)}$,
further specifying this set to be 
a list of 1-grams ordered by descending frequency of usage
$\frequency_{\elementsymbol}$
within
$\lexicallens(\systemsymbol)$.
For each 1-gram
$\elementsymbol$
in the lensed lexicon 
$\bigrank_{\lexicallens(\systemsymbol)}$,
we then straightforwardly determine
$\elementsymbol$'s normalized frequency as
$
\probsymbol_{\elementsymbol}
=
\frequency_{\elementsymbol}
/
\sum_{\elementsymbol'}
\frequency_{\elementsymbol'}.
$

In general, given a lexical lens $\lexicallens$,
the average
ousiometric score
of a text $\systemsymbol$ is:
\begin{equation}
  \avgmeaning
  \left(
  \systemsymbol;
  \lexicallens
  \right)
  =
  \sum_{\elementsymbol \in \bigrank_{\lexicallens(\systemsymbol)}}
  \probsymbol_{\elementsymbol}
  \meaningsymbol_{\elementsymbol},
  \label{eq:meaning.ousiometer}
\end{equation}
where
$\meaningsymbol_{\elementsymbol}$ is the average ousiometric score
for the 1-gram $\elementsymbol$
derived from the NRC VAD lexicon scores~\cite{mohammad2018b}.

\begin{revisionbar}

As an example,
in Fig.~\ref{fig:meaning.ousiometer-les-miserables},
we show ousiometric trajectory and time series for
Victor Hugo's ``Les Mis\'{e}rables.''
In doing so, we are continuing the development of our earlier computational work
on the measurement of emotional arcs in stories,
famously ventured by Kurt Vonnegut~\cite{vonnegut2010a,dodds2015a,reagan2016c,fudolig2023a}.

The main plot is an ousiometric trajectory of ``Les Mis\'{e}rables''
in the GPAD framework, oriented in the PD plane. To the right, we display
ousiometric time series for the GAS framework which interlocks with
the same for the PDS framework running along the bottom.
The colors indicate 10 narrative reading time blocks, as measured by 1-grams,
and help show the ousiometric trajectories path in time.

We observe that the \semdiff{bad}{good} time series is of similar
form to what we found using our hedonometer~\cite{reagan2016c}.
In Fig.~\ref{fig:meaning.hedonometer-les-miserables},
we show a screenshot for the happiness time series for
``Les Mis\'{e}rables''
taken from our interactive story viewer at \url{https://hedonometer.org}.
The agreement is satisfactory given that
the hedonometer and ousiometer instruments are built on two distinct word
lists using different evaluations (Likert vs.\ best-worst scaling).
An interactive visualization based on 
Fig.~\ref{fig:meaning.ousiometer-les-miserables}
would be a natural next step.

In Figs.~\ref{fig.meaning:figtelegnomic_timeseries_flipbook9017_001_10000_100_ousiometrics_GPADS100_LesMis_noname}--\ref{fig.meaning:figtelegnomic_timeseries_flipbook9017_025_10000_100_ousiometrics_GPADS100_LesMis_noname}
in the \suppmaterial,
we supply a flipbook which traces out the GPADS time series over 25 epochs.

A full analysis of our ousiometric time series for
``Les Mis\'{e}rables'' is well beyond the scope of our present work,
and we reserve it for future exploration.
Our purpose here is simply to show that we can readily build
an ousiometer for large-scale texts.

In connected work on the ousiometrics of literature~\cite{fudolig2023a},
we have used  empirical mode decomposition on ousiometric
time series to find that longer books are structured more like concatenations of shorter texts,
revealing characteristic fluctuations in power and danger.

Finally, in the \suppmaterial, we provide another example
use of an ousiometer for streaming text:
A retrospective analysis of Twitter for 2020/01 to 2021/01 inclusive
(Sec.~\ref{sec:meaning.ousiometer-twitter} and
Fig.~\ref{fig:meaning.ousiometer-twitter})~\cite{miller2011a,bermingham-smeaton-2011-using,gayo-avello2012a,jungherr2013forecasting,alshaabi2021d,dodds2021d}.

\end{revisionbar}

%%%%%%%%%%%%%%%%%%%%%%%%%%%%%%%%%%%%%%%%%%%%%%%%%%%%%%%%%%%%%%%%%%%%%%%%%%%%%%%%%%%%
%% Concluding remarks
%%%%%%%%%%%%%%%%%%%%%%%%%%%%%%%%%%%%%%%%%%%%%%%%%%%%%%%%%%%%%%%%%%%%%%%%%%%%%%%%%%%%

\begin{revisionbar}
  
\section{Concluding remarks}
\label{sec:meaning.concludingremarks}

%% \textbf{The power-danger framework of essential meaning:}

%% \textbf{The mismeasurement of meaning:}

\subsection{The mismeasurement of meaning}
\label{subsec:meaning.conclusion-mismeasurement}

The quantitative measurement of
essential meaning---ousiometrics---has been
properly engaged as a scientific challenge for close to a century.
Based on semantic differentials,
the three dimensional orthogonal framework of evaluation-potency-activation (EPA)
due to Osgood \etal~\cite{osgood1957a} has effectively remained
the leading conceptual framework, if not always by direct reference.
Research into the specific context of affect saw EPA adapted
as valence-arousal-dominance (VAD)~\cite{mehrabian1974a,mehrabian1974b}.
The VAD formalism has become widespread and not limited to studies of emotion,
even being used for general essential meaning studies
instead of EPA~\cite{bradley1999a,mohammad2018b}.

We have shown that $\sim$ 20,000 terms evaluated by
best-worst scaling in the VAD framework fails to reproduce
the orthogonal VAD framework itself.
We have contended that this  cannot be explained away by
participants misunderstanding bipolar adjectival pairs used
to define VAD dimensions.
Rather,
we have argued that a longstanding problem for ousiometrics
has been the difficulty of ascribing bipolar adjectival pairs to
accurately characterize dimensions derived from participants'
assessments of a larger set of semantic differentials~\cite{bakker2014a}.
As is, researchers tend to provide sets of bipolar adjectival pairs for fundamental
dimensions, making them overly blunt instruments that
have more likelihood of being correlated (see Tab.~\ref{tab:meaning.VADlimits}).
Even after exploring antonyms and antousionyms
(Sec.~\ref{sec:meaning.synousionyms}),
we continue to see this dimension characterization
problem as unavoidable.

We recommend
that ousiometric studies start from a larger set---on the order of 20 to 50---of
simple bipolar adjectival pairs and always perform dimensional reduction.
Standardizing such a set of clear bipolar adjectival pairs would be of great value to the field,
and our advice is independent of which instrument
is employed to rate semantic differentials (Likert scale,
best-worst scaling, etc.).
(Likert scales may have less inter-rater reliability
than best-worst scaling but do allow
datasets to be extended with independent studies.)
Such studies will be more expensive but will be far more robust.
Using ousiograms, which provide richly informative visualizations,
the extracted dimensions can then be examined and identified.
For lexicons sufficiently rich in types and corpora-matching in terms of tokens,
we expect that the axes of \semdiff{weak}{powerful} and \semdiff{safe}{dangerous} will emerge.

Automatically annotated histograms like our ousiograms could be used in any
sphere to compare two variables measured for a collection of
categorical entities (e.g., crime rates and median house prices for cities in the US).

\subsection{The GAS, PDS, and GPADS frameworks}
\label{subsec:meaning.conclusion-frameworks}

Here, we have found that essential
meaning does not conform to VAD but is instead well captured by
the two mutually interpretable coordinate systems spanned by
the broad semantic differentials
\semdiff{bad}{good} and \semdiff{gentle}{aggressive} (GAS),
and
\semdiff{weak}{powerful} and \semdiff{safe}{dangerous} (PDS).
Both share the dimension of essential meaning---\semdiff{structured}{unstructured}---which
we may interpret as the dimension of evolution.
We have argued in particular that the primary two dimensional plane,
which accounts for over 90\% of variance explained for types,
can be viewed as a kind of circumplex model giving us GPADS.
We note that one of the 50 semantic differentials used by Osgood et al.~\cite{osgood1957a}
in their foundational work was \semdiff{safe}{dangerous}. We now understand
that the dimension Activation in the EPA framework was an error. All dimensions
have an intensity level signified by vector magnitude.

%% in Fig.~\ref{fig:meaning.PD_sketch}.

\subsection{The safety bias of communication}
\label{subsec:meaning.conclusion-safety-bias}

Our finding of a safety bias in
diverse written and spoken language
generalizes our earlier work which revealed
a positivity bias~\cite{kloumann2012b,dodds2015a}---a linguistic instantiation
of the Pollyanna Principle~\cite{boucher1969a}.
In the GAS framework we have defined here,
the positivity bias is a goodness bias.
We have also found a complementary linguistic low-aggression bias in the GAS framework
(see Fig.~\ref{fig:meaning.ousiometer9310_googlebooks001}D).

We now understand that 
the linguistic goodness bias
and
the linguistic low-aggression bias
are shadows of an underlying linguistic safety bias---projections
of points in the two-dimensional $\Mpower$-$\Mdanger$ plane
onto the
orthogonal one-dimensional diagonal axes
of
goodness
and 
aggression.
The one dimensional map is not the two-dimensional territory.

Because of the safety bias,
congruences with other spaces like fictional archetypes,
and
the behavior of our prototype ousiometer,
we have demonstrated
that the PDS framework is the most
minimal, well-aligned description of essential meaning.
This does not however lead us to discarding the GAS framework
as the GPADS circumplex framework affords richer, more immediately
informative analyses than either alone.

\end{revisionbar}

\section{Data, code, and Materials Availability}
\label{sec:meaning.data-code-materials}

The GPADS framework data, scripts, and documentation reside on Gitlab at
\href{https://gitlab.com/petersheridandodds/ousiometry}{https://gitlab.com/petersheridandodds/ousiometry}.
The GPADS dataset is also on Zenodo~\cite{dodds2026ousiometrics-GPADS-dataset-zenodo}. We also provide a range of supporting material
at the paper's \onlineappendicesplain:\\
\onlineappendiceslocation.

\section*{Acknowledgments}
The authors appreciate discussions
with
Ryan~J.~Gallagher,
Lewis~Mitchell,
and
Aimee~Picchi.
PSD honors Kathleen Mary Standen with this work.
The authors are grateful for the computing resources provided by the
Vermont Advanced Computing Center
which was supported in part by
NSF awards \#~1827314 and \#~2117345;
foundational support from MassMutual;
National Science Foundation Award \#2242829
(Science of Online Corpora, Knowledge, and Stories);
Google Open Source
under the Open-Source Complex Ecosystems And Networks (OCEAN) project;
the  Alfred P. Sloan Foundation (G-2024-22498),
and an anonymous philanthropic gift.

%%%%%%%%%%%%%%%%%
%% changelog
%%%%%%%%%%%%%%%%%

\vfill

%% turn off for Science Advances
\section*{Changelog}

\begin{changelogbox}[2023/03/29]
\item
  Standardized expression 
  of semantical differentials
  throughout
  (e.g., \semdiff{dangerous}{safe})
\item 
  Reformatted for improved readability:
  Left justified, space between paragraphs.
\item
  Added a table of contents.
\item
  Added a figure showing the mapping between
  the compass of meaning and 
  Cipolla's law of human stupidity.
\item
  Added this changelog.
\end{changelogbox}

\medskip

\begin{changelogbox}[2025/06/12]
\item
  Renamed the Low-Energy to High-Energy axis as
  Gentleness to Aggression.
  Energy was a mistake carried on from the VAD framework.
\item
  Formally made the connection between GAS
  and PDS with a circumplex model for the two
  first dimensions leading to the combined framework of DAPGS.
\item
  Elevated the third dimension of structure in the paper's
  title and our general discussion.
\item
  Added the meaning cube representation
  in Fig.~\ref{fig:meaning.ousiometric_cube3d002}
  and the companion ranked word lists for the 13 opposing cube pairs in
  Figs.~\ref{fig:meaning.ousiometrics_opposite_cubes001_001}--\ref{fig:meaning.ousiometrics_opposite_cubes001_013}.
\item
  Included Fig.~\ref{fig:meaning.ousiogram-for-archetype-traits}
  and
  Tab.~\ref{tab:meaning.lists-for-archetype-traits}
  and accompanying discussion
  in Sec.~\ref{subsec:meaning.fictional-characters}
  to show
  the congruence with archetypes
  of fictional characters.
\item
  Improved discussion of
  Russell's circumplex model
  to show congruence with DAPG(S) framework.
\item
  Added material (figure, table, discussion)
  re the congruence between the three primary archetypes
  of fictional characters and the PDS framework.
\item
  Removed examples of PDS in D\&D alignment charts
  and Cipolla's Law of Human Stupidity.
\item
  Added an ousimetric analysis of
  Victor Hugo's ``Les Mis\'{e}rables''
  in~\ref{sec:meaning.ousiometer}.
\item
  Greatly expanded the \suppmaterial\ with new analytic figures and tables.
\item 
  Moved some material from previous manuscript to \suppmaterial.
\item
  Improved and simplified the title to better declare our findings.
\item
  Greatly reduced the conclusion.
\item
  Shared DAPGS data set on Zenodo.
\item
  General edits for improved clarity.
\end{changelogbox}

\medskip

\begin{changelogbox}[2026/06/02]
\item
  Reversed DAPG initialism as GPAD
  GPADS now aligns with ordering of subsequences PDS and GAS (G-P-A-D-S).
\item
  Updated data set on Zenodo with GPADS ordering~\cite{dodds2026ousiometrics-GPADS-dataset-zenodo}
\item
  Updated affiliations.
\item 
  General edits throughout.
\item
  Except for formatting and some minor differences,
  arXiv version is essentially the same as
  published Science Advances version.
\end{changelogbox}

%%%%%%%%%%%%%%%%%
%% references
%%%%%%%%%%%%%%%%%

\clearpage

\addcontentsline{toc}{section}{References}
\bibliography{true-dimensions-of-meaning}
%% \bibliography{everything.bib}

%% \clearpage

%% %% appendix
%%
%% %% optional: switch to single column
\onecolumn

%% following records the starting page number of the supplementary
%% section in a file called startsupp.txt
%% enables script-based breaking of manuscript and supplementary
\newwrite\tempfile
\immediate\openout\tempfile=startsupp.txt
\immediate\write\tempfile{\thepage}
\immediate\closeout\tempfile

\appendix

\begin{revisionbar}
\section*{Appendices}
%% \section*{Supplementary Material for ``\protect\input{true-dimensions-of-meaning.title.txt}\unskip.''}
\addcontentsline{toc}{section}{Appendices}
\end{revisionbar}

\renewcommand{\thesection}{A\arabic{section}}
\renewcommand{\thepage}{A\arabic{page}}
\renewcommand{\thefigure}{A\arabic{figure}}
\renewcommand{\thetable}{A\arabic{table}}
\setcounter{section}{0}
\setcounter{page}{1}
\setcounter{figure}{0}
\setcounter{table}{0}

%% \revisionnote{All supplementary is new except for the first section of large ousiograms in the three frameworks; Also, two sections have been moved from the original manuscript to here: Synousionyms/Antousionyms and Ousiometer for Twitter.}

\begin{figure*}[tp!]
  \centerfloat
  \includegraphics[width=1.1\textwidth]{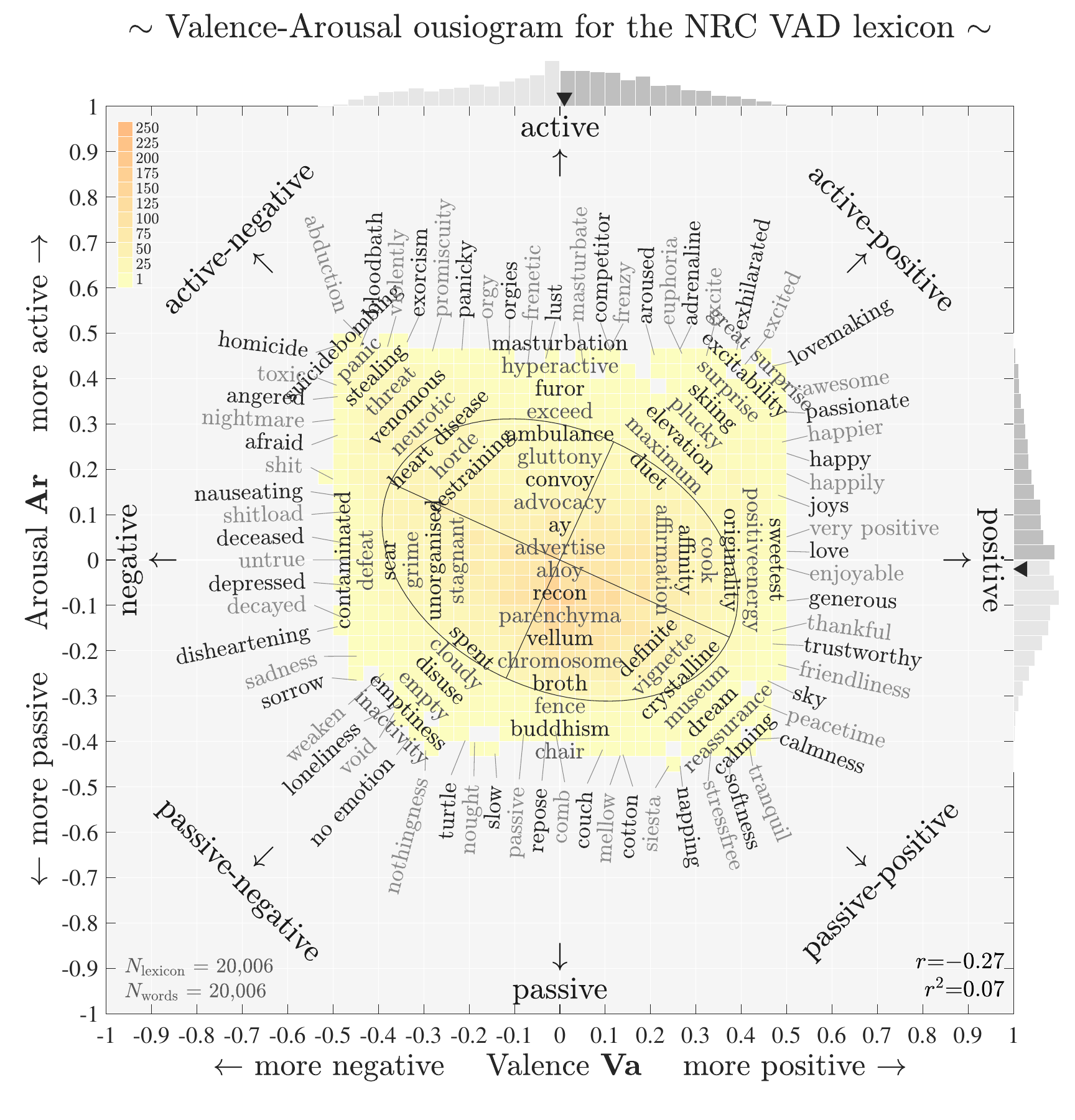}
  \caption{
    \textbf{
      Ousiogram for arousal vs.\ valence in the VAD framework.
    }
  }
  \label{fig:meaning.ousiometer9010_VAD012_1_supp1}
\end{figure*}

\begin{figure*}[tp!]
  \centerfloat
  \includegraphics[width=1.1\textwidth]{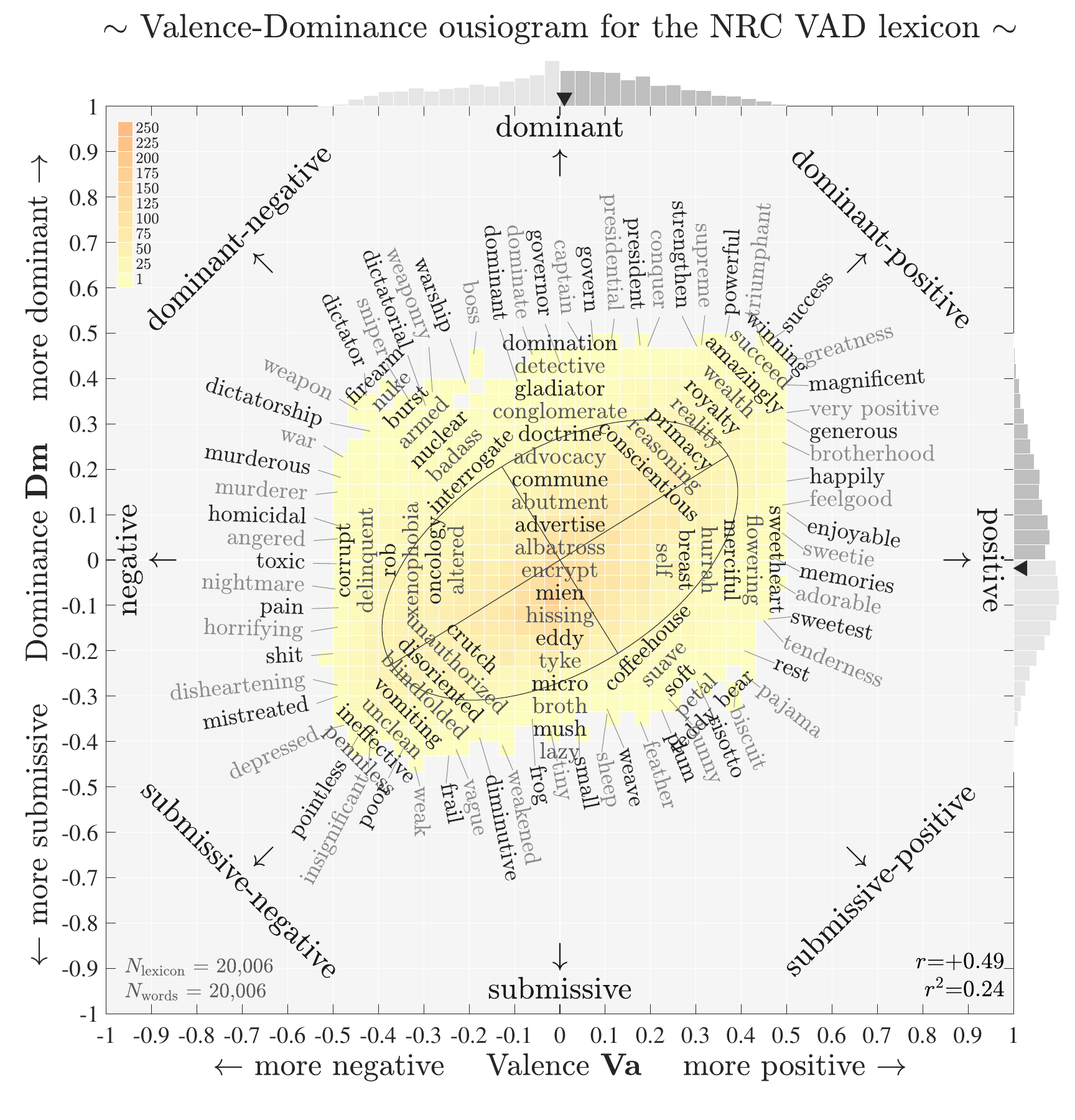}
  \caption{
    \textbf{
      Ousiogram for dominance vs.\ valence in the VAD framework.
      Matches Fig.~\ref{fig:meaning.ousiometer9010_VAD012_2_noname}
      in the main paper.
    }
  }
  \label{fig:meaning.ousiometer9010_VAD012_2_supp1}
\end{figure*}

\begin{figure*}[tp!]
  \centerfloat
  \includegraphics[width=1.1\textwidth]{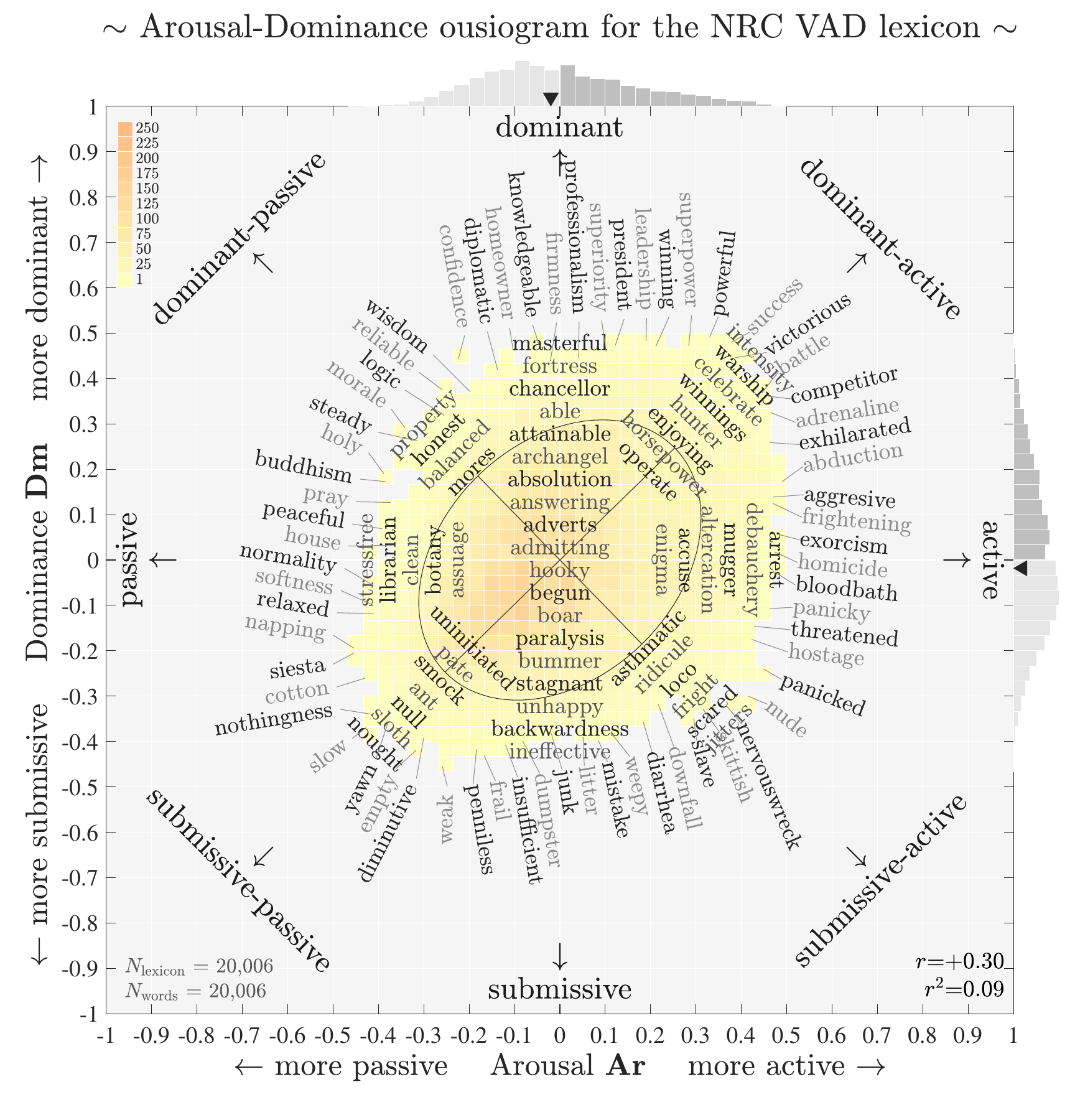}
  \caption{
    \textbf{
      Ousiogram for arousal vs.\ dominance in the VAD framework.
    }
  }
  \label{fig:meaning.ousiometer9010_VAD012_3_supp1}
\end{figure*}

\begin{figure*}[tp!]
  \centerfloat
  \includegraphics[width=1.1\textwidth]{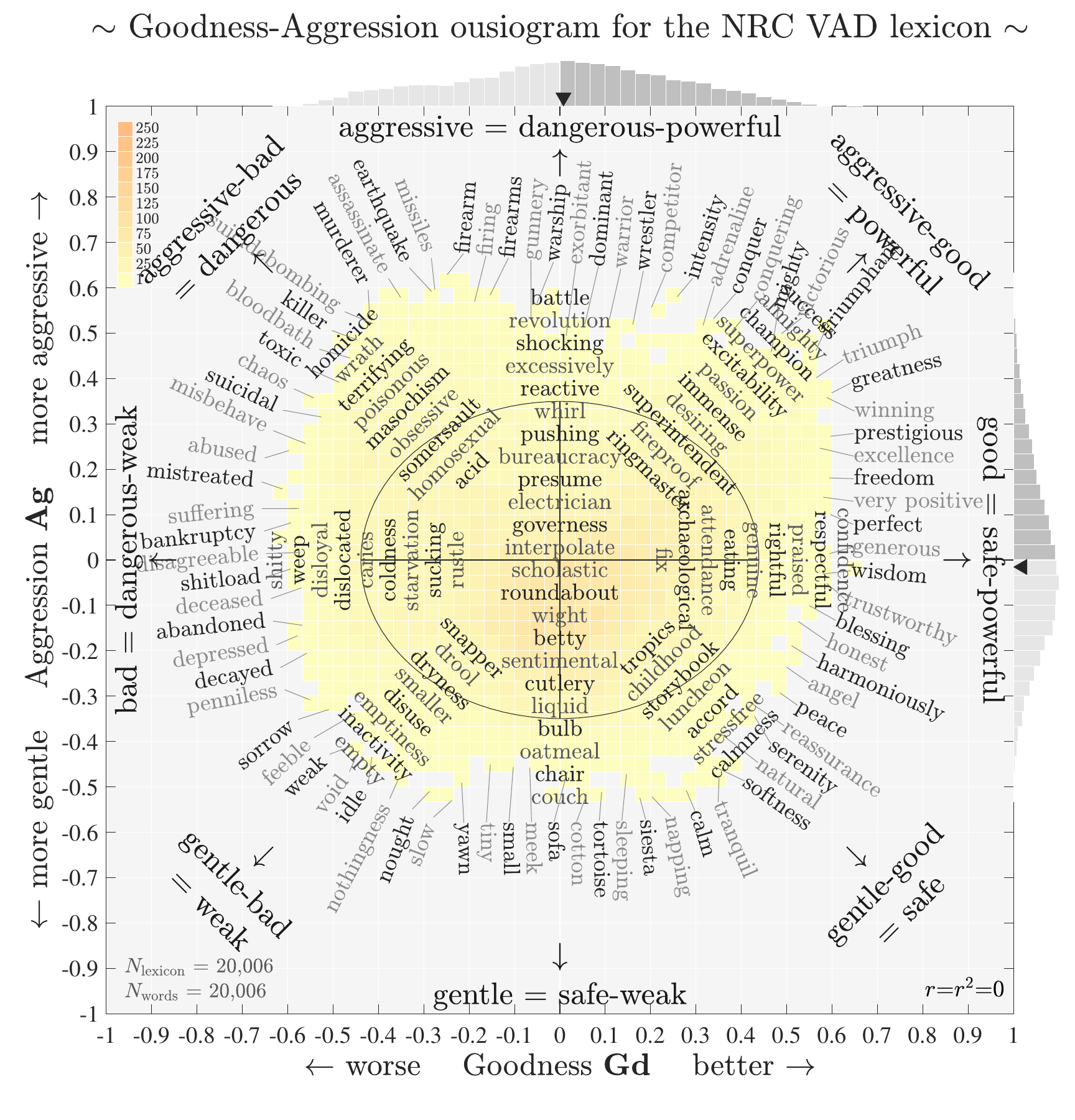}
  \caption{
    \textbf{
      Ousiogram for aggression vs.\ goodness in the GAS framework.
    }
  }
  \label{fig:meaning.ousiometer9010_GAS012_1_supp1}
\end{figure*}

\begin{figure*}[tp!]
  \centerfloat
  \includegraphics[width=1.1\textwidth]{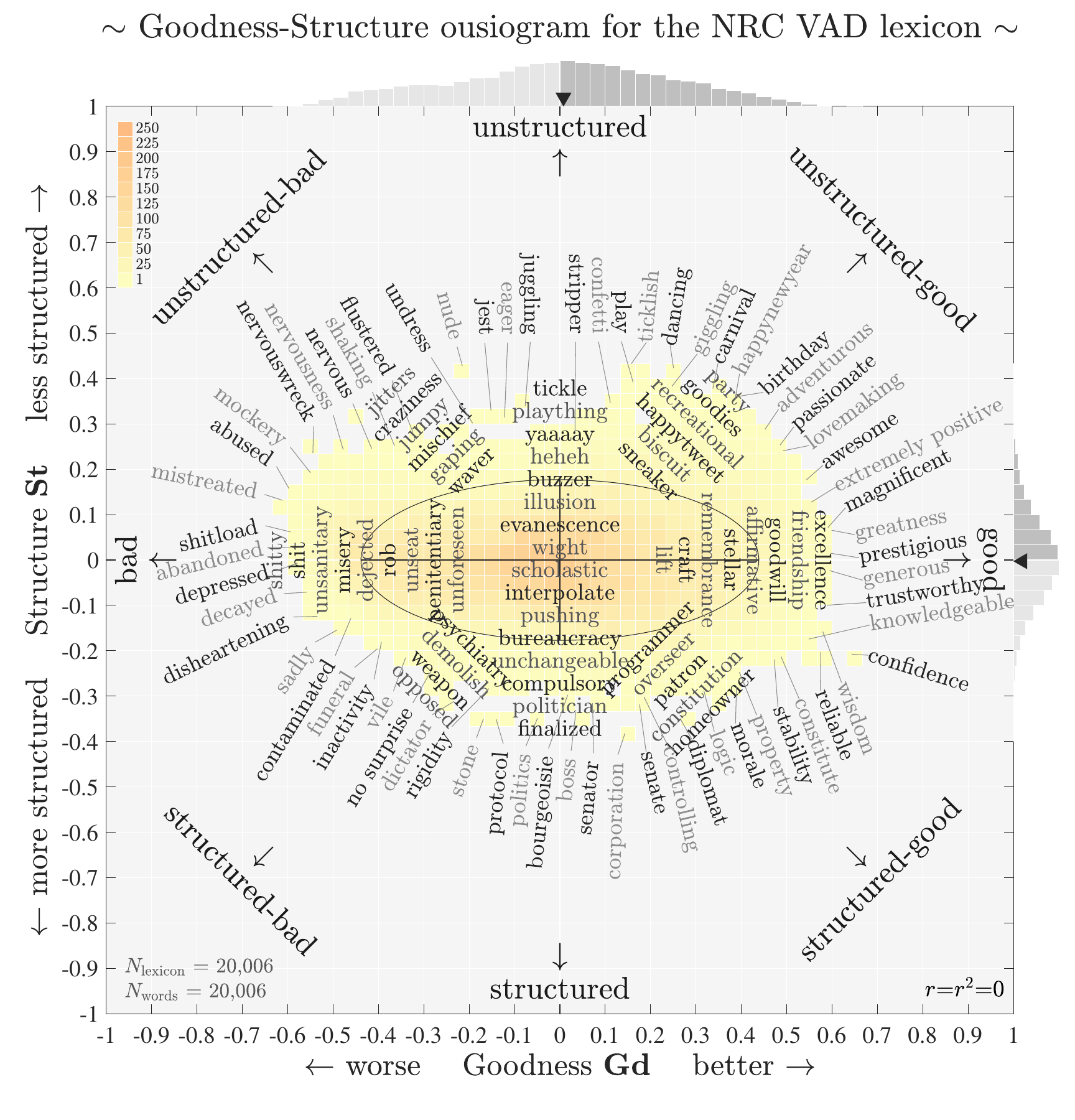}
  \caption{
    \textbf{
      Ousiogram for structure vs.\ goodness in the GAS framework.
    }
  }
  \label{fig:meaning.ousiometer9010_GAS012_2_supp1}
\end{figure*}

\begin{figure*}[tp!]
  \centerfloat
  \includegraphics[width=1.1\textwidth]{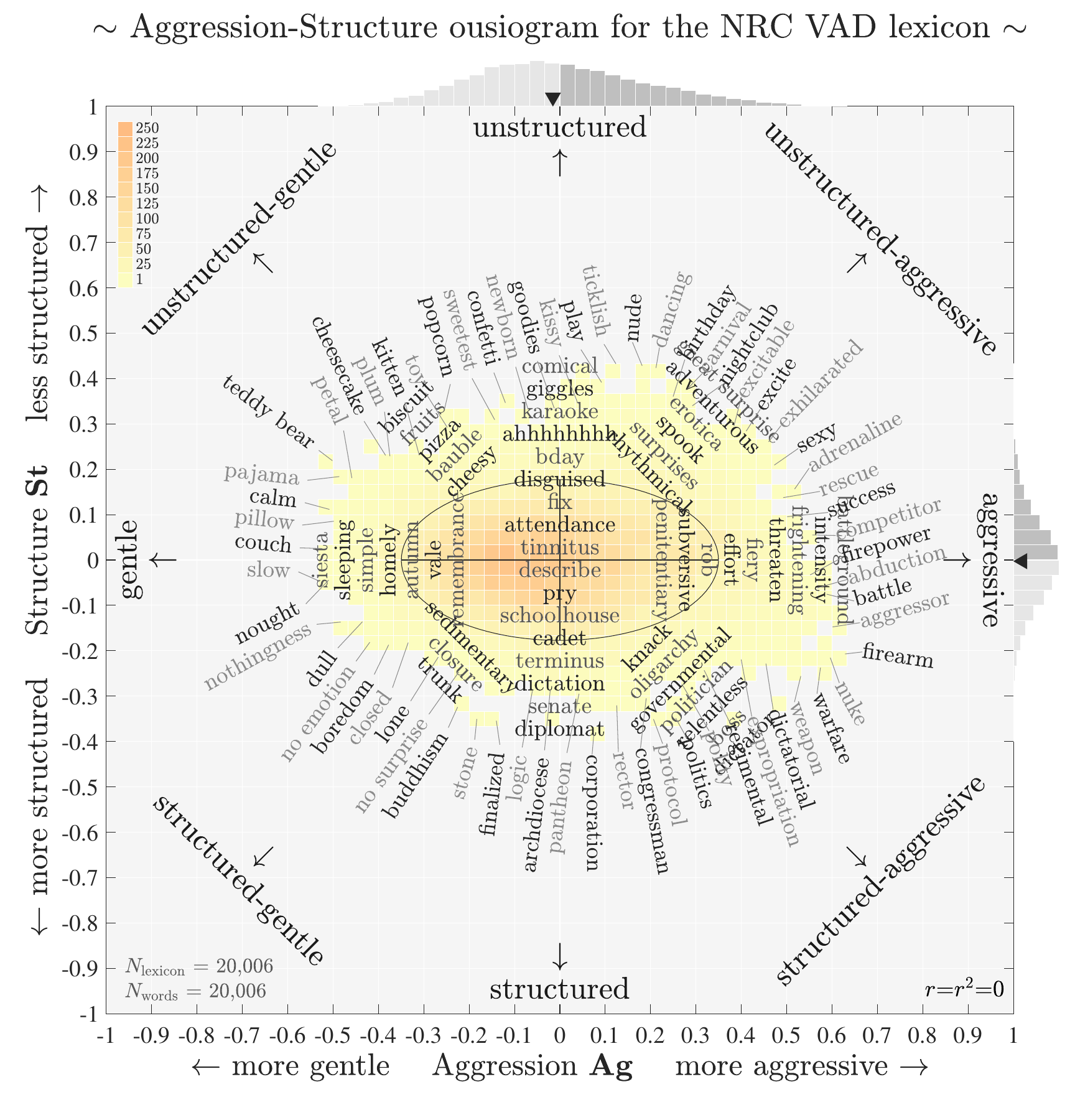}
  \caption{
    \textbf{
      Ousiogram for structure vs.\ aggression in the GAS framework.
    }
  }
  \label{fig:meaning.ousiometer9010_GAS012_3_supp1}
\end{figure*}

\begin{figure*}[tp!]
  \centerfloat
  \includegraphics[width=1.1\textwidth]{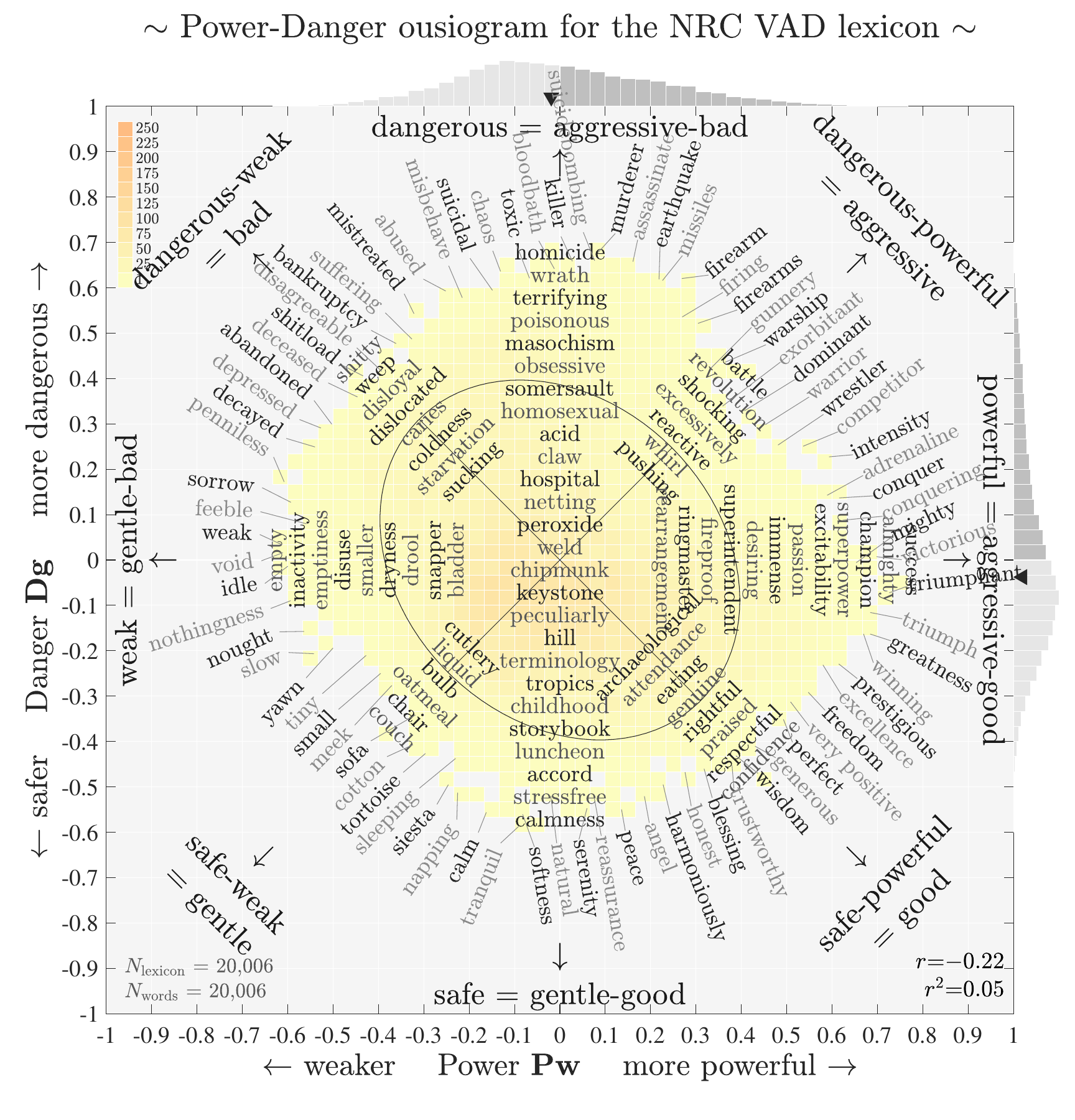}
  \caption{
    \textbf{
      Ousiogram for danger vs.\ power in the PDS framework.
      Matches Fig.~\ref{fig:meaning.ousiometer9010_PDS012_1}
      in the main paper.
    }
  }
  \label{fig:meaning.ousiometer9010_PDS012_1_supp1}
\end{figure*}

\begin{figure*}[tp!]
  \centerfloat
  \includegraphics[width=1.1\textwidth]{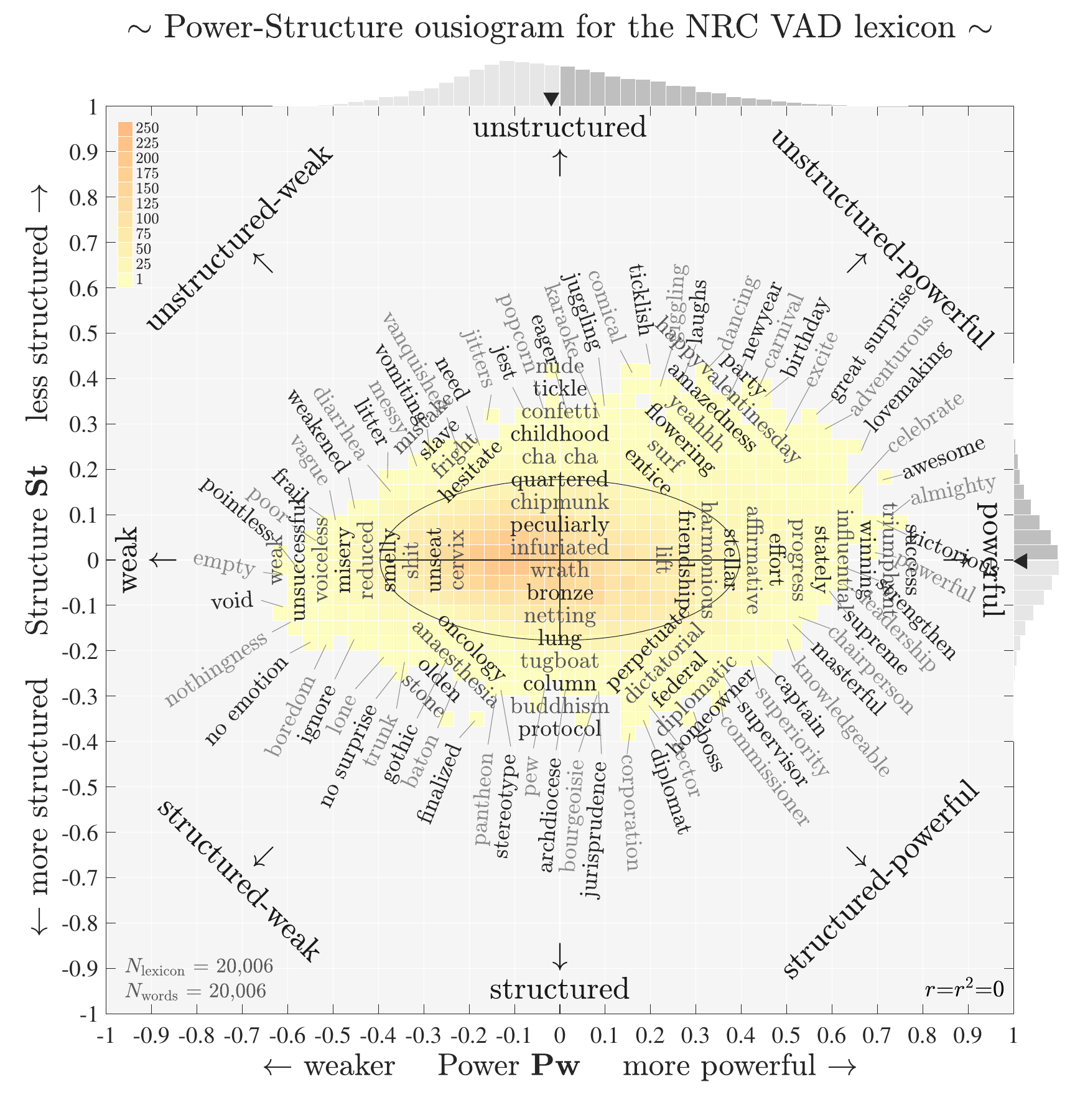}
  \caption{
    \textbf{
      Ousiogram for structure vs.\ power in the PDS framework.
    }
  }
  \label{fig:meaning.ousiometer9010_PDS012_2_supp1}
\end{figure*}

\begin{figure*}[tp!]
  \centerfloat
  \includegraphics[width=1.1\textwidth]{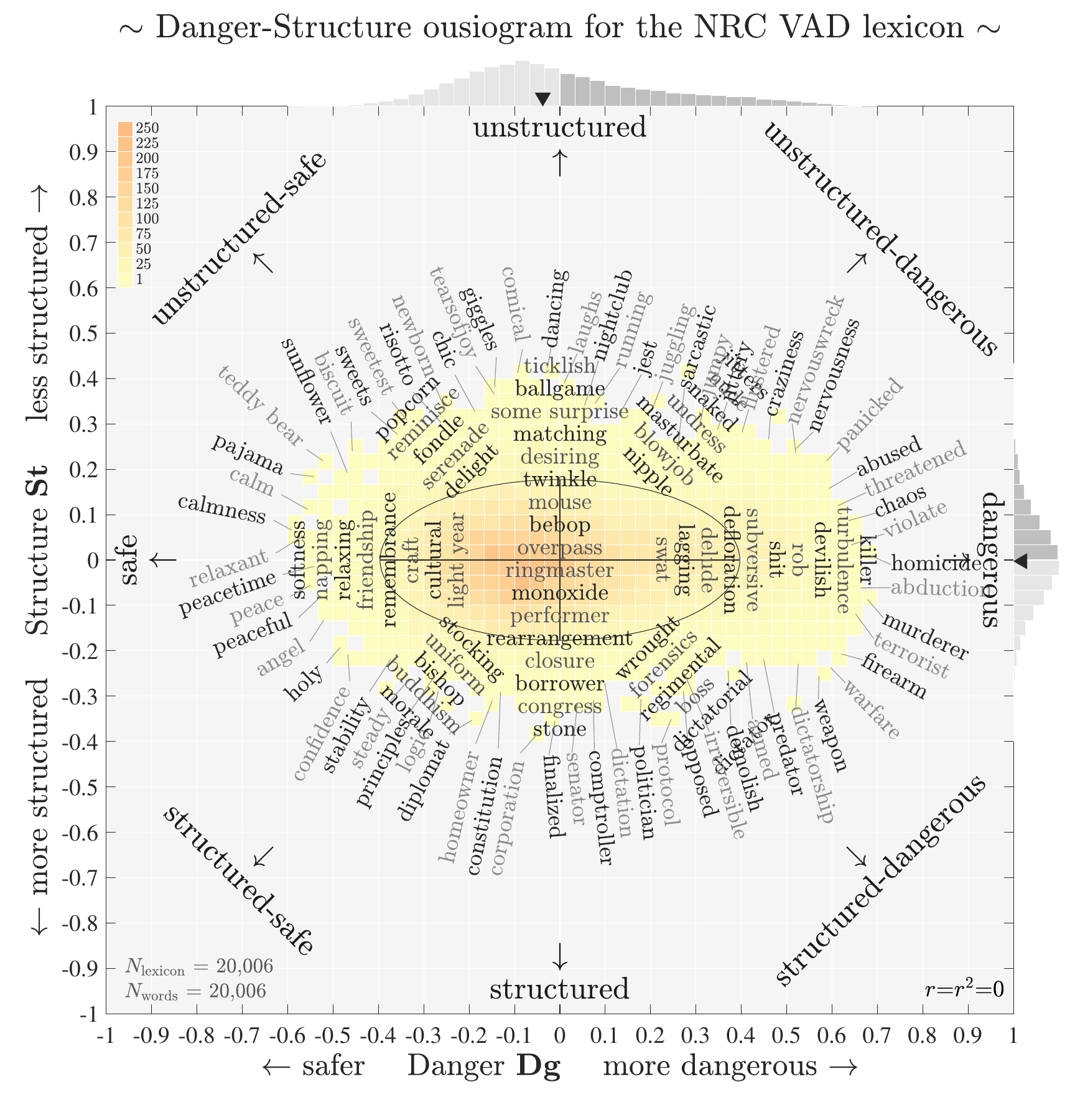}
  \caption{
    \textbf{
      Ousiogram for structure vs.\ danger in the PDS framework.
    }
  }
  \label{fig:meaning.ousiometer9010_PDS012_3_supp1}
\end{figure*}

\begin{figure*}[t]
  \includegraphics[width=\textwidth]{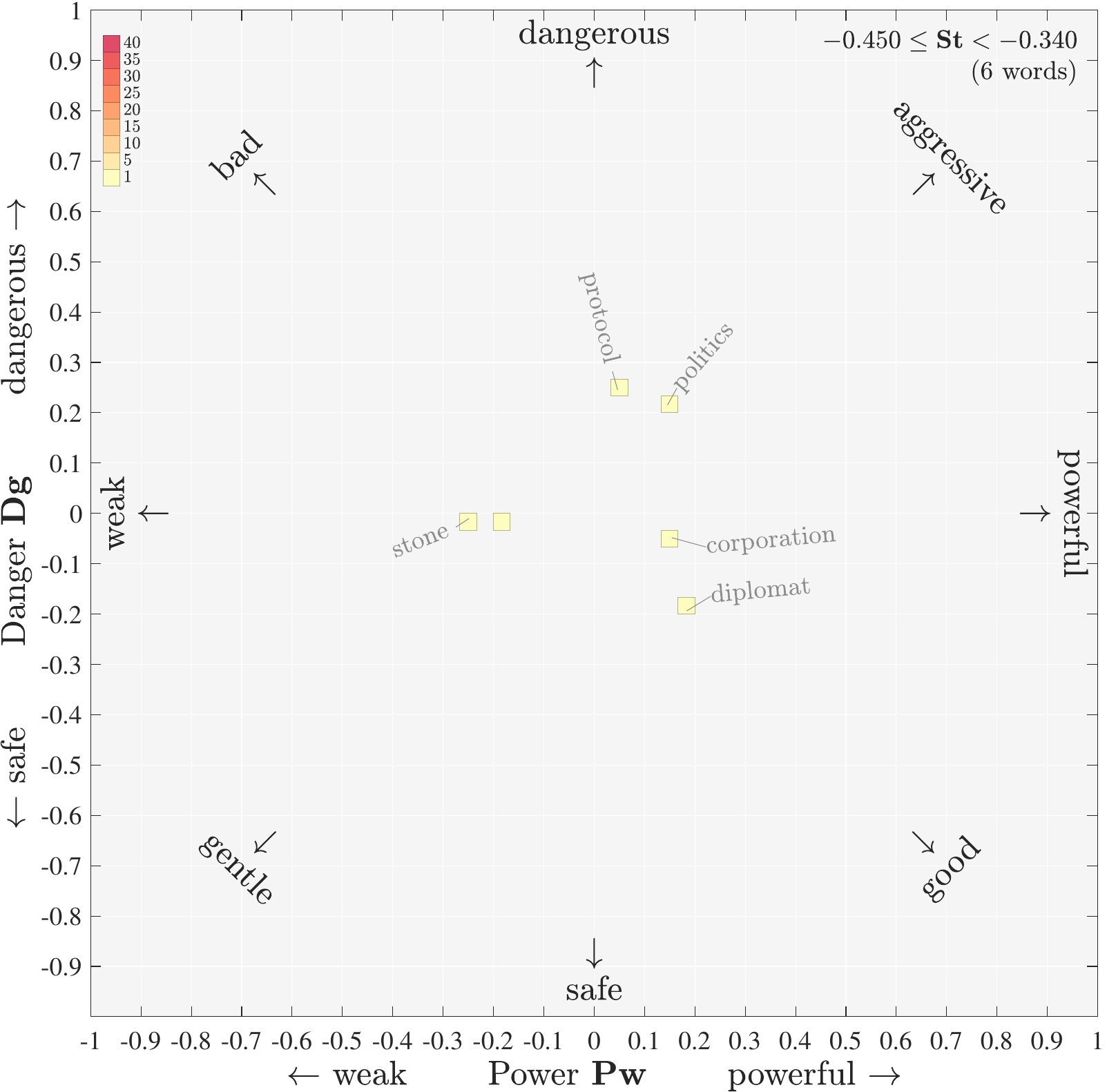}
  \caption{
    \textbf{
      Ousiometric slice for power-danger plane with structure: $-0.450 \le \textbf{St} < -0.340$.
    }
  }
  \label{fig.meaning:figousiometer_3d_MRI_slices100_01_noname}
\end{figure*}

\begin{figure*}[t]
  \includegraphics[width=\textwidth]{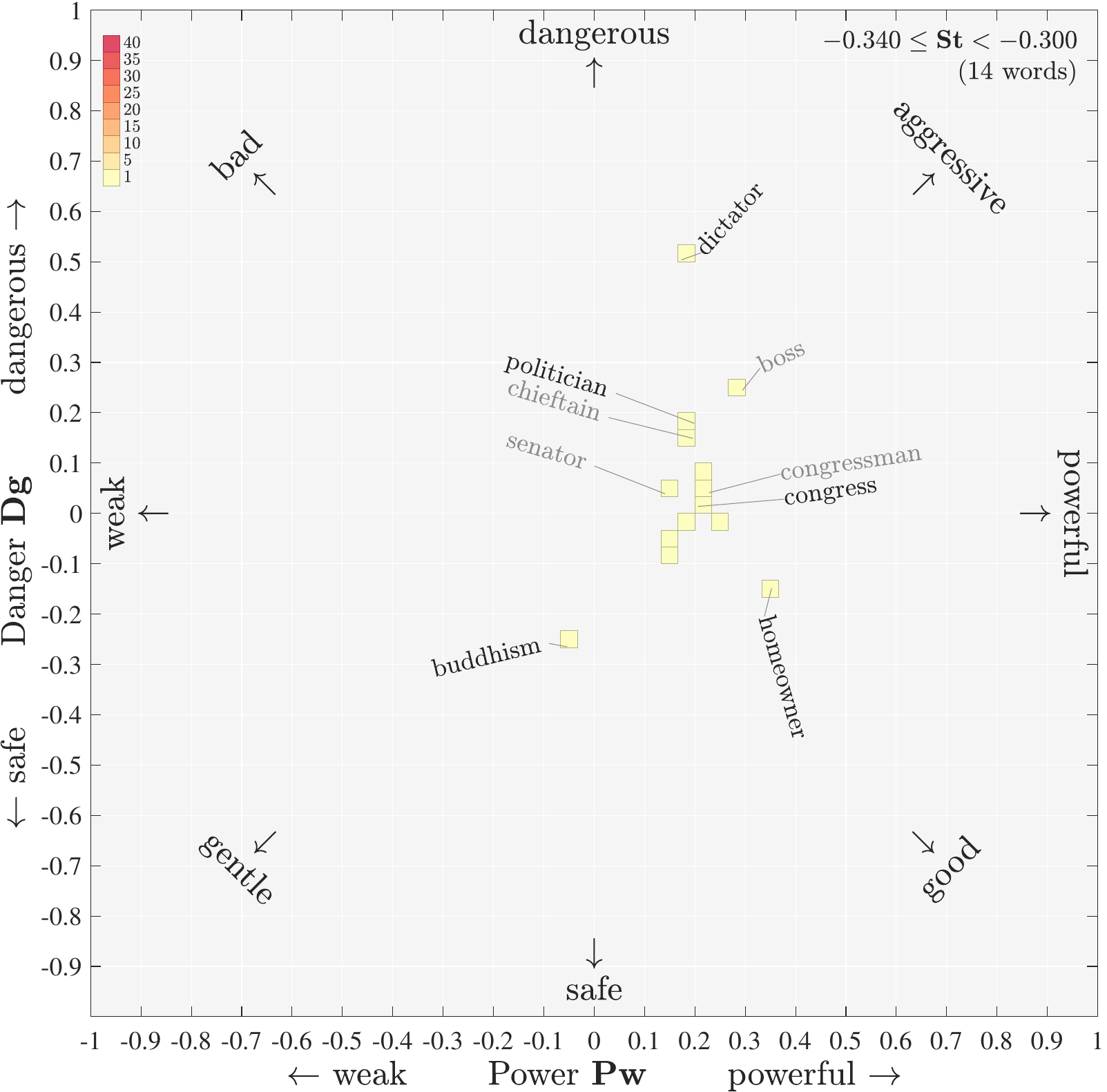}
  \caption{
    \textbf{
      Ousiometric slice for power-danger plane with structure: $-0.340 \le \textbf{St} < -0.300$.
    }
  }
  \label{fig.meaning:figousiometer_3d_MRI_slices100_02_noname}
\end{figure*}

\begin{figure*}[t]
  \includegraphics[width=\textwidth]{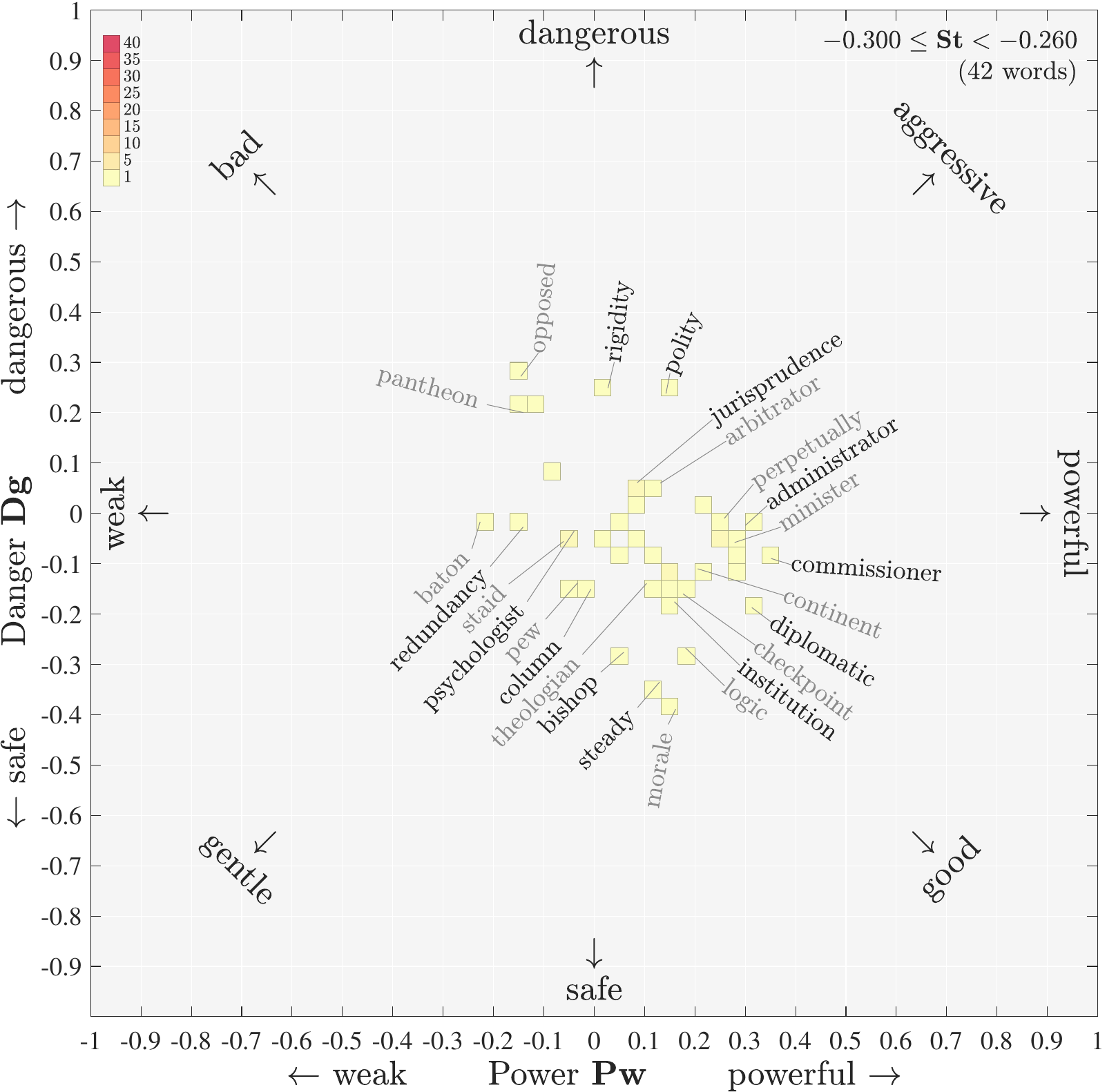}
  \caption{
    \textbf{
      Ousiometric slice for power-danger plane with structure: $-0.300 \le \textbf{St} < -0.260$.
    }
  }
  \label{fig.meaning:figousiometer_3d_MRI_slices100_03_noname}
\end{figure*}

\begin{figure*}[t]
  \includegraphics[width=\textwidth]{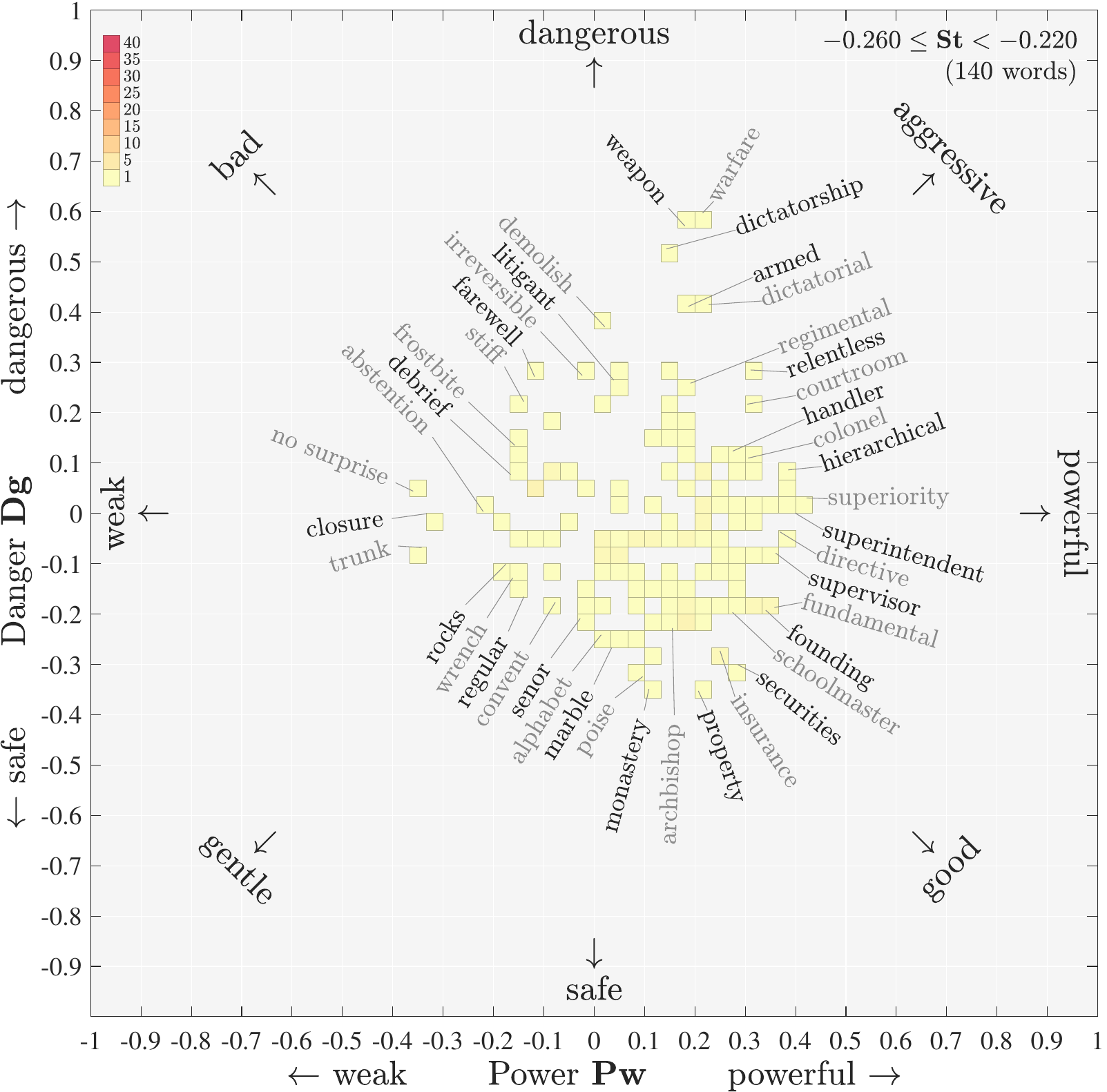}
  \caption{
    \textbf{
      Ousiometric slice for power-danger plane with structure: $-0.260 \le \textbf{St} < -0.220$.
    }
  }
  \label{fig.meaning:figousiometer_3d_MRI_slices100_04_noname}
\end{figure*}

\begin{figure*}[t]
  \includegraphics[width=\textwidth]{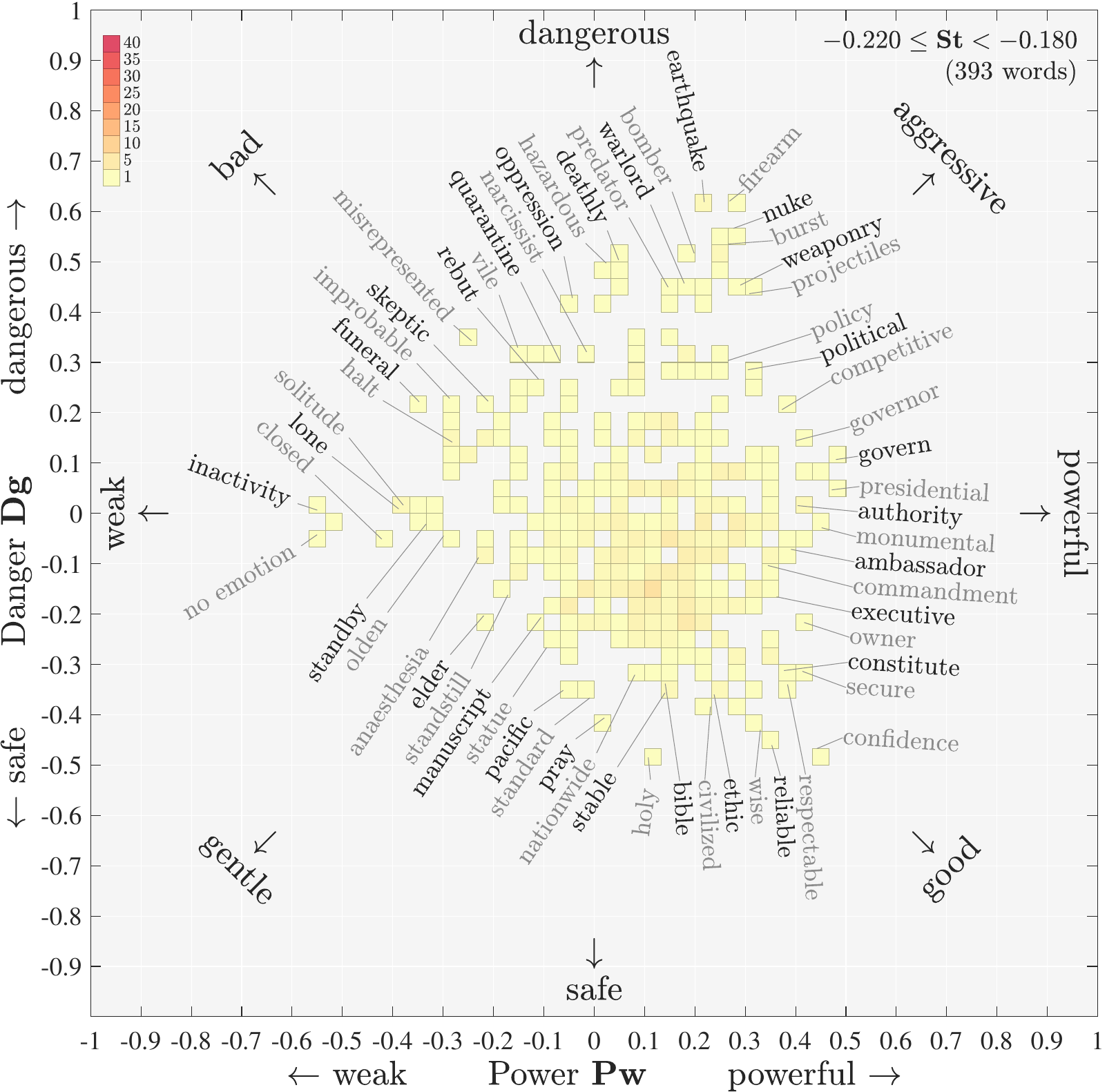}
  \caption{
    \textbf{
      Ousiometric slice for power-danger plane with structure: $-0.220 \le \textbf{St} < -0.180$.
    }
  }
  \label{fig.meaning:figousiometer_3d_MRI_slices100_05_noname}
\end{figure*}

\begin{figure*}[t]
  \includegraphics[width=\textwidth]{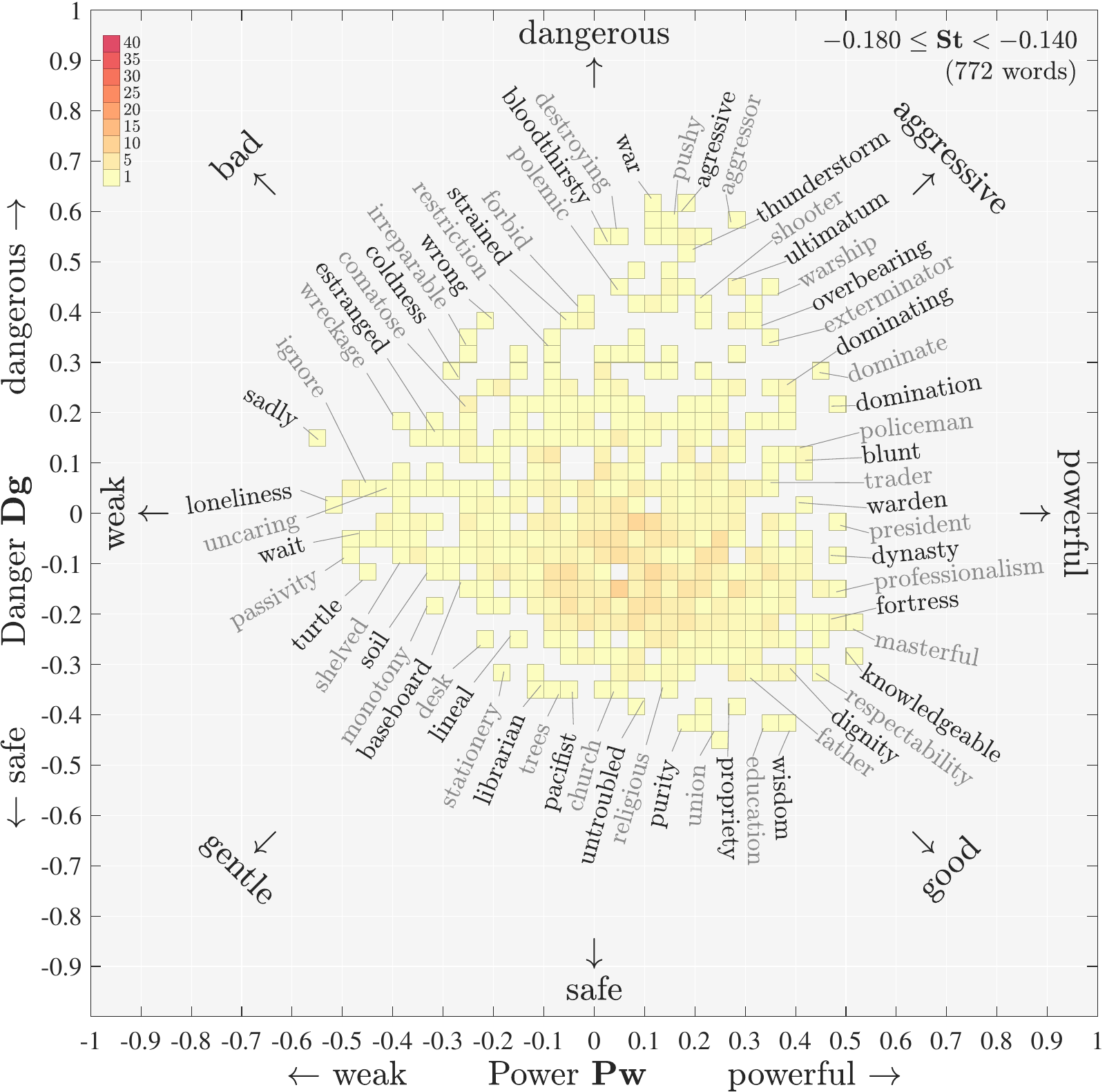}
  \caption{
    \textbf{
      Ousiometric slice for power-danger plane with structure: $-0.180 \le \textbf{St} < -0.140$.
    }
  }
  \label{fig.meaning:figousiometer_3d_MRI_slices100_06_noname}
\end{figure*}

\begin{figure*}[t]
  \includegraphics[width=\textwidth]{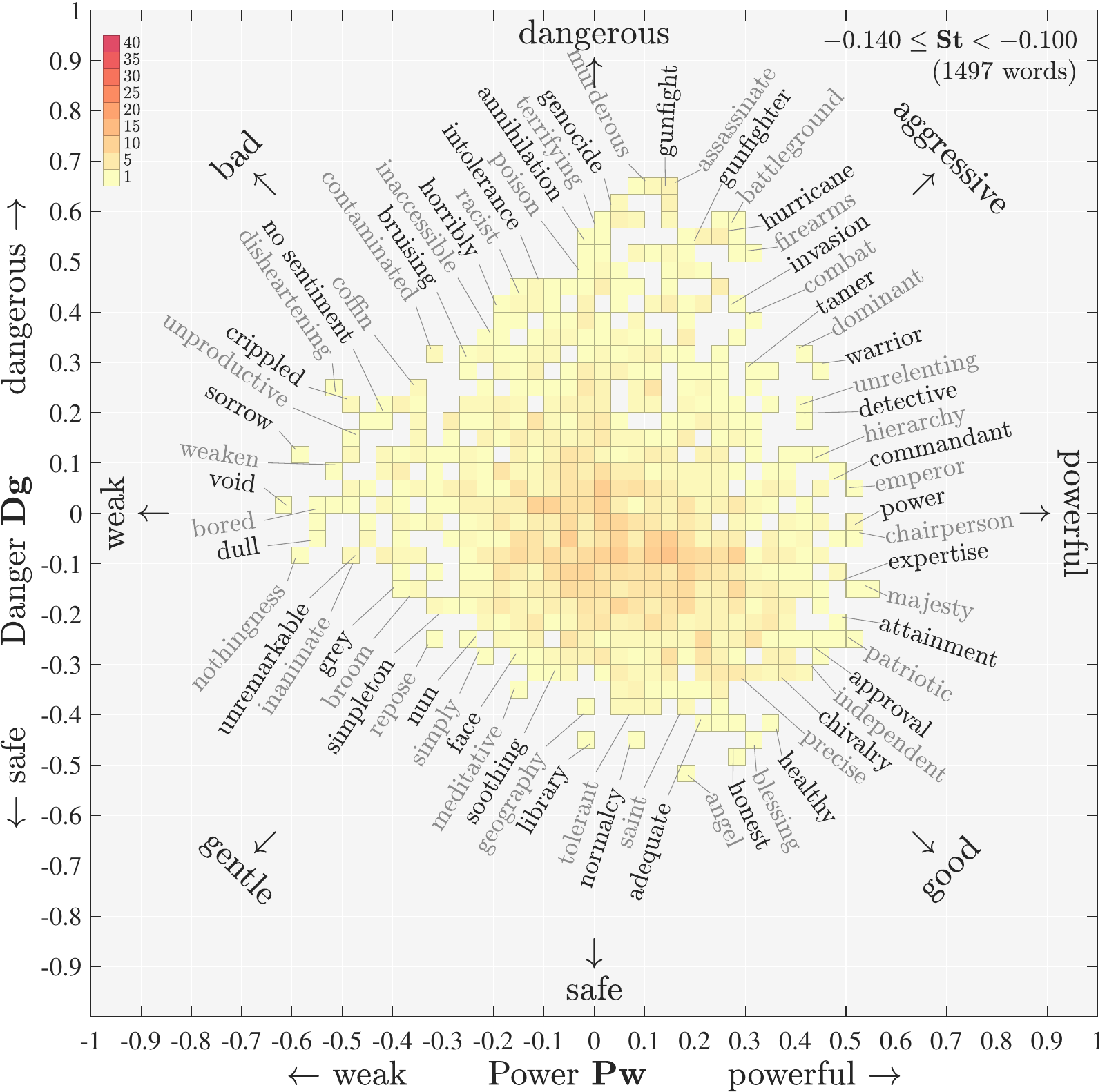}
  \caption{
    \textbf{
      Ousiometric slice for power-danger plane with structure: $-0.140 \le \textbf{St} < -0.100$.
    }
  }
  \label{fig.meaning:figousiometer_3d_MRI_slices100_07_noname}
\end{figure*}

\begin{figure*}[t]
  \includegraphics[width=\textwidth]{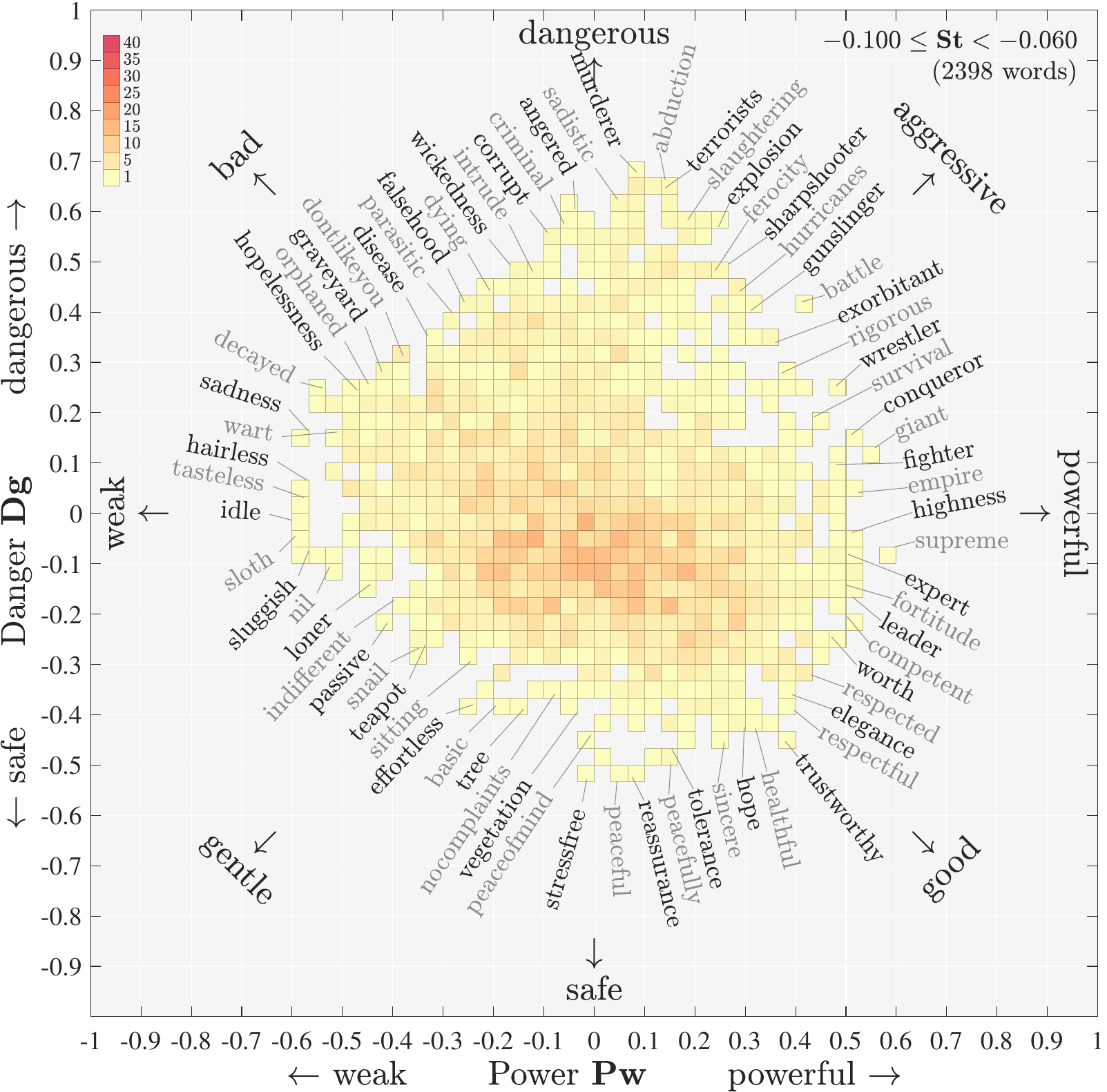}
  \caption{
    \textbf{
      Ousiometric slice for power-danger plane with structure: $-0.100 \le \textbf{St} < -0.060$.
    }
  }
  \label{fig.meaning:figousiometer_3d_MRI_slices100_08_noname}
\end{figure*}

\begin{figure*}[t]
  \includegraphics[width=\textwidth]{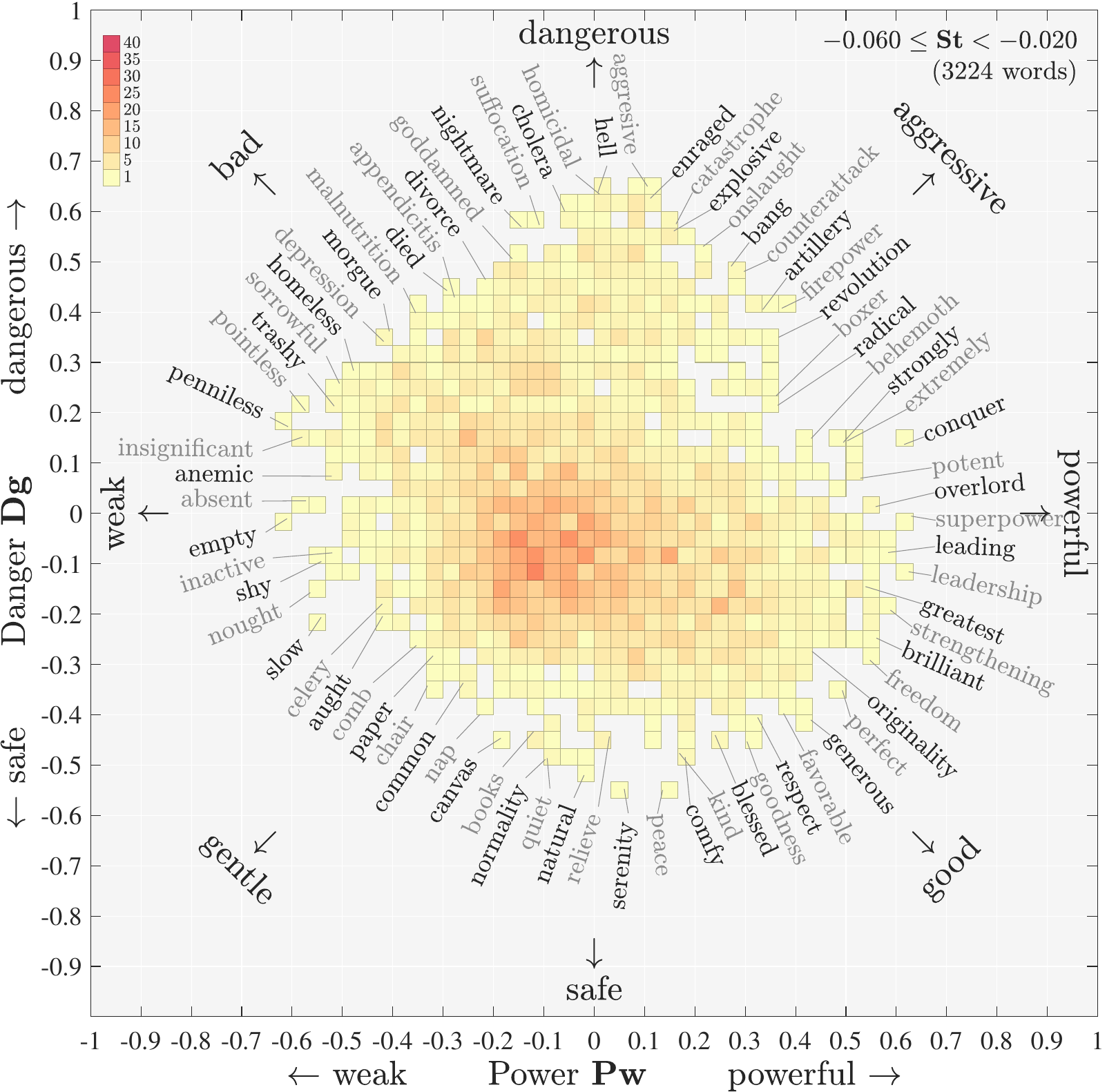}
  \caption{
    \textbf{
      Ousiometric slice for power-danger plane with structure: $-0.060 \le \textbf{St} < -0.020$.
    }
  }
  \label{fig.meaning:figousiometer_3d_MRI_slices100_09_noname}
\end{figure*}

\begin{figure*}[t]
  \includegraphics[width=\textwidth]{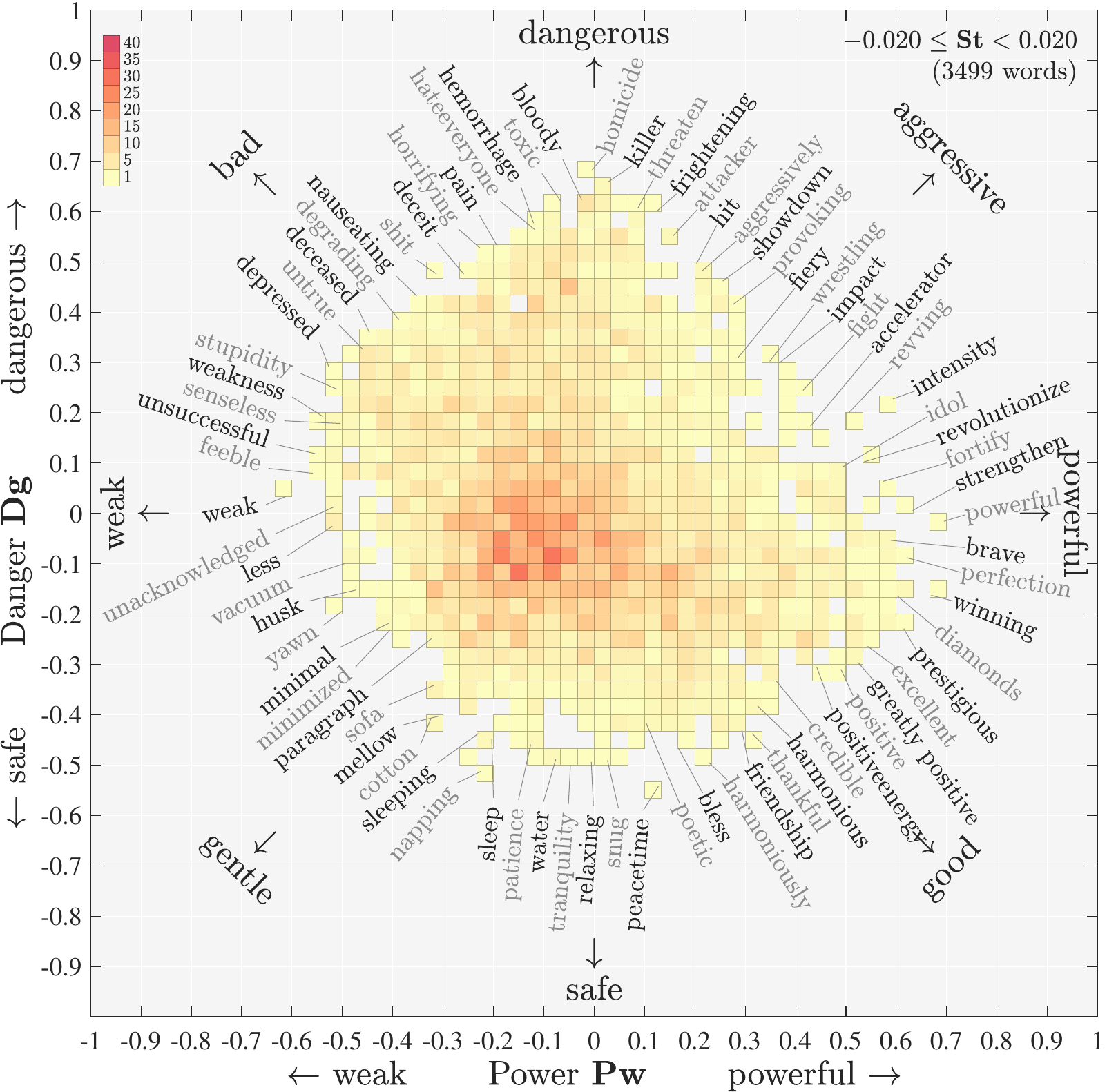}
  \caption{
    \textbf{
      Ousiometric slice for power-danger plane with structure: $-0.020 \le \textbf{St} < 0.020$.
    }
  }
  \label{fig.meaning:figousiometer_3d_MRI_slices100_10_noname}
\end{figure*}

\begin{figure*}[t]
  \includegraphics[width=\textwidth]{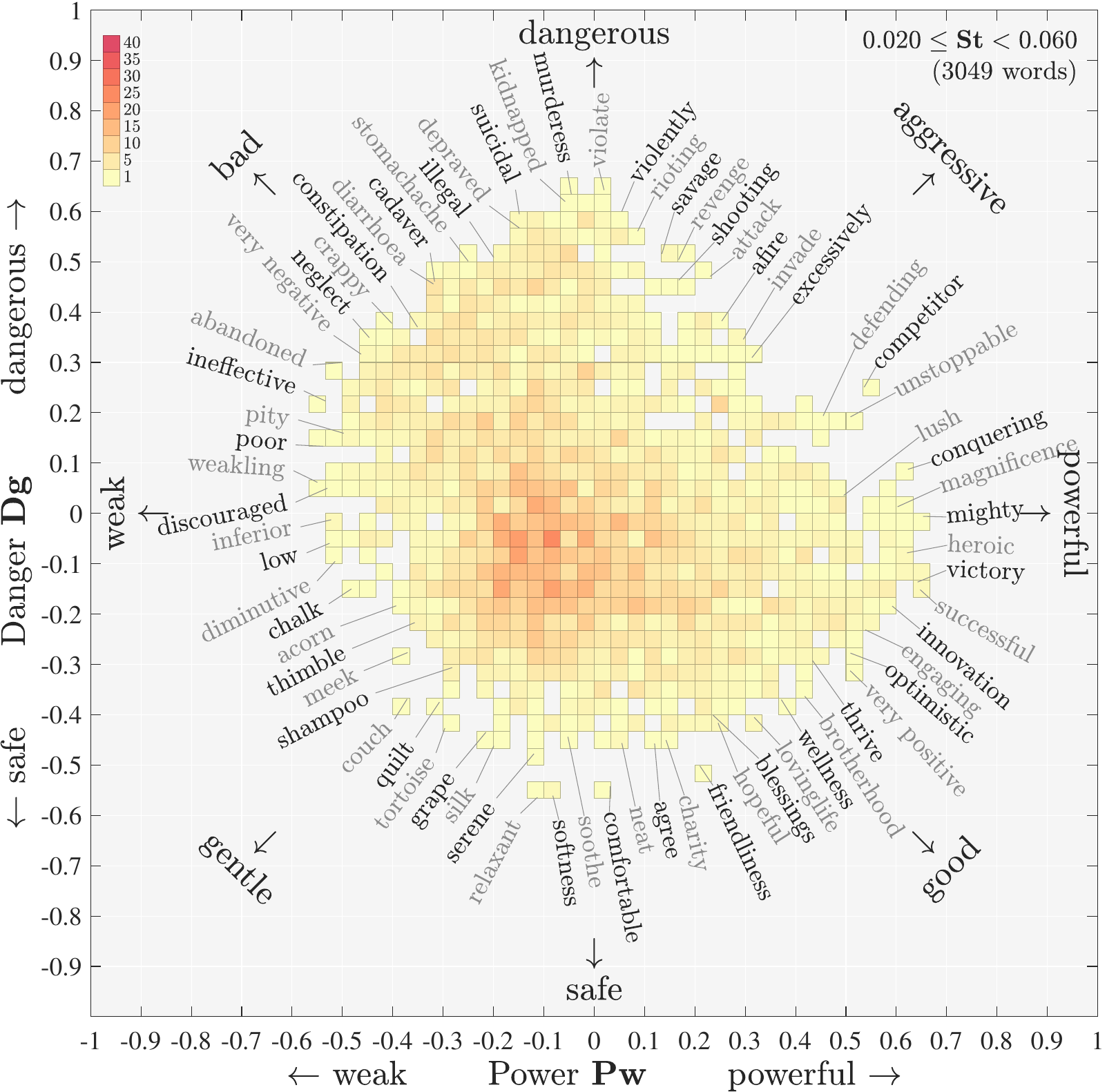}
  \caption{
    \textbf{
      Ousiometric slice for power-danger plane with structure: $0.020 \le \textbf{St} < 0.060$.
    }
  }
  \label{fig.meaning:figousiometer_3d_MRI_slices100_11_noname}
\end{figure*}

\begin{figure*}[t]
  \includegraphics[width=\textwidth]{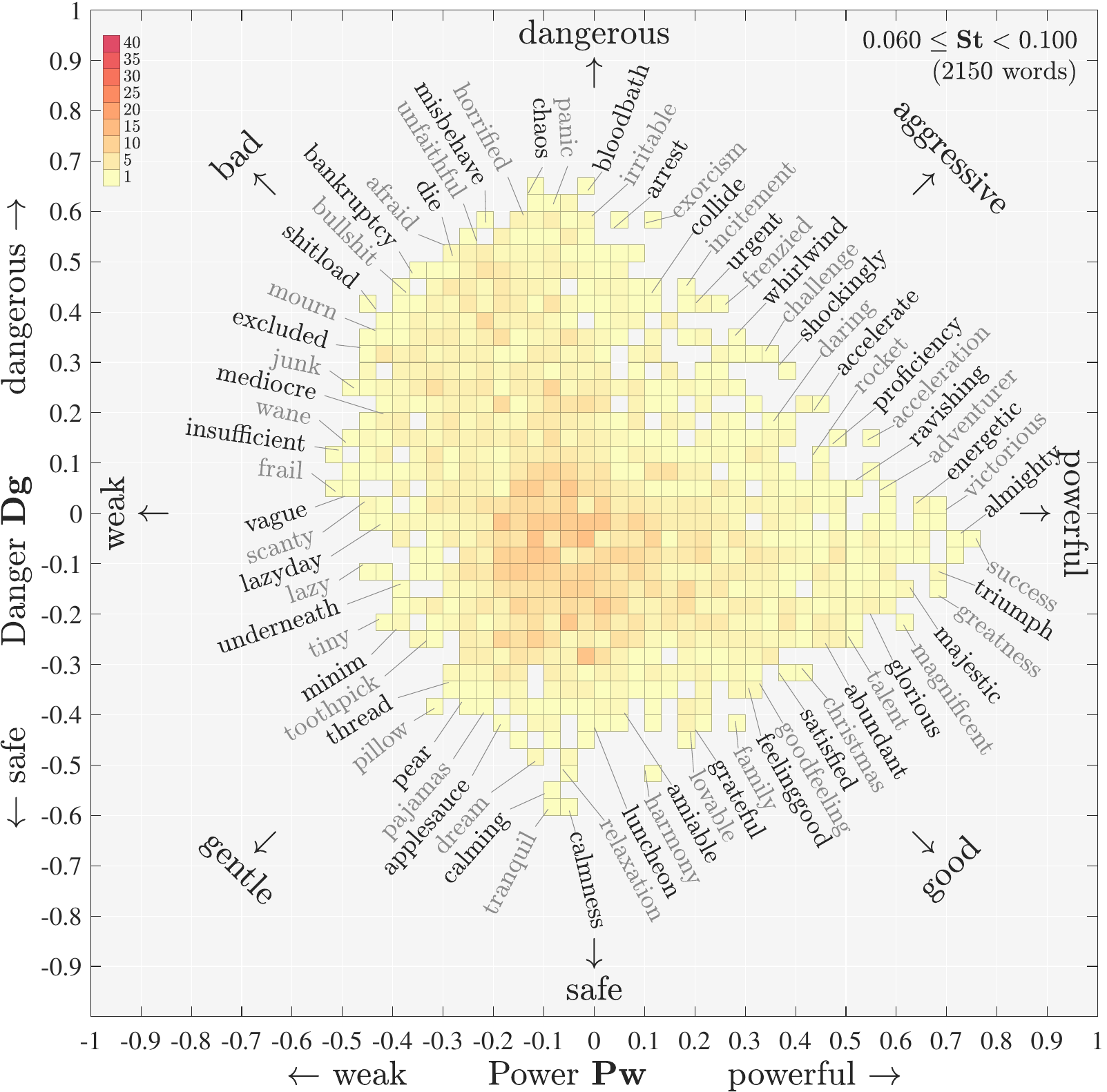}
  \caption{
    \textbf{
      Ousiometric slice for power-danger plane with structure: $0.060 \le \textbf{St} < 0.100$.
    }
  }
  \label{fig.meaning:figousiometer_3d_MRI_slices100_12_noname}
\end{figure*}

\begin{figure*}[t]
  \includegraphics[width=\textwidth]{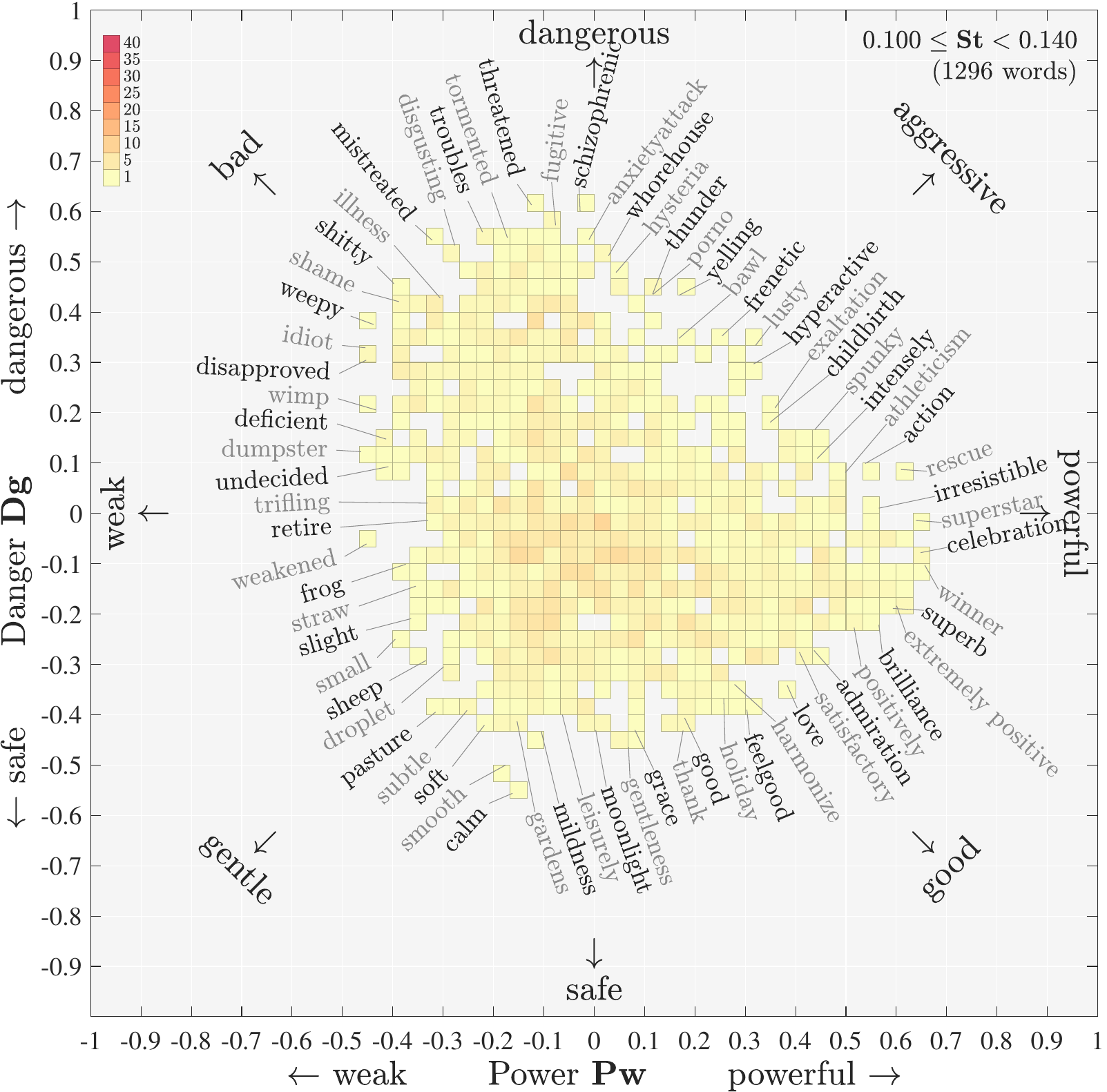}
  \caption{
    \textbf{
      Ousiometric slice for power-danger plane with structure: $0.100 \le \textbf{St} < 0.140$.
    }
  }
  \label{fig.meaning:figousiometer_3d_MRI_slices100_13_noname}
\end{figure*}

\begin{figure*}[t]
  \includegraphics[width=\textwidth]{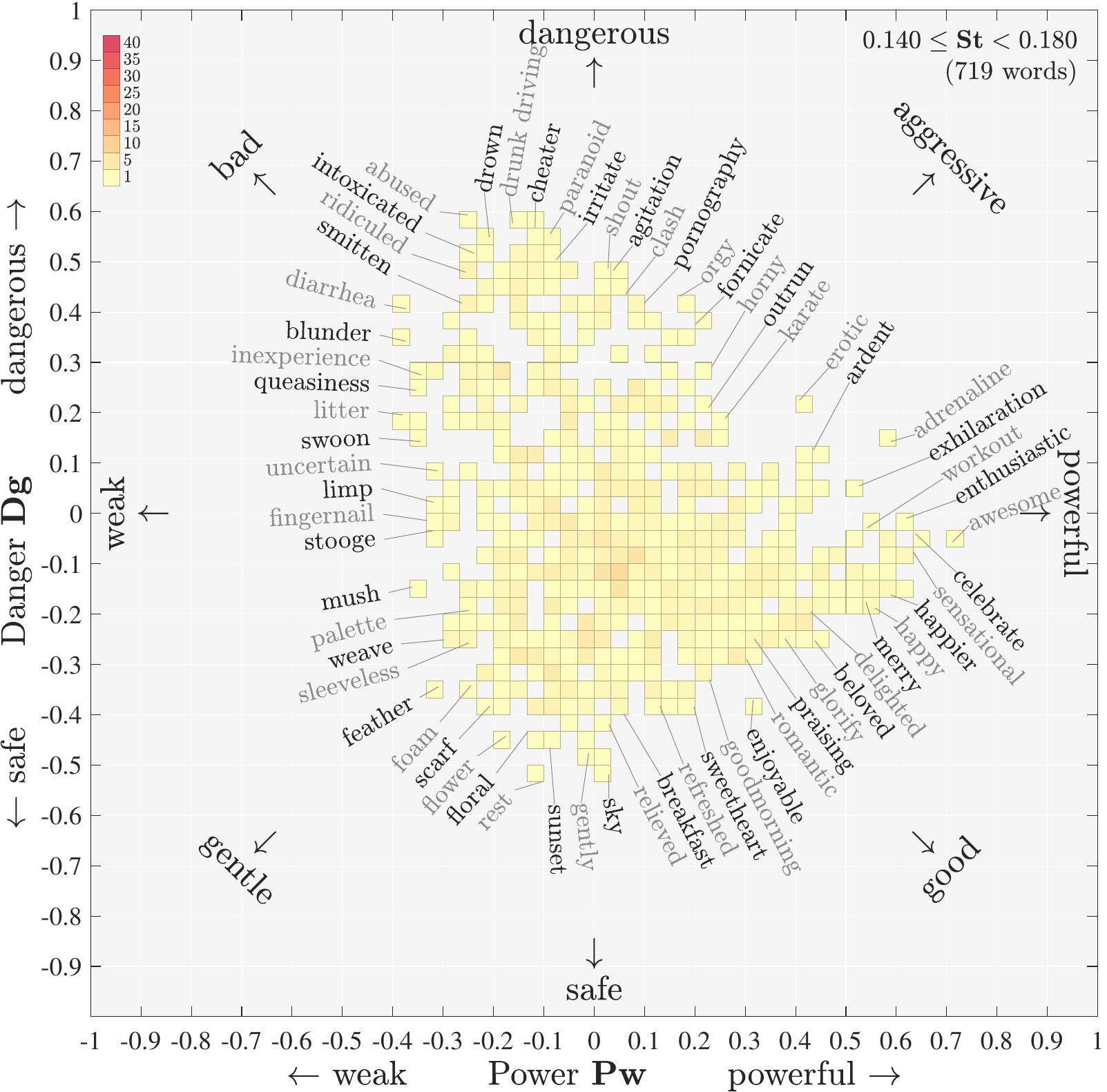}
  \caption{
    \textbf{
      Ousiometric slice for power-danger plane with structure: $0.140 \le \textbf{St} < 0.180$.
    }
  }
  \label{fig.meaning:figousiometer_3d_MRI_slices100_14_noname}
\end{figure*}

\begin{figure*}[t]
  \includegraphics[width=\textwidth]{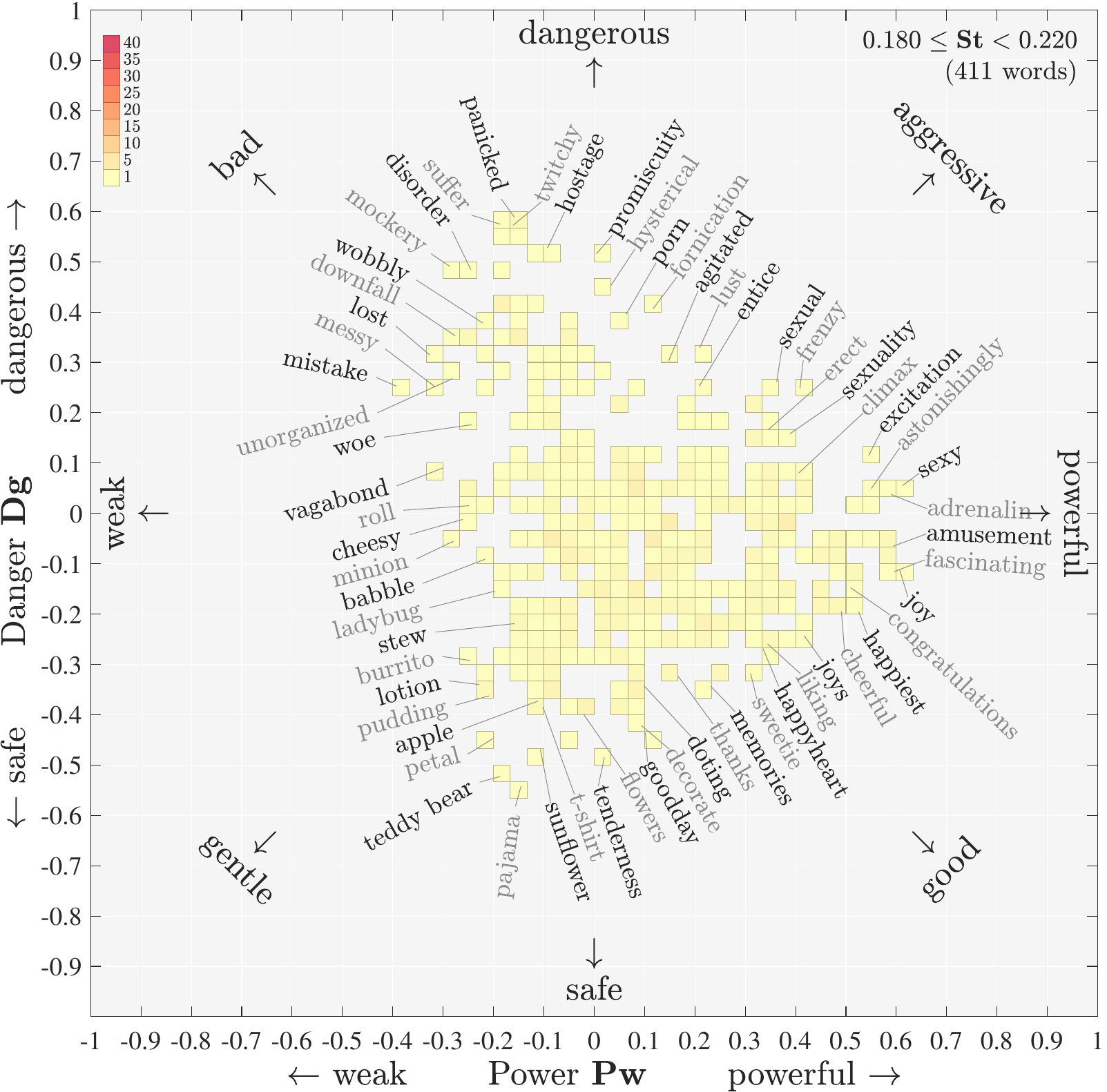}
  \caption{
    \textbf{
      Ousiometric slice for power-danger plane with structure: $0.180 \le \textbf{St} < 0.220$.
    }
  }
  \label{fig.meaning:figousiometer_3d_MRI_slices100_15_noname}
\end{figure*}

\begin{figure*}[t]
  \includegraphics[width=\textwidth]{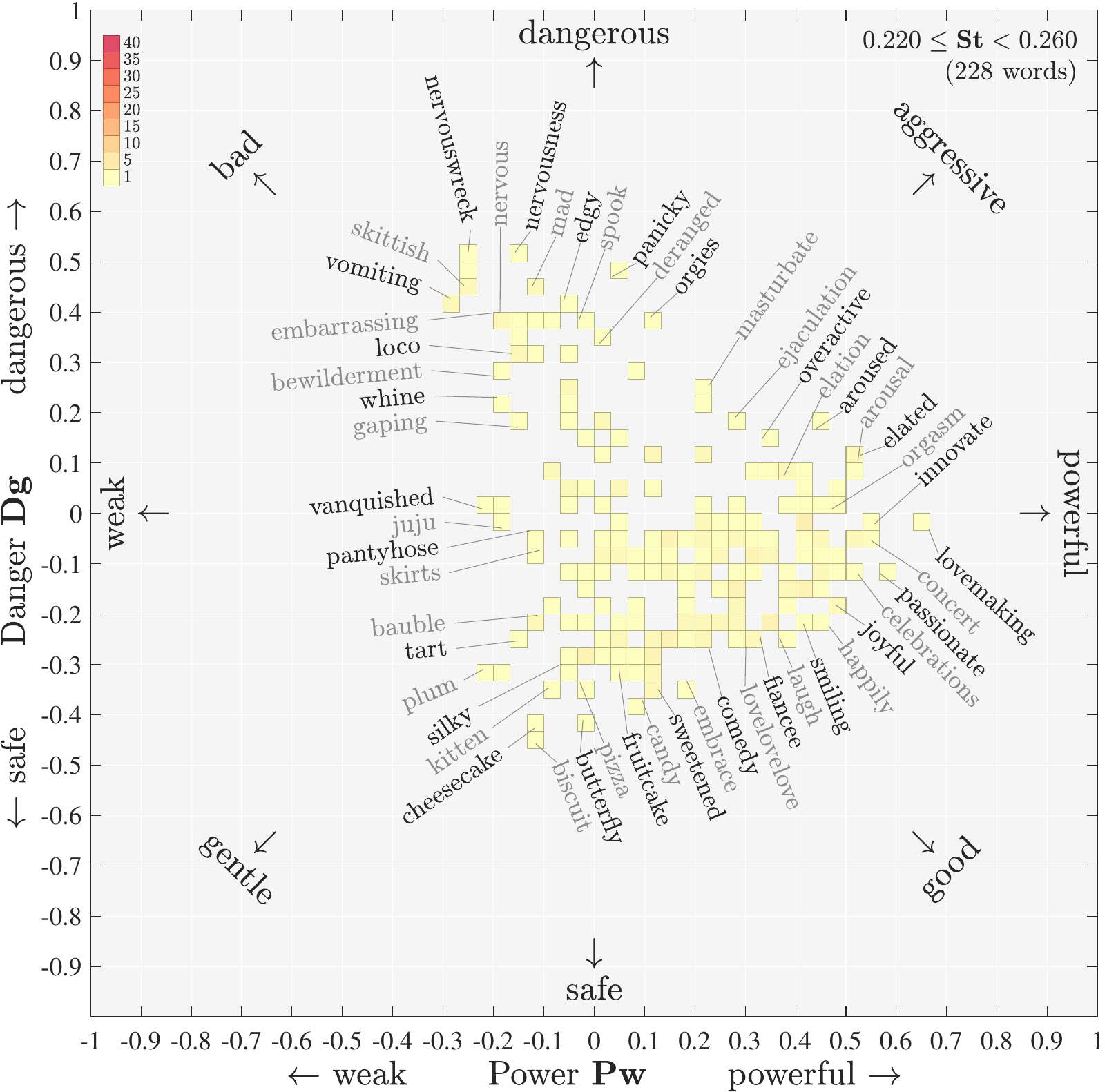}
  \caption{
    \textbf{
      Ousiometric slice for power-danger plane with structure: $0.220 \le \textbf{St} < 0.260$.
    }
  }
  \label{fig.meaning:figousiometer_3d_MRI_slices100_16_noname}
\end{figure*}

\begin{figure*}[t]
  \includegraphics[width=\textwidth]{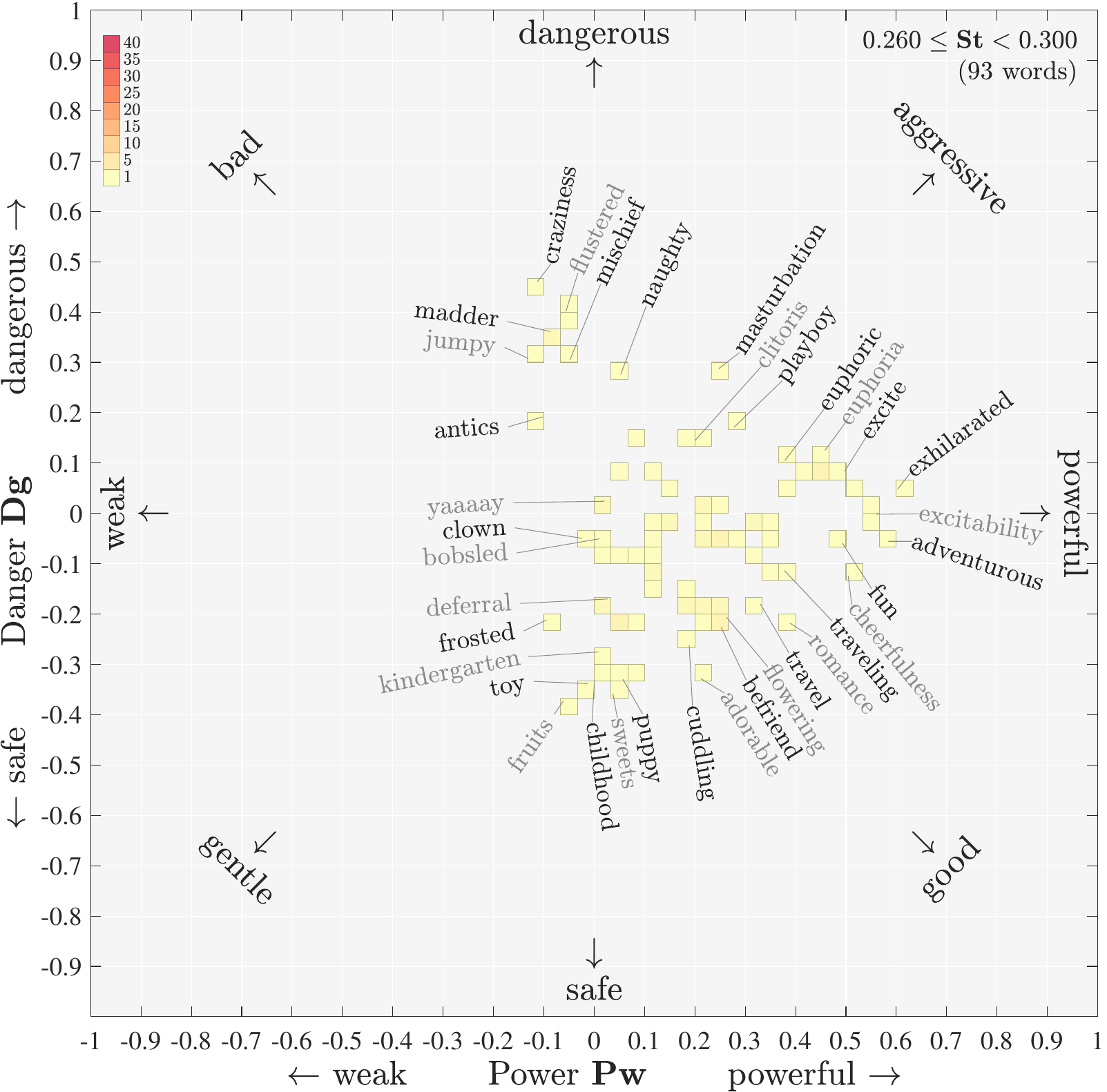}
  \caption{
    \textbf{
      Ousiometric slice for power-danger plane with structure: $0.260 \le \textbf{St} < 0.300$.
    }
  }
  \label{fig.meaning:figousiometer_3d_MRI_slices100_17_noname}
\end{figure*}

\begin{figure*}[t]
  \includegraphics[width=\textwidth]{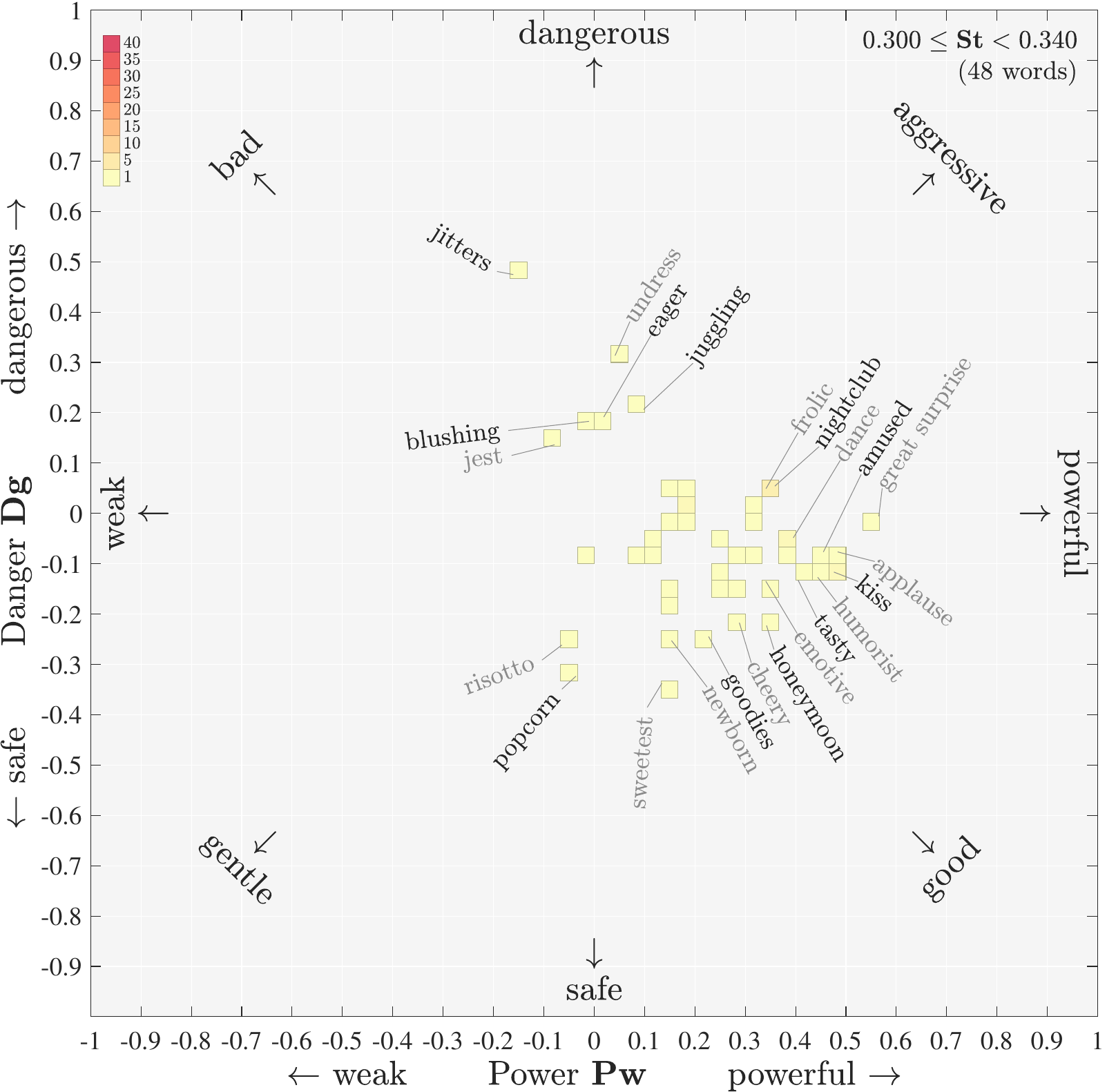}
  \caption{
    \textbf{
      Ousiometric slice for power-danger plane with structure: $0.300 \le \textbf{St} < 0.340$.
    }
  }
  \label{fig.meaning:figousiometer_3d_MRI_slices100_18_noname}
\end{figure*}

\begin{figure*}[t]
  \includegraphics[width=\textwidth]{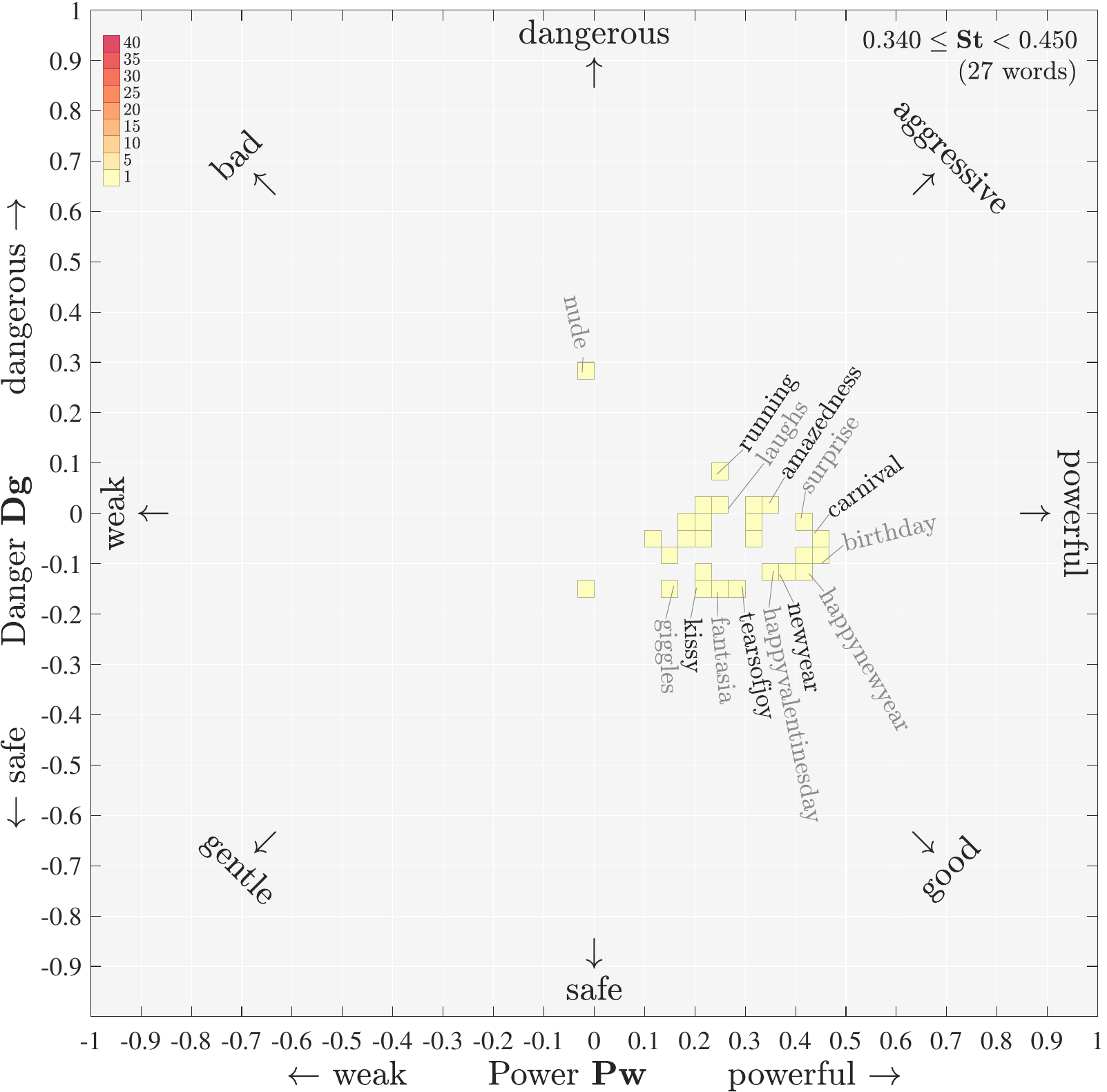}
  \caption{
    \textbf{
      Ousiometric slice for power-danger plane with structure: $0.340 \le \textbf{St} < 0.450$.
    }
  }
  \label{fig.meaning:figousiometer_3d_MRI_slices100_19_noname}
\end{figure*}

\begin{figure*}[tp!]
  \centerfloat
  \includegraphics[width=1.1\textwidth]{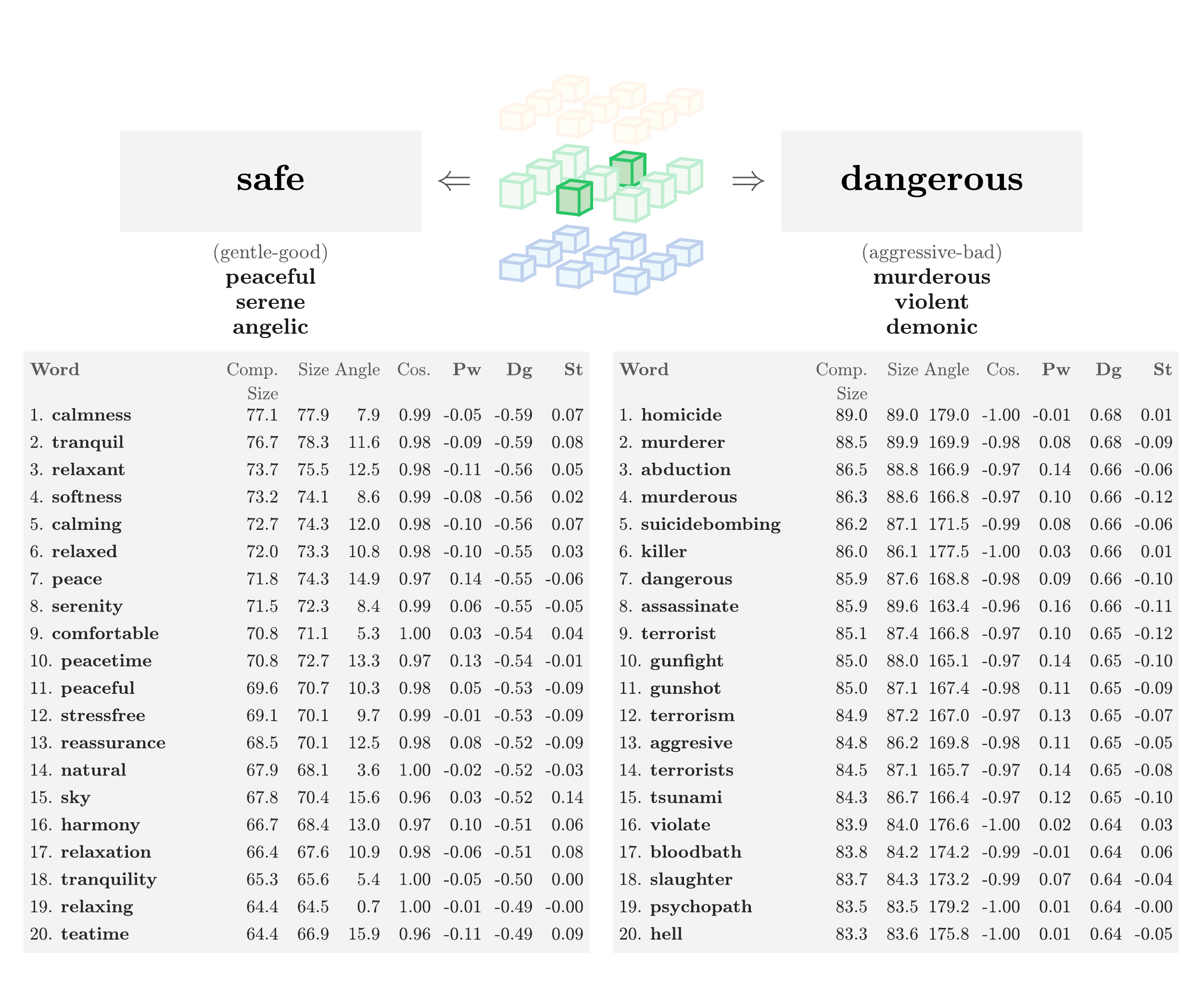}
  \caption{
    \textbf{
      Words with largest components in safe and dangerous directions,
      within a cone of half angle
      $
      \frac{1}{2}
      \frac{180}{\pi}
      \cos^{-1}(2/\sqrt{6})
      \simeq
      17.6\degree
      $.
    }
  }
  \label{fig:meaning.ousiometrics_opposite_cubes001_001}
\end{figure*}

\begin{figure*}[tp!]
  \centerfloat
  \includegraphics[width=1.1\textwidth]{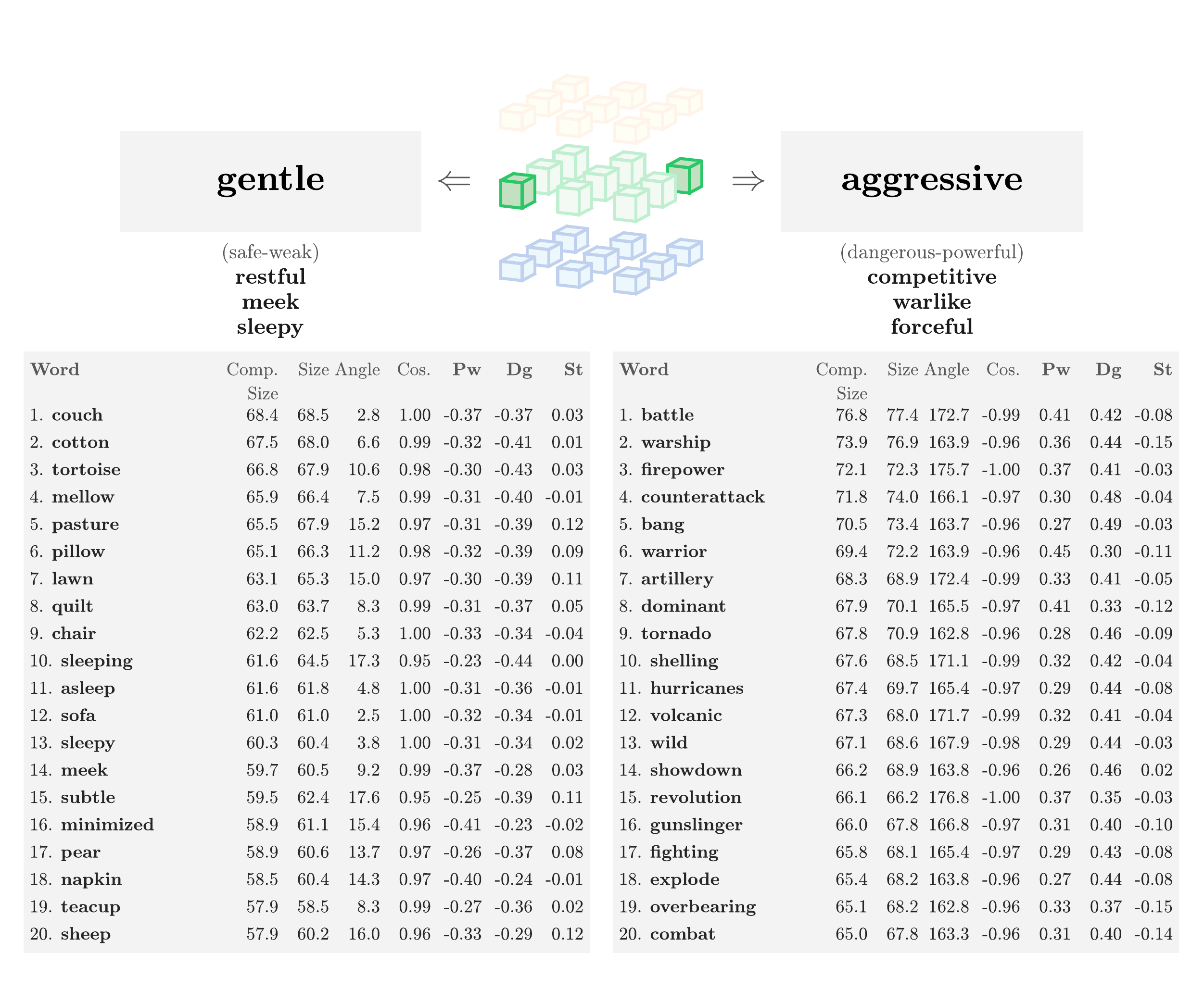}
  \caption{
    \textbf{
      Words with largest components in gentle and aggressive directions,
      within a cone of half angle
      $
      \frac{1}{2}
      \frac{180}{\pi}
      \cos^{-1}(2/\sqrt{6})
      \simeq
      17.6\degree
      $.
    }
  }
  \label{fig:meaning.ousiometrics_opposite_cubes001_002}
\end{figure*}

\begin{figure*}[tp!]
  \centerfloat
  \includegraphics[width=1.1\textwidth]{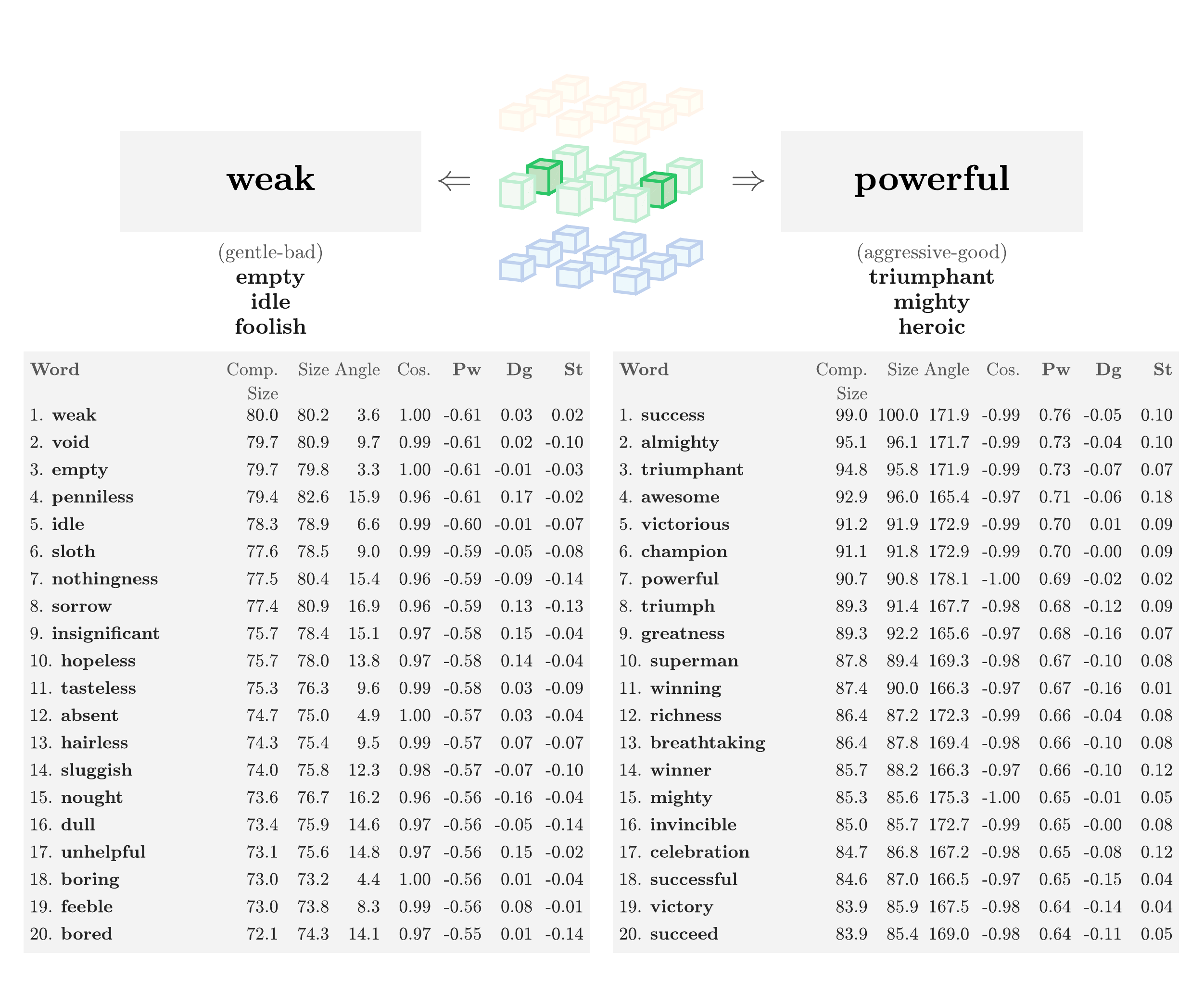}
  \caption{
    \textbf{
      Words with largest components in weak and powerful directions,
      within a cone of half angle
      $
      \frac{1}{2}
      \frac{180}{\pi}
      \cos^{-1}(2/\sqrt{6})
      \simeq
      17.6\degree
      $.
    }
  }
  \label{fig:meaning.ousiometrics_opposite_cubes001_003}
\end{figure*}

\begin{figure*}[tp!]
  \centerfloat
  \includegraphics[width=1.1\textwidth]{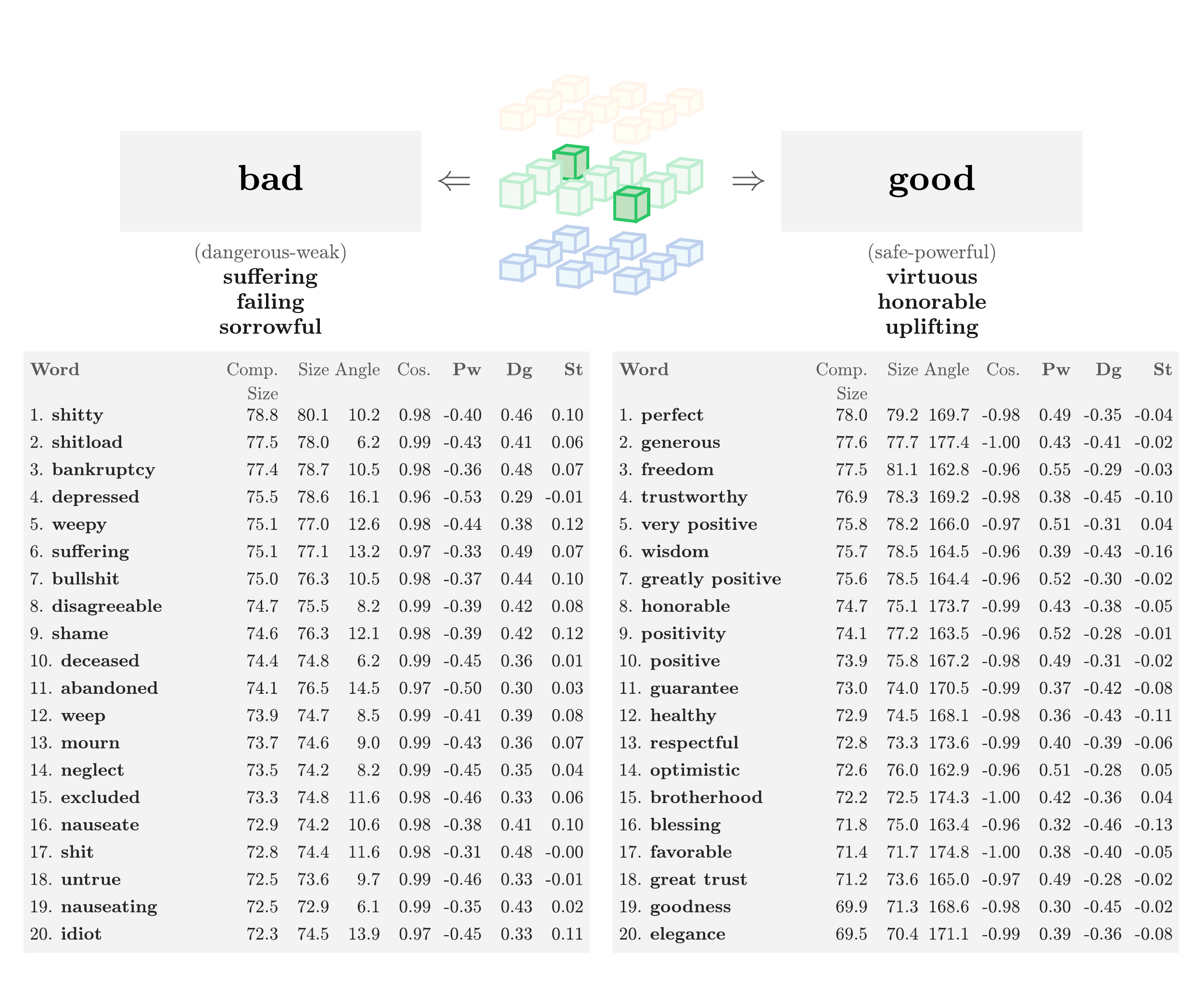}
  \caption{
    \textbf{
      Words with largest components in bad and good directions,
      within a cone of half angle
      $
      \frac{1}{2}
      \frac{180}{\pi}
      \cos^{-1}(2/\sqrt{6})
      \simeq
      17.6\degree
      $.\\
      ~
    }
  }
  \label{fig:meaning.ousiometrics_opposite_cubes001_004}
\end{figure*}

\begin{figure*}[tp!]
  \centerfloat
  \includegraphics[width=1.1\textwidth]{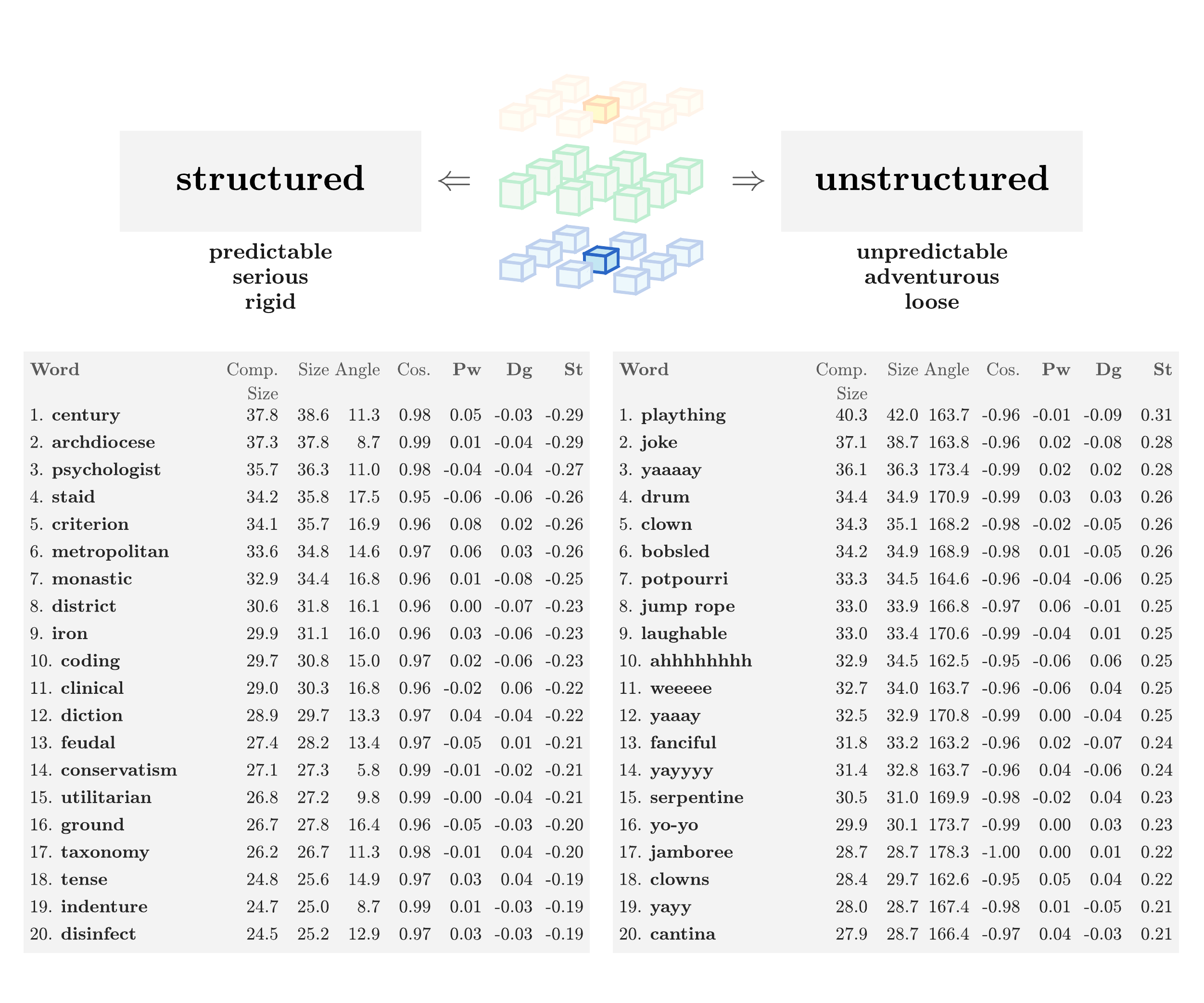}
  \caption{
    \textbf{
      Words with largest components in structured and unstructured directions,
      within a cone of half angle
      $
      \frac{1}{2}
      \frac{180}{\pi}
      \cos^{-1}(2/\sqrt{6})
      \simeq
      17.6\degree
      $.
    }
  }
  \label{fig:meaning.ousiometrics_opposite_cubes001_005}
\end{figure*}

\begin{figure*}[tp!]
  \centerfloat
  \includegraphics[width=1.1\textwidth]{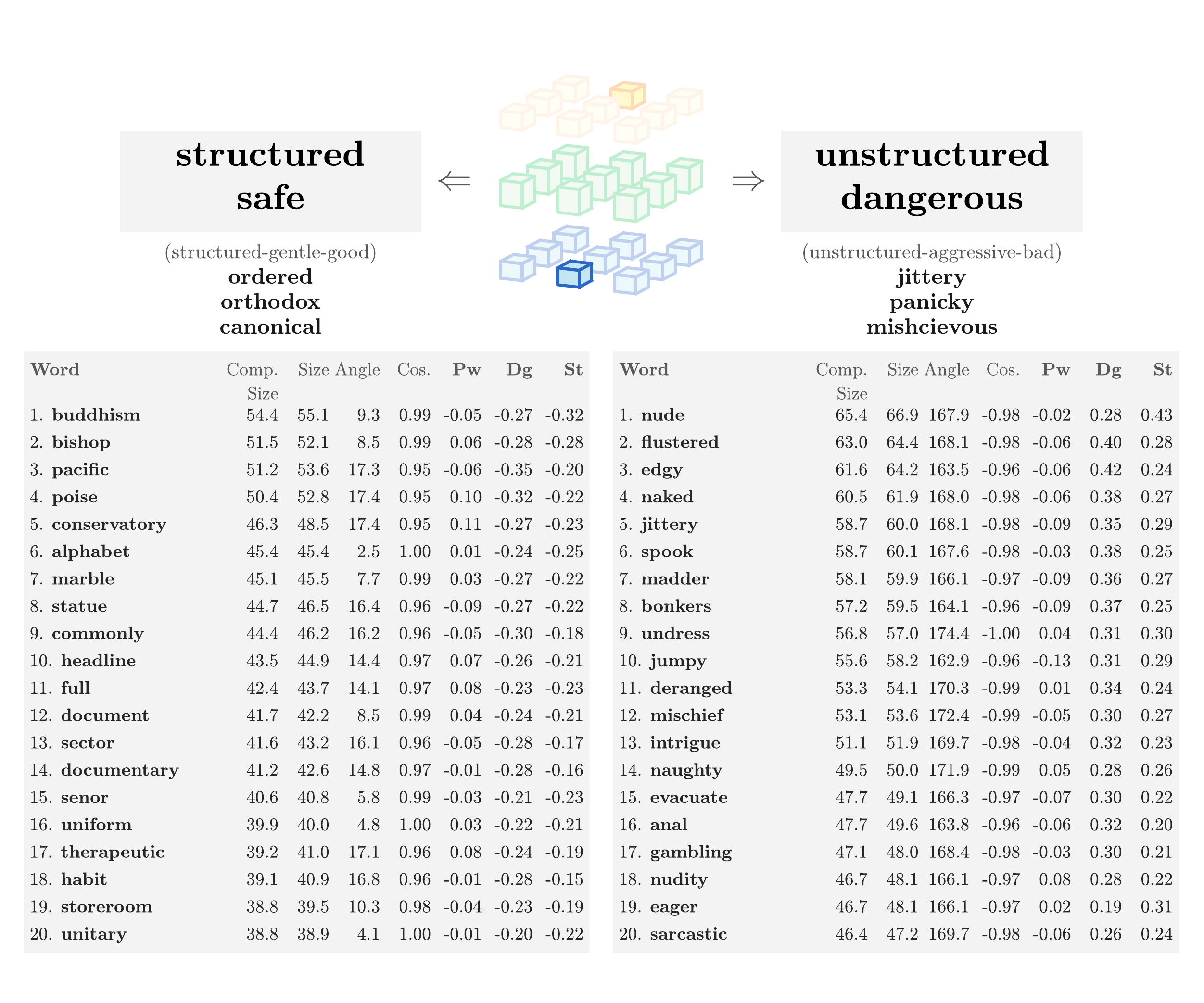}
  \caption{
    \textbf{
      Words with largest components in structured-safe and unstructured-dangerous directions,
      within a cone of half angle
      $
      \frac{1}{2}
      \frac{180}{\pi}
      \cos^{-1}(2/\sqrt{6})
      \simeq
      17.6\degree
      $.
    }
  }
  \label{fig:meaning.ousiometrics_opposite_cubes001_006}
\end{figure*}

\begin{figure*}[tp!]
  \centerfloat
  \includegraphics[width=1.1\textwidth]{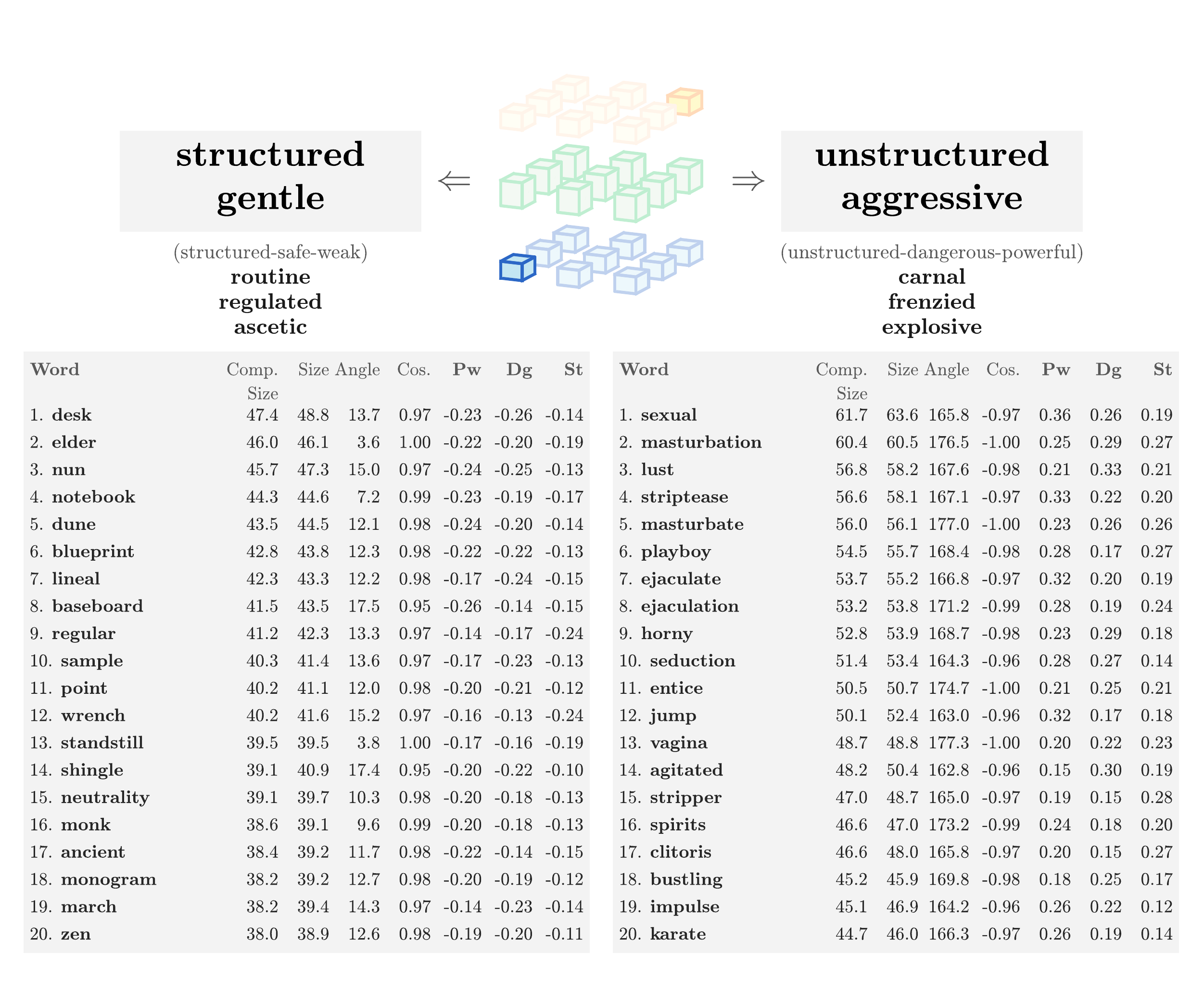}
  \caption{
    \textbf{
      Words with largest components in structured-gentle and unstructured-aggressive directions,
      within a cone of half angle
      $
      \frac{1}{2}
      \frac{180}{\pi}
      \cos^{-1}(2/\sqrt{6})
      \simeq
      17.6\degree
      $.
    }
  }
  \label{fig:meaning.ousiometrics_opposite_cubes001_007}
\end{figure*}

\begin{figure*}[tp!]
  \centerfloat
  \includegraphics[width=1.1\textwidth]{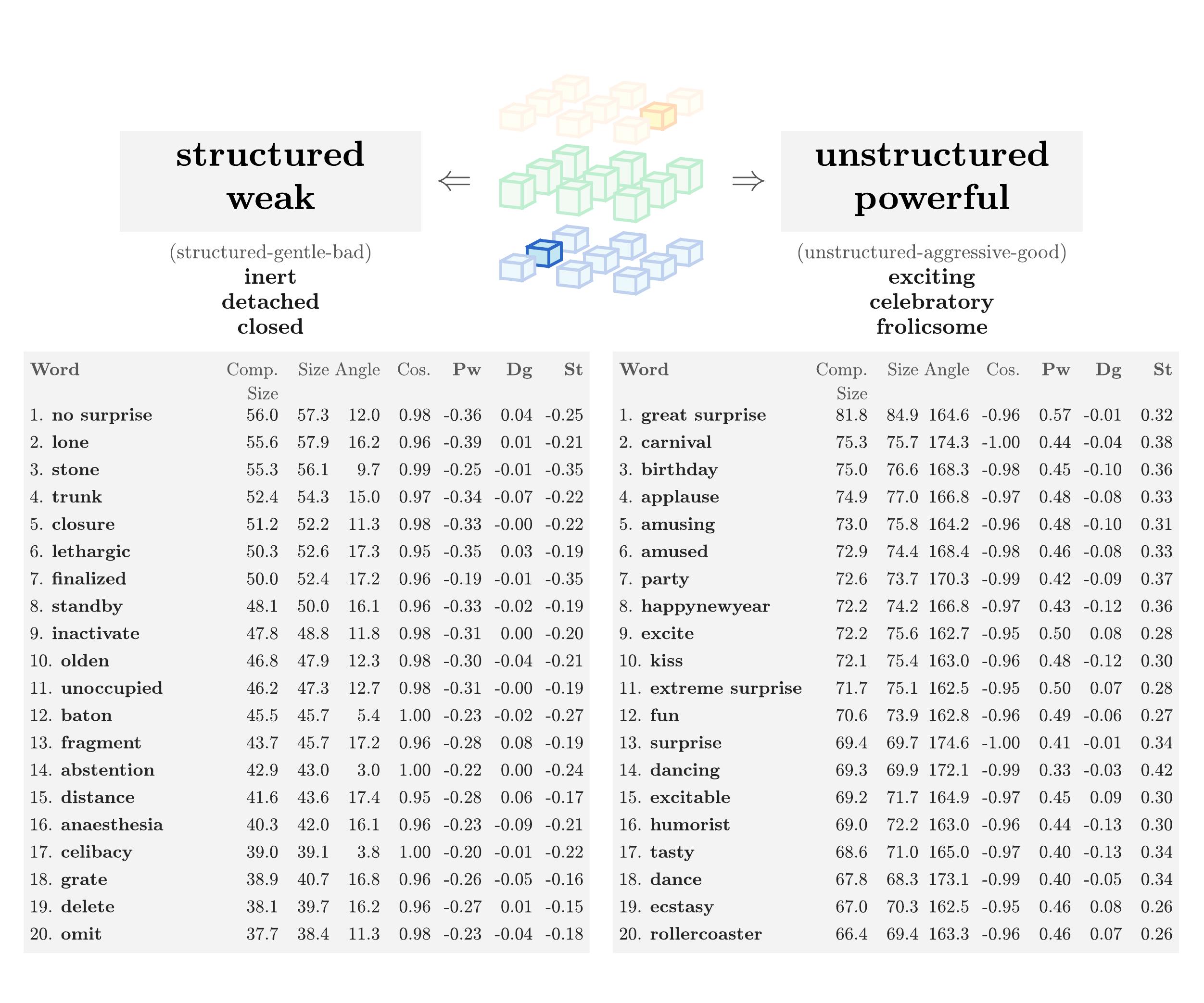}
  \caption{
    \textbf{
      Words with largest components in structured-weak and unstructured-powerful directions,
      within a cone of half angle
      $
      \frac{1}{2}
      \frac{180}{\pi}
      \cos^{-1}(2/\sqrt{6})
      \simeq
      17.6\degree
      $.
    }
  }
  \label{fig:meaning.ousiometrics_opposite_cubes001_008}
\end{figure*}

\begin{figure*}[tp!]
  \centerfloat
  \includegraphics[width=1.1\textwidth]{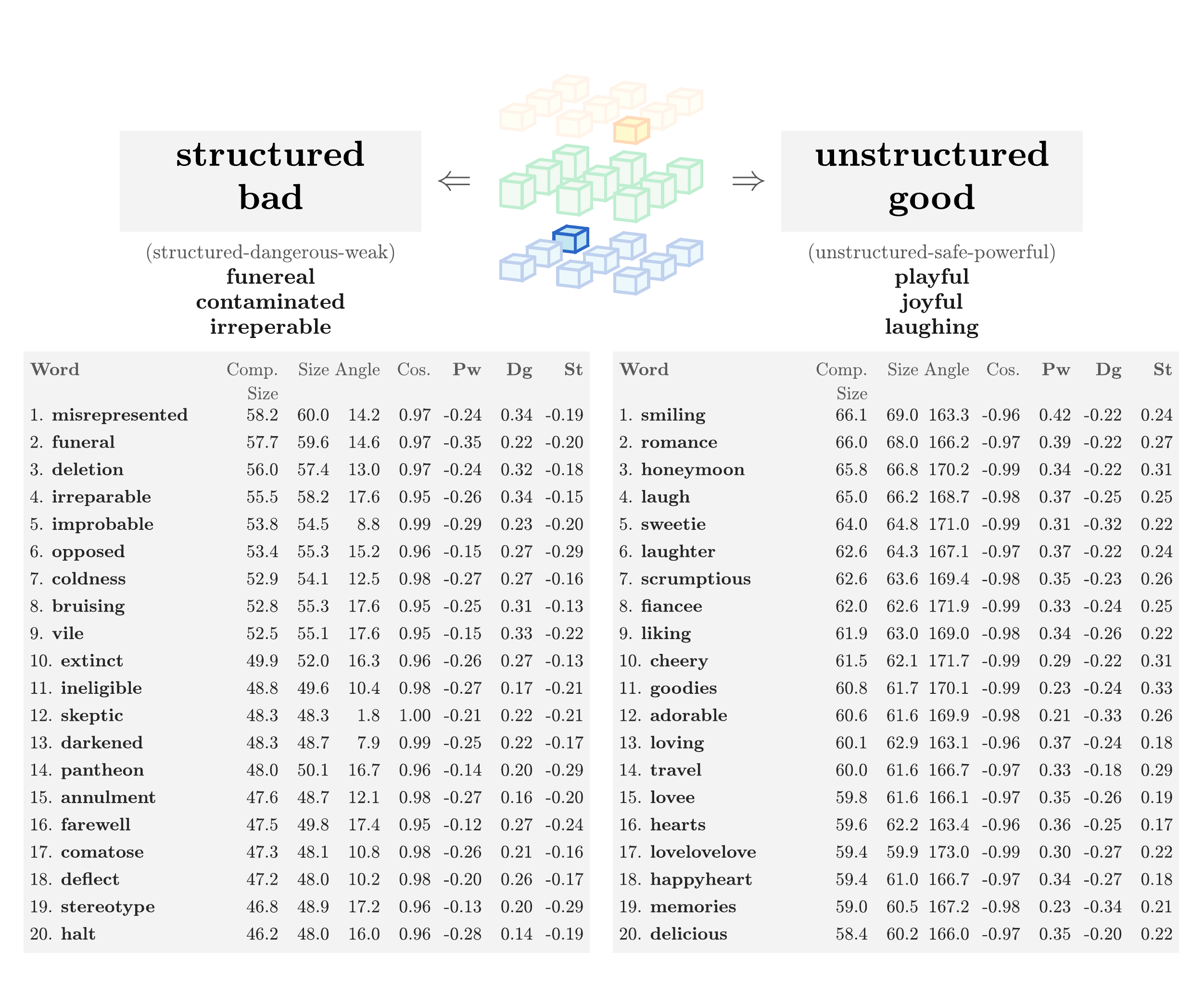}
  \caption{
    \textbf{
      Words with largest components in structured-bad and unstructured-good directions,
      within a cone of half angle
      $
      \frac{1}{2}
      \frac{180}{\pi}
      \cos^{-1}(2/\sqrt{6})
      \simeq
      17.6\degree
      $.
    }
  }
  \label{fig:meaning.ousiometrics_opposite_cubes001_009}
\end{figure*}

\begin{figure*}[tp!]
  \centerfloat
  \includegraphics[width=1.1\textwidth]{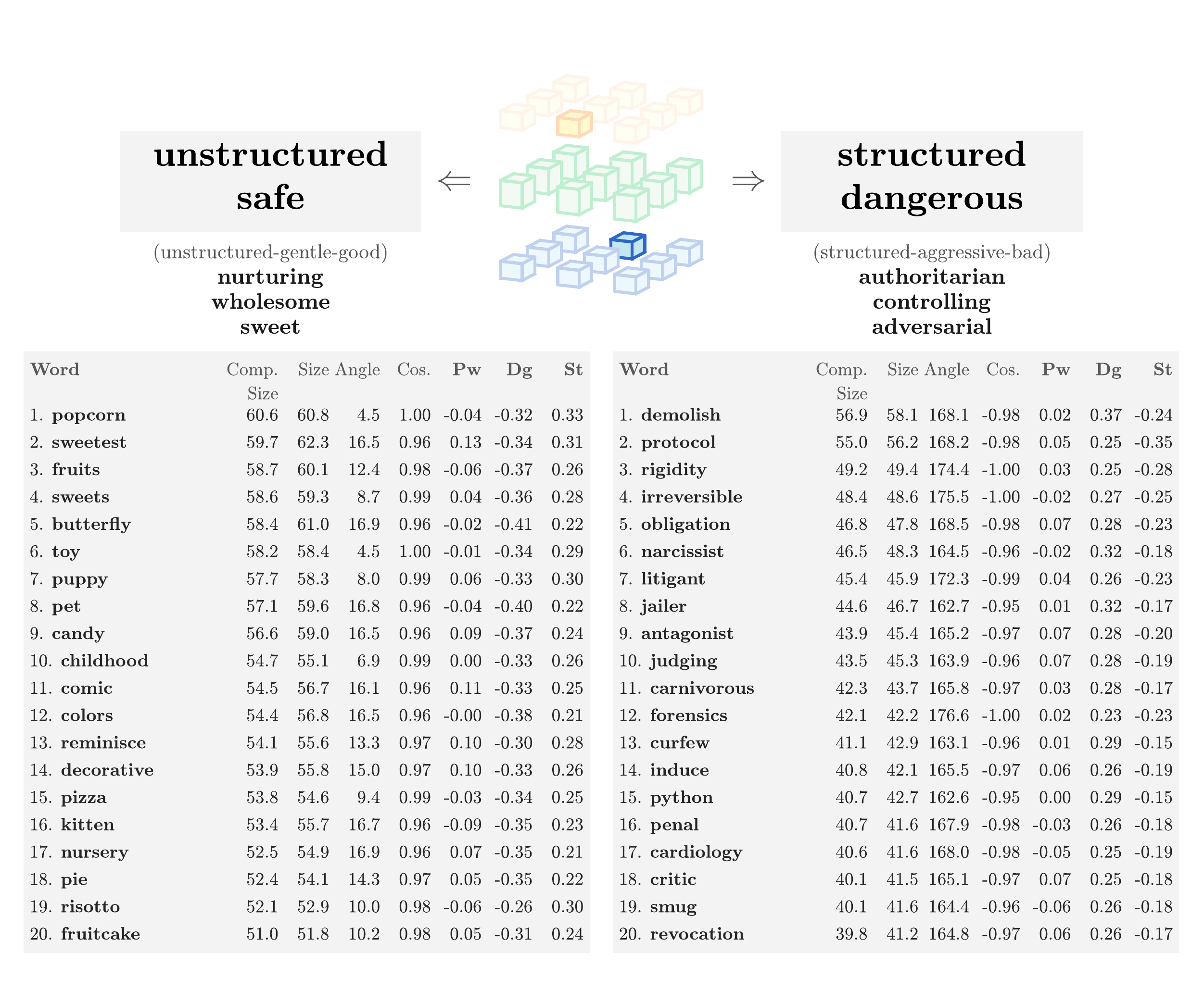}
  \caption{
    \textbf{
      Words with largest components in unstructured-safe and structured-dangerous directions,
      within a cone of half angle
      $
      \frac{1}{2}
      \frac{180}{\pi}
      \cos^{-1}(2/\sqrt{6})
      \simeq
      17.6\degree
      $.
    }
  }
  \label{fig:meaning.ousiometrics_opposite_cubes001_010}
\end{figure*}

\begin{figure*}[tp!]
  \centerfloat
  \includegraphics[width=1.1\textwidth]{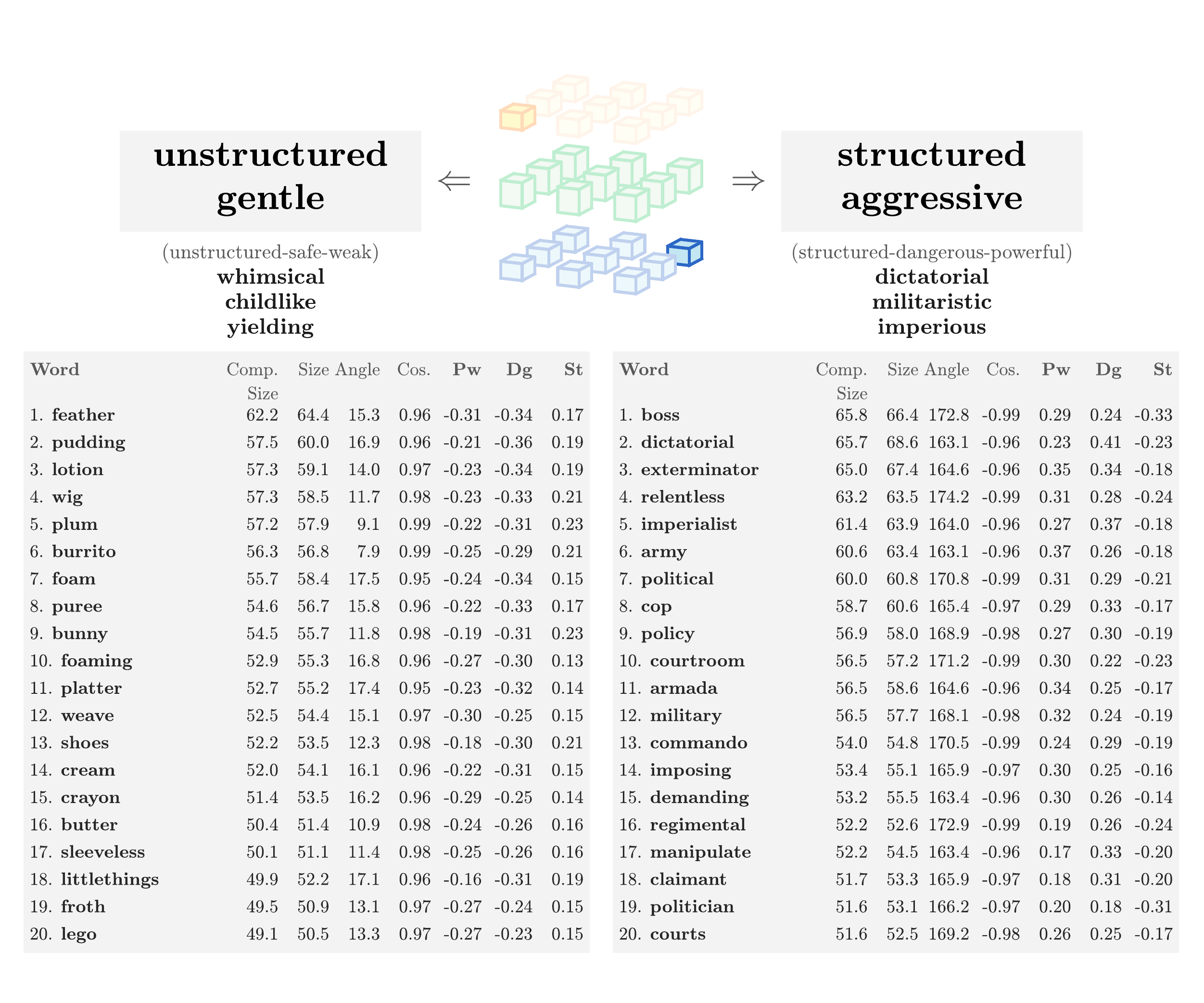}
  \caption{
    \textbf{
      Words with largest components in unstructured-gentle and structured-aggressive directions,
      within a cone of half angle
      $
      \frac{1}{2}
      \frac{180}{\pi}
      \cos^{-1}(2/\sqrt{6})
      \simeq
      17.6\degree
      $.
    }
  }
  \label{fig:meaning.ousiometrics_opposite_cubes001_011}
\end{figure*}

\begin{figure*}[tp!]
  \centerfloat
  \includegraphics[width=1.1\textwidth]{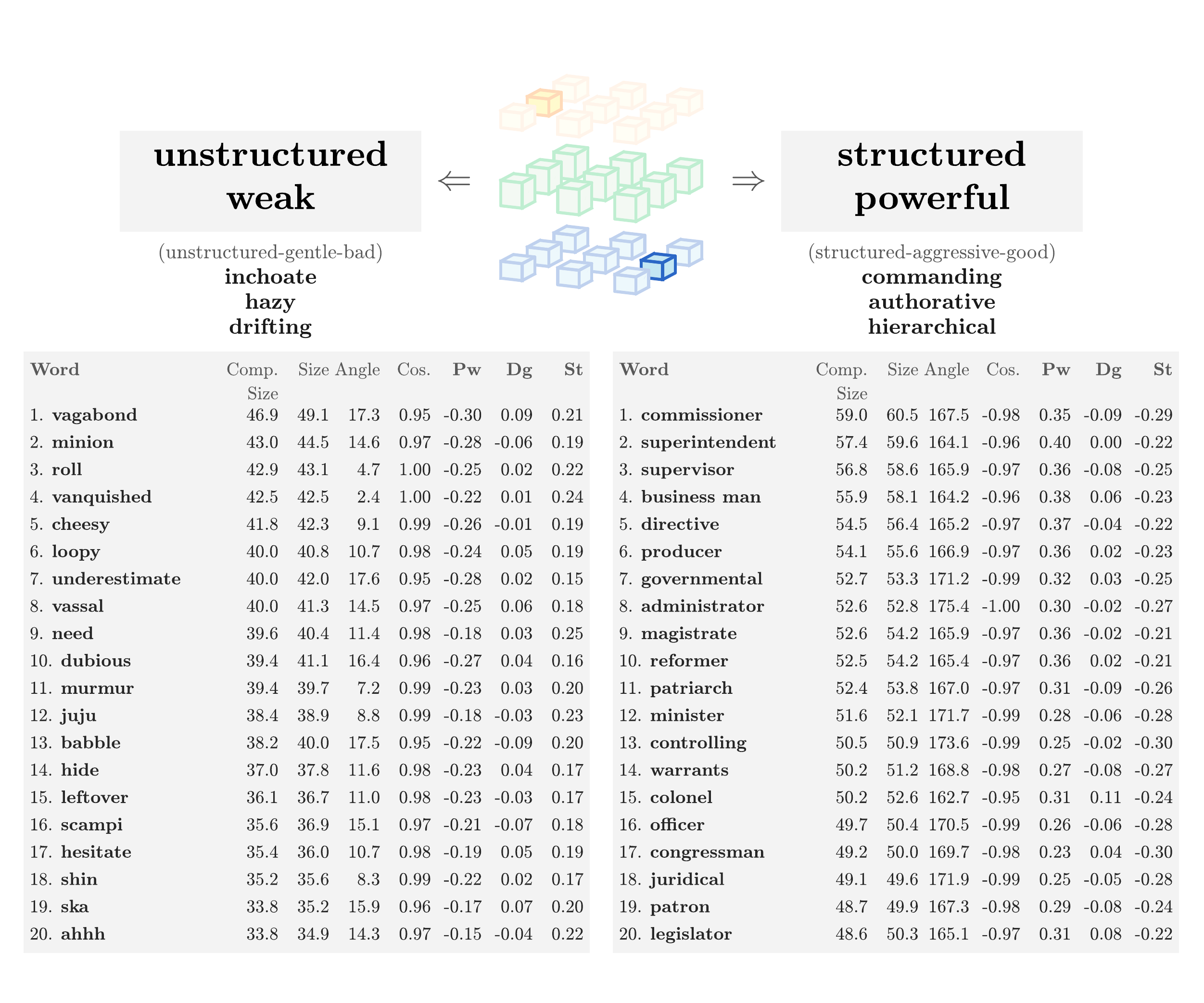}
  \caption{
    \textbf{
      Words with largest components in unstructured-weak and structured-powerful directions,
      within a cone of half angle
      $
      \frac{1}{2}
      \frac{180}{\pi}
      \cos^{-1}(2/\sqrt{6})
      \simeq
      17.6\degree
      $.
    }
  }
  \label{fig:meaning.ousiometrics_opposite_cubes001_012}
\end{figure*}

\begin{figure*}[tp!]
  \centerfloat
  \includegraphics[width=1.1\textwidth]{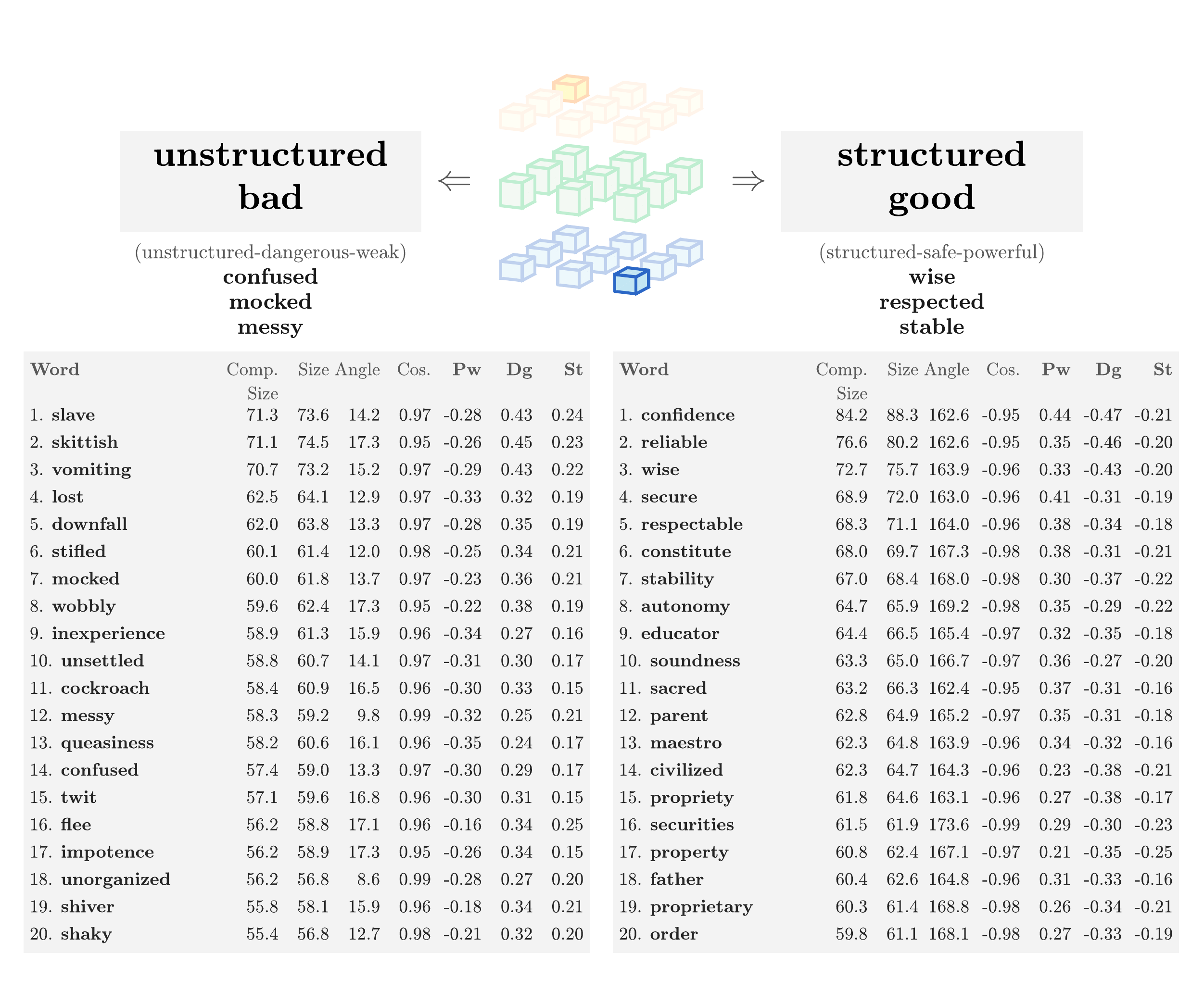}
  \caption{
    \textbf{
      Words with largest components in unstructured-bad and structured-good directions,
      within a cone of half angle
      $
      \frac{1}{2}
      \frac{180}{\pi}
      \cos^{-1}(2/\sqrt{6})
      \simeq
      17.6\degree
      $.
    }
  }
  \label{fig:meaning.ousiometrics_opposite_cubes001_013}
\end{figure*}

\begin{figure*}[tp!]
  \centerfloat
  \includegraphics[width=1.1\textwidth]{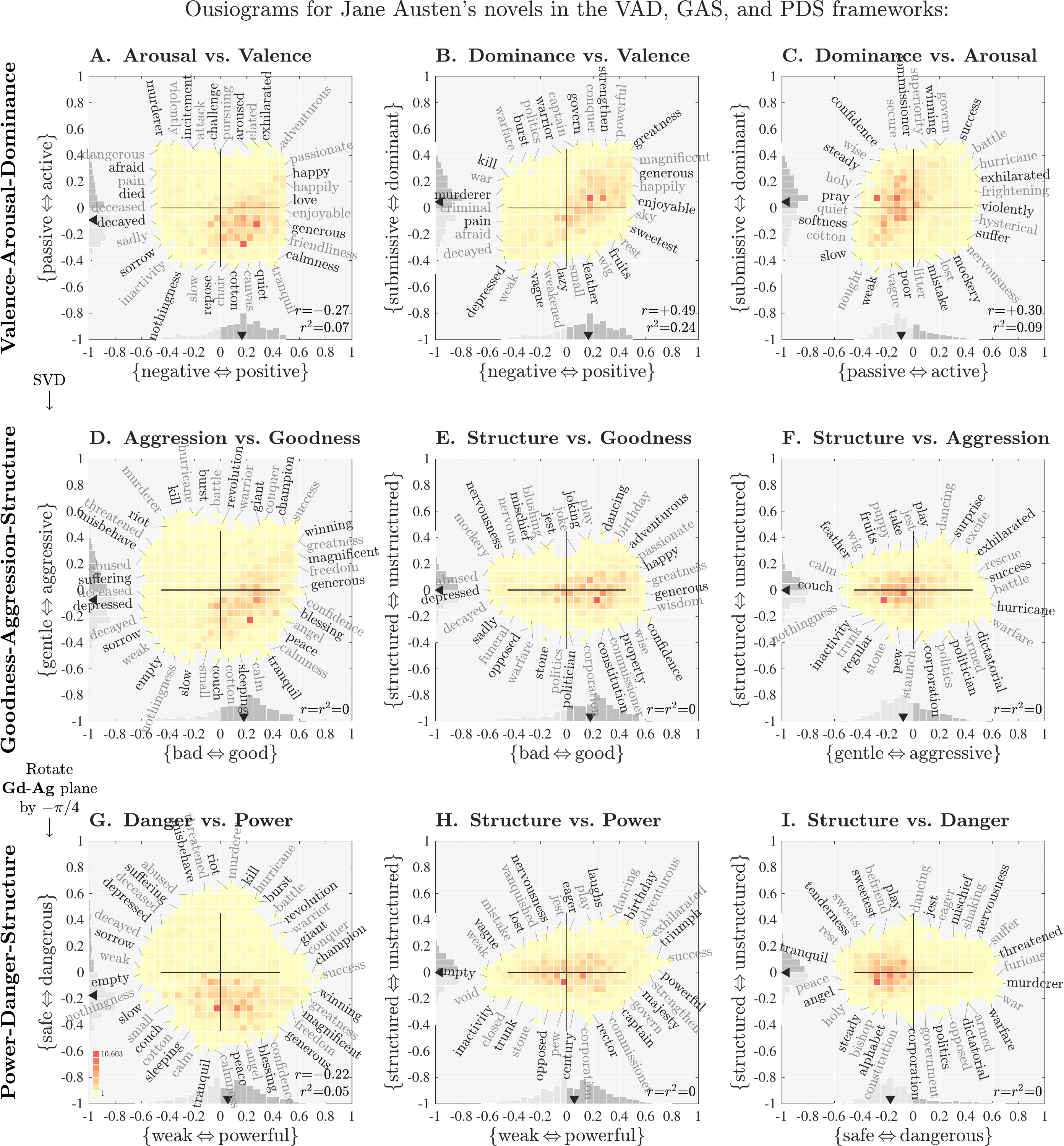}
  \caption{
    \textbf{
      Ousiograms showing the VAD-GAS-PDS analytic sequence for
      Jane Austen's writings.
    }
    The novels are
    ``Sense and Sensibility,''
    ``Pride and Prejudice,''
    ``Mansfield Park,''
    ``Emma,''
    ``Northanger Abbey,''
    and
    ``Persuasion,''
    published in 1811--1818.
    We obtained all novels from the Gutenberg Project:
    \url{http://www.gutenberg.org}.
    The underlying word frequency distribution is built by
    merging all books and then constructing a word frequency distribution.
    Panel G corresponds to
    Fig.~\ref{fig:meaning.ousiometer9400_PDS_tableau004}A.
    \\
    ~
  }
  \label{fig:meaning.ousiometer9310_jane_austen001}
\end{figure*}

\begin{figure*}[tp!]
  \centerfloat
  \includegraphics[width=1.1\textwidth]{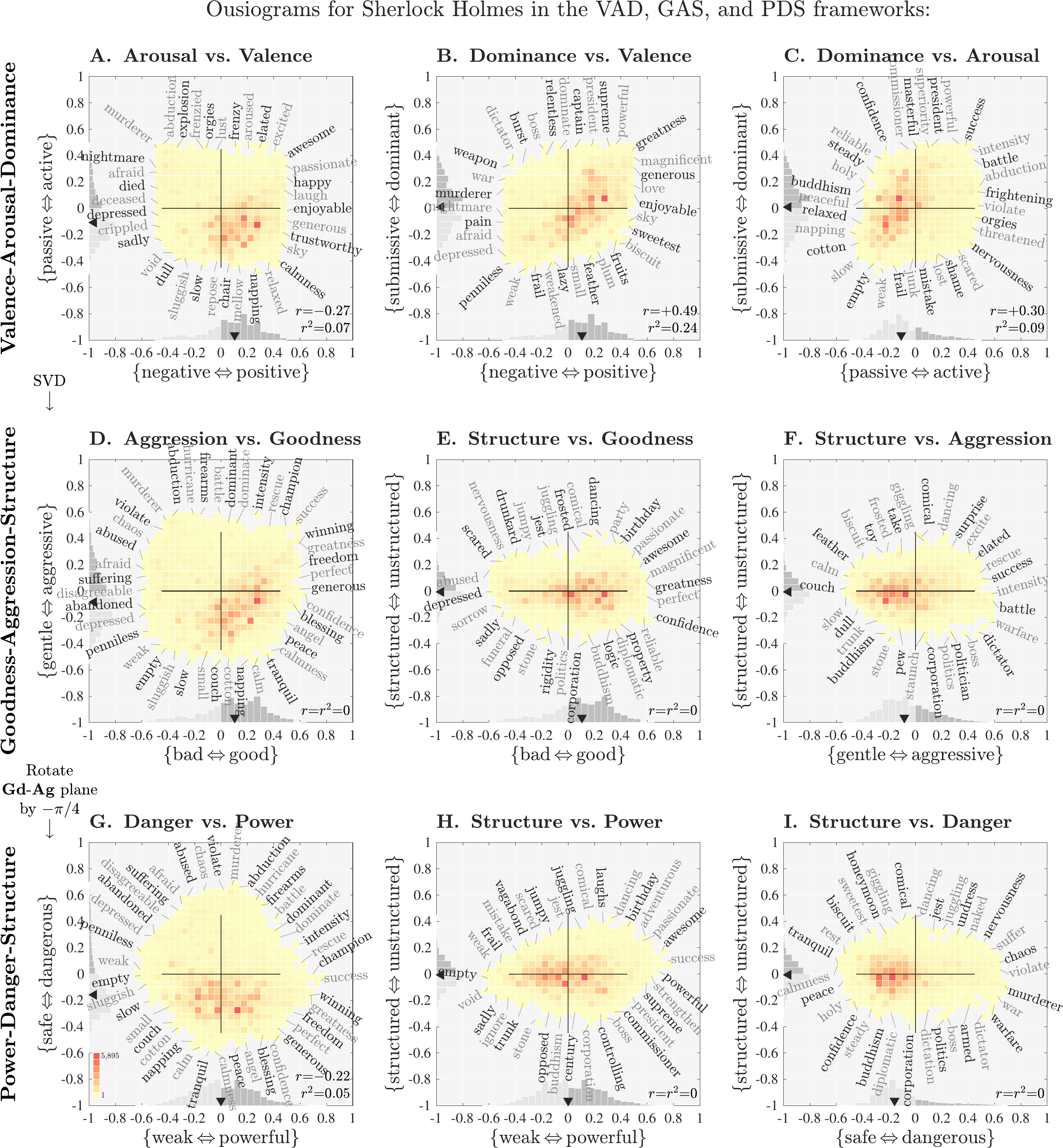}
  \caption{
    \textbf{
      Ousiograms showing the VAD-GAS-PDS analytic sequence for
      Sir Arthur Conan Doyle's Sherlock Holmes novels and short stories.
    }
    We obtained four novels and forty-four short stories from
    the complete Sherlock Holmes Canon
    \url{https://sherlock-holm.es/}
    (due to copyright, twelve short stories contained in 
    the ``Case-Book of Sherlock Holmes'' were not available from this source).
    The underlying word frequency distribution is built by
    merging all books and then constructing a word frequency distribution.
    Panel G corresponds to 
    Fig.~\ref{fig:meaning.ousiometer9400_PDS_tableau004}B.
  }
  \label{fig:meaning.ousiometer9310_sherlock_holmes001}
\end{figure*}

\begin{figure*}[tp!]
  \centerfloat
  \includegraphics[width=1.1\textwidth]{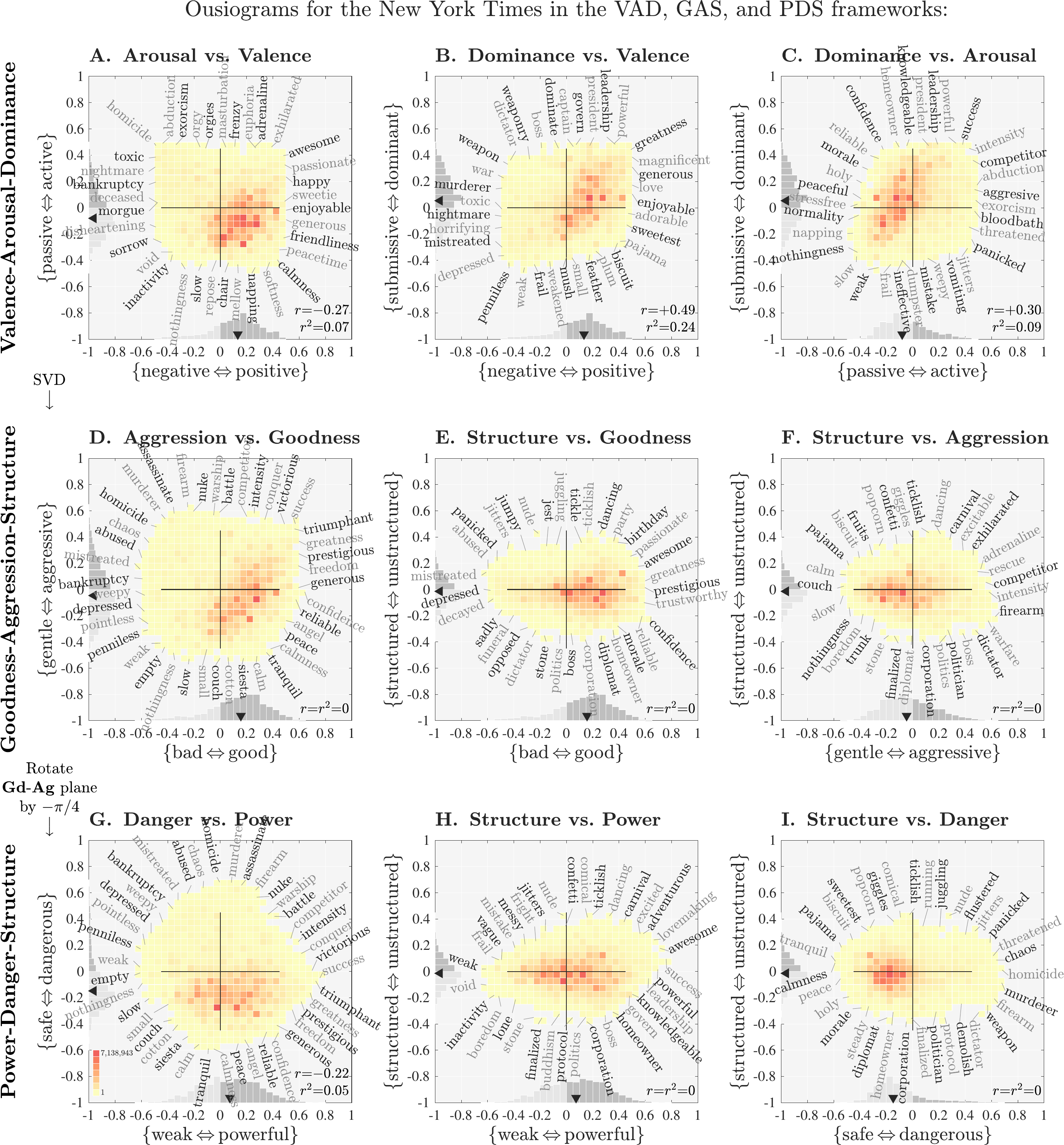}
  \caption{
    \textbf{
      Ousiograms showing the VAD-GAS-PDS analytic sequence for the New York Times.
    }
    The underlying word frequency distribution is built from
    a 1987--2007 annotated corpus~\cite{nytimescorpus2008a}.
    Panel G corresponds to 
    Fig.~\ref{fig:meaning.ousiometer9400_PDS_tableau004}C.
    \\
    ~
    \\
    ~
    \\
    ~
  }
  \label{fig:meaning.ousiometer9310_nyt001}
\end{figure*}

\begin{figure*}[tp!]
  \centerfloat
  \includegraphics[width=1.1\textwidth]{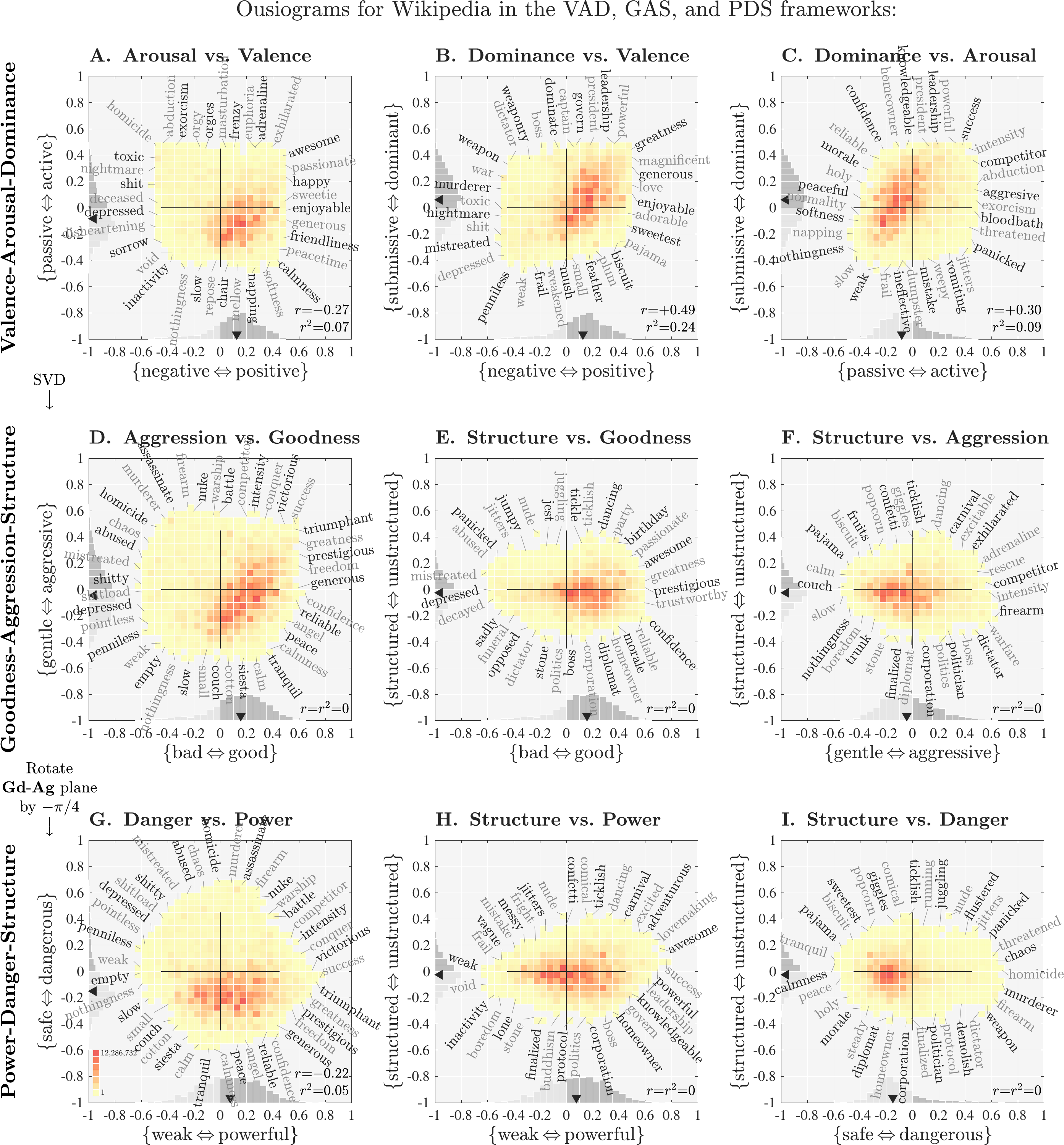}
  \caption{
    \textbf{
      Ousiograms showing the VAD-GAS-PDS analytic sequence for Wikipedia.
    }
    The underlying word frequency distribution
    is based on the March 2019 dump of the English Wikipedia~\cite{semenov2019a}.
    Panel G corresponds to 
    Fig.~\ref{fig:meaning.ousiometer9400_PDS_tableau004}D.
    \\
    ~
    \\
    ~
    \\
    ~
  }
  \label{fig:meaning.ousiometer9310_wikipedia001}
\end{figure*}

\begin{figure*}[tp!]
  \centerfloat
  \includegraphics[width=1.1\textwidth]{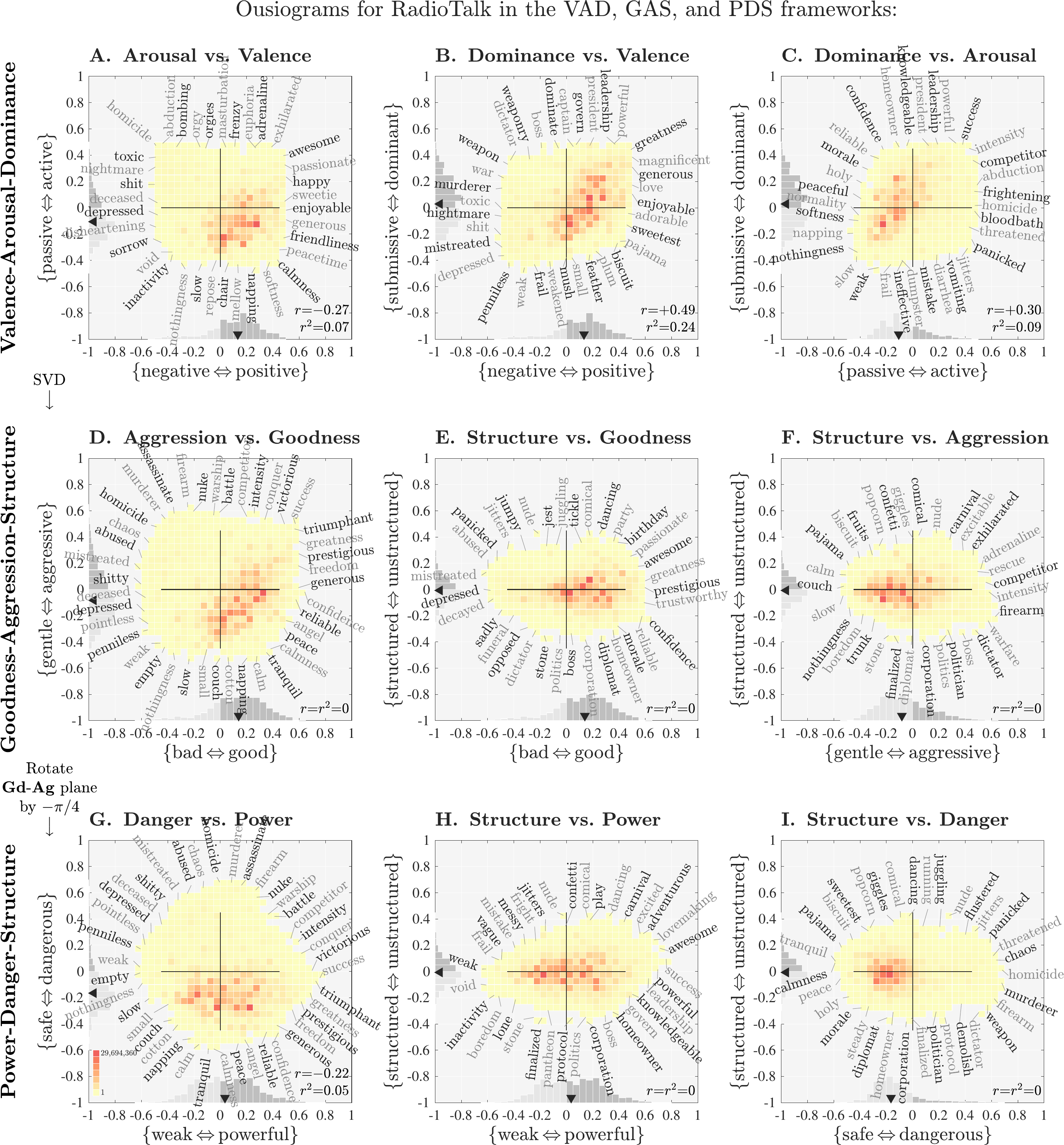}
  \caption{
    \textbf{
      Ousiograms showing the VAD-GAS-PDS analytic sequence for the RadioTalk corpus.
    }
    The underlying word frequency distribution
    is derived from a corpus of
    automated transcriptions of
    talk radio in the US covering the time period
    2018/10--2019/03~\cite{beeferman2019a}.
    Panel G corresponds to 
    Fig.~\ref{fig:meaning.ousiometer9400_PDS_tableau004}E.
    \\
    ~
    \\
    ~
  }
  \label{fig:meaning.ousiometer9310_radiotalk001}
\end{figure*}

\begin{figure*}[tp!]
  \centerfloat
  \includegraphics[width=1.1\textwidth]{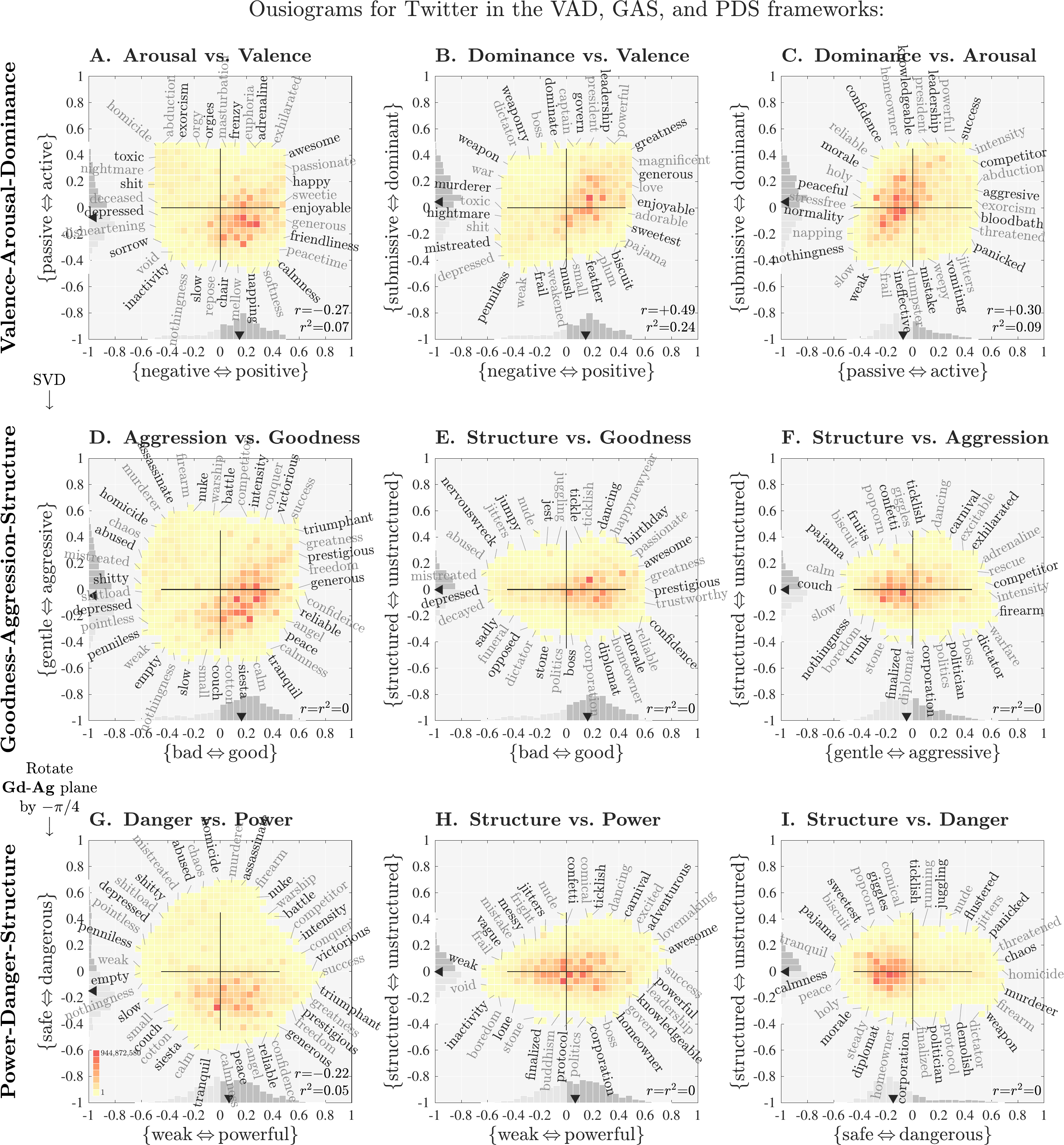}
  \caption{
    \textbf{
      Ousiograms showing the VAD-GAS-PDS analytic sequence for Twitter.
    }
    The underlying word frequency distribution is an equal weighting of day-scale word frequency distributions
    derived from approximately 10\% of English tweets in 2020~\cite{alshaabi2021c}.
    In contrast to the word frequency distributions obtained from `flat' corpora,
    the word frequency distribution for Twitter encodes a strong sense of popularity
    as social amplification is naturally included through retweets.
    Panel G corresponds to 
    Fig.~\ref{fig:meaning.ousiometer9400_PDS_tableau004}F.
    \\
    ~
  }
  \label{fig:meaning.ousiometer9310_twitter001}
\end{figure*}

\begin{figure*}[t]
  \includegraphics[width=\textwidth]{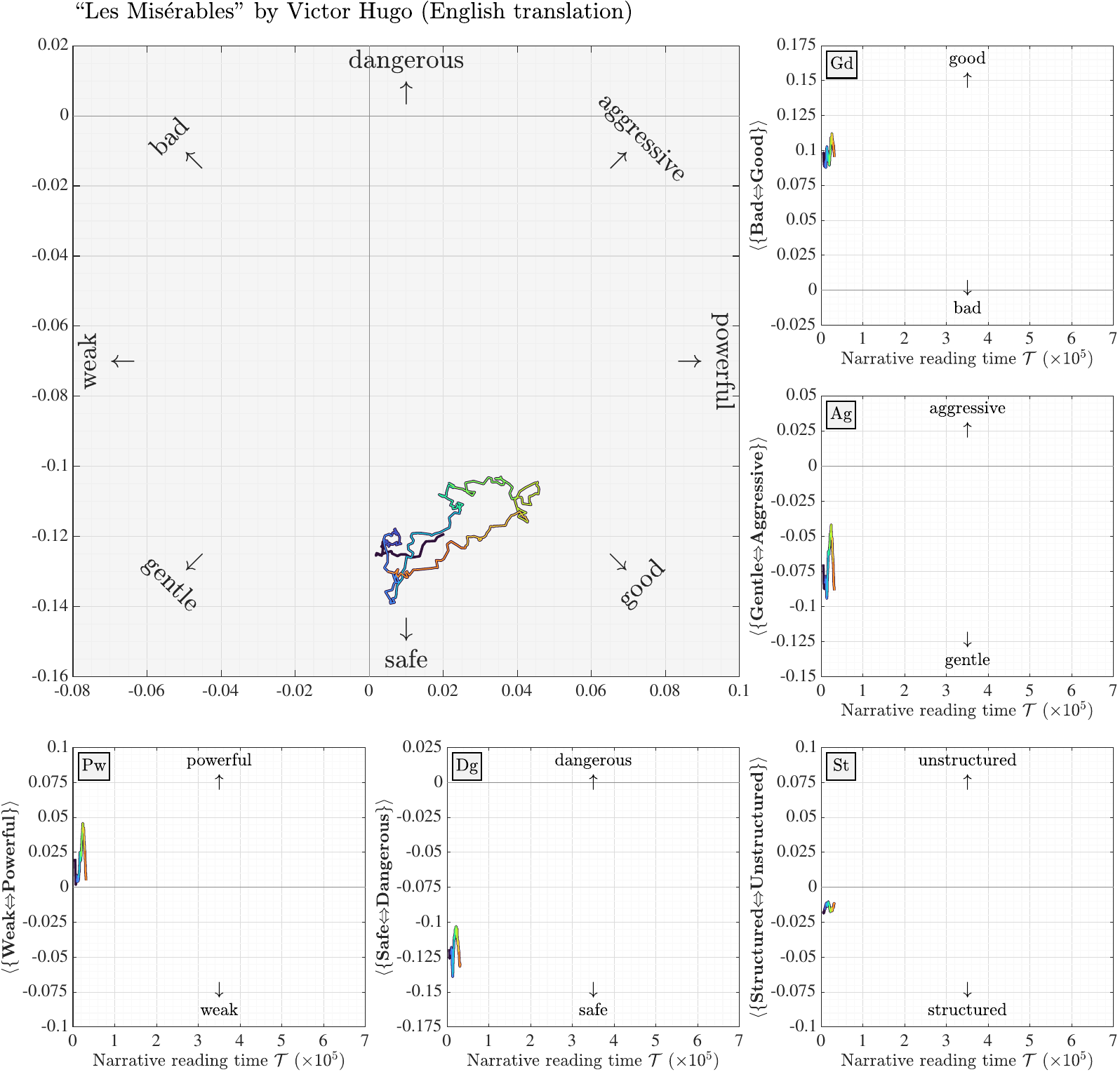}
  \caption{
    \textbf{
      Epoch 1 of 25 in Victor Hugo's ``Les Mis\'{e}rables.''
    }
  }
  \label{fig.meaning:figtelegnomic_timeseries_flipbook9017_001_10000_100_ousiometrics_GPADS100_LesMis_noname}
\end{figure*}

\begin{figure*}[t]
  \includegraphics[width=\textwidth]{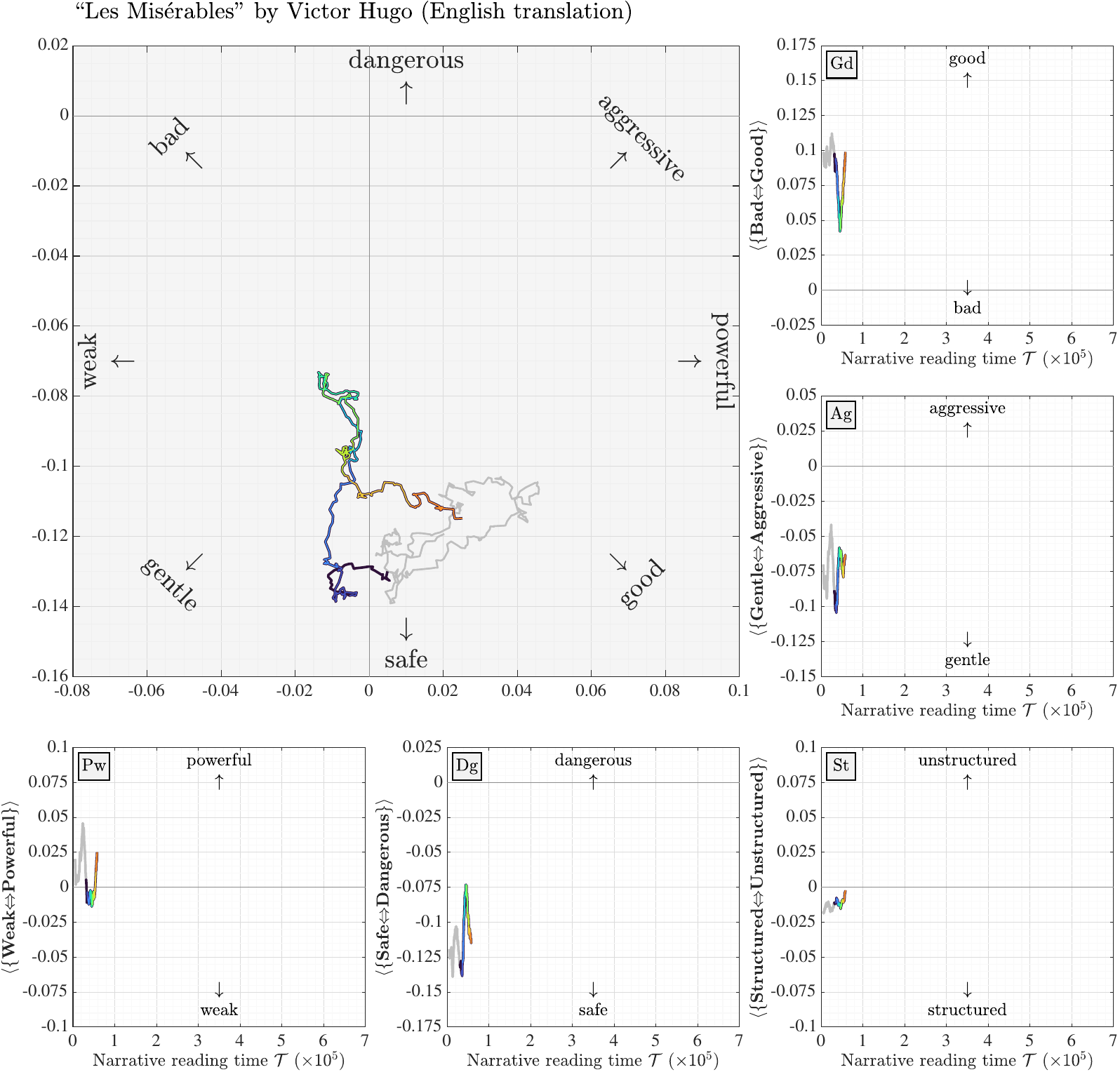}
  \caption{
    \textbf{
      Epoch 2 of 25 in Victor Hugo's ``Les Mis\'{e}rables.''
    }
  }
  \label{fig.meaning:figtelegnomic_timeseries_flipbook9017_002_10000_100_ousiometrics_GPADS100_LesMis_noname}
\end{figure*}

\begin{figure*}[t]
  \includegraphics[width=\textwidth]{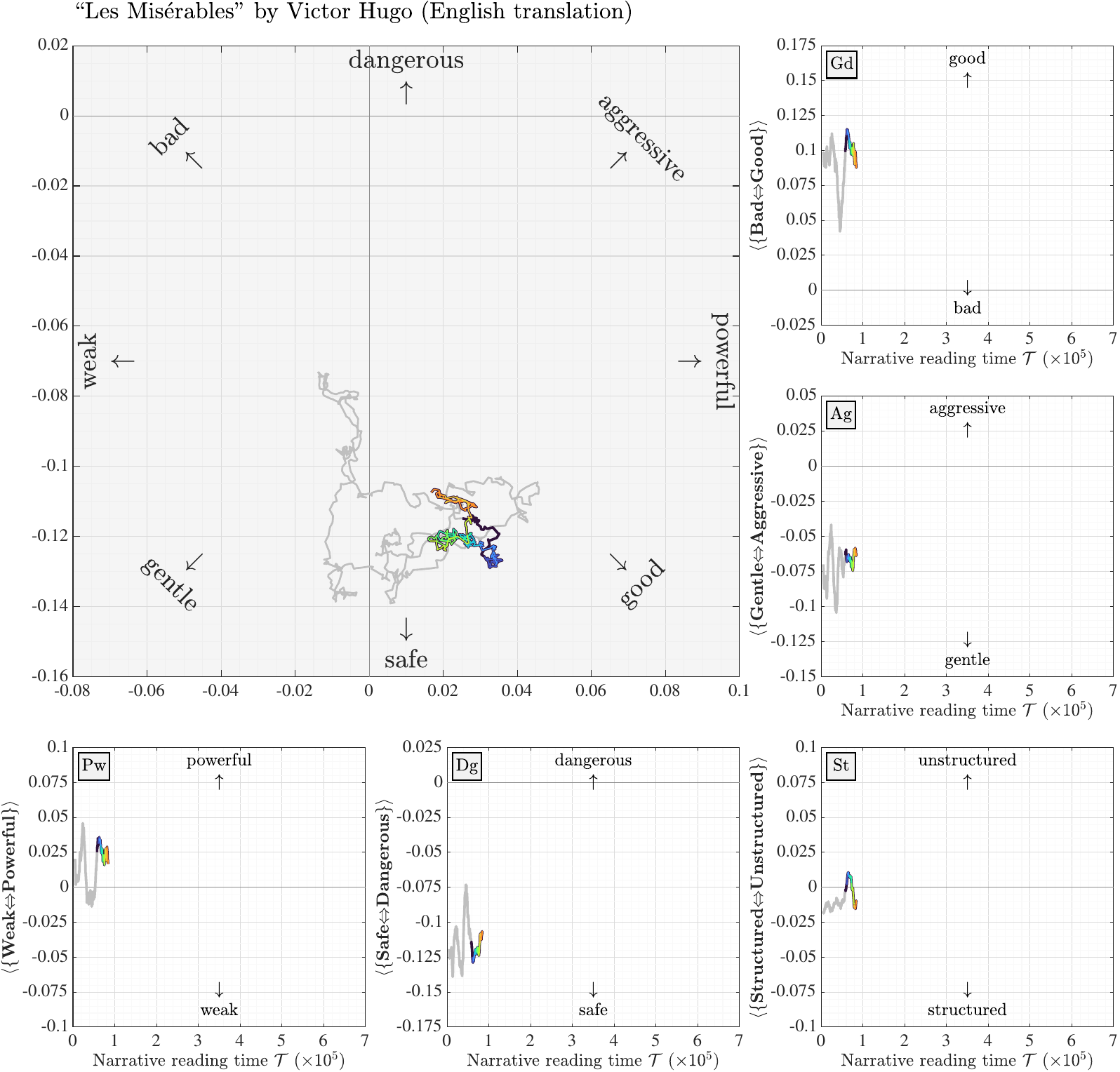}
  \caption{
    \textbf{
      Epoch 3 of 25 in Victor Hugo's ``Les Mis\'{e}rables.''
    }
  }
  \label{fig.meaning:figtelegnomic_timeseries_flipbook9017_003_10000_100_ousiometrics_GPADS100_LesMis_noname}
\end{figure*}

\begin{figure*}[t]
  \includegraphics[width=\textwidth]{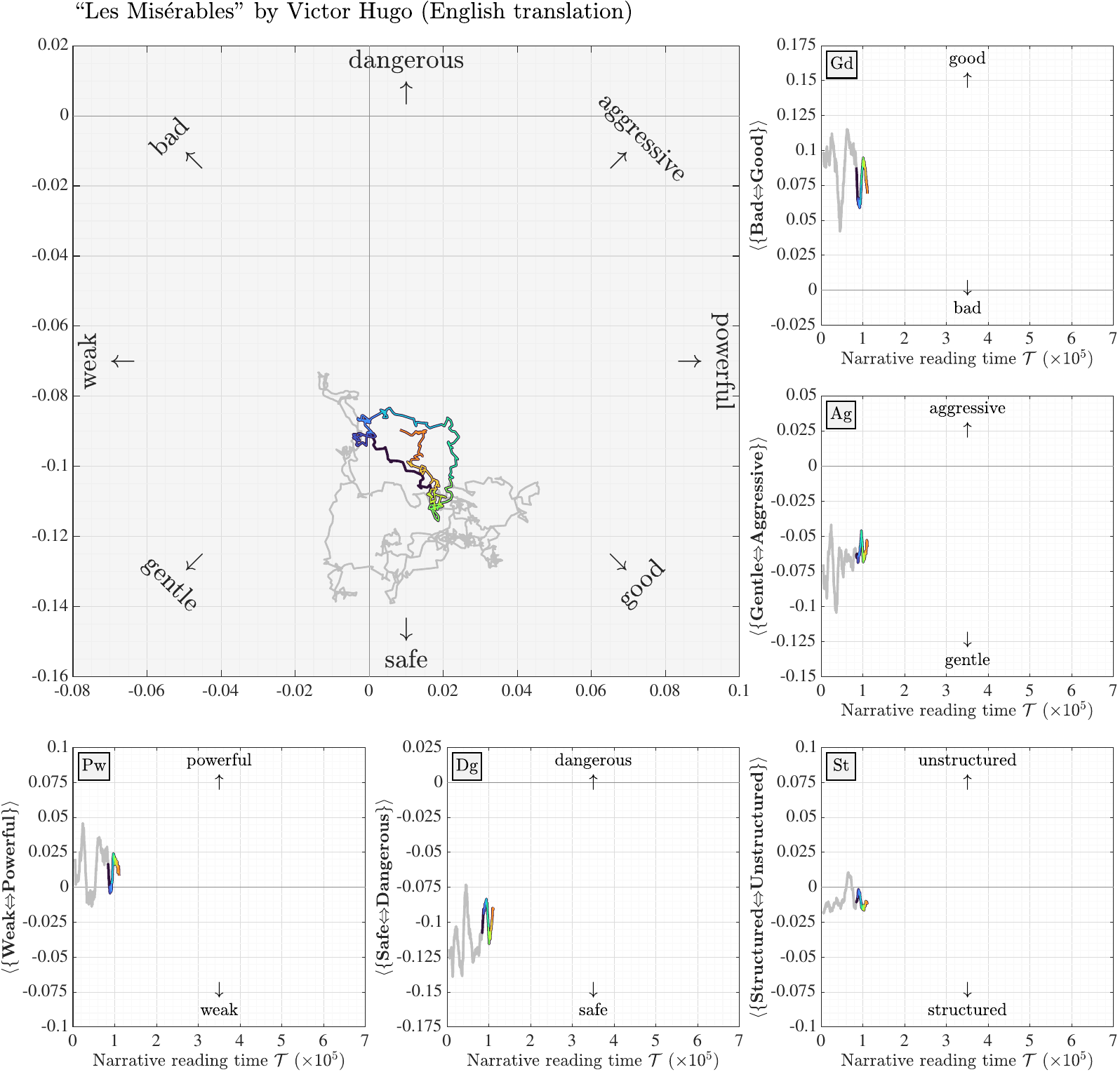}
  \caption{
    \textbf{
      Epoch 4 of 25 in Victor Hugo's ``Les Mis\'{e}rables.''
    }
  }
  \label{fig.meaning:figtelegnomic_timeseries_flipbook9017_004_10000_100_ousiometrics_GPADS100_LesMis_noname}
\end{figure*}

\begin{figure*}[t]
  \includegraphics[width=\textwidth]{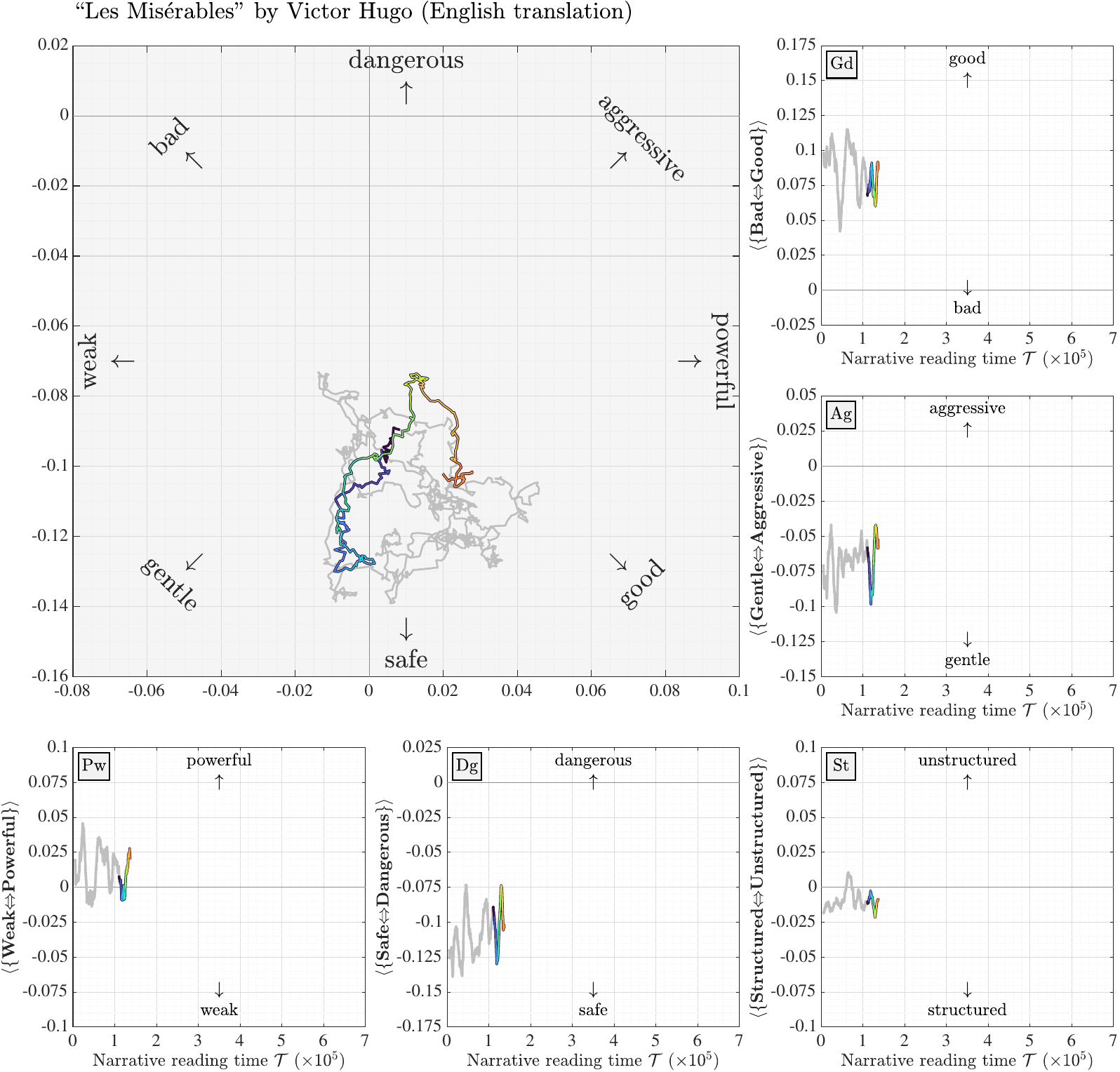}
  \caption{
    \textbf{
      Epoch 5 of 25 in Victor Hugo's ``Les Mis\'{e}rables.''
    }
  }
  \label{fig.meaning:figtelegnomic_timeseries_flipbook9017_005_10000_100_ousiometrics_GPADS100_LesMis_noname}
\end{figure*}

\begin{figure*}[t]
  \includegraphics[width=\textwidth]{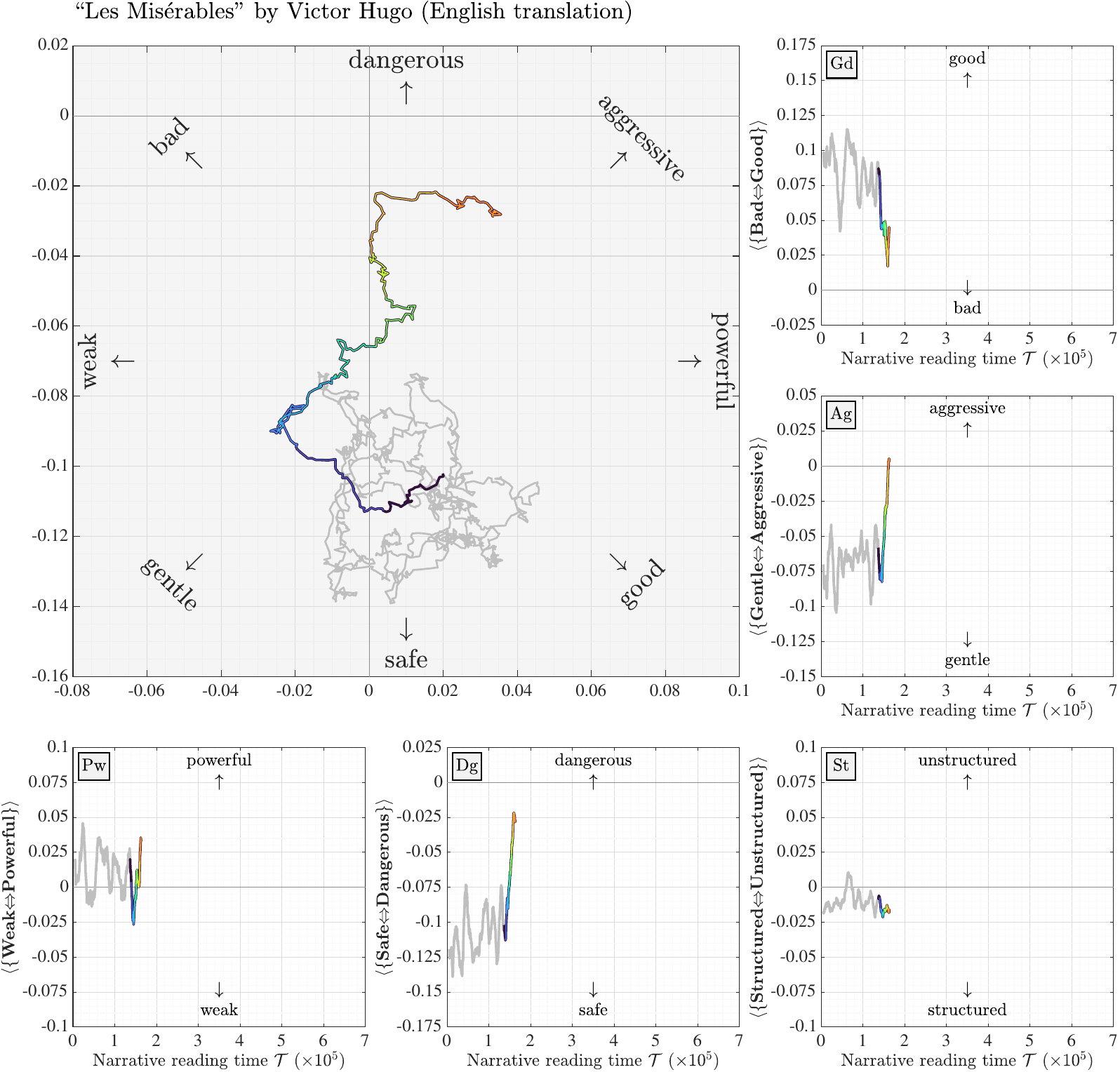}
  \caption{
    \textbf{
      Epoch 6 of 25 in Victor Hugo's ``Les Mis\'{e}rables.''
    }
  }
  \label{fig.meaning:figtelegnomic_timeseries_flipbook9017_006_10000_100_ousiometrics_GPADS100_LesMis_noname}
\end{figure*}

\begin{figure*}[t]
  \includegraphics[width=\textwidth]{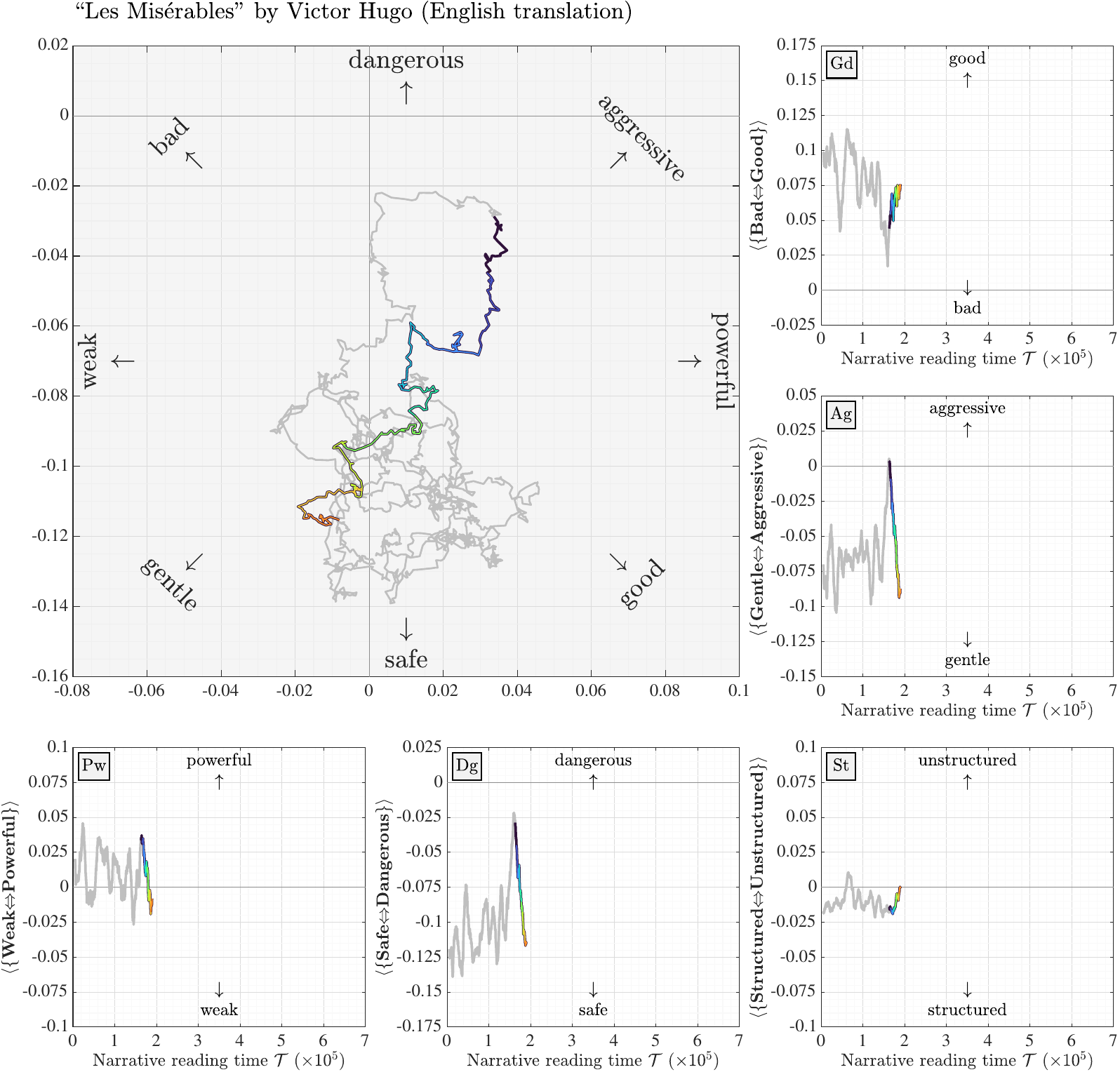}
  \caption{
    \textbf{
      Epoch 7 of 25 in Victor Hugo's ``Les Mis\'{e}rables.''
    }
  }
  \label{fig.meaning:figtelegnomic_timeseries_flipbook9017_007_10000_100_ousiometrics_GPADS100_LesMis_noname}
\end{figure*}

\begin{figure*}[t]
  \includegraphics[width=\textwidth]{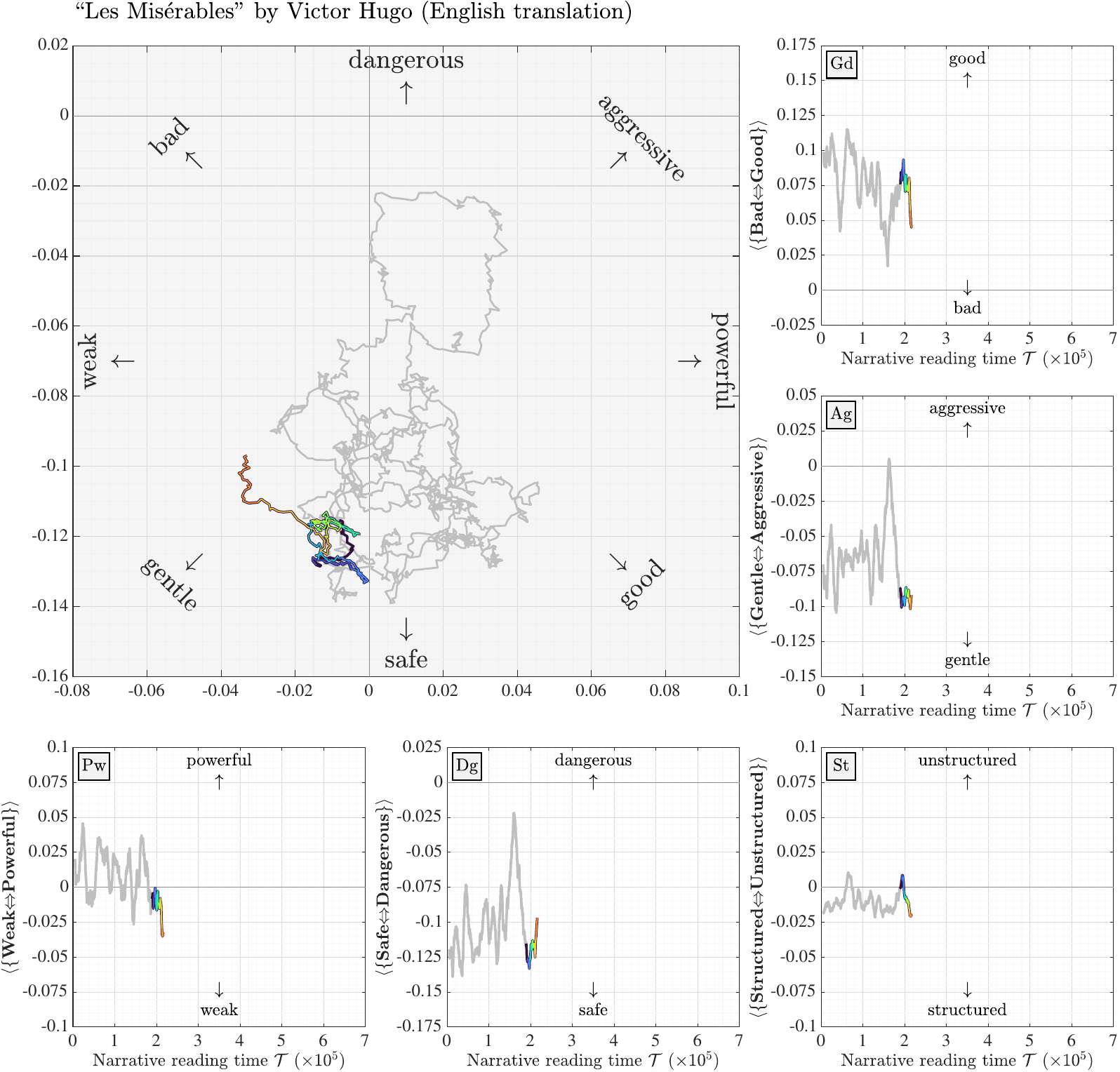}
  \caption{
    \textbf{
      Epoch 8 of 25 in Victor Hugo's ``Les Mis\'{e}rables.''
    }
  }
  \label{fig.meaning:figtelegnomic_timeseries_flipbook9017_008_10000_100_ousiometrics_GPADS100_LesMis_noname}
\end{figure*}

\begin{figure*}[t]
  \includegraphics[width=\textwidth]{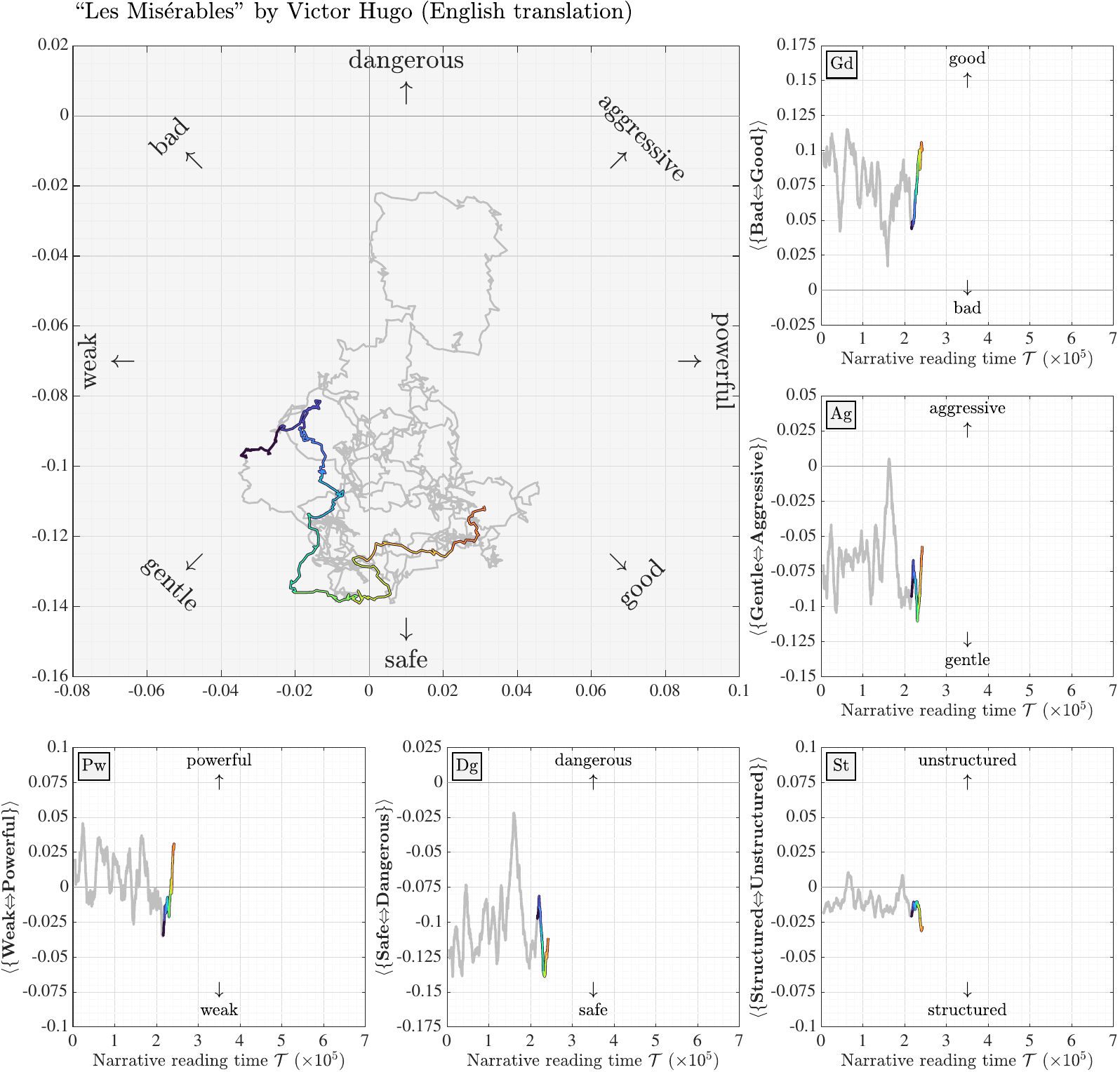}
  \caption{
    \textbf{
      Epoch 9 of 25 in Victor Hugo's ``Les Mis\'{e}rables.''
    }
  }
  \label{fig.meaning:figtelegnomic_timeseries_flipbook9017_009_10000_100_ousiometrics_GPADS100_LesMis_noname}
\end{figure*}

\begin{figure*}[t]
  \includegraphics[width=\textwidth]{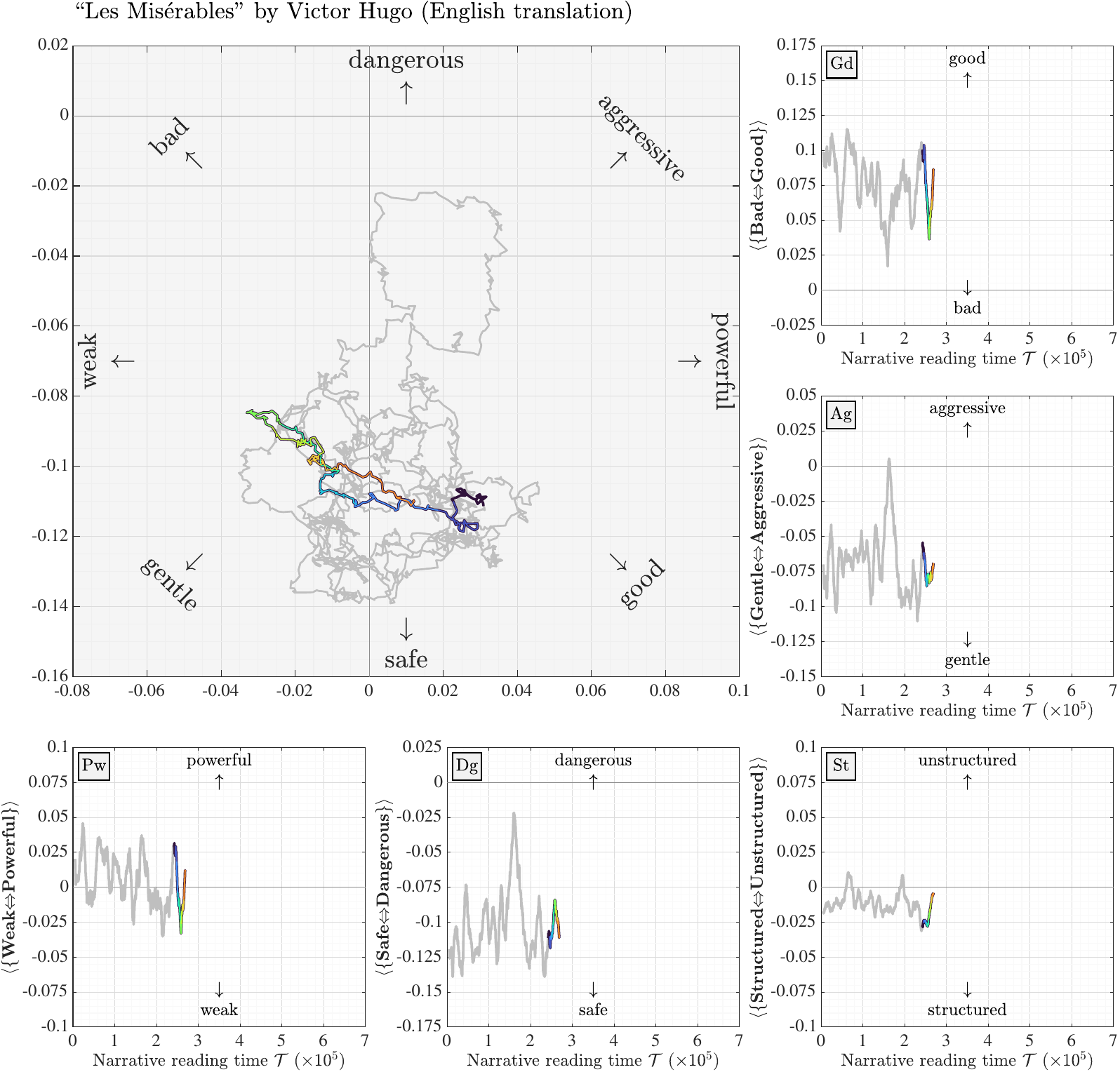}
  \caption{
    \textbf{
      Epoch 10 of 25 in Victor Hugo's ``Les Mis\'{e}rables.''
    }
  }
  \label{fig.meaning:figtelegnomic_timeseries_flipbook9017_010_10000_100_ousiometrics_GPADS100_LesMis_noname}
\end{figure*}

\begin{figure*}[t]
  \includegraphics[width=\textwidth]{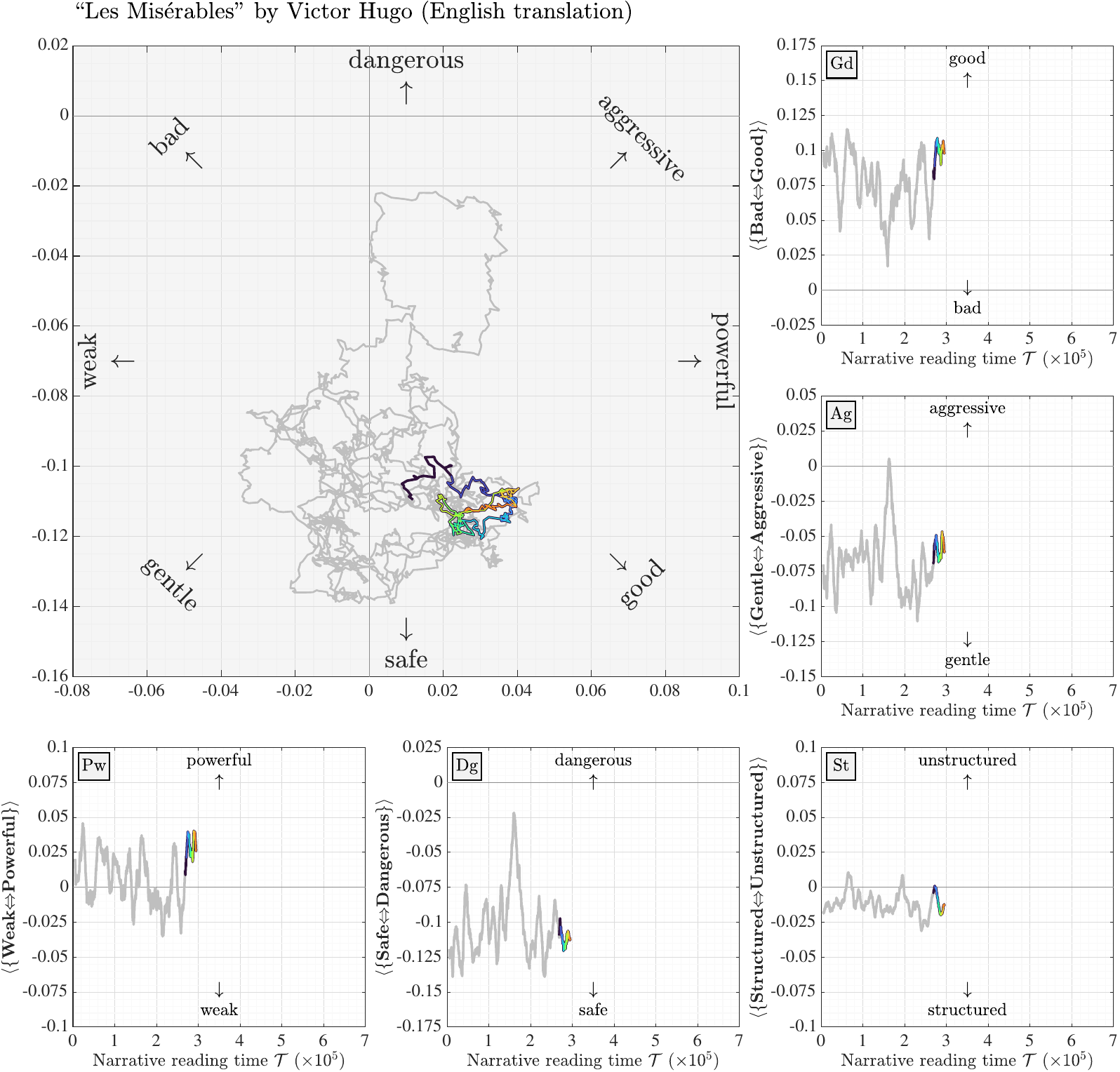}
  \caption{
    \textbf{
      Epoch 11 of 25 in Victor Hugo's ``Les Mis\'{e}rables.''
    }
  }
  \label{fig.meaning:figtelegnomic_timeseries_flipbook9017_011_10000_100_ousiometrics_GPADS100_LesMis_noname}
\end{figure*}

\begin{figure*}[t]
  \includegraphics[width=\textwidth]{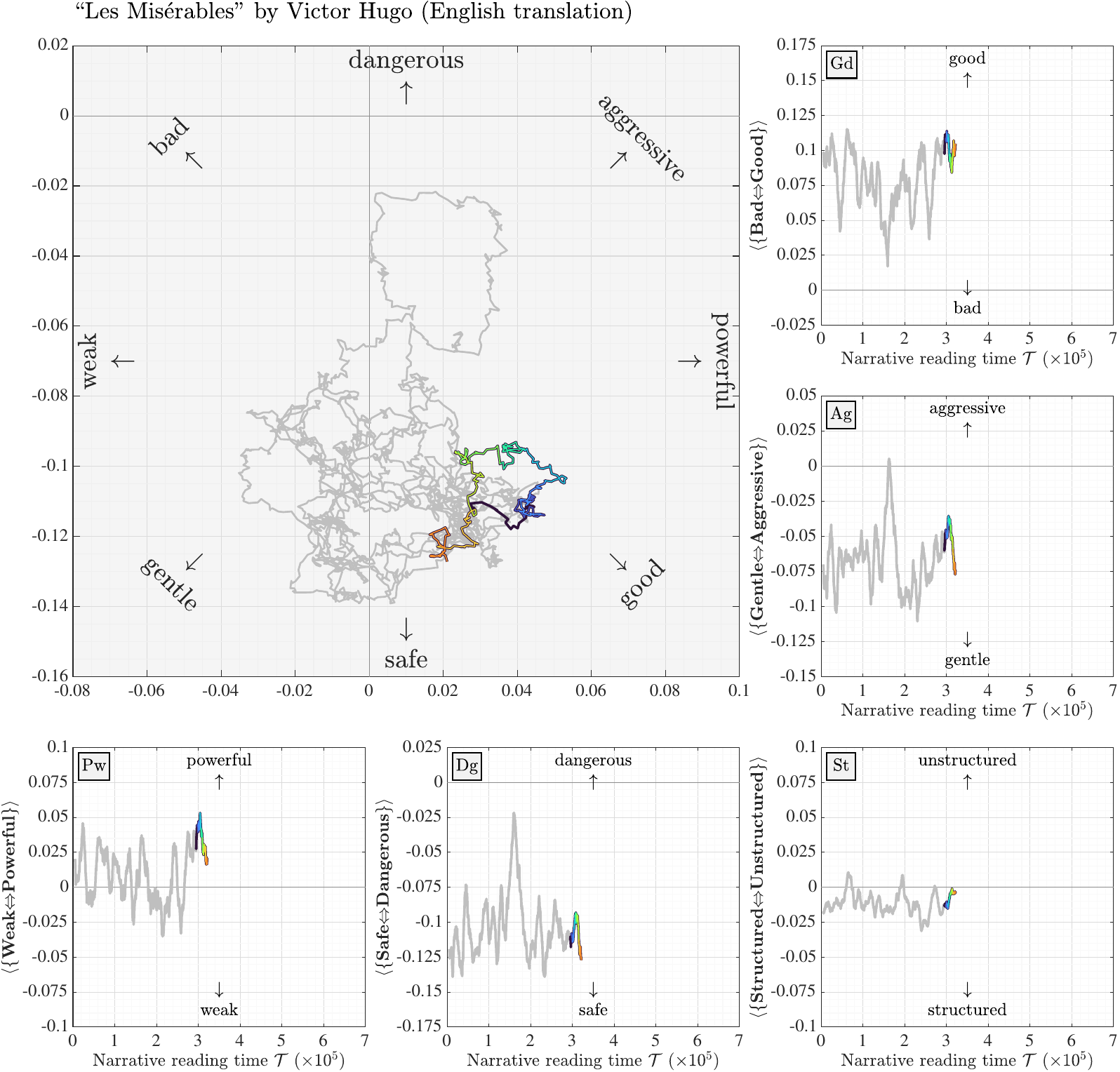}
  \caption{
    \textbf{
      Epoch 12 of 25 in Victor Hugo's ``Les Mis\'{e}rables.''
    }
  }
  \label{fig.meaning:figtelegnomic_timeseries_flipbook9017_012_10000_100_ousiometrics_GPADS100_LesMis_noname}
\end{figure*}

\begin{figure*}[t]
  \includegraphics[width=\textwidth]{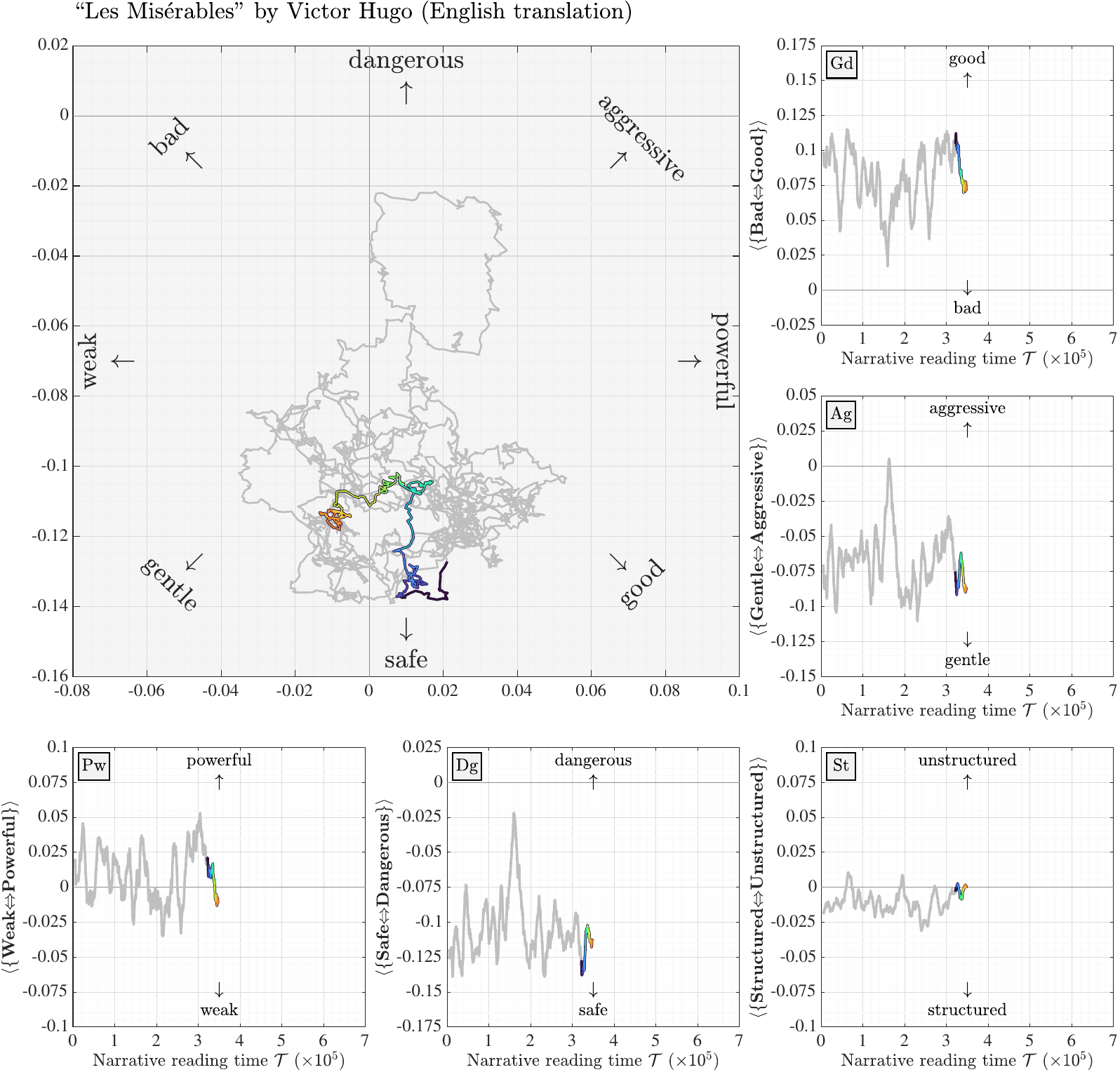}
  \caption{
    \textbf{
      Epoch 13 of 25 in Victor Hugo's ``Les Mis\'{e}rables.''
    }
  }
  \label{fig.meaning:figtelegnomic_timeseries_flipbook9017_013_10000_100_ousiometrics_GPADS100_LesMis_noname}
\end{figure*}

\begin{figure*}[t]
  \includegraphics[width=\textwidth]{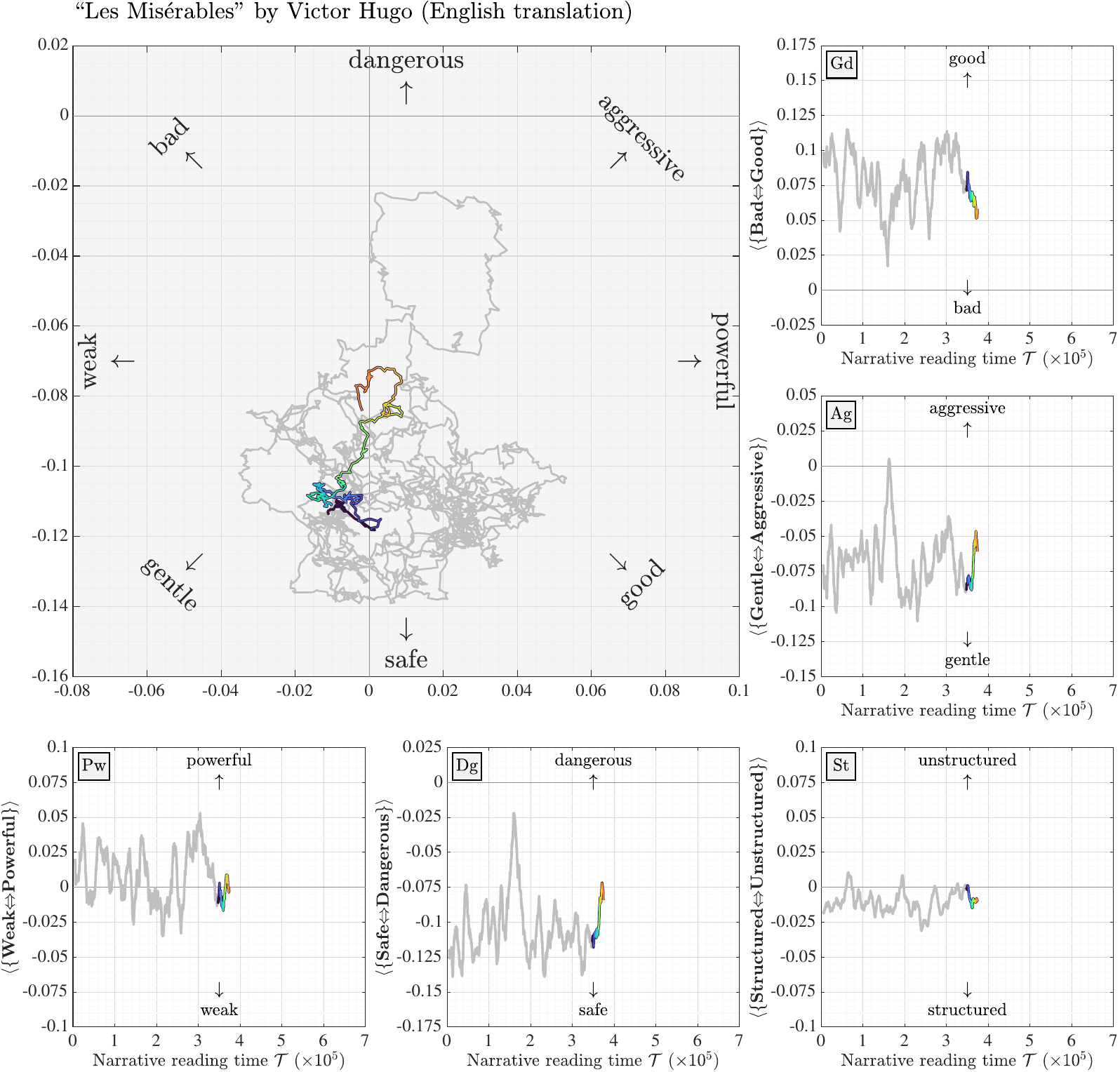}
  \caption{
    \textbf{
      Epoch 14 of 25 in Victor Hugo's ``Les Mis\'{e}rables.''
    }
  }
  \label{fig.meaning:figtelegnomic_timeseries_flipbook9017_014_10000_100_ousiometrics_GPADS100_LesMis_noname}
\end{figure*}

\begin{figure*}[t]
  \includegraphics[width=\textwidth]{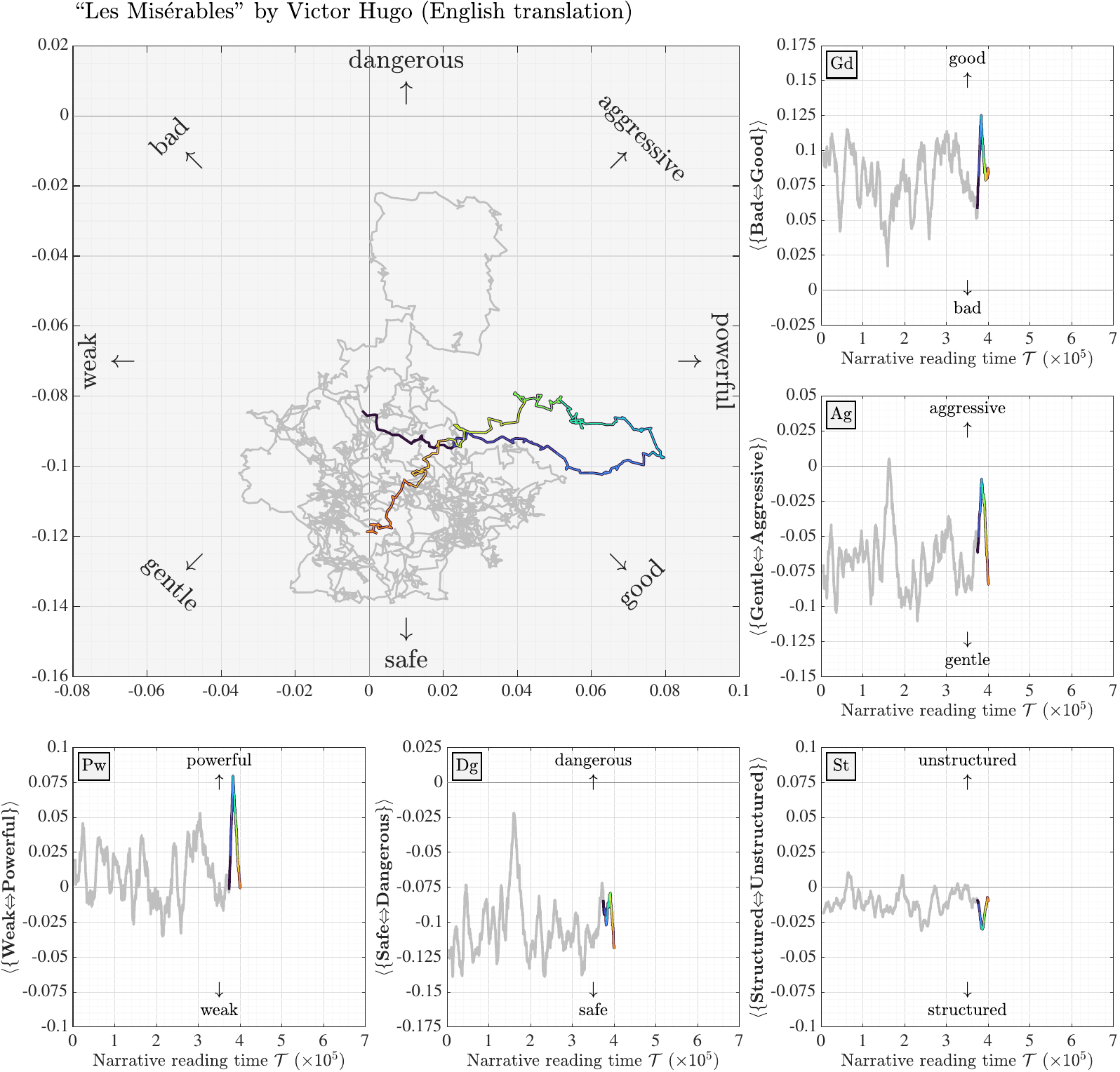}
  \caption{
    \textbf{
      Epoch 15 of 25 in Victor Hugo's ``Les Mis\'{e}rables.''
    }
  }
  \label{fig.meaning:figtelegnomic_timeseries_flipbook9017_015_10000_100_ousiometrics_GPADS100_LesMis_noname}
\end{figure*}

\begin{figure*}[t]
  \includegraphics[width=\textwidth]{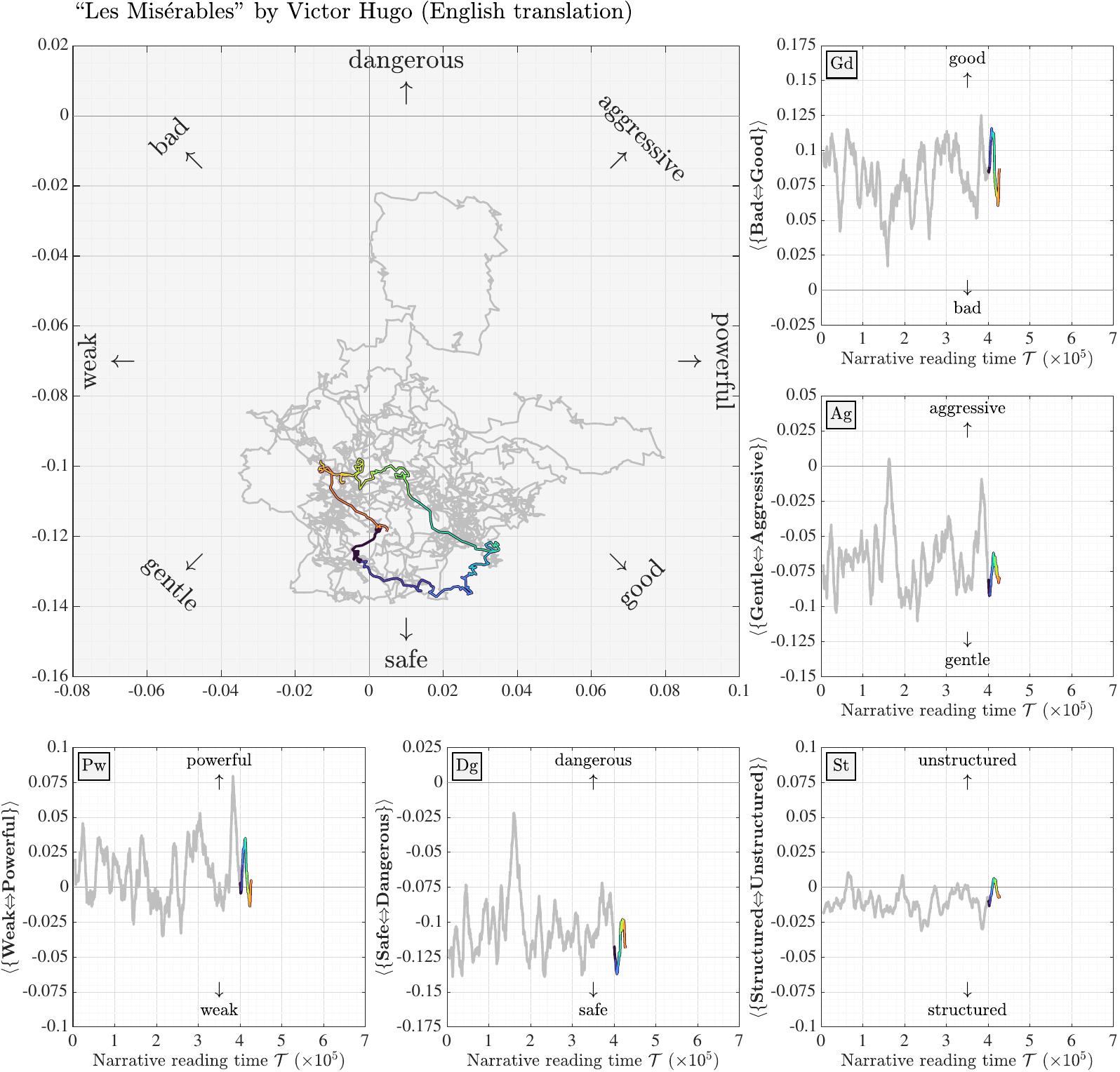}
  \caption{
    \textbf{
      Epoch 16 of 25 in Victor Hugo's ``Les Mis\'{e}rables.''
    }
  }
  \label{fig.meaning:figtelegnomic_timeseries_flipbook9017_016_10000_100_ousiometrics_GPADS100_LesMis_noname}
\end{figure*}

\begin{figure*}[t]
  \includegraphics[width=\textwidth]{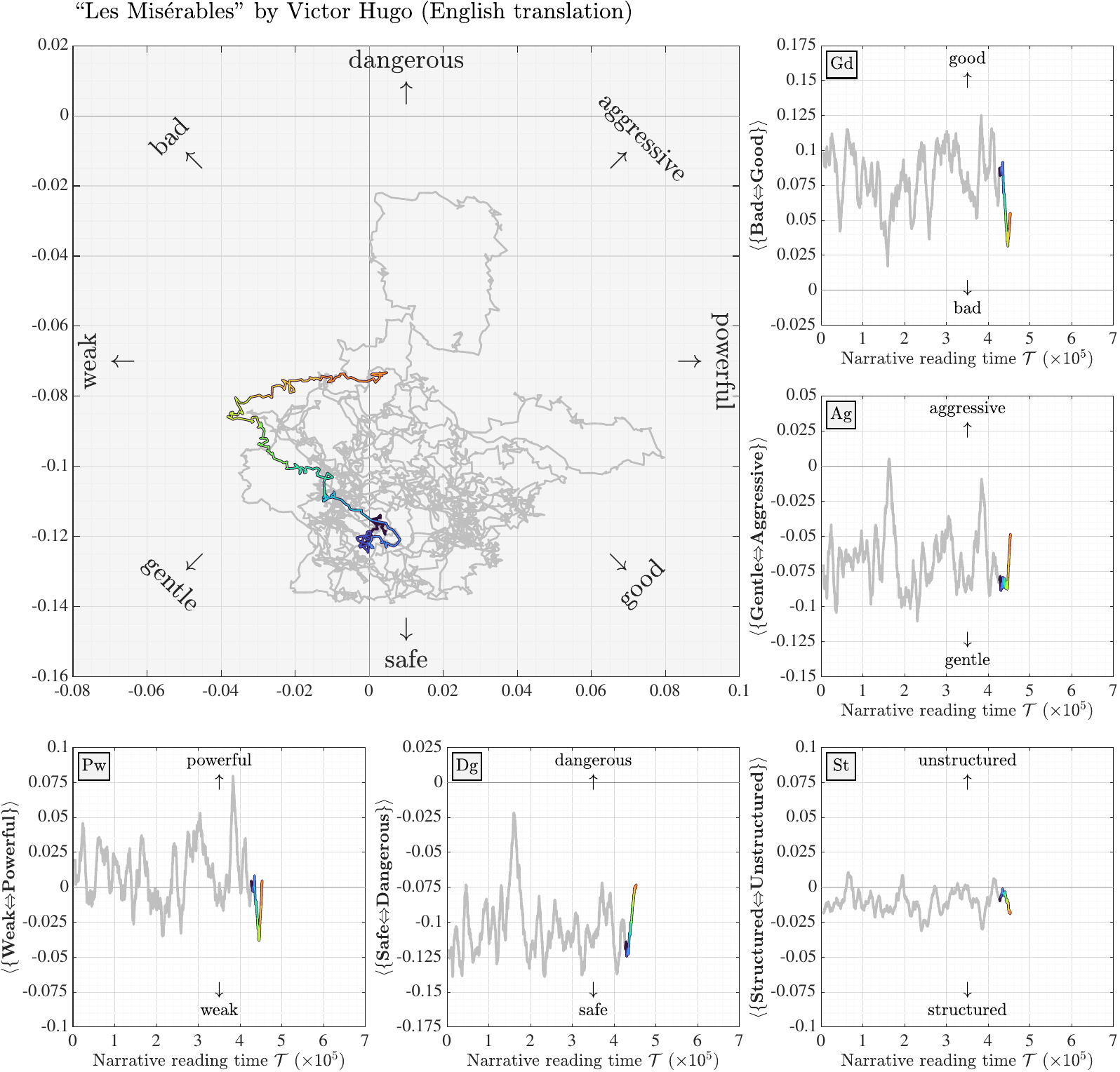}
  \caption{
    \textbf{
      Epoch 17 of 25 in Victor Hugo's ``Les Mis\'{e}rables.''
    }
  }
  \label{fig.meaning:figtelegnomic_timeseries_flipbook9017_017_10000_100_ousiometrics_GPADS100_LesMis_noname}
\end{figure*}

\begin{figure*}[t]
  \includegraphics[width=\textwidth]{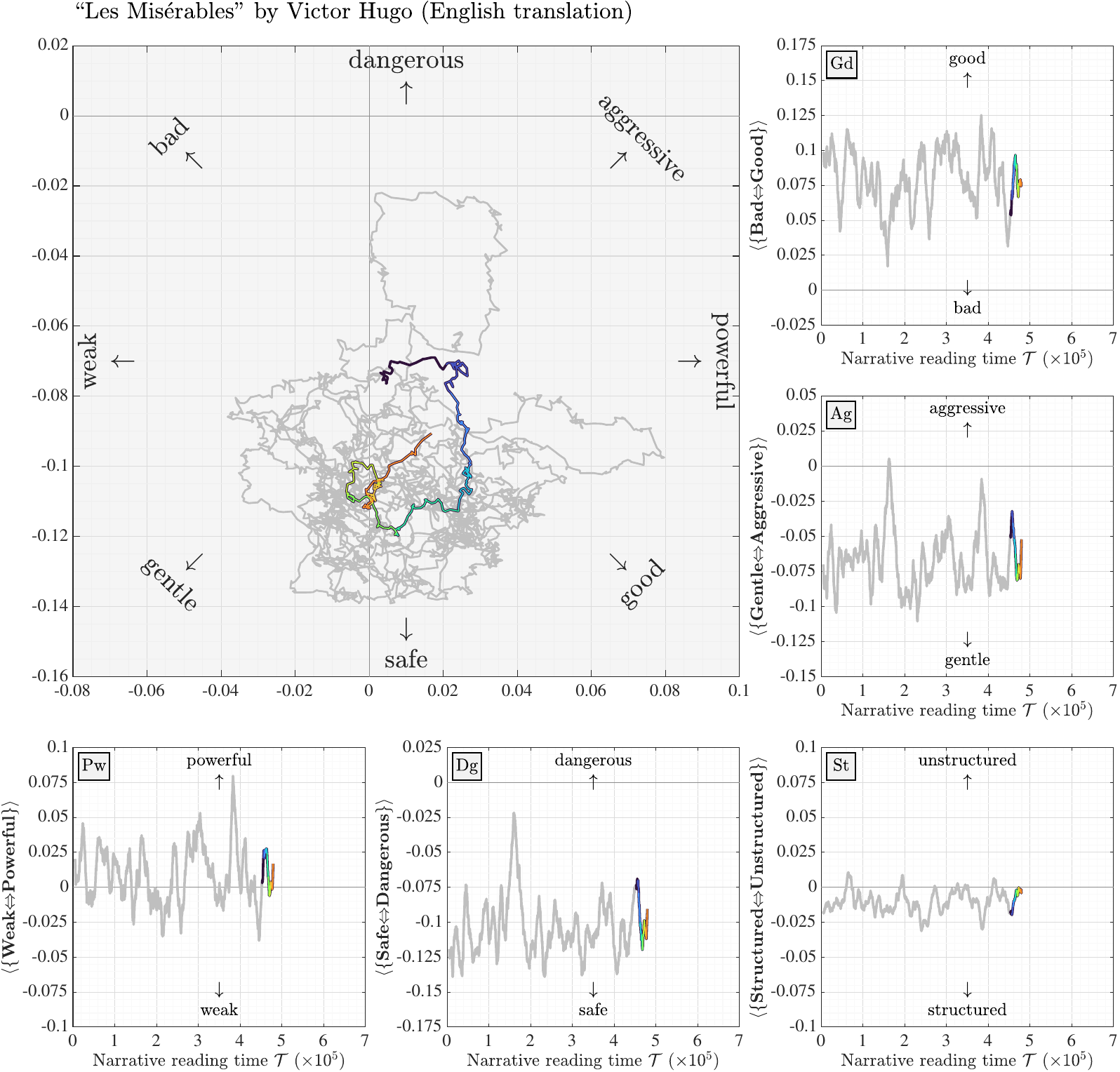}
  \caption{
    \textbf{
      Epoch 18 of 25 in Victor Hugo's ``Les Mis\'{e}rables.''
    }
  }
  \label{fig.meaning:figtelegnomic_timeseries_flipbook9017_018_10000_100_ousiometrics_GPADS100_LesMis_noname}
\end{figure*}

\begin{figure*}[t]
  \includegraphics[width=\textwidth]{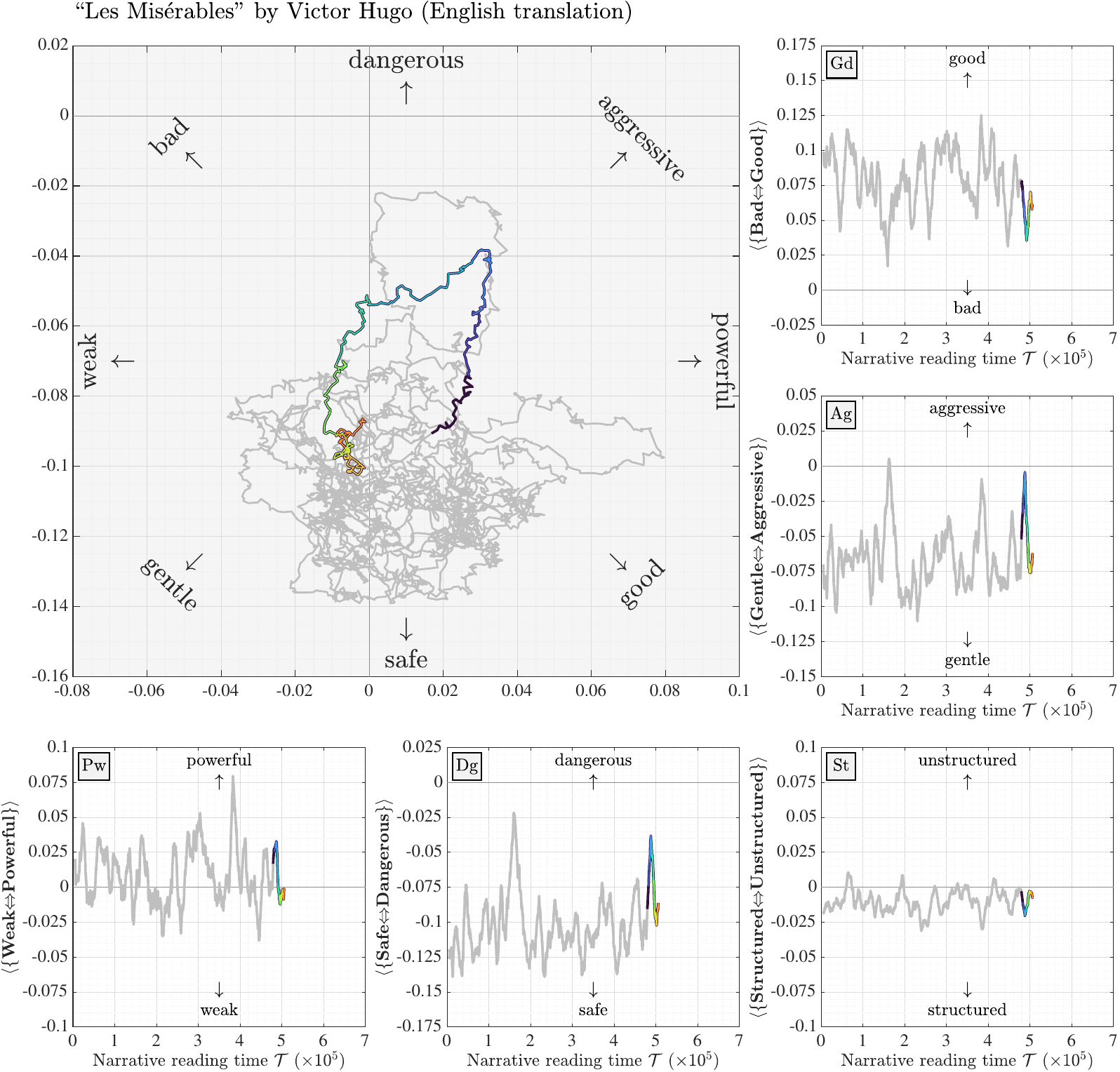}
  \caption{
    \textbf{
      Epoch 19 of 25 in Victor Hugo's ``Les Mis\'{e}rables.''
    }
  }
  \label{fig.meaning:figtelegnomic_timeseries_flipbook9017_019_10000_100_ousiometrics_GPADS100_LesMis_noname}
\end{figure*}

\begin{figure*}[t]
  \includegraphics[width=\textwidth]{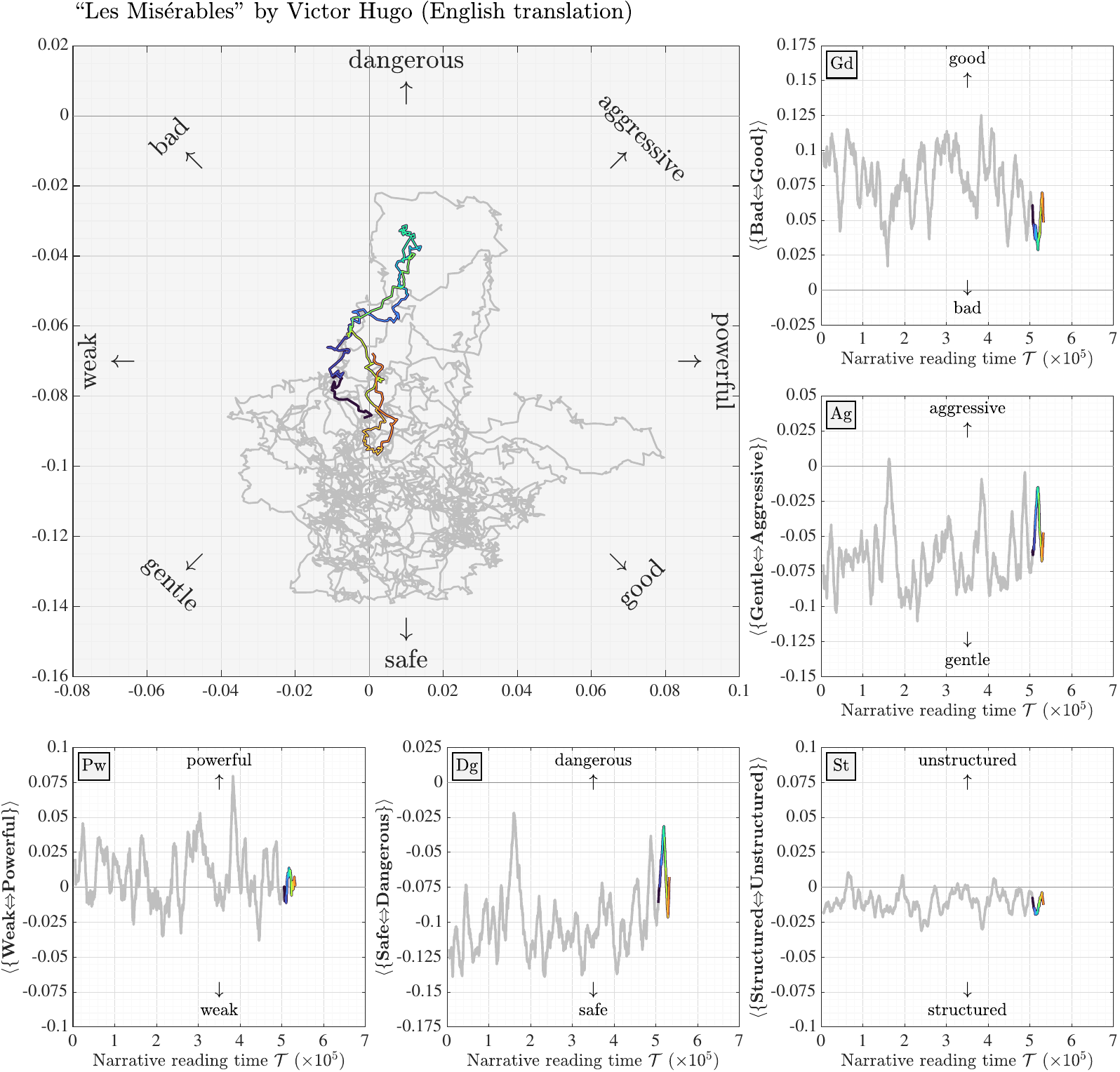}
  \caption{
    \textbf{
      Epoch 20 of 25 in Victor Hugo's ``Les Mis\'{e}rables.''
    }
  }
  \label{fig.meaning:figtelegnomic_timeseries_flipbook9017_020_10000_100_ousiometrics_GPADS100_LesMis_noname}
\end{figure*}

\begin{figure*}[t]
  \includegraphics[width=\textwidth]{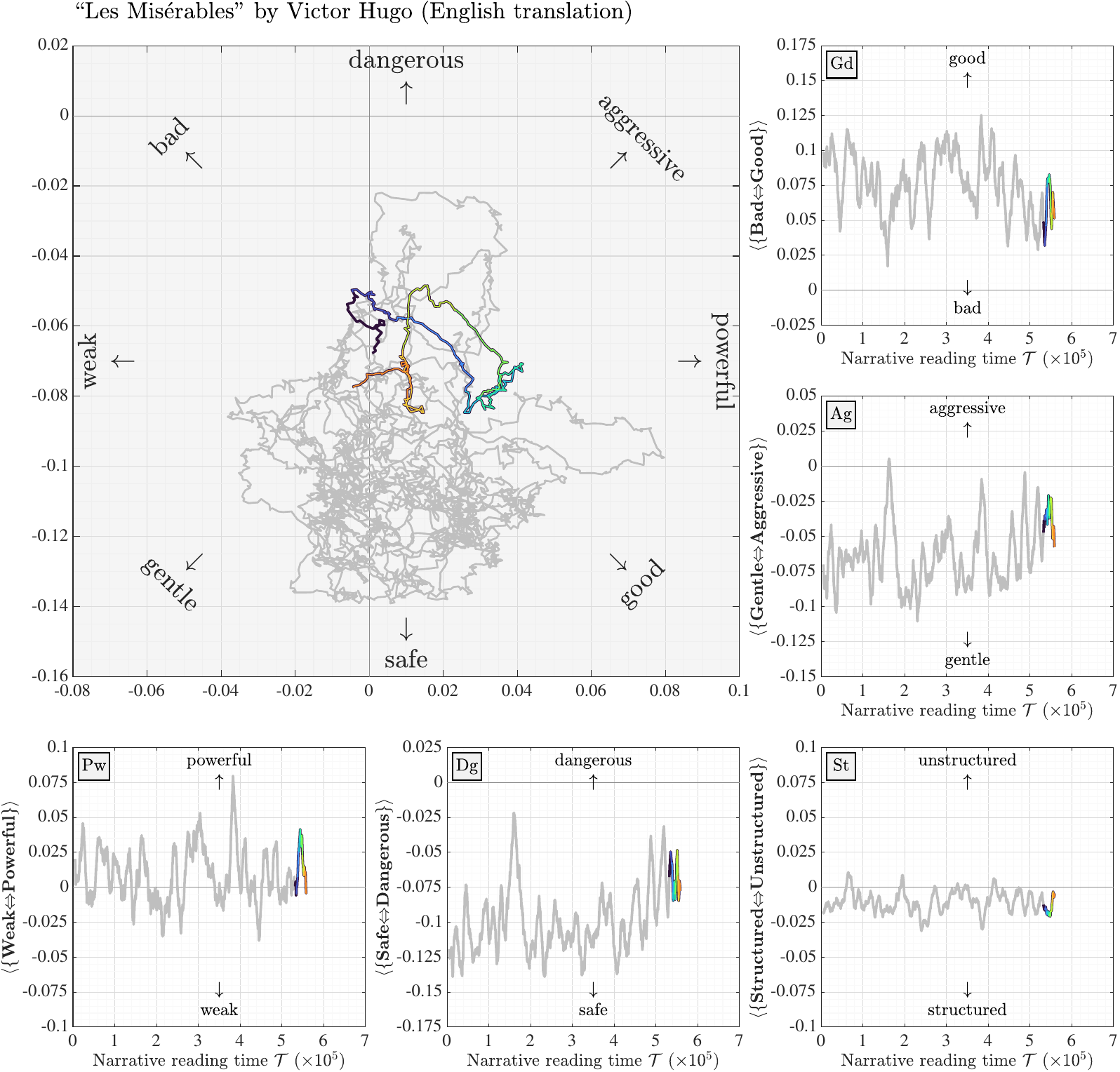}
  \caption{
    \textbf{
      Epoch 21 of 25 in Victor Hugo's ``Les Mis\'{e}rables.''
    }
  }
  \label{fig.meaning:figtelegnomic_timeseries_flipbook9017_021_10000_100_ousiometrics_GPADS100_LesMis_noname}
\end{figure*}

\begin{figure*}[t]
  \includegraphics[width=\textwidth]{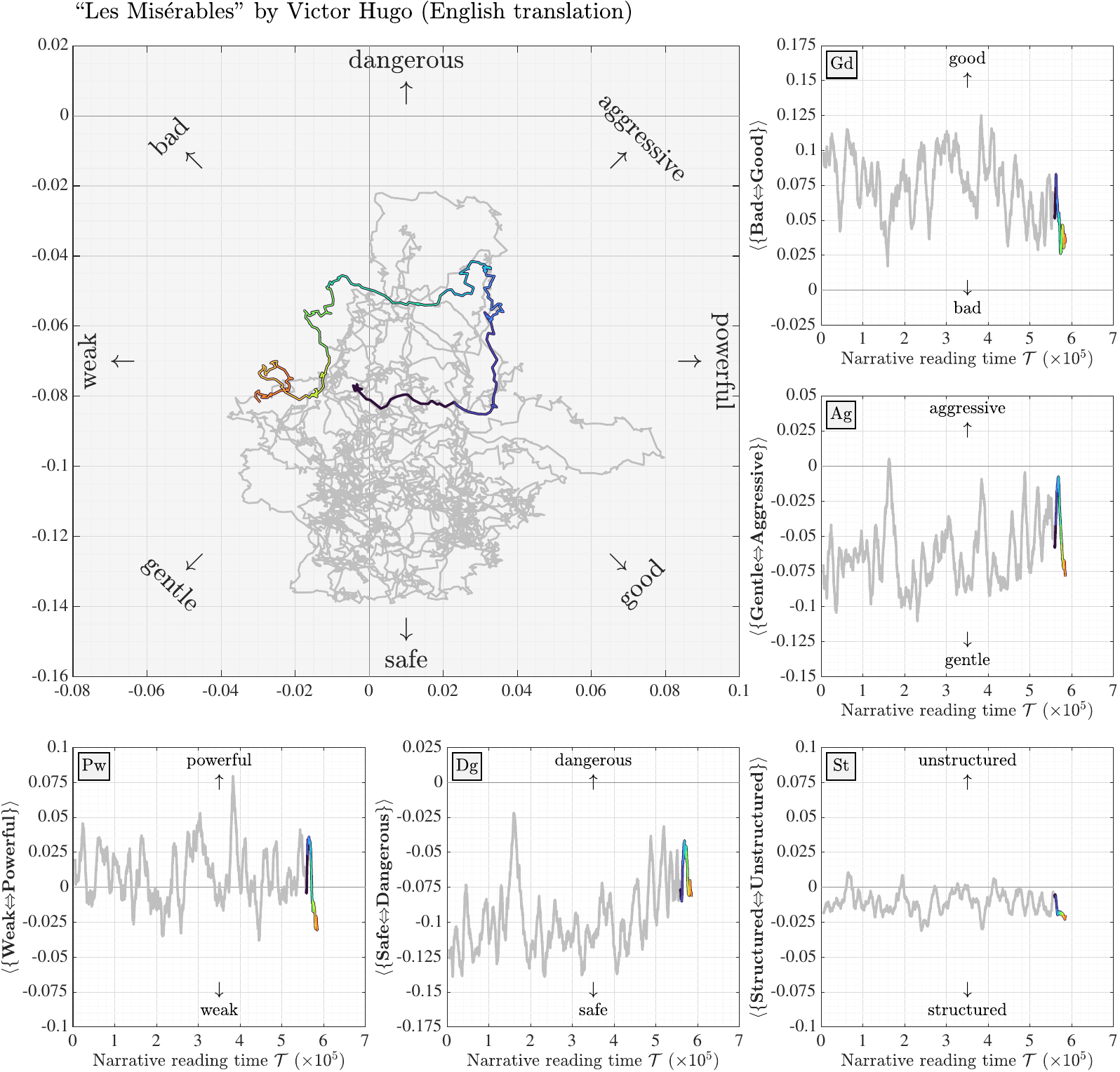}
  \caption{
    \textbf{
      Epoch 22 of 25 in Victor Hugo's ``Les Mis\'{e}rables.''
    }
  }
  \label{fig.meaning:figtelegnomic_timeseries_flipbook9017_022_10000_100_ousiometrics_GPADS100_LesMis_noname}
\end{figure*}

\begin{figure*}[t]
  \includegraphics[width=\textwidth]{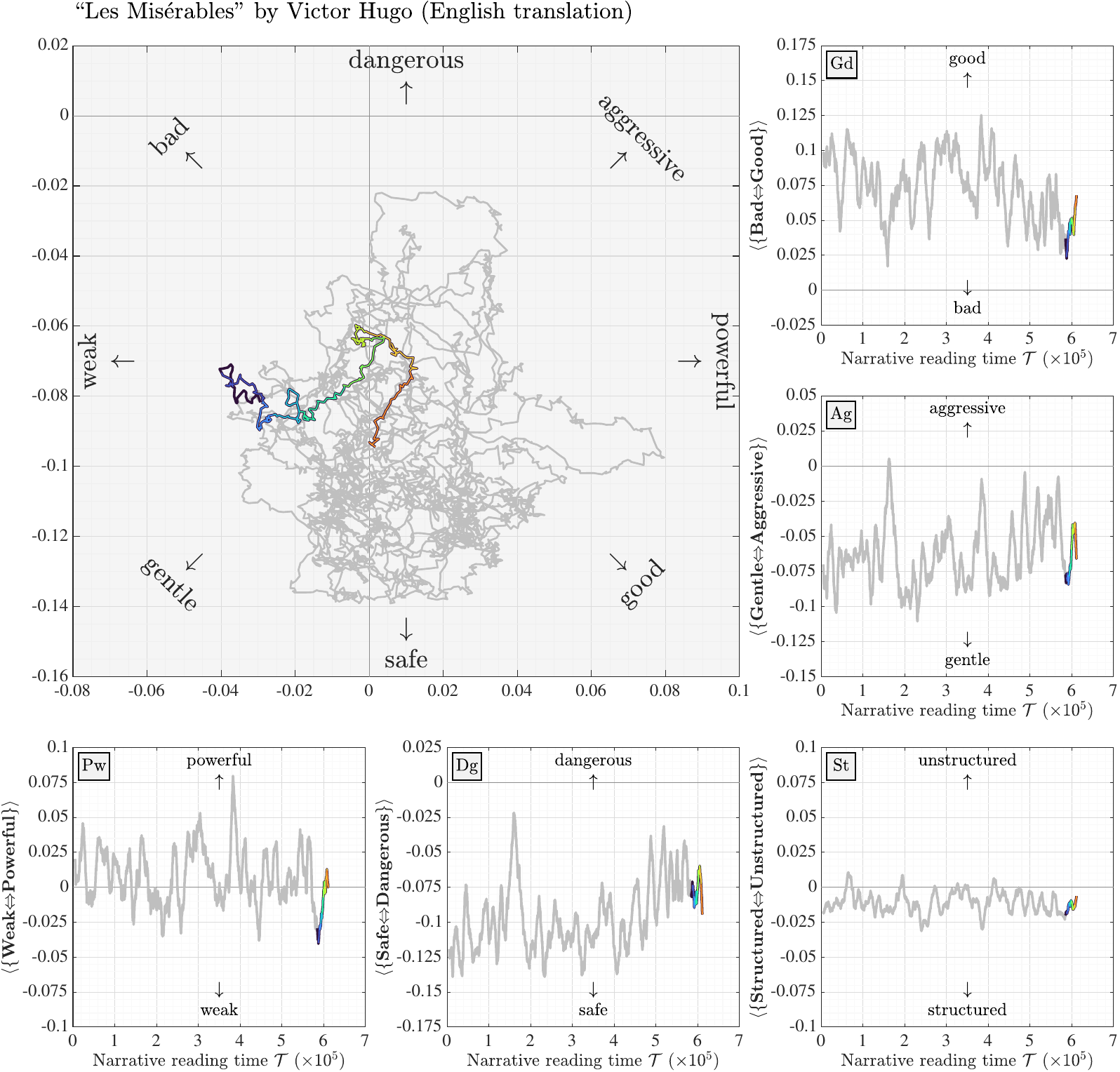}
  \caption{
    \textbf{
      Epoch 23 of 25 in Victor Hugo's ``Les Mis\'{e}rables.''
    }
  }
  \label{fig.meaning:figtelegnomic_timeseries_flipbook9017_023_10000_100_ousiometrics_GPADS100_LesMis_noname}
\end{figure*}

\begin{figure*}[t]
  \includegraphics[width=\textwidth]{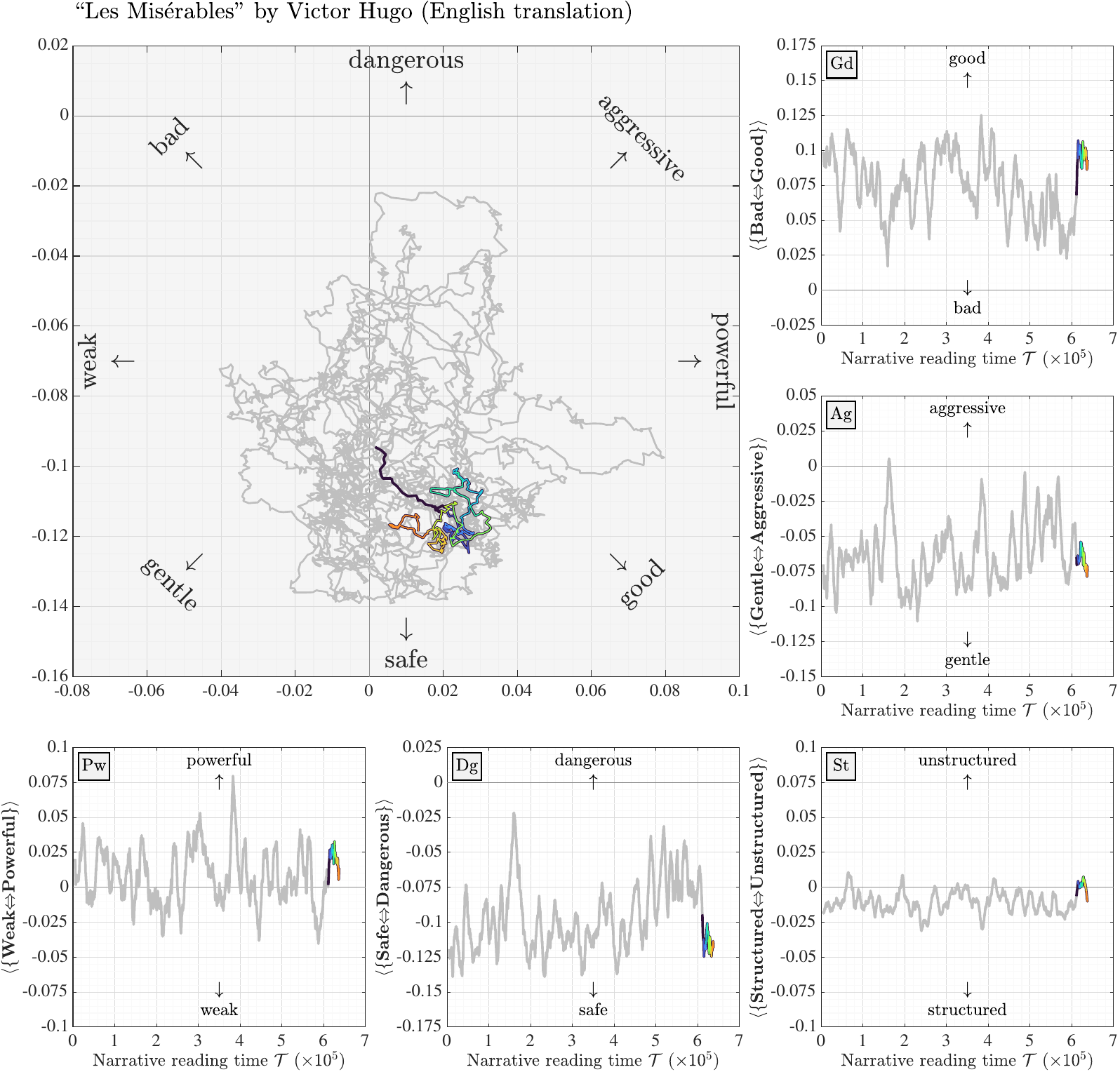}
  \caption{
    \textbf{
      Epoch 24 of 25 in Victor Hugo's ``Les Mis\'{e}rables.''
    }
  }
  \label{fig.meaning:figtelegnomic_timeseries_flipbook9017_024_10000_100_ousiometrics_GPADS100_LesMis_noname}
\end{figure*}

\begin{figure*}[t]
  \includegraphics[width=\textwidth]{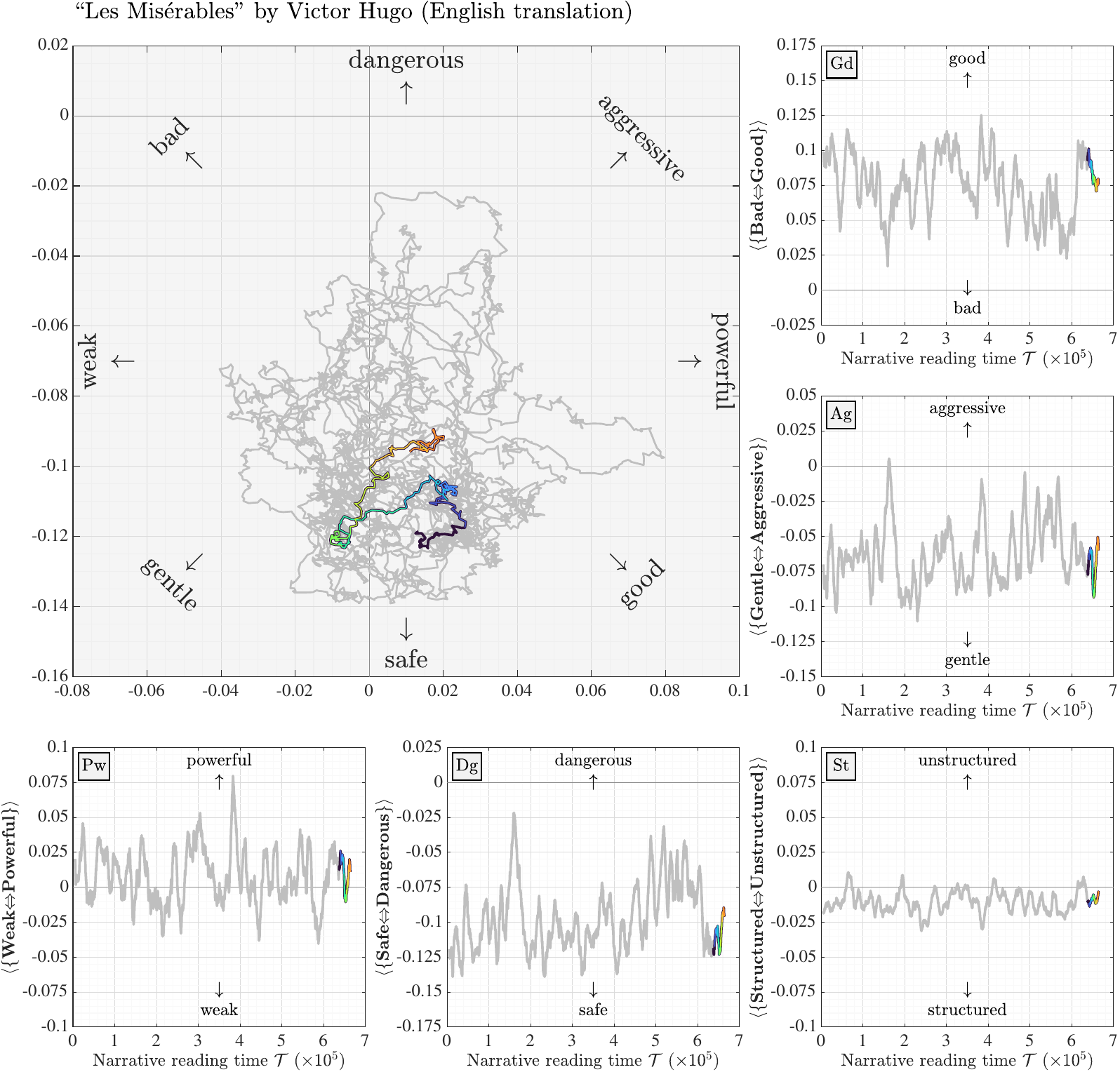}
  \caption{
    \textbf{
      Epoch 25 of 25 in Victor Hugo's ``Les Mis\'{e}rables.''
    }
  }
  \label{fig.meaning:figtelegnomic_timeseries_flipbook9017_025_10000_100_ousiometrics_GPADS100_LesMis_noname}
\end{figure*}

\begin{revisionbar}

  \textbf{Note:} The appendices are best viewed in single page mode
  rather than continuous scroll.

  In particular, these sequences of figures are aligned so
  that their sections function as flipbooks:
  \begin{itemize}
  \item
    Sec.~\ref{sec:meaning.appendix-ousiogram-three-frameworks}:
    Large ousiograms for the VAD, GAS, and PDS frameworks:
    Figs.~\ref{fig:meaning.ousiometer9010_VAD012_1_supp1}--\ref{fig:meaning.ousiometer9010_PDS012_3_supp1}.
  \item
    Sec.~\ref{sec:meaning.MRIs}:
    `MRI' slices of GPAD plane with Structure $\Mstructure$ varying:
    \ref{fig.meaning:figousiometer_3d_MRI_slices100_01_noname}--\ref{fig.meaning:figousiometer_3d_MRI_slices100_19_noname}.
  \item
    Sec.~\ref{sec:meaning.appendix-cube-tables}:
    Tables of words with largest components in GPADS framework for semantic differentials in cube model:
    Figs.~\ref{fig:meaning.ousiometrics_opposite_cubes001_001}--\ref{fig:meaning.ousiometrics_opposite_cubes001_013}.
  \item
    Sec.~\ref{sec:meaning.appendix-ousiogram-analytic-sequences}:
    Analytic sequence demonstrating the safety bias for six distinct corpora:
    Figs.~\ref{fig:meaning.ousiometer9310_jane_austen001}--\ref{fig:meaning.ousiometer9310_twitter001}.
  \item
    Sec.~\ref{sec:meaning.les-miserables-ousiometric-flipbook}:
    Epoch sequence of ousiometric time series
    for Victor Hugo's ``Les Mis\'{e}rables'':
    Figs.~\ref{fig.meaning:figtelegnomic_timeseries_flipbook9017_001_10000_100_ousiometrics_GPADS100_LesMis_noname}--\ref{fig.meaning:figtelegnomic_timeseries_flipbook9017_025_10000_100_ousiometrics_GPADS100_LesMis_noname}.
  \end{itemize}

\end{revisionbar}

\clearpage

\section{Large ousiograms for VAD, GAS, and PDS}
\label{sec:meaning.appendix-ousiogram-three-frameworks}

\clearpage

\clearpage

\twocolumn

\section{Synousionyms and antousionyms, and the problems with prescribing ousiometric axes through end-point descriptors}
\label{sec:meaning.synousionyms}

\setlength{\tabcolsep}{6pt}
\begin{table*}[tbp!]
  \small
  
  \textbf{Safe-Powerful (Good) to Dangerous-Weak (Bad) axis:}
  \rowcolors{1}{gray!15}{white}
  \begin{tabular}{r|ccc|ccc|ccc}
    \hline
    Synousionyms & Valence & Arousal & Dominance & Goodness & Aggression & Structure & Power & Danger & Structure \\
    \hline
    \hline
    \textbf{Anchor: wisdom} & 0.430 & -0.198 & 0.371  & 0.579 & -0.031 & -0.158 & 0.388 & -0.432 & -0.158 \\
    education & 0.396 & -0.225 & 0.340  & 0.539 & -0.065 & -0.167 & 0.336 & -0.427 & -0.167 \\
    healthy & 0.438 & -0.181 & 0.318  & 0.558 & -0.047 & -0.108 & 0.362 & -0.428 & -0.108 \\
    trustworthy & 0.469 & -0.185 & 0.324  & 0.589 & -0.052 & -0.100 & 0.379 & -0.453 & -0.100 \\
    reliable & 0.412 & -0.259 & 0.375  & 0.575 & -0.076 & -0.202 & 0.353 & -0.460 & -0.202 \\
    \hline
    Antousionyms & Valence & Arousal & Dominance & Goodness & Aggression & Structure & Power & Danger & Structure \\
    \hline
    \hline
    bullshit & -0.458 & 0.176 & -0.317 & -0.575 & 0.046 & 0.095 & -0.373 & 0.439 & 0.095 \\
    shitty & -0.480 & 0.179 & -0.337 & -0.604 & 0.042 & 0.100 & -0.397 & 0.456 & 0.100 \\
    nauseate & -0.438 & 0.160 & -0.324 & -0.558 & 0.026 & 0.101 & -0.376 & 0.413 & 0.101 \\
    weeping & -0.418 & 0.188 & -0.332 & -0.549 & 0.042 & 0.131 & -0.359 & 0.418 & 0.131 \\
    shame & -0.440 & 0.170 & -0.345 & -0.572 & 0.023 & 0.120 & -0.388 & 0.421 & 0.120 \\
    diarrhea & -0.408 & 0.184 & -0.357 & -0.552 & 0.023 & 0.151 & -0.374 & 0.407 & 0.151 \\
    \hline
  \end{tabular}

  \medskip
  
  \textbf{Powerful (Aggressive-Good) to Weak (Gentle-Bad) axis:}
  \smallskip\\
  \rowcolors{1}{gray!15}{white}
  \begin{tabular}{r|ccc|ccc|ccc}
    \hline
    Synousionyms & Valence & Arousal & Dominance & Goodness & Aggression & Structure & Power & Danger & Structure \\
    \hline
    \hline
    \textbf{Anchor: success} & 0.459 & 0.380 & 0.481  & 0.571 & 0.501 & 0.095 & 0.758 & -0.050 & 0.095 \\
    almighty & 0.438 & 0.374 & 0.458  & 0.543 & 0.487 & 0.098 & 0.728 & -0.040 & 0.098 \\
    triumphant & 0.449 & 0.337 & 0.472  & 0.565 & 0.462 & 0.073 & 0.726 & -0.072 & 0.073 \\
    champion & 0.390 & 0.380 & 0.445  & 0.494 & 0.492 & 0.087 & 0.698 & -0.001 & 0.087 \\
    victorious & 0.384 & 0.386 & 0.446  & 0.489 & 0.499 & 0.087 & 0.698 & 0.007 & 0.087 \\
    \hline
    Antousionyms & Valence & Arousal & Dominance & Goodness & Aggression & Structure & Power & Danger & Structure \\
    \hline
    \hline
    sorrow & -0.448 & -0.265 & -0.336 & -0.509 & -0.329 & -0.127 & -0.593 & 0.127 & -0.127 \\
    tasteless & -0.354 & -0.304 & -0.352 & -0.430 & -0.385 & -0.092 & -0.576 & 0.032 & -0.092 \\
    idle & -0.321 & -0.333 & -0.388 & -0.414 & -0.434 & -0.068 & -0.600 & -0.014 & -0.068 \\
    empty & -0.312 & -0.317 & -0.419 & -0.424 & -0.439 & -0.033 & -0.610 & -0.011 & -0.033 \\
    void & -0.365 & -0.337 & -0.370 & -0.443 & -0.420 & -0.103 & -0.611 & 0.016 & -0.103 \\
    \hline
  \end{tabular}

  \medskip

  \textbf{Dangerous-Powerful (Aggressive) to Safe-Weak (Gentle) axis:}
  \begin{tabular}{r|ccc|ccc|ccc}
    \hline
    Synousionyms & Valence & Arousal & Dominance & Goodness & Aggression & Structure & Power & Danger & Structure \\
    \hline
    \hline
    \textbf{Anchor: volcanic} & -0.156 & 0.410 & 0.281  & -0.061 & 0.515 & -0.045 & 0.322 & 0.407 & -0.045 \\
    shelling & -0.163 & 0.417 & 0.273  & -0.072 & 0.518 & -0.039 & 0.316 & 0.417 & -0.039 \\
    artillery & -0.150 & 0.412 & 0.294  & -0.050 & 0.523 & -0.050 & 0.335 & 0.405 & -0.050 \\
    wild & -0.188 & 0.422 & 0.250  & -0.105 & 0.514 & -0.032 & 0.289 & 0.438 & -0.032 \\
    rifles & -0.163 & 0.364 & 0.265  & -0.068 & 0.470 & -0.062 & 0.284 & 0.380 & -0.062 \\
    \hline
    Antousionyms & Valence & Arousal & Dominance & Goodness & Aggression & Structure & Power & Danger & Structure \\
    \hline
    \hline
    couch & 0.094 & -0.418 & -0.302 & -0.002 & -0.524 & 0.025 & -0.372 & -0.369 & 0.025 \\
    mellow & 0.133 & -0.431 & -0.235 & 0.066 & -0.504 & -0.009 & -0.310 & -0.403 & -0.009 \\
    pillow & 0.163 & -0.372 & -0.305 & 0.049 & -0.498 & 0.085 & -0.317 & -0.387 & 0.085 \\
    tortoise & 0.173 & -0.422 & -0.250 & 0.092 & -0.511 & 0.025 & -0.297 & -0.427 & 0.025 \\
    quilt & 0.143 & -0.377 & -0.274 & 0.048 & -0.482 & 0.052 & -0.307 & -0.375 & 0.052 \\
    cotton & 0.139 & -0.429 & -0.260 & 0.059 & -0.517 & 0.012 & -0.324 & -0.407 & 0.012 \\
    \hline
  \end{tabular}

  \medskip

  \textbf{Dangerous (Aggressive-Bad) to Safe (Gentle-Good) axis:}
  \smallskip\\
  \rowcolors{1}{gray!15}{white}
  \begin{tabular}{r|ccc|ccc|ccc}
    \hline
    Synousionyms & Valence & Arousal & Dominance & Goodness & Aggression & Structure & Power & Danger & Structure \\
    \hline
    \hline
    \textbf{Anchor: homicide} & -0.490 & 0.473 & 0.018  & -0.485 & 0.478 & 0.011 & -0.005 & 0.681 & 0.011 \\
    killer & -0.459 & 0.471 & 0.043  & -0.446 & 0.485 & 0.008 & 0.028 & 0.658 & 0.008 \\
    psychopath & -0.460 & 0.443 & 0.036  & -0.446 & 0.458 & -0.003 & 0.009 & 0.640 & -0.003 \\
    bloodshed & -0.452 & 0.442 & 0.025  & -0.444 & 0.450 & 0.008 & 0.004 & 0.633 & 0.008 \\
    violate & -0.439 & 0.470 & 0.019  & -0.440 & 0.468 & 0.033 & 0.020 & 0.642 & 0.033 \\
    \hline
    Antousionyms & Valence & Arousal & Dominance & Goodness & Aggression & Structure & Power & Danger & Structure \\
    \hline
    \hline
    natural & 0.354 & -0.382 & -0.019 & 0.354 & -0.382 & -0.026 & -0.020 & -0.520 & -0.026 \\
    tranquil & 0.417 & -0.406 & -0.145 & 0.351 & -0.480 & 0.078 & -0.091 & -0.588 & 0.078 \\
    softness & 0.375 & -0.414 & -0.098 & 0.338 & -0.455 & 0.021 & -0.082 & -0.561 & 0.021 \\
    serenity & 0.400 & -0.378 & 0.057 & 0.429 & -0.345 & -0.054 & 0.060 & -0.547 & -0.054 \\
    comfortable & 0.427 & -0.337 & -0.027 & 0.406 & -0.361 & 0.039 & 0.032 & -0.542 & 0.039 \\
    calmness & 0.434 & -0.395 & -0.106 & 0.383 & -0.453 & 0.065 & -0.049 & -0.591 & 0.065 \\
    \hline
  \end{tabular}
  \caption{
    \textbf{
      Example synousionyms and antousionyms for the four axes
      of the GAS and PDS frameworks.
    }
    For four anchor words
    `wisdom',
    `success',
    `volcanic',
    and
    `homicide',
    we list
    their four closest synousionyms
    and five antousionyms.
    For all words, we record
    scores in the three frameworks of VAD, GAS, and PDS.
    See the linear transformations of \Req{eq:meaning.VAD-to-GAS}
    and \Req{eq:meaning.transformVAD-to-PDS} for
    how VAD connects with GAS and PDS.
  }
  \label{tab:meaning.synousionyms}
\end{table*}

%% Before moving from types to tokens,
%% We add a further consideration on
%% the naming of the essential dimensions of meaning.

Which words and terms match in terms of essence of meaning?
We define synousionym and antousionym as
the ousiometric equivalents of
synonym and antonym.

To determine a word's synousionyms,
we find the words closest in PDS-space
(the specific framework does not matter).
For antousionyms, we find words
closest to the negated point in PDS-space
$(-\Mpower,-\Mdanger,-\Mstructure)$
(we could equivalently use GAS).

Distilling words to their essential meaning may affect
synonym and antonym pairs in opposite ways.
Words that are not synonyms may be
synousionyms,
while words that are antonyms may not be antousionyms.

For example, the word `failure',
$(\Mpower,\Mdanger,\Mstructure)$
=
(-0.39, 0.28, 0.13),
is not the antousionym of `success',
$(\Mpower,\Mdanger,\Mstructure)$
=
(0.76, -0.05, 0.09).
Within the NRC VAD lexicon,
the closest antousionym for `success' is
`empty',
$(\Mpower,\Mdanger,\Mstructure)$
=
(-0.61, -0.01, -0.03).
In Tab.~\ref{tab:meaning.synousionyms}, we show
the closest four
synousionyms as well as five antousionyms
for
the words
`wisdom',
`success',
`volcanic',
and
`homicide'.
These words are examples of four extreme points of the power-danger ousiogram:
safe-powerful,
powerful,
dangerous-powerful,
and
dangerous.

In Sec.~\ref{subsec:meaning.DVousiogram}, we noted that choosing names of
ousiometric dimensions may be problematic,
going beyond the issues of end-point descriptors.
For one example,
the word `goodness'
has the following VAD, GAS, and PDS scores:
(0.47, -0.18, 0.21),
(0.54, -0.11, -0.02),
and
(0.30, -0.45, -0.02).
We see that `goodness' has a non-neutral low aggression component
and is not purely aligned with the Goodness axis.
%% Goodness itself does not have equal components of power and danger;
%% it is fact slightly below (low aggression) what we called the goodness axis.
The five closest synousionyms of `goodness' are
`thankful',
`friendship',
`motherly',
`hope',
and
`graciously'
while
the five top antousionyms of `goodness' are
`frustrating',
`cadaver',
`displease',
`shameful',
and
`disrespectful'.
The antonym `badness' is not a close antousionym of `goodness'
with VAD, GAS, and PDS scores of
(-0.406, 0.323, -0.037),
(-0.417, 0.311, 0.008),
and
(-0.075, 0.515, 0.008).
Within the PDS framework,
while `badness' is  aligned with the danger axis,
`goodness' is in the safe-powerful quadrant.
Some close synousionyms for `badness'
are `rabid', `shatter', and `tremor'
and for antousionyms, we find
`comfortable',
`homestead',
and
`peacetime'.

A further complication for determining end-point descriptors
is that due to the asymmetric, point coverage of essential meaning space,
the closest antousionym may not be reflexive.
For example, 
`chaos' has PDS scores
(-0.13, 0.67, 0.09).
The closest antousionym for `chaos' is 
`angel'
(0.19, -0.52, -0.13)
whose closest antousionym is
`shattered'
(-0.19, 0.49, 0.11).

These observations again point to the difficulties of
prescribing dimensions for participants in surveys.
The solution is to see end-point descriptors as guides only
and to always examine how participants responded using SVD.

For the PDS framework,  `powerful' and `dangerous' align well
with the end-points of their respective axes with PDS
scores of
(0.70, -0.02, 0.02)
and
(0.09, 0.66, 0.10).
The word `weak' similarly aligns well with the negative power axis
with PDS scores of (-0.61, 0.03, 0.02).
And `powerful' and `weak' are both antonyms and close antousionyms of each other.

The descriptor `safe' does not perform as cleanly however,
as it connotes more-than-neutral power
with PDS scores of (0.29, -0.41, -0.09).
Antousionyms for `dangerous' are
`relaxed',
`softness',
`calming',
`relaxant',
and
`calmness'.
(The closest antousionym for `safe' is `seasick'.)
\begin{revisionbar}
  However semantic differentials also must divide a space into two halves,
  and \semdiff{safe}{dangerous} does this well when we consider
  words above and below the \semdiff{weak}{powerful} axis.
  We also feel `safe' functions well conceptually
  as an end-point descriptor
  as it is an easily reached antonym of `dangerous',
  if not also an antousionym.
\end{revisionbar}

We note that in developing our work,
we entertained a number of alternative names for the PDS framework
including
Success, Stress, and Structure,
and
Power, Peril, and Play.
Ultimately, both of these choices are limited as truly general ousiometric frameworks
with success and play in particular eliciting people-centric themes.
And in any case, while alliteration may appeal to some,
the confusion of variables starting with the same letter
would be problematic.

The full space of synousionyms and antousionyms can
be explored using VAD, GAS, and PDS scores for all words and terms
in the \suppmaterial.

\onecolumn

\clearpage

\section{Flipbook `MRIs' in power-danger-structure framework}
\label{sec:meaning.MRIs}

\clearpage

\clearpage

\clearpage

\clearpage

\clearpage

\clearpage

\clearpage

\clearpage

\clearpage

\clearpage

\clearpage

\clearpage

\clearpage

\clearpage

\clearpage

\clearpage

\clearpage

\clearpage

\clearpage

\clearpage

\section{Tables for the 13 semantic differential pairs within cube-based ousiometric framework}
\label{sec:meaning.appendix-cube-tables}

\clearpage

\clearpage

\section{Ousiogram analysis sequences for real corpora}
\label{sec:meaning.appendix-ousiogram-analytic-sequences}

\clearpage

\clearpage

\section{Ousiometric time series and trajectory for Les Mis\'{e}rables: Flipbook}
\label{sec:meaning.les-miserables-ousiometric-flipbook}

\clearpage

%% read in
%% figures/2025-06ousiometric-time-series-books/figtelegnomic_timeseries_flipbook9017.tex

\clearpage

\clearpage

\clearpage

\clearpage

\clearpage

\clearpage

\clearpage

\clearpage

\clearpage

\clearpage

\clearpage

\clearpage

\clearpage

\clearpage

\clearpage

\clearpage

\clearpage

\clearpage

\clearpage

\clearpage

\clearpage

\clearpage

\clearpage

\clearpage

\clearpage

\section{Hedonometer for Les Mis\'{e}rables}
\label{sec:meaning.les-miserables-hedonometer}

\clearpage

\begin{figure*}[tp!]
  \includegraphics[width=\textwidth]{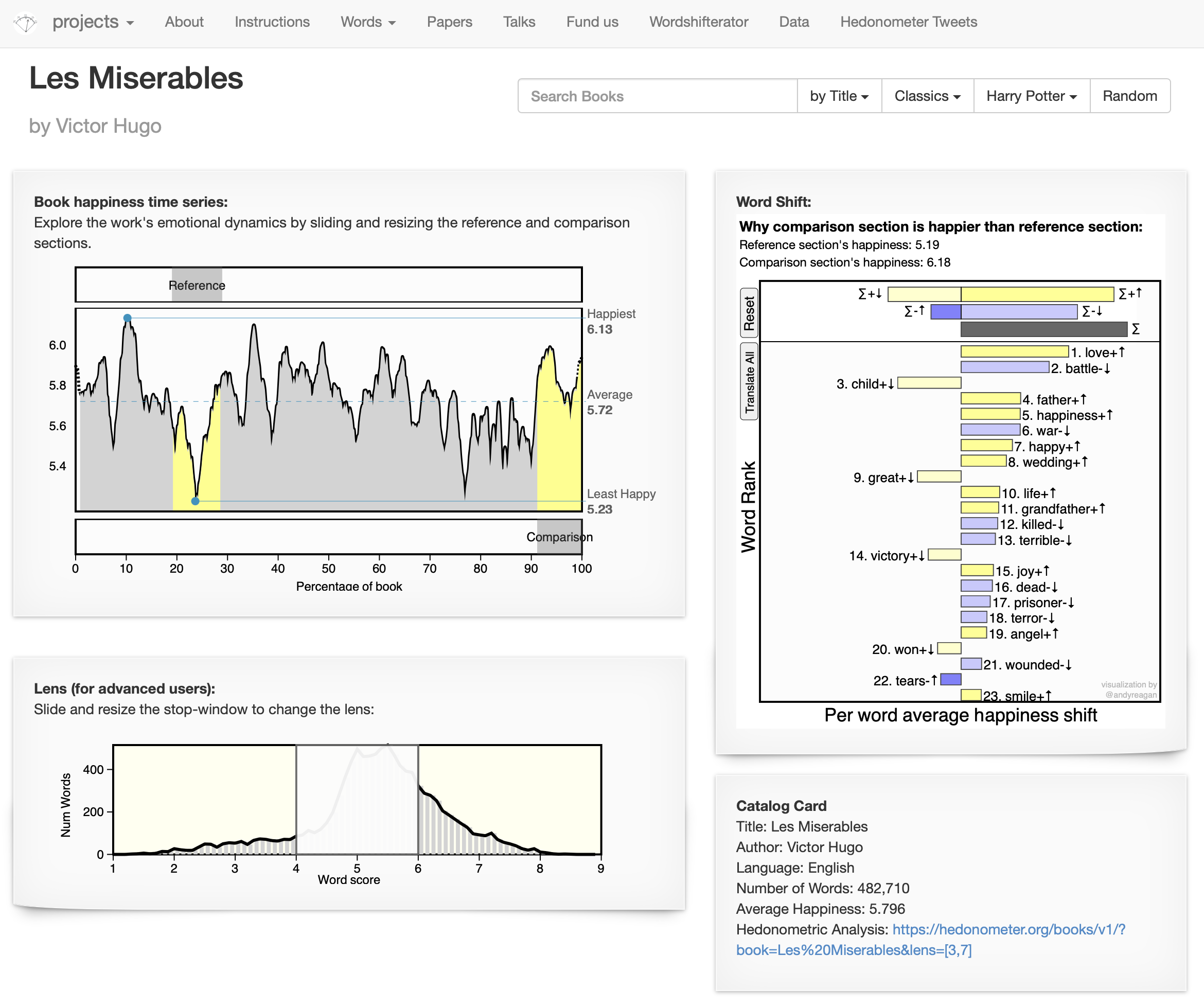}
  \caption{
    \textbf{
      Hedonometric time series for Les Mis\'{e}rables.
    }
    The time series 
    shows a similar profile to the
    ousiometric
    \semdiff{bad}{good} time series in
    Fig.~\ref{fig:meaning.ousiometer-les-miserables}
    in Sec.~\ref{sec:meaning.ousiometer}.
    Screenshot from interactive viewer at \url{https://hedonometer.org}
    based on Ref.~\cite{reagan2016c}.
  }
  \label{fig:meaning.hedonometer-les-miserables}
\end{figure*}

\clearpage

\section{Ousiometer prototype for high frequency social media}
\label{sec:meaning.ousiometer-twitter}

\twocolumn

\begin{figure*}[tp!]
  \includegraphics[width=\textwidth]{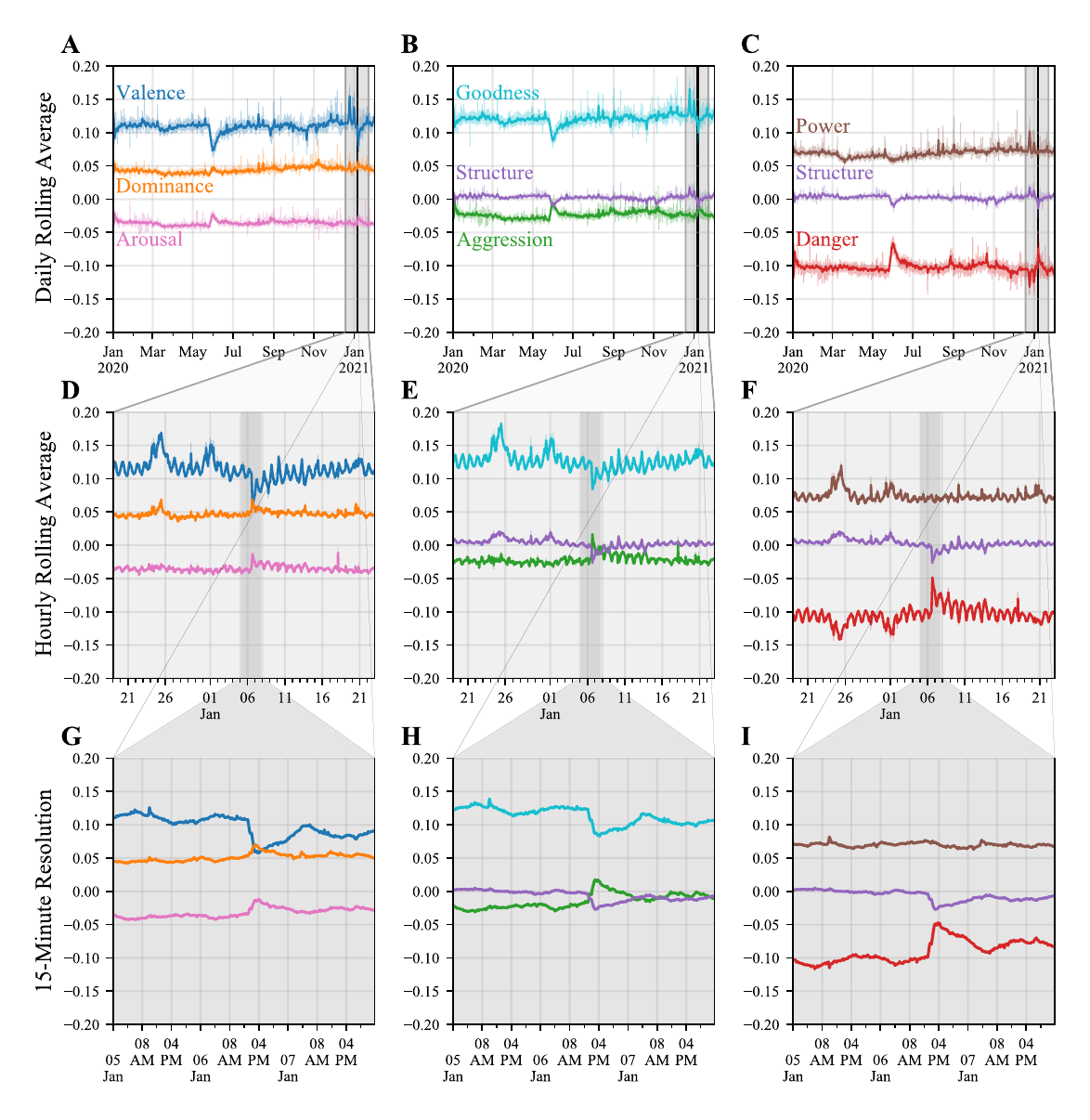}
  \caption{
    \textbf{
      The ousiometer:
      Example essential meaning time series for Twitter, 2020/01--2021/01.
    }
    The three columns correspond to average meaning
    scores for the 
    frameworks of VAD, GAS, and PDS,
    computed per \Req{eq:meaning.ousiometer}.
    The first row shows time series for
    the 13 months covering all of 2020 and January, 2021.
    The second and third rows focus in on the
    attack on the US Capitol on 2021/01/06 by supporters of President Trump.
    The scale for the second row is
    5 weeks (2020/12/19 to 2021/01/23)
    and
    3 days (2021/01/05 to 2021/01/07) for the third row.
    All underlying time series are 15 minute time scales
    with day-scale and hour-scale smoothing overlaid in
    the first and second rows.
    Major events with spikes and/or durable memory
    are the US's assassination of the Iranian general Soleimani,
    the COVID pandemic,
    George Floyd's murder,
    and
    events related to the 2020 US presidential election,
    including the attack on the US Capitol.
    Because dominance is relatively stable throughout,
    the GAS and PDS dimensions effectively vary as functions
    only of valence and arousal
    (see Eqs.~\ref{eq:meaning.VAD-to-GAS}
    and~\ref{eq:meaning.transformVAD-to-PDS}).
    In particular, goodness and aggression track
    valence and arousal closely.
    For the 2021/01/06 attack, the danger time series spikes
    while power remains stable (panels F and I).
    Structure drops indicating increased seriousness.
    In total, even though only three are independent,
    power, danger, goodness, aggression, and structure
    are all valuable time series.
    Notes:
    We constructed the
    Twitter $1$-gram corpus
    from approximately 10\% of all English tweets~\cite{alshaabi2021b,alshaabi2021c},
    with  all $1$-grams moved to lower case.
    We form a lexical lens $\lexicallens$ by taking
    1-grams from the NRC VAD lexicon
    and adding a hashtag version of each 1-gram.
    As such, the ousiometer is not specifically tailored for Twitter
    during the time period covered.
    As we have done for the hedonometer~\cite{dodds2011e,alshaabi2021d},
    our ousiometer could be readily
    improved by expanding the lexical lens to incorporate missing
    salient 1-grams.
  }
  \label{fig:meaning.ousiometer-twitter}
\end{figure*}

For an example corpus, we assess English language
Twitter~\cite{alshaabi2021b,alshaabi2021c} for
the historically turbulent time period 2020/01/01 to 2021/01/31~\cite{dodds2021d}.
In Fig.~\ref{fig:meaning.ousiometer-twitter},
we show ousiometric time series 
for the three frameworks of VAD, GAS, and PDS.
We explain how we compute these time series
and then briefly discuss how they track specific historical events.

We now apply our prototype ousiometer to
English language Twitter at a base
resolution of 15 minutes~\cite{alshaabi2021b,alshaabi2021c}.
We use~\Req{eq:meaning.ousiometer} to
generate the ousiometric time series
in Fig.~\ref{fig:meaning.ousiometer-twitter}.
The three columns of
Fig.~\ref{fig:meaning.ousiometer-twitter}
correspond
to the VAD, GAS, and PDS frameworks.
The rows from top to bottom
move from the year scale of 2020 
and the start of 2021, focusing
in on the attack on the US Capitol
by supporters of President Trump on 2021/01/06.
The specific time frames are
13 months (2020/01/01 to 2021/01/31), 
5 weeks (2020/12/19 to 2021/01/23),
and
3 days (2021/01/05 to 2021/01/07).
We overlay day-scale and hour-scale smoothing
for the first two rows respectively.

Looking across all panels, we see the various
ousiometric biases in the context of Twitter.
Valence, dominance, goodness, and power all show positive biases,
while arousal, aggression, and danger present negative averages.
Structure is the only neutral dimension.

At the year scale
in Figs.~\ref{fig:meaning.ousiometer-twitter}A--C,
the three frameworks show evidence of major shocks, trends,
and daily fluctuations, all to varying degrees.
The two major events in the first half
of 2020---those leading to long-lasting societal effects---were the global realization
of the COVID-19 pandemic in mid March and the murder of George Floyd
at the end of May and subsequent Black Lives Matter protests~\cite{alshaabi2021d}.
A number of other events also stand out
including
the assassination of the Iranian general Soleimani by the US on 2020/01/03,
which led to talk of war.

We only see the COVID-19-response shock
in four dimensions---valence, goodness, power, and danger---while
the shock of George Floyd's murder registers in all eight dimensions.
The COVID-19 shock is muted 
in part because
we are (understandably) missing key words in the NRC VAD lens
such as 
`coronavirus',
and
`covid'.
The word `pandemic' points directly to danger
with PDS scores (0.00,0.45,-0.03),
as does
`virus' with (-0.04, 0.32, 0.06).
As we discuss below, expanding the NRC VAD lexicon
is an evidently needed step for improving the ousiometer.

Moving to the five weeks of the second row
of Fig.~\ref{fig:meaning.ousiometer-twitter}, 
the main signal deviations are due to
Christmas, New Year's Eve and Day,
and the 2021/01/06 attack on the US Capitol.
We also now see a daily cycle across all dimensions,
reminiscent of what we found when measuring happiness (valence)
on Twitter using the hedonometer~\cite{dodds2011e,miller2011a}.

Finally, the time series in
the bottom row of Fig.~\ref{fig:meaning.ousiometer-twitter}
show, in high temporal resolution,
the collective shock expressed on Twitter
in response to the attack on the US Capitol.
For over roughly two hours starting after midday on 2021/01/06,
we see the strongest
shocks occur in
valence (decreasing, Fig.~\ref{fig:meaning.ousiometer-twitter}G)
and danger (increasing, Fig.~\ref{fig:meaning.ousiometer-twitter}I).

For the main dimensions of the orthogonal frameworks, GAS and PDS,
it is danger $\Mdanger$ that is the real dimension of change.
In the PDS framework,
while danger rises, power $\Mpower$ remains relatively constant throughout the attack.
In the GAS framework, the time series for
goodness and aggression mirror each other
and are projections of the danger signal.
We also observe an increase in more rigid and serious 1-grams,
as the structure score $\Mstructure$ drops through the attack.

While we have presented the GAS and PDS time series as distinct sets
and notwithstanding that they are of course linear transformations of each other,
we suggest that showing all five time series is of value.
The eight cardinal and intercardinal points of the power-danger plane are
all meaningful, and it is helpful to reflect on which one might be dominating.
We are after all plotting time series that represent the harder-to-visualize
trajectory of a curve in PDS space.

For a deeper analysis of all time series,
and beyond the scope of the present paper,
we would use
word shift graphs~\cite{dodds2009b,dodds2011e,dodds2015a,reagan2017a,gallagher2021a}
to illuminate which 1-grams drive changes in ousiometric scores.

We note that these instruments are not inherently predictive
for social phenomena, but rather extract
real-time signals of essential meaning from online text.
Analysis of such signals in pursuit of
prediction is itself a separate, massive,
and fraught enterprise~\cite{bermingham-smeaton-2011-using,gayo-avello2012a,jungherr2013forecasting}.
%% More success can be found in spaces where the analysis is across complete time series,
%% such as the measurement of emotional arcs suggested by Kurt Vonnegut~\cite{reagan2016c,fudolig2023a}.

\onecolumn

\end{document}